%% file: arbeit.tex
\documentclass[a4paper,twoside]{report}

 	\usepackage{printlen}
 	\usepackage{url}
 	\usepackage{multirow}
	 
	\usepackage[german, ngerman, english]{babel}
	
	\usepackage{amsmath}
	\usepackage{mathtools}
	\usepackage{amssymb}
	\usepackage{bbm}
	\usepackage{mathtools}
		
	\usepackage{bm}
		
	\usepackage{xfrac}
	
	\usepackage{tikz}
	\usetikzlibrary{arrows}
	
	\usetikzlibrary{calc}
	\usetikzlibrary{shapes.geometric}
	
	\usepackage{gnuplot-lua-tikz}
	
	\usepackage{enumerate}

	\usepackage[section]{placeins}
	\makeatletter
	\AtBeginDocument{%
  		\expandafter\renewcommand\expandafter\subsection\expandafter{%
   			\expandafter\@fb@secFB\subsection
  		}%
	}
	\makeatother
	
	\makeatletter
	\AtBeginDocument{%
  		\expandafter\renewcommand\expandafter\subsubsection\expandafter{%
   			\expandafter\@fb@secFB\subsubsection
  		}%
	}
	\makeatother
	
	\usepackage{float}
	
	\usepackage{caption}

	\usepackage{graphicx}

	\usepackage{pgfplots}	
	\pgfplotsset{compat = 1.13}
	\usepgfplotslibrary{groupplots}
	\pgfplotsset{empty/.style={%
	width	=	7.25cm,
	height	=	5.50cm,
	yticklabel style={text width=0em,align=right},
	yticklabels={,,},
	xticklabels={,,},  
	xmin=-1.6,xmax=1.6,
	ymin=-1.6,ymax=1.6,}}
	
	\makeatletter
	\pgfmathdeclarefunction{erf}{1}{%
  	\begingroup
    \pgfmathparse{#1 > 0 ? 1 : -1}%
    \edef\sign{\pgfmathresult}%
    \pgfmathparse{abs(#1)}%
    \edef\x{\pgfmathresult}%
    \pgfmathparse{1/(1+0.3275911*\x)}%
    \edef\t{\pgfmathresult}%
    \pgfmathparse{%
      1 - (((((1.061405429*\t -1.453152027)*\t) + 1.421413741)*\t 
      -0.284496736)*\t + 0.254829592)*\t*exp(-(\x*\x))}%
    \edef\y{\pgfmathresult}%
    \pgfmathparse{(\sign)*\y}%
    \pgfmath@smuggleone\pgfmathresult%
 	\endgroup
	}
	
	\usepackage{array}
	\newcolumntype{x}[1]{>{\centering\arraybackslash\hspace{0pt}}p{#1}}
	\usepackage{arydshln}

	\usepackage{nameref}
	\usepackage[toc,page]{appendix}	

	\RequirePackage{natbib}
	\renewcommand\cite{\citep}	
	 
	\title{Word Embeddings and Stability}
	\author{Lucas Rettenmeier\\Heidelberg Institute for Theoretical Studies\\ }		  

	\usepackage[width = 345pt, height = 598pt, centering]{geometry}

	\newcommand\frontmatter{%
  	\if@openright
    \cleardoublepage
  	\else
    \clearpage
  	\fi
  	\pagenumbering{roman}}

	\newcommand\mainmatter{%
  	\if@openright
    \cleardoublepage
  	\else
    \clearpage
  	\fi
 	\pagenumbering{arabic}}
	\newcommand\backmatter{
  	\if@openright
    \cleardoublepage
  	\else
    \clearpage
 	\fi
  	\pagenumbering{roman}
   	}

	\makeatother

	\usepackage[hang, flushmargin]{footmisc}
	\input{chapters/toc.tex}

	\DeclarePairedDelimiterX{\infdivx}[2]{(}{)}{%
  	#1\;\delimsize\|\;#2%
	}

	\newcommand{\wtv}{\textbf{word2vec}}
	\newcommand{\glv}{\textbf{GloVe}}
	\newcommand{\ftt}{\textbf{fastText}}

\begin{document}


	\pagestyle{empty}
	
	\input{chapters/titlepage.tex}

	
	\input{chapters/abstract.tex}


	\frontmatter
	\setcounter{page}{1}
	\tableofcontents

	\input{chapters/introduction.tex}
	
	
	\mainmatter
	\pagestyle{plain}
	\setcounter{page}{1}	
	
	\input{chapters/1_prerequisites.tex}
	\input{chapters/2_instability.tex}
	\input{chapters/3_reducing.tex}
	\input{chapters/4_semantic.tex}

	\input{chapters/5_discussion.tex}
	

	\backmatter
	\setcounter{page}{1}
	\input{chapters/appendix.tex}

\end{document}

%% file: chapters/titlepage.tex
\begin{titlepage}
	\begin{center}
 
	\Large\textbf{Department of Physics and Astronomy\\
	University of Heidelberg}

	\vspace{14cm}

	\normalsize
	Master Thesis in Physics\\
	submitted by\\
	\vspace{0.5cm}
	\textbf{Lucas Rettenmeier}\\
	
	\vspace{0.5cm}
	born in Aalen (Germany)\\
	\vspace{0.5cm}
	\textbf{2020}

	\newpage

	\Large\textbf{Word Embeddings}\\
	\Large Stability and Semantic Change

	\vspace{16cm}

	\normalsize
    This Master Thesis has been carried out by Lucas Rettenmeier at the\\
    Heidelberg Institute for Theoretical Studies under the supervision of\\
    Prof. Dr. Michael Strube and Prof. Dr. Fred Hamprecht.
	\vfill
	\end{center}
\end{titlepage}

%% file: chapters/abstract.tex
\begin{abstract}
Word embeddings are computed by a class of techniques within \textsl{natural language processing} (NLP), that create continuous vector representations of words in a language from a large text corpus. The stochastic nature of the training process of most embedding techniques can lead to surprisingly strong instability, i.e. subsequently applying the same technique to the same data twice, can produce entirely different results \cite{hellrich-hahn-2016-assessment, antoniak2018, wendlandt2018}. In this work, we present an experimental study on the instability of the training process of three of the most influential embedding techniques of the last decade: \wtv\ \cite{mikolov2013b}, \glv\ \cite{pennington-etal-2014-glove} and \ftt\ \cite{bojanowski2016}. Based on the experimental results, we propose a statistical model to describe the instability of embedding techniques and introduce a novel metric to measure the instability of the representation of an individual word. Finally, we propose a method to minimize the instability -- by computing a modified average over multiple runs -- and apply it to a specific linguistic problem: The detection and quantification of \textsl{semantic change}, i.e. measuring changes in the meaning and usage of words over time.
\end{abstract}

\begin{otherlanguage}{ngerman}
	\begin{abstract}
Word-Embeddings sind das Ergebnis einer Klasse von Methoden in der Computerlinguistik, mit denen kontinuierliche Vektordarstellungen von W\"ortern einer Sprache aus einem gro{\ss}en Textcorpus konstruiert werden. Die stochastische Natur der Trainingsprozesse dieser Methoden kann zu \"uberraschend gro{\ss}er Instabilit\"at f\"uhren, das hei{\ss}t, die zweimalige Anwendung einer Methode auf einen Textcorpus kann stark variierende Ergebnisse liefern \cite{hellrich-hahn-2016-assessment, antoniak2018, wendlandt2018}. In dieser Arbeit pr\"asentieren wir eine experimentelle Studie zur Instabilit\"at der Trainingsprozesse drei der bedeutendsten Word-Embedding Methoden des letzten Jahrzehnts: \wtv\ \cite{mikolov2013b}, \glv\ \cite{pennington-etal-2014-glove} und \ftt\ \cite{bojanowski2016}. Auf Basis der experimentellen Resultate entwickeln wir ein statistisches Modell zur Beschreibung der Instabilit\"at der Methoden und f\"uhren eine neue Metrik zur Messung der Instabilit\"at der Vektordarstellung einzelner Worte ein. Schlie{\ss}lich erarbeiten wir ein Verfahren, um die Instabilit\"at zu minimieren: Das Bilden eines modifizierten Mittelwerts \"uber mehrere Trainingsl\"aufe. Abschlie{\ss}end wird dieses Verfahren auf eine spezifische linguistische Problemstellung angewandt: Bedeutungswandel -- das hei{\ss}t \"Anderungen in der Bedeutung und Nutzung von W\"ortern -- zu erkennen und zu messen.
	\end{abstract}
\end{otherlanguage}

%% file: chapters/introduction.tex
\chapter*{Introduction}

In the 1950s, linguists like \citet{joos1950}, \citet{harris1954} and \citet{firth1957}, formulated the \textsl{distributional hypothesis} -- the idea, that words that frequently occur in the same contexts tend to have similar meanings. This was popularized by Firth's claim ``a word is characterized by the company it keeps'', which is widely accepted by linguists today.

Vector semantics, a key area within NLP research in the last decades, is based on this hypothesis: The aim is to learn representations, usually in the form of real-valued $d$-dimensional vectors, of the meaning of individual words (also called \textsl{embeddings} or \textsl{word vectors}) from their distributions in (large) text corpora. The first techniques to produce dense vectors that represent the meaning of words were introduced by \citet{deerwester1989}, and shortly later recast as \textbf{LSA}: Latent semantic analysis \cite{deerwester1990}.

More than a decade later, \citet{bengio2003} applied a neural network model, using the back-propagation technique of \citet{rumelhart1986} to statistical language modelling, specifically, to the task of predicting a word given the two words to the left and to the right. too this approach, one does not only obtain the language model, but also the parameters of the model -- dense word representations -- that may be used for other, potentially unrelated, tasks. The approach was further improved by \citet{bengio2007}, \citet{collobert2008} and \citet{mnih2009}, but it took another decade before word embeddings started to rise to the level of relevance they inhibit today.

Research interest grew rapidly after \cite{mikolov2013b} published a neural network-based model, called \wtv, that allowed very efficient training hence enabled the use of training corpora up to a size of $10^{11}$ words. \citet{mikolov-etal-2013-linguistic} made the somewhat surprising observation, that these distributed representations capture syntactic and semantic regularities in linear relationships. For example, the vector operation:
\begin{align}
\vec{v}\,(\,\texttt{king}\,) - \vec{v}\,(\,\texttt{man}\,) + \vec{v}\,(\,\texttt{woman}\,)
\end{align}
yields a vector that is closer to the representation of \texttt{queen} than of any other word.

Thereafter, numerous models inspired by the approach of \citet{mikolov2013b} were published: \citet{pennington-etal-2014-glove} developed \glv, a count-based method with a similar optimization objective to \wtv, which according to the authors, leverages the statistical information more efficiently than the prediction-based, neural network models. More recently, \citet{bojanowski2016} introduced \ftt, applying the \wtv\ model to sub-word structures (character-$n$-grams), instead of words. 

This surge in research interest was accompanied and driven by an increasing number of downstream NLP applications that were found to benefit from the use of embeddings. Today, \citet{jurafsky_book} go as far as to say: ``These representations are used in every NLP application that makes use of meaning''. A few prominent examples are text classification \cite{sebastiani2002, lilleberg2015, zhang2015b}, question answering \cite{tellex2003, yih-etal-2014-semantic}, named entity recognition \cite{siencnik2015, habibi2017} and information retrieval \cite{manning2008, zuccon2015}.

Today, contextualized embeddings like \textbf{ELMo} \cite{peters2018} and \textbf{BERT} \cite{devlin2018} -- neural network-based models, that calculate the conditional representation of a word given its context -- outperform the earlier models mentioned above on most tasks. One of the main advantages of these models is the ability to differentiate between different word senses of a homonym, e.g. the representation of the word \texttt{bank} will be vastly different for the two contexts listed below:
\begin{enumerate}[align=left]
	\itemsep0em
    \item [\bfseries{Context 1:}] \texttt{He is sitting on the \underline{bank} of the river.}
    \item [\bfseries{Context 2:}] \texttt{She made a deposit at the \underline{bank} earlier this morning.}  
\end{enumerate}
Non-contextualized embedding techniques, on the other hand, assign the same representation to the word \texttt{bank} in both contexts. 

One task where non-contextualized word embeddings -- despite their shortcomings -- are still widely used today is the detection and measurement of \textsl{semantic change}, i.e. how the meaning of words changes over time \cite{tang_2018, kutuzov2018, tahmasebi2018}. \citet{hamilton2016} used \wtv\ embeddings, trained on historical corpora to derive statistical laws of semantic change, e.g. that less frequently used words tend to have higher rates of semantic change than more frequently used ones, but \citet{dubossarsky2017} contested these findings and argued, that they are actually artifacts of the inherent instability of the embedding techniques.

The problem of the instability of embedding techniques, i.e. the variance between two models that are subsequently trained with the same technique on the same training corpus, was raised repeatedly in the last years \cite{hellrich-hahn-2016-assessment, antoniak2018, wendlandt2018}. However, we found little research on how to minimize the instability and prevent the issues that are caused by it.

In this work, we examine the instability of different techniques for the training of non-contextualized word embeddings, propose a method to create more stable embeddings and apply our findings to the evaluation of semantic change in different languages.

In Chapter 1 the embedding techniques and text corpora that are relevant for our experiments, as well as conventions on notation are introduced. 

In Chapter 2 we present the -- to our knowledge -- largest study to date on the stability of non-contextualized word embeddings: We performed experiments on three of the most popular methods for non-contextualized word embeddings in recent years (\wtv, \glv\ and \ftt), by performing multiple subsequent training runs on Wikipedia corpora in seven languages with numerous configurations -- training more than 10,000 models in total. We were able to describe the resulting variability accurately with a simple statistical model and introduce a novel measure to quantify the distance between word embeddings, that circumvents several problems of the approaches used in most of the previous work. Finally, we propose a novel distinction between two types of instability, that yields insights on the internal structure of the different embedding techniques.

In Chapter 3 the influence of the choice of the embedding technique as well as hyper-parameter settings on the instability is described, before we propose a novel approach to reduce the instability of the embeddings -- by averaging over aligned samples -- that is supported by the statistical model introduced in Chapter 2 and delivered promising results in our experiments.

Finally, we apply this novel approach to two different problems in the context of semantic change -- outlined in Chapter 4: Firstly, task 1 of the SemEval 2020 Workshop \cite{schlechtweg2020semeval}, where our best submission ranks 7th and 6th out of 34 participating teams, on the two sub-tasks respectively. And secondly, we used the instability-reducing approach to differentiate between true semantic change on a large historical corpus and the artifacts found by \citet{dubossarsky2017} to confirm the \textsl{law of conformity} proposed by \citet{hamilton2016}.

%% file: chapters/1_prerequisites.tex
\chapter{Prerequisites}
In this chapter, we introduce the embedding techniques and text corpora that are relevant for our experiments; as well as the conventions on notation that are used in this work.

\section{Notation}

\begin{itemize}

\item The \textsl{vocabulary}, i.e. the set of all words for which the respective model contains a representation is written as $\mathcal{V}$, and its size as $|\mathcal{V}|=:v$. If we refer to any word of the vocabulary, it is spelled in typewriter font, e.g. \texttt{cat}.

\item Lowercase letters with arrows, like $\vec{u},\vec{v}\in\mathbb{R}^d$ refer to row vectors of \textsl{dimension} $d$. A model represents every word $w\in\mathcal{V}$ of the vocabulary as a \textsl{word embedding} (or word vector) of this shape. 

\item We use \textsl{bold capital letters to denote matrices}, like an embedding space $\mathbf{V}_i\in\mathbb{R}^{v\times d}$, i.e. the stack of embeddings (row vectors) of all words of the vocabulary, or a matrix-transformation $\mathbf{A}\in \mathbb{R}^{d\times d}$.

\item \textsl{Corpora}, i.e. collections of texts, including the specific preprocessing that was applied to them, are denoted as $\mathcal{C}$.

\item \textsl{Embedding techniques}, like \wtv,\ \glv\ or \ftt, along with all the respective choices for the free parameters of these techniques are denoted as $\mathcal{T}$.

\item We use Greek capital letters for \textsl{probability distributions}; the Normal distribution is denoted as $\mathcal{N}$.

\item Bold lowercase letters are used for \textsl{distance metrics} $\mathbf{d}$, which capture the difference between the embeddings of a word $w$ in the two embedding spaces $\mathbf{V}_i$ and $\mathbf{V}_j$. 

\end{itemize}

\section{Experimental Setup}
\label{sec_exp_setup}
\subsection{Embedding Techniques}

All results and findings described below are based on the following experimental setup: We chose three of the most influential techniques for non-contextualized word embeddings in the last decade, namely \wtv\ \cite{mikolov2013c}, \glv\ \cite{pennington-etal-2014-glove} and \ftt\ \cite{bojanowski2016}. As outlined in Table \ref{tab_embedding_models}, these models cover three distinct classes of embedding techniques. For every technique, we used the latest implementations provided by the original authors. All models were trained with the default parameters\footnote{For \wtv\ and \ftt, the \texttt{skip-gram} setting was used. For \ftt, one change was made to the default parameters: The initial learning rate \texttt{lr} was set to $0.1$, as we found this setting to increase the score on word analogy tasks for most evaluated languages. And finally, when training \glv\ on the English corpus, we had to restrict the number of iterations to 25, due to technical limitations.} and a 300-dimensional embedding space.

\begin{table}[htbp]
\center
\input{tables/tab_embedding_models.tex}
\caption{Classification of the three different models for non-contextualized word embeddings that were used within the scope of this work. As one can see, all three models have distinct and qualitatively different characteristics.}
\label{tab_embedding_models}
\end{table}

\subsubsection{word2vec}
\label{sec_w2v}
As mentioned before, the introduction of \wtv\ by \citet{mikolov2013b} lead to a surge of interest within NLP research in the distributed representations of words, or word embeddings.\footnote{Later, similar models were used to obtain representations of $n$-grams, byte-pairs, sentences and documents.} The approach of \citet{mikolov2013b} -- because of its model architecture -- is far more efficient in the training of the embeddings than earlier prediction-based techniques, like the ones developed by \citet{bengio2003} or \citet{collobert2008}, hence enabling the use of larger training corpora, up to a size of $10^{11}$ words. This is reflected in a significant increase in the quality of the embeddings, compared to the earlier methods -- as measured on word analogy and similarity tasks -- and contributed to the increasing use of the representations in ``every NLP application that makes use of meaning'' \cite{jurafsky_book}.

\begin{figure}[htbp]
\center
\input{tikz/fig_skip_gram.tex}
\caption{Illustration of the skip-gram model architecture introduced by \citet{mikolov2013b}. In the example, the input embedding $\vec{v}_i(w_t)$ of the target word $w_t$ (=\texttt{sat}) is used to \textsl{predict} its context, i.e. the output embeddings $\vec{v}_o(w_{t+j})$ of the surrounding words. At any training step, the input and output embeddings are updated with the objective of maximizing $\log p\left(w_{t+j}\,|\,w_t\right)$.}
\label{fig_skip_gram}
\end{figure}
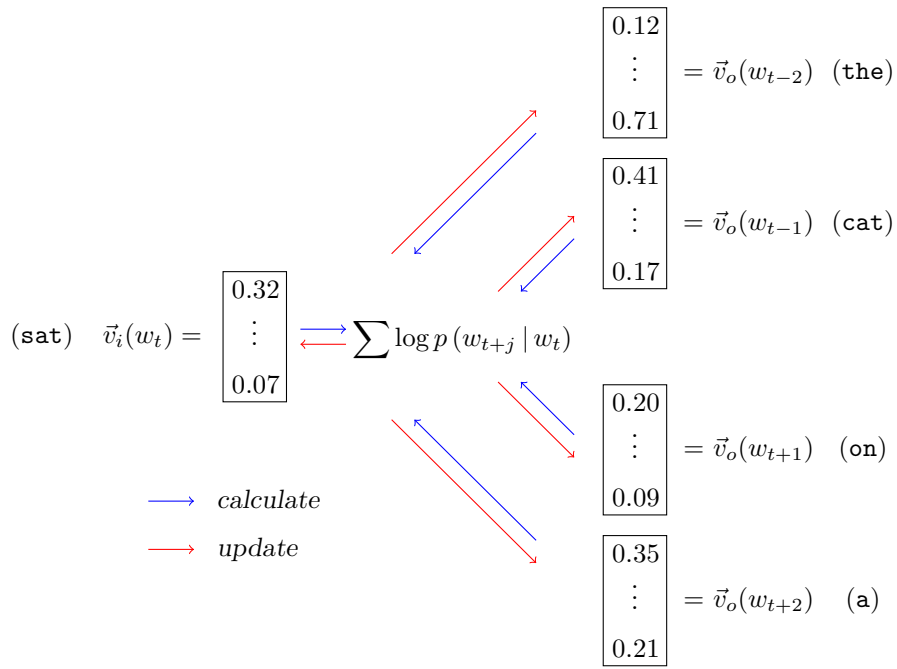

\citet{mikolov2013b} introduced two distinct flavors of the \wtv\ model: Continuous bag-of-words and skip-gram. We focus on the latter, since it was more commonly featured in previous work on instability and semantic change \cite{hamilton2016,antoniak2018}. The model architecture is illustrated in Figure \ref{fig_skip_gram}: The input embedding $\vec{v}_i(w_t)\in\mathbb{R}^d$ of the target word $w_t$ is used to \textsl{predict} its context, i.e. the output embeddings $\vec{v}_o(w_{t+j})\in\mathbb{R}^d$ of the surrounding words, hence the model is classified as \textsl{prediction-based} (see Table \ref{tab_embedding_models}).

The training of the input and output embeddings -- which is based on iterating over all words of the corpus -- is illustrated in Figure \ref{fig_skip_gram}. For any pair\footnote{The number of context words per target word, i.e. the size of the context is an adjustable parameter of the model. Typically, the $c=5$ words before and after the target word are used.} of target word $w_t$ and context word $w_{t+j}$ the training objective is to \textsl{maximize the logarithm of the predicted probability} $p\left(w_{t+j}\,|\,w_t\right)$ to observe $w_{t+j}$ in the context of $w_{t}$, which \citet{mikolov2013b} define -- in the basic formulation of the model -- as the normalized exponential function (\textsl{softmax}) of the dot product of the two embeddings:
\begin{align}
\label{eq_skipgram}
p\left(w_{t+j}\,|\,w_t\right)=\frac{\exp\left[\vec{v}_o(w_{t+j})\cdot \vec{v}_i(w_{t})^{\top}\right]}{\sum_{w\in\mathcal{V}}\exp\left[\vec{v}_o(w) \cdot \vec{v}_i(w_{t})^\top\right]}
\end{align}
However, since the evaluation of the denominator in the equation above requires calculating $v=|\mathcal{V}|$ vector products, this approach is computationally very expensive. The authors provide two more efficient alternatives to approximate $p\left(w_{t+j}\,|\,w_t\right)$: \textsl{Hierarchical softmax} and \textsl{negative sampling}. As mentioned above, all models in our experiments were trained with the default parameters, hence negative sampling was used. This is a modification of the \textsl{noise contrastive estimation} introduced by \citet{gutmann2012}: The term $\sum_{w\in\mathcal{V}}\exp\left[\vec{v}_o(w)\cdot\vec{v}_i(w_{t})^\top\right]$ in Equation (\ref{eq_skipgram}) is estimated by randomly drawing $k$ ``noise words'' from the vocabulary. With typical values of $5\leq k\leq 25$, this approach is several orders of magnitude faster than the softmax approach.

The optimization objective for every training step of the negative sampling model is to maximize the following expression (in this context, $\sigma$ refers to the sigmoid function):
\begin{align}
\label{eq_sgns}
\log \sigma\left[\vec{v}_o(w_{t+j})\cdot\vec{v}_i(w_{t})^\top\right] + \sum_{n=1}^{k}\mathbb{E}_{w_n\sim P_n(w)}\left\{\log\sigma\left[-\vec{v}_o(w_{n})\cdot \vec{v}_i(w_{t})^\top\right]\right\}
\end{align}
The second term is calculated by drawing $k$ noise words $w_n$ from the vocabulary, according to the noise distribution $P_n(w)$: The unigram distribution raised to the power $3/4$.\footnote{\citet{mikolov2013b} found this shape of the noise distribution to significantly outperforms other conceivable approaches.}

Another measure introduced by the authors to improve the training efficiency is the \textsl{sub-sampling} of frequent words: Any appearance of a word $w$ in the training corpus, whose frequency $f(w)$ exceeds a threshold $t$ (typically $t\approx 10^{-5}$) is discarded with the probability $p_d(w)$:
\begin{align}
p_d(w)=1-\sqrt{\frac{t}{f(w)}}
\end{align} 
Apart from accelerating the training process, \citet{mikolov2013b} found this setting to significantly improve the quality of the learned embeddings of rare words.

Finally, after the training -- which generally consists of several epochs over the full corpus with a continuously decreasing learning rate -- is completed, the embedding $\vec{v}(w)\in\mathbb{R}^d$ of any word $w$ is defined as:
\begin{align}
\vec{v}(w):= \vec{v}_i(w)
\end{align}
The output embeddings $\vec{v}_o(w)$ are discarded.

\subsubsection{GloVe}
\citet{pennington-etal-2014-glove} argued, that the \wtv\ technique presented above, poorly utilizes statistical information of the training corpus, since the embeddings are trained subsequently on word-context pairs, instead of global co-occurrence counts, such as methods like latent semantic analysis (\textbf{LSA}), introduced by \citet{deerwester1990}. However, given the desirable vector space properties of the \wtv\ embeddings, manifested in the best performance on word analogy and word similarity tasks of all techniques at the time, they proposed a global regression model with a similar optimization objective to the one used by \citet{mikolov2013b} and called it \textsl{Global Vectors for Word Representation} (\glv).

In the first step, the global word-word co-occurrence matrix $\mathbf{X}\in \mathbb{N}^{v\times v}$ is constructed from the training corpus, where the entry $X_{kl}$ corresponds to the number of times the word $w_l$ appears in the context of the target word $w_k$.\footnote{In our experiments, a context of 15 words to the left and 15 words to the right of the target word, was used.}

The optimization objective of the technique is to minimize the following expression:
\begin{align}
\label{eq_glove_optimization}
\sum_{k,l=1}^{v}f\left(X_{kl}\right)\left[\vec{v}_i(w_k)\cdot \vec{v}_o(w_l)^\top + b_i(w_k) + b_o(w_l)-\log X_{kl}\right]^2
\end{align}
Where -- similarly to \wtv\ -- $\vec{v}_i(w_k)$ and $\vec{v}_o(w_l)$ refer to the input and output embeddings of the word $w$ respectively; $b_i(w_k)$ and $b_o(w_l)$ are word-dependent bias terms. \citet{pennington-etal-2014-glove} define the \textsl{weighting function} $f:\mathbb{R}\to\mathbb{R}$ as:
\begin{align}
f = 
\begin{cases}
\left(x/x_{\text{max}}\right)^{\alpha} 	&	\text{ if } x < x_{\text{max}} \\
\multicolumn{1}{@{}c@{\quad}}{1}			& 	\text{ otherwise}
\end{cases}
\end{align}
With typical values of $x_{\text{max}}=100$ and $\alpha = 3/4$. Since $f(0)=0$, this function allows zero entries in the co-occurrence matrix that correspond to $\log X_{kl}\to \infty$. In similar approaches like \textbf{LSA}, that do not have a weighting function, an artificial offset must be added to $\mathbf{X}$ instead.

Intuitively, Equation (\ref{eq_glove_optimization}) means, that we minimize $[\vec{v}_i(w_k)\cdot \vec{v}_o(w_l)^\top-\log X_{kl}]$ while allowing for a fixed bias per word and allowing larger deviations for pairs with fewer than $x_{\text{max}}$ co-occurrences.

Technically, the adaptive gradient method introduced by \citet{duchi2011} is used to solve the optimization problem, considering only non-zero elements of $\mathbf{X}$, to obtain the embeddings $\vec{v}_i(w)$ and $\vec{v}_o(w)$ for every word $w\in\mathcal{V}$.

Finally, given that Equation (\ref{eq_glove_optimization}) is invariant under the exchange of $k$ and $l$, we expect the input and output embeddings to coincide -- apart from random fluctuations. Hence, the authors define the vector representation $\vec{v}(w)\in\mathbb{R}^d$ of any word $w$ as:
\begin{align}
\vec{v}(w):= \vec{v}_i(w) + \vec{v}_o(w)
\end{align}

\subsubsection{fastText}
The \ftt\ technique, developed by \citet{bojanowski2016} is based on the skip-gram model with negative sampling, introduced in Section \ref{sec_w2v}. The authors claim, that one of the main limitations of this -- and other popular models -- is that they ignore the morphology of words, by assigning a distinct vector to each word. Hence, they propose to represent each word as a bag of character-$n$-grams, and to train vector representations of these character-$n$-grams. Compared to the conventional skip-gram model, the modified approach comes with faster training times, allows to compute representations for words that did not appear in the training data and achieves slightly higher scores on most word analogy tasks \cite{bojanowski2016}.

In the model, each word $w$ of the vocabulary is represented by a set of tokens $\mathcal{Z}(w)$ that contains all character-$n$-grams with $n_l\leq n\leq n_u$, as well as the word itself. The boundary symbols $<$ and $>$ are added to the beginning and end of the word to distinguish prefixes and suffixes. Taking the word \texttt{chair} as an example, with $n_l = 3$ and $n_u=4$, the representation looks like this:
\begin{align*}
\texttt{chair} \rightarrow \{\underbrace{\texttt{<ch}, \texttt{ cha}, \texttt{ hai}, \texttt{ air}, \texttt{ ir>}}_{n=3 \text{ grams}},\ \underbrace{\texttt{<cha}, \texttt{ chai}, \texttt{ hair}, \texttt{ air>}}_{n=4 \text{ grams}},\ \underbrace{\texttt{<chair>}}_{\text{word}}\}
\end{align*}
In practice, $n_l=3$ and $n_u=6$ are most commonly used. The representation $\vec{v}(w)$ of a word $w$ is then defined as the sum of the vector representations $\vec{z}(g)$ of all $n$-grams $g\in\mathcal{Z}(w)$:
\begin{align}
\vec{v}(w)=\sum_{g\in\mathcal{Z}(w)}\vec{z}(g)
\end{align}
The training is similar to the skip-gram approach, with the same optimization objective -- outlined in Equation (\ref{eq_sgns}). However, the calculation of the product of two word embeddings $\vec{v}_o(w_{1})$ and $\vec{v}_i(w_2)$ is based on their sub-word embeddings:
\begin{align}
\vec{v}_o(w_{1})\cdot \vec{v}_i(w_2)^\top = \sum_{g_1\in\mathcal{Z}(w_1)} \ \sum_{g_2\in\mathcal{Z}(w_2)} \vec{z_o}(g_1)\cdot \vec{z_i}(g_2)^\top
\end{align}
And after every training step, the sub-word embeddings $\vec{z}_o(g)$ and $\vec{z}_i(g)$ are updated, instead of $\vec{v}_o(w)$ and $\vec{v}_i(w)$, as in the model of \citet{mikolov2013b}.

As mentioned above, this technique allows to obtain representations of words that did not occur in the training data, by calculating the sum over the respective character-$n$-grams. However -- to ensure comparability with the word-based approaches -- we did not make use of this functionality in our experiments.
\subsection{Corpora}
\label{sec_wiki_coprora}
All our experiments on the stability of word embeddings, are based on models trained on Wikipedia corpora in one of seven different languages. Wikipedia is the largest free online encyclopaedia, available in more than 200 different languages. Because the articles are curated, high text quality is ensured. Please refer to Table \ref{tab_embedding_models} for a list of the languages along with the sizes of the respective corpora and vocabularies. Apart from English, which is a natural choice, the languages were selected based on two criteria:

\begin{enumerate}[(I)]

\item Firstly, as one would naturally expect and \citet{antoniak2018} have claimed, smaller corpora tend to be less stable than larger ones. Furthermore, we are interested in detecting semantic change over time, hence we need to train word embeddings for specific epochs, for which there are often only comparatively small corpora available. Therefore, the focus of our investigations is on languages with a limited size of training data.

\item Secondly, to ensure the validity of our preprocessing and training setup, we want to compare the quality of our embeddings in any language against published baselines. Word analogy tasks, i.e. datasets composed of word 4-tuples of the form \textsl{Man : Woman :: King : Queen } have become the de facto standard to evaluate the quality of non-contextualized word embeddings in recent years.

\end{enumerate}

Hence, we selected the languages with the smallest sized Wikipedia, for which an analogy dataset as well as a baseline score for at least one of the three embedding techniques used in this work, which was trained on a comparable corpus, has been published. 

\begin{table}[htbp]
\center
\input{tables/tab_embedding_languages.tex}
\caption{Outline of the seven different Wikipedia corpora used to train word embeddings. The XML Wikipedia dumps that were used in our experiments were created on the 1st of September 2019. The most current ones can be obtained from \url{https://dumps.wikimedia.org/}. As per the default settings of our embedding models, only words with five or more occurrences are included in the vocabulary.}
\label{tab_embedding_languages}
\end{table}

\subsubsection{Preprocessing}
As explained in more detail in Section \ref{sec_comparison}, the scores published by \citet{grave-etal-2018-learning} are used as a baseline on the word analogy tasks for any language apart from English. To ensure that the scores are comparable, we follow the preprocessing pipeline outlined in their work, which consists of three steps:

\begin{description}

\item[Text Extraction] The text content of the XML Wikipedia dumps is extracted with a modified version of Matt Mahoney's \texttt{wikifil.pl} script\footnote{\url{http://mattmahoney.net/dc/textdata.html}.}. The most notable deviation from the original script is the following: Letters are not lowercased, which means in practice that capitalized and non-capitalized occurrences of a word (e.g. \texttt{The} and \texttt{the}) are presented to the embedding models as two distinct tokens.

\item[Deduplication] The second step of the pipeline comprises the removal of duplicate lines from the data. We used the tool published by \citet{grave-etal-2018-learning}, which computes a hash of each line and removes all lines with identical hashes.\footnote{While this approach might -- in theory -- lead to non-duplicate lines being deleted, the small probability associated with an incident of this sort means that the quality of the embeddings is not impaired.} Overall, around 20\% of the data is removed in this step.

\item[Tokenization] Finally, the de-duplicated text data is tokenized. We used the Stanford word segmenter \citep{chang2008} for Chinese, the ICU tokenizer for Hindi, and the tokenizer from the Europarl preprocessing tools \citep{koehn2005} for the remaining languages. 

\end{description}

\subsubsection{Repeated Runs with Random Document Sampling}

Finally, as we want to understand the nature of the random processes in the training of word embeddings, every model in every language was trained at least $128$ times for three different types of document sampling. The three sampling methods \textsl{fixed}, \textsl{shuffled} and \textsl{bootstrapped}, which were introduced to this scope by \citet{antoniak2018} are outlined in Table \ref{tab_sampling_methods}.

\begin{table}[htbp]
\center
\input{tables/tab_sampling_methods.tex}
\caption{To examine the random nature of the different embedding techniques, we trained every model in every language at least 128 times for the three different types of document sampling listed above. This allows us to measure the influence of the document order, as well as the presence of individual documents on the variability of the resulting embeddings.}
\label{tab_sampling_methods}
\end{table}

\subsection{Implementation}

We wrote a Python module to store, compare, and analyse word embedding spaces independent of the underlying technique. For the training, the original implementations of \wtv, \glv, and \ftt\ are called from within the module. The code is published on GitHub.\footnote{\url{https://github.com/lucasrettenmeier/word-embedding-stability}}

\subsection{Comparison to Baseline on Word Analogy Tasks}
\label{sec_comparison}

To validate the corpora, our preprocessing pipeline, and proper training of the respective models, we compare the performance on \textsl{word analogy tasks} with previously published baselines. Word analogy tasks are datasets composed of word 4-tuples of the form \textsl{Man : Woman :: King : Queen} and have become the de facto standard to evaluate the quality of non-contextualized word embeddings in recent years.

For English we use the dataset published by \citet{mikolov2013b}, that of \citet{svoboda2016} for Czech, that of \citet{chen2015} for Chinese, that of \citet{venekoski-vankka-2017-finnish} for Finnish, that of \citet{hartman2017} for Portuguese\footnote{The dataset consists of a European as well as a Brazilian variant, only the European variant was used in the scope of this work.}, and finally the datasets proposed by \citet{grave-etal-2018-learning} for Hindi and Polish. For all languages listed above, apart from English, we use the scores published by \citet{grave-etal-2018-learning} on the same task as a baseline. The performance of the English embeddings is measured against the results of \citet{bojanowski2016}.

Tables \ref{tab_ftt_baseline_comparison} and \ref{tab_wtv_baseline_comparison} show the scores that we obtained in comparison to the different baselines. Following \citet{grave-etal-2018-learning}, the vocabulary of each model was restricted to the 200,000 most frequent words from the training data before evaluating the model on the word analogy dataset.\footnote{Although \citet{bojanowski2016} did not explicitly mention it, we confirmed with the authors that the scores they reported are also based on restricting the vocabulary to the 200,000 most frequent words.} This means, that a fraction of the questions of the analogy tasks is not answered, as they contain out-of-vocabulary words. Therefore, to compare the reported scores one also needs to take the coverage, i.e. the percentage of answered questions, into account. Table \ref{tab_coverage_comparison} shows the comparison between our evaluation and the results published by \citet{grave-etal-2018-learning}. \citet{bojanowski2016} did not publish these numbers.

Overall, the scores we obtained on the word analogy tasks \textsl{agree with the previously published results}. Our most relevant observations are:

\begin{itemize}

\item The scores show a \textsl{significant variance} over the 128 runs on shuffled corpora. This supports the argument, that every time a score on a word analogy dataset -- or any task that depends on word embeddings for that matter -- is published, it should be obtained by \textsl{averaging over a sufficient number of subsequent runs.} The current practice in research is to provide only one number, without any information on its variance and our data indicates that this is insufficient.

\item For most languages and techniques, our scores obtained are slightly higher than the ones reported previously, especially for the languages with comparatively little Wikipedia data (Hindi and Finnish). We attribute this improvement mainly to the fact, that we were able to use slightly larger corpora to train the embeddings, as new articles are written on Wikipedia daily while existing ones are edited and extended. 

\item Notable exceptions, i.e. cases, where our results are lower than previously published scores, are Chinese and English. For Chinese, we suspect a problem with the specific tokenization procedure to be the cause. In the case of English, we can only compare our scores to the ones published by \citet{bojanowski2016}, who used a different preprocessing that includes lowercasing the training data and did not report the coverage on the analogy tasks. Therefore, one cannot expect perfect accordance of the results.

\end{itemize}

\begin{table}[htbp]
\center
\input{tables/tab_ftt_baseline_comparison.tex}
\caption{Scores of our \ftt\ models (skip-gram) on the word analogy tasks for different languages compared to the results published by \citet{bojanowski2016} and \citet{grave-etal-2018-learning}. The results of this work, noted in the rightmost column, state the mean $\mu$, standard deviation $\sigma$, and highest score of 128 runs on independently shuffled corpora.}
\label{tab_ftt_baseline_comparison}
\end{table}

\begin{table}[htbp]
\center
\input{tables/tab_wtv_baseline_comparison.tex}
\caption{Scores of our \wtv\ models (skip-gram) on the word analogy tasks for different languages compared to the results published by \citet{bojanowski2016}. The results of this work, noted in the rightmost column, state the mean $\mu$, standard deviation $\sigma$, and highest score of 128 runs on independently shuffled corpora.}
\label{tab_wtv_baseline_comparison}
\end{table}

\begin{table}[htbp]
\center
\input{tables/tab_coverage_comparison.tex}
\caption{Coverage of our models on the word analogy tasks for different languages compared to the results published by \citet{grave-etal-2018-learning}.}
\label{tab_coverage_comparison}
\end{table}

\subsection{Historical Corpora}
As mentioned in the introduction, the second part of this work is focused on detecting and measuring \textsl{semantic change}, i.e. i.e. differences in the meaning of words between distinct time periods. In order to analyse these differences, historical corpora are required, i.e. at least two corpora in a given language, consisting of documents from separate epochs. 

All our experiments on semantic change are based on two datasets: First, the Corpus of Historical American English, or COHA \cite{davies2015}, a $400$ million word corpus comprising documents written in American English between 1810 and 2010. And second, the dataset provided for Task 1 of the 14th International Workshop on Semantic Evaluation, taking place in Barcelona in the fall of 2020 \cite{schlechtweg2020semeval}: This dataset consists of documents in four different languages -- English, German, Latin and Swedish -- and the documents in each language are split into two distinct sets based on their date of origin. We outline the specifics of the two datasets in more detail in the sections below.

\subsubsection{Corpus of Historical American English (COHA)}
\label{sec_COHA_corpus}

The Corpus of Historical American English contains 400 million words in more than 100,000 texts which date from the 1810s to the 2000s. The corpus contains texts from different genres, namely fiction, magazines newspapers and non-fiction books and is balanced by genre from decade to decade \cite{davies2015}. The distribution of the text size over the 20 decades from 1810 to 2009 is illustrated in Table \ref{tab_coha_corpus_numbers}.

\begin{table}[htbp]
\begin{center}
\input{tables/tab_coha_corpus_numbers.tex}
\caption{The Corpus of Historical American English (COHA) consists of documents from the 20 decades between 1810 and 2010 and comprises nearly 400 million words in total. The table shows the total number of tokens for each of the 20 decades, as well as the size of the respective vocabulary (any word with less than 5 appearances is discarded from the vocabulary).}
\label{tab_coha_corpus_numbers}
\end{center}
\end{table}

Because of its size, temporal range and robustness (genre-balanced, lemmatized), the corpus has been used regularly in previous work on semantic change \cite{hamilton2016,egger2016,kutuzov2018,tahmasebi2018}. We used the lemmatized version of the corpus and, hence, only applied minimal preprocessing -- removing punctuation and lowercasing all words.

\subsubsection{SemEval 2020 Task 1: Unsupervised Lexical Semantic Change Detection}
\label{sec_SEMEVAL_corpus}

Task 1 of the SemEval 2020 workshop involves the unsupervised detection of semantic change on diachronic corpora in four different languages: English, German, Latin and Swedish. The full problem definition is outlined in Section \ref{sec_semeval}. In this section, we focus exclusively on the corpora provided by the organizers for this task.

Table \ref{tab_semeval_corpus_numbers} illustrates the main properties of the corpora in the four different languages. The organizers did not compile the corpora from scratch but relied on existing ones: The English corpora are based on COHA (see Section \ref{sec_COHA_corpus}). The German data is a combination of three newspaper corpora (Deutsches Textarchiv, Berliner Zeitung and Neues Deutschland). For Latin, the LatinISE corpus is used \cite{latinISE} and for Swedish, the KubHist corpus \cite{kubhist}. We did not apply any specific preprocessing to the corpora.

\begin{table}[htbp]
\begin{center}
\input{tables/tab_semeval_corpus_numbers.tex}
\caption{Properties of the diachronic corpora in English, German, Latin and Swedish that were provided for Task 1 of the SemEval 2020 workshop. }
\label{tab_semeval_corpus_numbers}
\end{center}
\end{table} 

%% file: tables/tab_embedding_models.tex
\begin{tabular}{l|cc} 
Trained on 	& Count-Based 	& Prediction-Based \\
\hline 
Words 		& \glv\ (2014) 	& \wtv\ (2013) \\ 
Sub-Words 	&  				& \ftt\ (2016) \\ 
\end{tabular}

%% file: tikz/fig_skip_gram.tex
\begin{tikzpicture}


\node[text width = 1 cm, align = center]			at 	(-5.3,+0.0)	{(\texttt{sat})};
\node[text width = 1 cm, align = center] 			at 	(-4.0,+0.0)	{$\vec{v}_i(w_t)=$};
\node[draw, text width = 0.6 cm, align = center] 	at	(-2.5,+0.0)	{$0.32$ \\ $\vdots$ \\ $0.07$};


\node[text width = 1 cm, align = center] 			at	(+5.5,+3.5)	{(\texttt{the})};
\node[text width = 2 cm, align = center] 			at 	(+4.0,+3.5)	{$=\vec{v}_o(w_{t-2})$};
\node[draw, text width = 0.6 cm, align = center] 	at	(+2.5,+3.5)	{$0.12$ \\ $\vdots$ \\ $0.71$};

\node[text width = 1 cm, align = center] 			at	(+5.5,+1.5)	{(\texttt{cat})};
\node[text width = 2 cm, align = center] 			at 	(+4.0,+1.5)	{$=\vec{v}_o(w_{t-1})$};
\node[draw, text width = 0.6 cm, align = center] 	at	(+2.5,+1.5)	{$0.41$ \\ $\vdots$ \\ $0.17$};

\node[text width = 1 cm, align = center] 			at	(+5.5,-1.5)	{(\texttt{on})};
\node[text width = 2 cm, align = center] 			at 	(+4.0,-1.5)	{$=\vec{v}_o(w_{t+1})$};
\node[draw, text width = 0.6 cm, align = center] 	at	(+2.5,-1.5)	{$0.20$ \\ $\vdots$ \\ $0.09$};

\node[text width = 1 cm, align = center] 			at	(+5.5,-3.5)	{(\texttt{a})};
\node[text width = 2 cm, align = center] 			at 	(+4.0,-3.5)	{$=\vec{v}_o(w_{t+2})$};
\node[draw, text width = 0.6 cm, align = center] 	at	(+2.5,-3.5)	{$0.35$ \\ $\vdots$ \\ $0.21$};


\node[align = center] at (+0.2,+0.0)	{$\begin{aligned}\sum \log p\left(w_{t+j}\,|\,w_t\right)\end{aligned}$};


\draw[->, blue] 	(-1.9,+0.1) -- (-1.3,+0.1);
\draw[->, red] 		(-1.3,-0.1) -- (-1.9,-0.1);

\draw[->, blue] 	(+1.7,-1.3) -- (+1.0,-0.6);
\draw[->, red] 		(+0.7,-0.6) -- (+1.7,-1.6);

\draw[->, blue] 	(+1.2,-2.7) -- (-0.4,-1.1);
\draw[->, red] 		(-0.7,-1.1) -- (+1.2,-3.0);

\draw[->, blue] 	(+1.7,+1.3) -- (+1.0,+0.6);
\draw[->, red] 		(+0.7,+0.6) -- (+1.7,+1.6);

\draw[->, blue] 	(+1.2,+2.7) -- (-0.4,+1.1);
\draw[->, red] 		(-0.7,+1.1) -- (+1.2,+3.0);

\node[text width = 3 cm, align = left] at (-1.5,-2.5)	{\textsl{calculate} \\ \textsl{update}};
\draw[->, blue] 	(-3.9,-2.18) -- 	(-3.3,-2.18);
\draw[->, red]		(-3.9,-2.82) -- 	(-3.3,-2.82);
\end{tikzpicture}

%% file: tables/tab_embedding_languages.tex
\begin{tabular}{lccc}
\textbf{Language} & \textbf{Shortcut} & \textbf{Tokens} $\left[\times 10^6\right]$ &  \textbf{Vocabulary} $\left[\times 10^4\right]$ \\
\hline
Hindi 			& \textsc{Hi} & 48		& 19	\\
Finnish 		& \textsc{Fi} & 155		& 97	\\
Chinese 		& \textsc{Zh} & 215		& 96 	\\
Czech 			& \textsc{Cs} & 225		& 85 	\\
Polish 			& \textsc{Pl} & 469		& 137 	\\
Portuguese 		& \textsc{Pt} & 489		& 87 	\\
English 		& \textsc{En} & 4501	& 398	\\
\end{tabular}

%% file: tables/tab_sampling_methods.tex
\begin{tabular}{x{1.55cm}|p{6.35cm}|cc}
\textbf{Method} & \multicolumn{1}{c|}{\textbf{Description}} & \textbf{Run 1} & \textbf{Run 2}\\
\hline
\textbf{fixed} &
Documents are sampled in \textsl{fixed} order, the variability of the resulting word embeddings is a result of the inherent random processes of the respective technique.& 
$d_1\ d_2\ d_3$ & $d_1\ d_2\ d_3$ \\
\rule{0pt}{6ex} \textbf{shuffled} &
Documents are randomly \textsl{shuffled}, to measure the influence of the document order on the variability of the embeddings. &
$d_1\ d_3\ d_2$ & $d_2\ d_1\ d_3$ \\
\rule{0pt}{6ex} \textbf{bootstrapped} &
Documents are randomly sampled with replacement, to observe the variability due to the presence of individual documents.
& $d_3\ d_2\ d_3$ & $d_2\ d_2\ d_1$ \\
\end{tabular}

%% file: tables/tab_ftt_baseline_comparison.tex
\begin{tabular}{c|c|c|ccc}
\multirow{2}{*}{\textbf{Language}} & \multirow{2}{*}{\textbf{Bojanowski (2016)}} & \multirow{2}{*}{\textbf{Grave (2018)}} & \multicolumn{3}{c}{\textbf{This Work}} \\
&	&	& $\mu$	& $\sigma$	& \textsl{Best}	\\
\hline
\textsc{Hi} & -				& 10.6 			& $17.06$	& $0.46$	& $\textbf{18.24}$ 	\\
\textsc{Fi} & -				& 35.9 			& $42.84$	& $1.33$	& $\textbf{47.71}$ 	\\
\textsc{Zh} & -				& \textbf{60.2}	& $57.01$	& $1.25$	& $59.50$ 			\\
\textsc{Cs} & -				& 63.1		 	& $62.90$	& $0.55$	& $\textbf{64.36}$ 	\\
\textsc{Pl} & -				& 53.4 			& $58.16$	& $0.78$	& $\textbf{60.20}$ 	\\
\textsc{Pt} & -				& 54.0 			& $56.52$	& $0.42$	& $\textbf{57.67}$ 	\\
\textsc{En} & \textbf{76.2}	& -				& $74.21$	& $0.21$	& $74.83$			\\
\end{tabular}

%% file: tables/tab_wtv_baseline_comparison.tex
\begin{tabular}{c|c|ccc}
\multirow{2}{*}{\textbf{Language}} & \multirow{2}{*}{\textbf{Bojanowski (2016)}} & \multicolumn{3}{c}{\textbf{This Work}} \\
&	& $\mu$	& $\sigma$	& \textsl{Best}	\\
\hline
\textsc{Cs}		& 45.8 			& $48.57$	& $0.50$	& $\textbf{49.80}$ 	\\
\textsc{En} 	& \textbf{73.9} & $71.89$	& $0.20$	& $72.35$			\\
\end{tabular}

%% file: tables/tab_coverage_comparison.tex
\begin{tabular}{c|ccccccc}
						& \textsc{Hi} 	& \textsc{Fi} 	& \textsc{Zh} 	& \textsc{Cs} 	& \textsc{Pl} 	& \textsc{Pt}	& \textsc{En} 	\\
\hline
\textbf{Grave (2018)} 	& 70.8			&	94.6		&	100.0		&	76.9		&	69.5		&	79.2		&	-  			\\
\textbf{This Work}		& 72.0			&	94.6		&	96.6		&	83.2		&	70.3		&	79.2		&	97.5  		\\
\end{tabular}

%% file: tables/tab_coha_corpus_numbers.tex
\begin{tabular}{cx{2cm}x{2.2cm}}
\multirow{2}{*}{\textbf{Decade}}	& \textbf{Tokens}	& \textbf{Vocabulary} 	\\
& $\left[ \times 10^5 \right]$		& $\left[ \times 10^3 \right]$				\\
\hline
$1810 - 1819$	& $11$	& $12$ \\
$1820 - 1829$	& $65$	& $27$ \\
$1830 - 1839$	& $129$	& $39$ \\
$1840 - 1849$	& $150$	& $43$ \\
$1850 - 1859$	& $154$	& $43$ \\
$1860 - 1869$	& $157$	& $48$ \\
$1870 - 1879$	& $173$	& $47$ \\
$1880 - 1889$	& $188$	& $51$ \\
$1890 - 1899$	& $190$	& $53$ \\
$1900 - 1909$	& $253$	& $68$ \\
$1910 - 1919$	& $212$	& $56$ \\
$1920 - 1929$	& $238$	& $63$ \\
$1930 - 1939$	& $229$	& $63$ \\
$1940 - 1949$	& $227$	& $64$ \\
$1950 - 1959$	& $229$	& $67$ \\
$1960 - 1969$	& $223$	& $67$ \\
$1970 - 1979$	& $221$	& $68$ \\
$1980 - 1989$	& $234$	& $75$ \\
$1990 - 1999$	& $327$	& $98$ \\
$2000 - 2009$	& $275$	& $83$ \\
\end{tabular}

%% file: tables/tab_semeval_corpus_numbers.tex
\begin{tabular}{l|cx{2cm}x{2.2cm}}
\multirow{2}{*}{\textbf{Language}} & \multirow{2}{*}{\textbf{Time Period}}
& \textbf{Tokens}	& \textbf{Vocabulary}								 		\\
&&	$\left[ \times 10^6 \right]$ & $\left[ \times 10^3 \right]$				\\
\hline
\multirow{2}{*}{English} 	& $t_1=1810 - 1860$ 		& $65$		& $23$ 	\\
									& $t_2=1960 - 2010$ 		& $66$		& $33$ 	\\
\hdashline[0.3pt/2.2pt]
\multirow{2}{*}{German} 	& $t_1=1800 - 1899$ 		& $690$		& $219$ \\
									& $t_2=1946 - 1990$ 		& $697$		& $265$ \\
\hdashline[0.3pt/2.2pt]
\multirow{2}{*}{Latin} 	& $t_1=200\,\text{BC} - 0$ 	& $17$		& $14$ 	\\
									& $t_2=0 - 2000$ 			& $91$		& $50$ 	\\
\hdashline[0.3pt/2.2pt]
\multirow{2}{*}{Swedish} 	& $t_1=1790 - 1830$ 		& $671$		& $278$	\\
									& $t_2=1895 - 1903$ 		& $1086$	& $251$	\\
\end{tabular}

%% file: chapters/2_instability.tex
\chapter{The Stability of Word Embeddings}
\label{chp_measuring}

A glance into Table \ref{tab_ftt_baseline_comparison} should suffice to explain why understanding and quantifying the random nature of word embeddings is a pressing matter for research in NLP: Two word embedding models, trained with the same embedding technique and default parameters on basically identical, but independently shuffled Wikipedia corpora, can yield entirely different scores on the standard word analogy task of the respective language. Table \ref{tab_ftt_baseline_differences} shows the full extent of this problem -- the relative differences of the score of seemingly identical models can exceed $20\%$. 

\begin{table}[htbp]
\center
\input{tables/tab_ftt_baseline_differences.tex}
\caption{Lowest and highest scores on the word analogy tasks for different languages observed in 128 subsequent runs of \ftt\ trained on independently shuffled Wikipedia corpora as outlined in Section \ref{sec_exp_setup}.}
\label{tab_ftt_baseline_differences}
\end{table}

In the last few years, these scores were used on various occasions to argue that one embedding technique is superior to another \cite{mikolov2013c, pennington-etal-2014-glove, levy2014, bojanowski2016}. However, in none of the work we know of, could we find more than one result on these tasks, obtained from the subsequent training of multiple models. This raises serious doubts on the significance of some of these results, and we recommend for any future research to report the mean and variance of the score over -- at least five -- subsequent runs.

Furthermore, since the largest value of word embeddings for NLP does not lie within the embeddings themselves, but in their use for various downstream tasks, this problem is amplified: It is hard to estimate the influence of the variability of word embeddings on the performance on these downstream tasks. This makes it even more important to understand why and how unstable word embeddings are.

The origin of the randomness within the process of training word embeddings seems to be well understood: \citet{yin2018} pointed out that all popular techniques for non-contextualized word embeddings can be formulated as either \textsl{implicit or explicit matrix factorization}, i.e. the low-rank matrix approximation of a signal matrix. \textbf{LSA} \cite{deerwester1990,Landauer1997AST} and \textbf{PPMI} \cite{levy2014} are examples for explicit matrix factorization, whereas \glv\ \cite{pennington-etal-2014-glove}, just like skip-gram based methods, e.g. \wtv\ \cite{mikolov2013c} and \ftt\ \cite{bojanowski2016} have been shown to implicitly perform matrix factorization. Whereas the signal matrix can in principle be unambiguously obtained from the corpus, the embedding space constructed by any of the techniques presented above, is only an \textsl{approximation of the semantic information captured in the corpus}.

Thus, any embedding space provides a somewhat distorted view on the semantics of the corpus it was derived from \cite{hellrich-hahn-2017-fool}. In practice, this distortion is the result of several random processes: First and foremost, the random initialization of an embedding vector for every word in the vocabulary. Second, the order in which the documents are processed during training (this does not apply to count-based techniques). And finally, the random sub-sampling of frequent words, which could technically be omitted, but is very common in practice.

We could replace these random processes by deterministic alternatives; however, this would only replace the random distortion by a fixed one, thus creating a false sense of reliability.

In the following section, we study the influence of these random processes, or in other words, the \textsl{instability of word embedding spaces}. This instability is measured by the \textsl{variability of the embedding spaces}, that are derived in independent runs of the same technique on the same corpus.

\section{The Random Nature of Word Embedding Techniques}
\label{sec_quantifying_stability}

The random nature of the creation of word embeddings implies, that any time an embedding technique $\mathcal{T}$ with a specific set of parameters is applied to a corpus $\mathcal{C}$, an embedding space $\mathbf{V}_i\in\mathbb{R}^{v\times d}$ (where $v$ is the size of the vocabulary $\mathcal{V}$ and $d$ the dimension of the embeddings) is sampled from a probability distribution $\Omega$:
\begin{align}
\mathbf{V}_i\sim \Omega(\mathcal{T},\mathcal{C})
\end{align}
We define instability as the variability of independently obtained embeddings of the same technique on the same corpus.

Now, we assume to have a \textsl{well-defined distance metric} $\mathbf{d}$ which, given a word $w$ (e.g. \texttt{cat}) as well as two embedding spaces $\mathbf{V}_i$ and $\mathbf{V}_j$ returns a measure of the difference between the embedding of $w$ in the two spaces $\mathbf{V}_i$ and $\mathbf{V}_j$. Then, the instability $\mathcal{I}$ of the embedding of $w$ in the distribution $\Omega$ can be written as the average of $\mathbf{d}(w,\mathbf{V}_i,\mathbf{V}_j)$ over an infinite number of pairs $\mathbf{V}_i,\mathbf{V}_j$ sampled from $\Omega(\mathcal{T},\mathcal{C})$:
\begin{align}
\mathcal{I}(w,\Omega) = \lim_{N\to\infty} \frac{2}{N(N+1)} \sum_{i\neq j=0}^{N} \mathbf{d}(w,\mathbf{V}_i,\mathbf{V}_j) \quad \text{with} \quad \mathbf{V}_i,\mathbf{V}_j\sim \Omega(\mathcal{T},\mathcal{C})
\end{align}

One could estimate this instability $\mathcal{I}(w,\Omega)$ in practice by drawing a sufficient number of samples $\mathbf{V}_i$ from $\Omega$ (i.e. by applying the embedding technique $\mathcal{T}$ to the corpus $\mathcal{C}$ multiple times), provided a metric $\mathbf{d}$ as specified above, that measures the distance between two embeddings of the same word in different embedding spaces.

\subsection{Random Orientation of Embedding Spaces}
\label{sec_random_orientation_of_embedding_spaces}

However, the inherent characteristics of word embeddings make it difficult to capture this distance: It is widely accepted today, that non-contextualized word embeddings are essentially invariant under rotations \cite{hamilton2016,artetxe2016,smith2017,yin2018}. This means, that two embedding spaces are essentially identical -- and can be substituted for one another in any practical application -- if one can be obtained from the other by applying an orthogonal transformation $\mathbf{A}\in\mathbb{R}^{d\times d}: \mathbf{A}\mathbf{A}^\top = \mathbf{I}$.

For example, applying an orthogonal transformation $\mathbf{A}\in\mathbb{R}^{d\times d}$ to an embedding space does not influence the cosine similarity of any two embeddings $\vec{u},\vec{v}$ which we denote as row vectors of size $d$:
\begin{equation}
\begin{aligned}
\cos\left(\angle (\vec{u}\mathbf{A},\vec{v}\mathbf{A})\right) &= \frac{(\vec{u}\mathbf{A})(\vec{v}\mathbf{A})^\top}{\left((\vec{u}\mathbf{A})(\vec{u}\mathbf{A})^\top\right)\cdot\left((\vec{v}\mathbf{A})(\vec{v}\mathbf{A})^\top\right)} \\
&= \frac{\vec{u}\mathbf{A}\mathbf{A}^\top\vec{v}^\top}{\left(\vec{u}\mathbf{A}\mathbf{A}^\top\vec{u}^\top\right)\cdot\left(\vec{v}\mathbf{A}\mathbf{A}^\top\vec{v}^\top\right)}\\
&= \frac{\vec{u}\vec{v}^\top}{\left(\vec{u}\vec{u}^\top\right)\cdot\left(\vec{v}\vec{v}^\top\right)} = \cos\left(\angle (\vec{u},\vec{v})\right)
\end{aligned}
\label{eq_rotational_invariance_of_embeddings}
\end{equation}

\begin{figure}[htbp]
\center
\input{tikz/fig_random_orientation_of_embeddings.tex}
\caption{Two-dimensional illustration of the random orientation of an embedding space over two consecutive runs of the same technique on the same corpus.}
\label{fig_random_orientation_of_embeddings}
\end{figure}
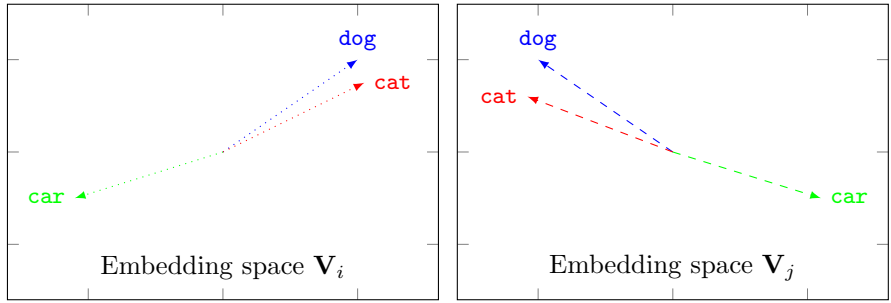

In practice, we observe that this rotation-invariance in combination with the random initialization means that every time an embedding space $\mathbf{V}_i$ is sampled from the distribution $\Omega(\mathcal{T},\mathcal{C})$ (i.e. on every run), \textsl{it is randomly oriented} (see Figure \ref{fig_random_orientation_of_embeddings}).

This is the reason why finding a metric $\mathbf{d}(w,\mathbf{V}_i,\mathbf{V}_j)$ to measure the distance of the embeddings of one word $w$ within two different spaces $\mathbf{V}_i$ and $\mathbf{V}_j$ is not trivial: Simply applying a $\mathbb{R}^n$ metric, like the \textsl{Euclidean distance} or the \textsl{cosine similarity} to the two different embeddings of $w$, namely $\vec{v}_i(w)$ and $\vec{v}_j(w)$ would not yield any meaningful result, because of the random orientation illustrated above.

\subsection{Rotation-Invariant Quantities}
\label{sec_invariant_quantities}

Therefore, it is no surprise that all previous work on quantifying stability is ultimately based on \textsl{rotation-invariant quantities} of the embedding spaces, i.e. quantities which remain unchanged under an orthogonal transformation \citep{hellrich-hahn-2016-assessment,hellrich-hahn-2016-bad,hellrich-hahn-2017-fool,hellrich-etal-2019-influence,antoniak2018,Chugh2018StabilityOW,wendlandt2018,pierrejean-tanguy-2018-predicting}. We already demonstrated that the cosine similarity of two embeddings $\vec{u}$, $\vec{v}$ is such a quantity -- see Equation (\ref{eq_rotational_invariance_of_embeddings}). However, it is important to be clear that the embeddings themselves are \textsl{not rotation-invariant} as $\vec{u}\neq\vec{u}\mathbf{A}$ for most $\vec{u}\in \mathbb{R}^d$ and $\mathbf{A}\in\mathbb{R}^{v\times d}$ orthogonal. 

A rotation-invariant quantity will \textsl{not be influenced by the random orientation} of the embedding space on every run. Therefore, it is expected to be consistent over multiple runs, i.e. to be constant apart from the anticipated random deviations. This allows us to compare the quantity over multiple runs and to use it to study the stability of the underlying technique.

\subsection{Distribution of the Cosine Similarity of Two Arbitrary Embeddings}
\label{sec_cosine_similarity_gaussian}

To get a better understanding of the inherent randomness of the process of constructing word embeddings from a corpus $\mathcal{C}$ we examine the distribution of a rotation-invariant quantity, namely the cosine similarity of the embeddings $\vec{v}_i$ and $\vec{v}_j$ of two arbitrary words $w_i$ and $w_j$ respectively. 

The cosine similarity of the embeddings of the two words \texttt{cat} and \texttt{dog} for the 128 runs of \wtv\ we conducted on independently shuffled versions of the English Wikipedia is illustrated below:
\begin{align*}
&\text{First run:}				& \cos\left[\angle \left(\vec{v}_1(\texttt{cat}),\vec{v}_1(\texttt{dog})\right)\right]			&=0.6805 \\
&\text{Second run:} 				& \cos\left[\angle \left(\vec{v}_2(\texttt{cat}),\vec{v}_2(\texttt{dog})\right)\right]			&=0.6840 \\
&\text{Third run:} 				& \cos\left[\angle \left(\vec{v}_3(\texttt{cat}),\vec{v}_3(\texttt{dog})\right)\right]			&=0.6782 \\
&\quad \quad ... \quad				& ... \quad \quad \quad \quad 																	& \\
&128^{\text{th}}\text{ run:} 		& \cos\left[\angle \left(\vec{v}_{128}(\texttt{cat}),\vec{v}_{128}(\texttt{dog})\right)\right]	&=0.6837
\end{align*}
These handful of results seem in line with our expectation, that the rotation-invariant cosine similarity will deviate around a particular value for the individual runs. We are interested in how this distribution looks like in more detail, i.e., if any particular shape can be recognized. Figure \ref{fig_gauss_cat_dog} shows a histogram of the distribution of $\cos\left[\angle \left(\vec{v}_i(\texttt{cat}),\vec{v}_i(\texttt{dog})\right)\right]$ for the 128 runs we performed. 

\begin{figure}[htbp]
\center
\input{"./plots/gauss_cat_dog.pgf"}
\vspace{-0.7cm}
\caption{Distribution of the cosine similarity $\cos\left[\angle \left(\vec{v}_i(\texttt{cat}),\vec{v}_i(\texttt{dog})\right)\right]$ of the embeddings of the two words \texttt{cat} and \texttt{dog} for 128 runs of \wtv\ on independently shuffled versions of the English Wikipedia.}
\label{fig_gauss_cat_dog}
\end{figure}

Given the comparably small size of 128 samples, we cannot draw any immediate conclusions on the shape of the underlying distribution, and producing more samples is very resource-intensive, as this requires a full run of \wtv\ on the English Wikipedia corpus, which -- at the time of our experiments, in the fall of 2019 -- was equivalent to around 23 GB of raw text. 

However, we can efficiently compute the same distributions for many word pairs, other than \texttt{cat} and \texttt{dog}, from the existing data. Visual inspection of these distributions leads us to two work hypotheses, that we examine with a statistical analysis below:

\begin{enumerate}

\item The cosine similarities $\cos\left[\angle \left(\vec{v}_i(w_1),\vec{v}_i(w_2)\right)\right]$ and $\cos\left[\angle \left(\vec{v}_i(w_3),\vec{v}_i(w_4)\right)\right]$ of any two non-overlapping pairs of words $(w_1,w_2)$ and $(w_3,w_4)$ are statistically independent of each other.

\item The cosine similarity $\cos\left[\angle \left(\vec{v}_i(w_1),\vec{v}_i(w_2)\right)\right]$ of any word pair follows a Normal distribution.

\end{enumerate}

\begin{figure}[htbp]
\center
\vspace{-0.6cm}
\hspace*{-0.5cm}\input{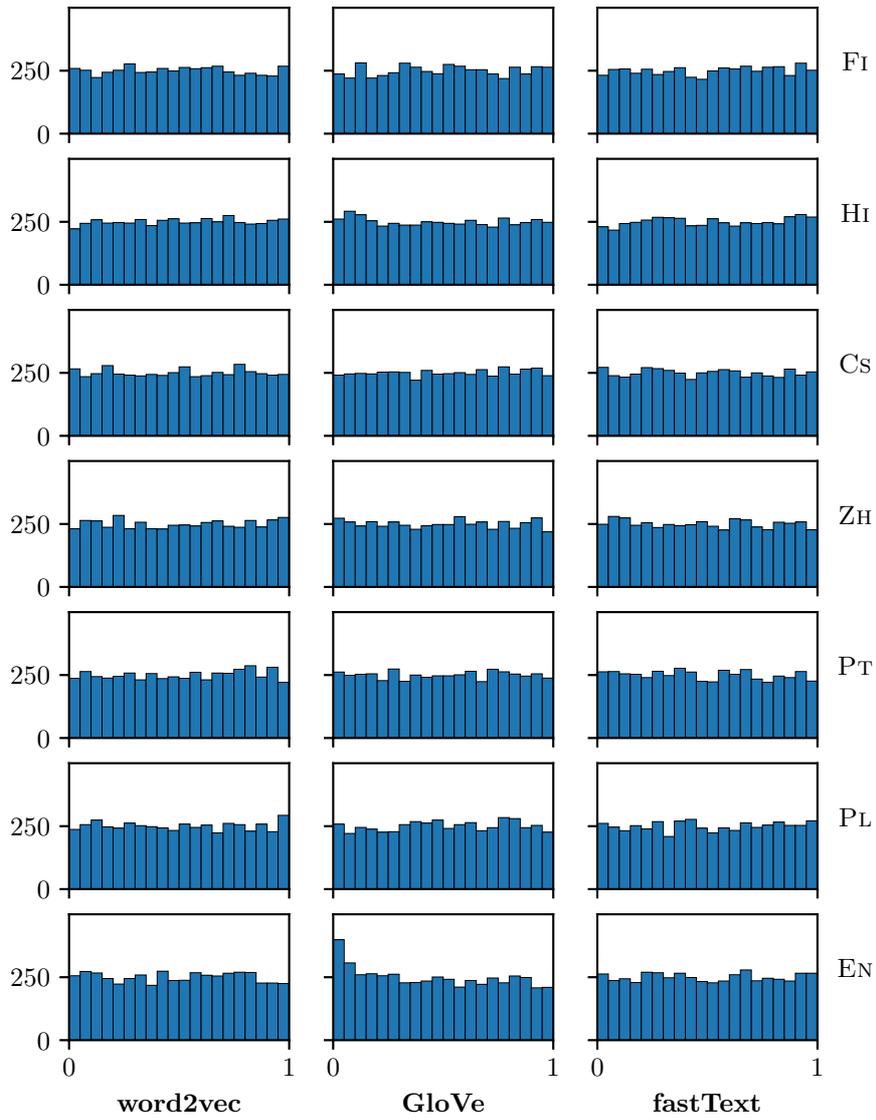}
\vspace{-1.8cm}
\caption{Distribution of significance values ($p$) of the Spearman rank correlation test on the distribution for 5,000 independent pairs of cosine similarity values over 128 runs of \ftt\ (skip-gram), \glv\ and \wtv\ (skip-gram) in the languages outlined in Section \ref{sec_exp_setup}.}
\label{fig_p_values_independence}
\end{figure}

To examine these hypotheses, we randomly sample 20,000 words $w_1,w_2,...$ from the vocabulary of our models in any language and compute the cosine similarity for the 10,000 word pairs $(w_1,w_2)$,$(w_3,w_4)$, etc. As we are not aware of a way to \textsl{prove} statistical independence in this situation, we examine a closely related measure to verify the first hypothesis: The correlation of the cosine similarity values for independent word pairs. Out of the 10,000 word pairs, we can construct 5,000 pairs of word pairs, and measure the significance value (or $p$-value) of the Spearman rank correlation test \cite{spearman1904} between the 128 samples of the cosine similarity of each pair. If the null hypothesis is correct, and the cosine similarity values for the different word pairs are not related to each other, we would expect a uniform distribution of the $p$-values in the interval $[0,1]$. Figure \ref{fig_p_values_independence} shows the distribution of the $p$-values for all languages and techniques described in Section \ref{sec_exp_setup}. For every configuration, we observe a homogeneous distribution, which is a strong indicator that the null hypothesis is correct, and the \textsl{cosine similarities of arbitrary word pairs are independent of each other.}

\begin{figure}[htbp]
\center
\vspace{-0.6cm}
\hspace*{-0.5cm}\input{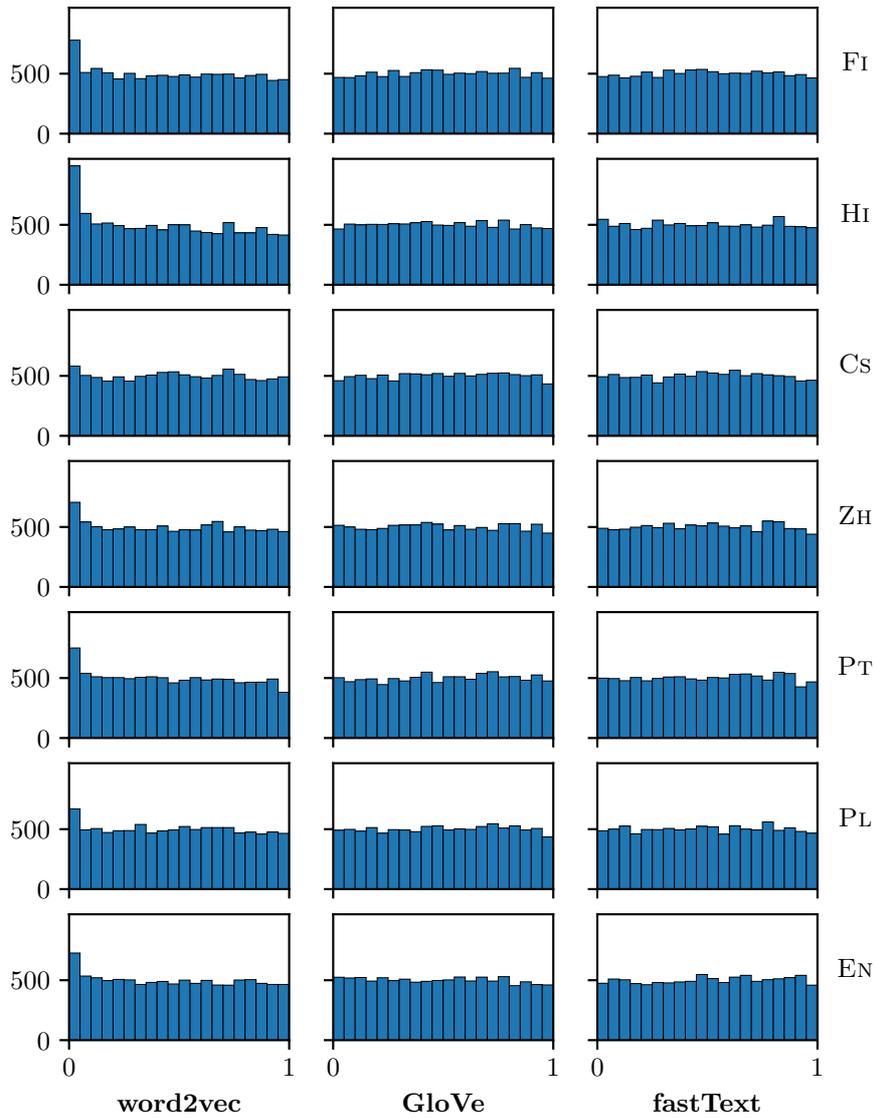}
\vspace{-1.8cm}
\caption{Distribution of significance values ($p$) of the Shapiro-Wilk-Test on the distribution of the cosine similarity for 10,000 independent word pairs over 128 runs of \ftt\ (skip-gram), \glv\ and \wtv\ (skip-gram) in the languages outlined in Section \ref{sec_exp_setup}.}
\label{fig_p_values_shapiro}
\end{figure}

Assuming the first hypothesis is valid, we can examine the second hypothesis with a statistical test on the normality of a given distribution: The Shapiro-Wilk test \cite{shapiro1965}. We calculate the significance of this test over the 128 samples for each of the 10,000 word pairs. If the null hypothesis is correct, and the cosine similarity values for the different word pairs are normally distributed, one would again expect a uniform distribution of the $p$-values in the interval $[0,1]$. Hence, the results in Figure \ref{fig_p_values_shapiro} are a strong indicator that the cosine similarity values for any word pair for most languages and embeddings techniques follow a Normal distribution. The leftmost column of Figure \ref{fig_p_values_shapiro} shows why we need to be careful with an absolute statement: The \wtv\ models, especially for Hindi and Finnish, show an accumulation of significance values in the leftmost interval, which implies a deviation from the normal distribution. Nevertheless, we can assume both statements above to hold in nearly all practical situations, as shown on multiple occasions in the sections below.

\begin{figure}[htbp]
\center
\input{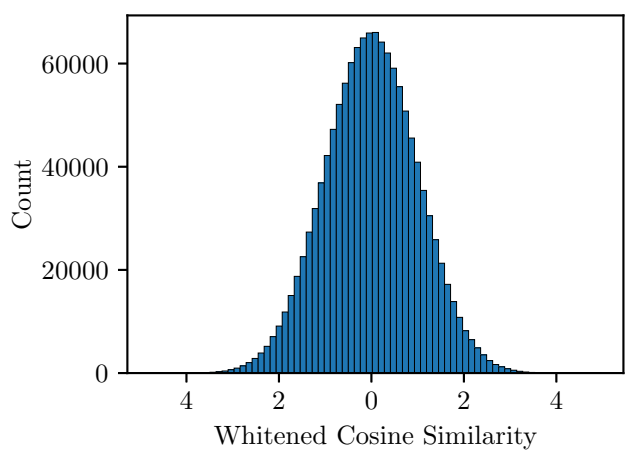}
\vspace{-0.7cm}
\caption{Accumulated distribution of the cosine similarity over 10,000 word pairs for 128 runs of \glv\ on independently shuffled versions of the Polish Wikipedia. The samples are ``whitened'' before accumulating them, i.e. the distributions for each of the 10,000 word pairs are shifted so that $\mu=0$ and stretched until $\sigma^2=1$.}
\label{fig_gauss_accumulated}
\end{figure}

Finally, one can also obtain an illustration of the shape of the distribution of the cosine similarity values, by accumulating samples over different word pairs. This should not be seen as a mathematical proof, but rather as a graphical representation of our findings: Figure \ref{fig_gauss_accumulated} shows this accumulation takes the shape of the well-known Gaussian bell curve.

\section[Measuring Distances Between Embedding Spaces]{Measuring Distances Between Embedding\\ Spaces}
\label{sec_measuring_distances}

As outlined above, we metric that captures the distance of the embedding of a word between two embedding spaces, to quantify the instability of embedding techniques. In this section, we compare several established and novel approaches for this metric and decide on the one that is best suited for the task at hand.

\subsection{Requirements for a Distance Metric}
\label{sec_requirements}

Before we look at a selection of different approaches for this distance metric, we want to establish a framework to evaluate and compare these approaches, i.e. a set of requirements that a metric should fulfil.

\begin{enumerate}[(I)]

\item\textbf{Formal Criteria}
The metric $\mathbf{d}$ needs to assign a distance, i.e. a real number to the difference of the embeddings of a word $w$ in the two spaces $\mathbf{V}_i$ and $\mathbf{V}_j$: 
\begin{align}
\mathbf{d}:\mathcal{V}\times \mathbb{R}^{v\times d} \times \mathbb{R}^{v\times d} \to [0,1], \quad (w,\mathbf{V}_i, \mathbf{V}_j) \mapsto \mathbf{d}(w,\mathbf{V}_i, \mathbf{V}_j)
\end{align}
We demand that the codomain of the metric $\mathbf{d}$ is limited to the interval $[0,1]$, since this allows us to transform any distance metric $\mathbf{d}$ into a corresponding \textsl{similarity metric} $\mathbf{s_d}$ -- and vice versa -- with:
\begin{align}
\begin{split}
\mathbf{s_d}&:\mathcal{V}\times \mathbb{R}^{v\times d} \times \mathbb{R}^{v\times d} \to [0,1], \quad (w,\mathbf{V}_i, \mathbf{V}_j) \mapsto \mathbf{s_d}(w,\mathbf{V}_i, \mathbf{V}_j) \\
\mathbf{s_d}&(w,\mathbf{V}_i, \mathbf{V}_j)=1-\mathbf{d}(w,\mathbf{V}_i, \mathbf{V}_j)
\end{split}
\end{align}

\item\textbf{Consistency}
Some of the metrics introduced below have a free parameter $f\in M$ and there is no undisputed ``right'' choice for the value of this parameter. This means that the approach does not define one specific metric $\mathbf{d}$, but describes a class of metrics, comprising all possible choices of the free parameter:
\begin{align}
\{\mathbf{d}_f\ |\ f\in M\}
\end{align}
This is not a problem per se, but if this is the case, we expect \textsl{the results based on different choices of the free parameter to be consistent} with one another. This means, if we have two words $w_1$ and $w_2$ and two embedding spaces $\mathbf{V}_1$ and $\mathbf{V}_2$ and the metric $\mathbf{d}_{f_1}$ based on the choice $f_1$ for the free parameter, yields:
\begin{align}
\mathbf{d}_{f_1}(w_1,\mathbf{V}_1,\mathbf{V}_2) > \mathbf{d}_{f_1}(w_2,\mathbf{V}_1,\mathbf{V}_2)
\end{align}
Then, we would expect the metric $\mathbf{d}_{f_2}$ with $f_2\neq f_1$ to give a consistent result:
\begin{align}
\mathbf{d}_{f_2}(w_1,\mathbf{V}_1,\mathbf{V}_2) > \mathbf{d}_{f_2}(w_2,\mathbf{V}_1,\mathbf{V}_2)
\end{align}
In practice, we can check how well this requirement is fulfilled by calculating $\mathbf{d}_{f}(w,\mathbf{V}_i,\mathbf{V}_j)
$ for a set of randomly sampled words $w$ and determining how well the order of the results is preserved over different values of $f$.

\item\textbf{Independence} Based on the observation, that several different approaches were already proposed for the task at hand, each with its strengths and weaknesses, it should be clear that there is no straightforward solution to measuring the distance between the embeddings of a word in two different spaces. Therefore, we may need to content ourselves with measuring a different quantity, that is somehow related to the distance. In this case, we want the measured quantity to be \textsl{largely independent of other factors} that are proven to be unrelated to the distance. 

\end{enumerate}

We do not claim that list of requirements is exhaustive, but it nonetheless constitutes a helpful framework that we apply to compare different approaches in the sections below.

\subsection{Nearest-Neighbor Based Approaches}
\label{sec_previous_approaches}

As mentioned above, all previous work on the instability of word embeddings is based on rotation-invariant quantities, and exclusively on the cosine similarity of two embeddings: The stability of the embedding of a specific target word $w$ is quantified by \textsl{comparing the list of $n$ nearest neighbors} of the target word (i.e. embeddings with the largest cosine similarity), over multiple runs (typically with $5\leq n \leq 25$).

Two slightly different variants of how exactly the stability is calculated were proposed in the past: Firstly, by dividing the size of the overlap of the two lists with $n$ \citep{hellrich-hahn-2017-fool,wendlandt2018,pierrejean-tanguy-2018-predicting}. We refer to this quantity as $p_{\text{@}n}$. Secondly, by calculating the Jaccard coefficient \citep{jaccard-1912} of the two lists \citep{hellrich-hahn-2016-assessment,hellrich-hahn-2016-bad,hellrich-etal-2019-influence,antoniak2018,Chugh2018StabilityOW}, which we denote as $j_{\text{@}n}$. The principle is illustrated in Table \ref{tab_momentum_nn}. 

\begin{table}[htbp]
\center
\input{tables/tab_momentum_nn.tex}
\captionsetup{singlelinecheck=off}
\caption[.]{
Most similar words to the target word \texttt{momentum} for two independent runs of \wtv\ (skip-gram, default parameters) trained on the English Wikipedia, with a reduced vocabulary size of 200,000 words. \\In the example above, eight out of the ten most similar words from run 1 are again found in the top ten of run 2: Only \texttt{angular} and \texttt{relativistic} are dropped. This yields:
\begin{equation}
p_{\text{@}10} = \frac{8}{10} = 0.8 \quad \quad \quad j_{\text{@}10} = \frac{8}{12} \approx 0.667
\end{equation}
As soon as we extend the scope of the comparison to the fifteen most similar words, \texttt{angular} is again found in the list of both runs. But in addition to \texttt{relativistic}, the three words \texttt{eigenstate}, \texttt{spin} and \texttt{accelerating} are not found in the top fifteen of run 2, which means:
\begin{equation}
p_{\text{@}15} = \frac{11}{15} \approx 0.733 \quad \quad \quad j_{\text{@}15} = \frac{11}{19} \approx 0.579
\end{equation}
}
\label{tab_momentum_nn}
\end{table}

The two metrics are rather similar and can be converted into each other. To illustrate this, let $m$ denote the number of items that two lists of length $n$ have in common. Then:
\begin{align}
\begin{split}
p_{\text{@}n} &= \frac{m}{n} \iff m = n\cdot p_{\text{@}n} \\
j_{\text{@}n} &= \frac{m}{n + (n-m)} = \frac{m}{2n - m} = \frac{ n\cdot p_{\text{@}n}}{2 n - n\cdot p_{\text{@}n}} = \frac{p_{\text{@}n}}{2-p_{\text{@}n}}
\end{split}
\end{align}
Now we examine if these metrics fulfil the criteria introduced above.

\subsubsection*{(I) Formal Criteria}

Both metrics $p_{\text{@}n}$ and $j_{\text{@}n}$ measure the similarity of the embedding of a word $w$ between two embedding spaces $\mathbf{V}_i$ and $\mathbf{V}_j$. The minimum and maximum values the two metrics can assume are $0$ and $1$ respectively. Therefore, both fulfil the formal criteria for a \textsl{similarity metric} $\mathbf{s_d}$ as defined above, which allows us to transform them into a corresponding distance metric $\mathbf{d}$, with:
\begin{align}
\mathbf{d}(w,\mathbf{V}_i,\mathbf{V}_j)=1-\mathbf{s_d}(w,\mathbf{V}_i,\mathbf{V}_j)
\end{align} 
which fulfil all formal criteria that we introduced above.

\subsubsection*{(II) Consistency} 
\label{sec_nn_consistency}

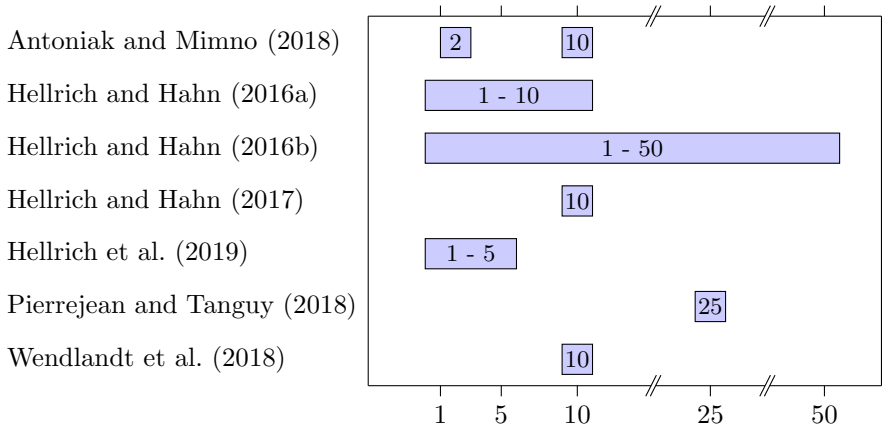
\begin{figure}[htbp]
\center
\input{tikz/fig_previous_work_values_n.tex}
\caption{Illustration of the different values for $n$, the number of nearest neighbors of a target word, which were used in previous work to evaluate the stability of word embeddings with the metrics $p_{\text{@}n}$ and $j_{\text{@}n}$.}
\label{fig_previous_work_values_n}
\end{figure}

For both metrics that were previously used to measure the stability of word embeddings, $p_{\text{@}n}$ and $j_{\text{@}n}$, one must pick an arbitrary value for $n$, i.e. the number of nearest neighbors, that are compared over subsequent runs. So far, there is no consensus on a value of $n$ which is best suited for the task, as Figure \ref{fig_previous_work_values_n} illustrates.

We analyse the consistency of the metrics over different values of $n$, for \wtv, \glv, and \ftt\ embeddings trained on 16 independently shuffled versions of Wikipedia corpora in different languages as outlined in Section \ref{sec_exp_setup}:

\begin{enumerate}
\itemsep0em

\item For each language, 1000 target words are randomly sampled from the joint vocabulary of all runs in the respective language.

\item For each of the runs, the 50 nearest neighbors by cosine distance are calculated for every target word.

\item Finally, the average of $p_{\text{@}n}$ and $j_{\text{@}n}$ for every target word at $n\in\{2,5,10,25,50\}$ is calculated over the 120 pairs that can be constructed out of the 16 runs taken into consideration for every language and technique. This reduces random fluctuations and allows to draw conclusions on the underlying distribution,\footnote{We show in Appendix \ref{app_average_size_suffices} that the sample size of 16 subsequent runs is sufficient.} i.e. the mean values of $p_{\text{@}n}$ and $j_{\text{@}n}$ for every target word.

\end{enumerate} 

If $p_{\text{@}n}$ and $j_{\text{@}n}$ were to be considered \textsl{consistent over the free parameter $n$}, we would expect that a target word, which is identified as comparably stable based on $n=5$, also ranks among the more stable words for $n=50$. Table \ref{tab_dependance_of_pj_on_n} shows the Spearman correlation \cite{spearman1904} of $p_{\text{@}n}$ and $j_{\text{@}n}$ for the 1000 target words between different values of $n$. We generally observe values significantly smaller than $1$ and in some cases even less than $0.5$, which indicates only a loose correlation between the metrics at different values of $n$. 

Let us summarize the first problem we identified with using the nearest-neighbor based metrics $p_{\text{@}n}$ and $j_{\text{@}n}$ to capture the stability of word embeddings: The metrics are inconsistent over the free variable $n$, namely the number of nearest neighbors that are evaluated, and so far there is no consensus on a particularly suitable value for $n$. This leaves us with large uncertainties when using the metrics.

\begin{table}[htbp]
\begin{center}
\input{tables/tab_dependance_of_pj_on_n.tex}
\caption{Spearman correlation of the metrics $p_{\text{@}n}$ and $j_{\text{@}n}$ for 1000 target words for different values of $n\in \{ 2,5,10,25,50 \}$ for \wtv, \glv, and \ftt, obtained as outlined in Section \ref{sec_nn_consistency}. For each of the techniques, we show the average of the correlation for all languages mentioned in Section \ref{sec_exp_setup}.}
\label{tab_dependance_of_pj_on_n}
\end{center}
\end{table}

\subsubsection*{(III) Independence} 
To determine if the metrics $p_{\text{@}n}$ and $j_{\text{@}n}$ are influenced by any quantities which are proven to be unrelated to the distance of the embeddings, we first need to understand what the metrics exactly capture. Therefore, we \textsl{attempt to predict the measurements of $p_{\text{@}n}$ and $j_{\text{@}n}$} for randomly sampled target words from different corpora $\mathcal{C}$ and various embedding techniques $\mathcal{T}$, only based on a few assumptions about the underlying probability distribution $\mathbf{V}_i\sim \Omega(\mathcal{T}, \mathcal{C})$ of the embedding spaces $\mathbf{V}_i$ (please refer to Section \ref{sec_quantifying_stability} for more details).\\

As outlined in Section \ref{sec_cosine_similarity_gaussian}, the cosine similarity of the embeddings of any two words $w_1$ and $w_2$ in the embedding space $\mathbf{V}_i$ -- that we denote as $\cos(w_1,w_2)_{i}$ -- follows a distinct probability distribution $\Psi(\mathcal{T},\mathcal{C},w_1,w_2)$. The shape of this distribution is close to a \textsl{Normal distribution}:\footnote{Since the values of the cosine similarity of two vectors are limited to the interval $[0,1]$ whereas the Normal distribution is non-zero the real axis, this cannot be a true equality. However, we found that it is a good approximation for all practical purposes.}
\begin{align}
\cos(w_1,w_2)_{i} \sim \Psi(\mathcal{T},\mathcal{C},w_1,w_2) \approx \mathcal{N}\left(\mu_{12},\sigma_{12}^2 \right)
\end{align}
Now we assume to know the parameters of this distribution, namely the mean and standard deviation for a specific target word $w_t$ with every other word in the vocabulary:
\begin{align}
\mathcal{P}(w_t)=:\{(\mu_{ts},\sigma_{ts})\ | \ w_s\in \mathcal{V}\}
\end{align}

In practice, we can estimate these parameters by sampling a set of embedding spaces $\{\mathbf{V}_i\text{ for }i=1,...,r\}$ from the distribution $\Omega(\mathcal{T},\mathcal{C})$, i.e. applying the same embedding technique $\mathcal{T}$ to the corpus $\mathcal{C}$ subsequently for $r$ times. Then, we measure $\cos(w_t,w_s)_{i}$ for each run and finally use the formulas below to obtain the maximum-likelihood estimation of the parameters of the underlying Normal distribution:
\begin{align}
\mu_{ts} = \frac{1}{r} \sum_{i=1}^{r} \cos(w_t,w_s)_i \quad \quad \quad \sigma_{ts} = \sqrt{\frac{1}{r} \sum_{i=1}^{r} \left[\cos(w_t,w_s)_i-\mu_{ts}\right]^2}
\end{align}
An excerpt of an estimation of $\mathcal{P}(w_t)$ for an exemplary target word $w_t$ is shown in Table \ref{tab_gaussian_parameters}.

\begin{table}[htbp]
\center
\input{tables/tab_gaussian_parameters.tex}
\caption{Estimations of the parameters of the distribution $\Psi(\mathcal{T},\mathcal{C},w_t,w_s)$ for the target word $w_t=\ \texttt{momentum}$ and different query words $w_s$, sorted by similarity. The estimation is based on 32 runs of \wtv\ (skip-gram) trained on the English Wikipedia, with a reduced vocabulary size of 200,000 words. The two columns on the right describe the predicted probability for the different query words to appear as the nearest neighbors of the target word based on Equations (\ref{eq_probability_top1_final}) and (\ref{eq_top2}).}
\label{tab_gaussian_parameters}
\end{table}

For any word $w_s\in\mathcal{V}$, let $p_{\text{\#}n}(w_t,w_s)$ denote the probability that $w_s$ is one of the $n$ nearest neighbors of $w_t$, for a randomly sampled embedding space $\mathbf{V}_i\sim \Omega(\mathcal{T},\mathcal{C})$. Deriving this probability is becoming increasingly complex for larger $n$ so, for the moment, we focus on the special case $n=1$: The probability that $w_s$ is \textsl{the nearest neighbor} of $w_t$ for any run $i$.\footnote{Please refer to Appendix \ref{app_prediction_n=2} for an outlook on the derivation for $n>1$.} This is the case if, and only if:
\begin{align}
\cos(w_t,w_s)_{i} > \cos(w_t,w_{s'})_i \quad \forall w_{s'}\in \mathcal{V}\setminus\{w_t,w_s\}
\label{eq_condition_nearest}
\end{align}
For the sake of readability, we fix an arbitrary pair $w_t$, $w_s$ and introduce the following notation:
\begin{align}
\begin{split}
\cos(w_t,w_s)&=:\tilde{x}\sim\mathcal{N}\left(\tilde{\mu},\tilde{\sigma}^2\right) \\
\cos(w_t,w_{s'})&=:x_j \sim\mathcal{N}\left(\mu_j,\sigma_j^2\right) \quad \text{with} \quad j \in \{1,2,...,v-2\}
\end{split}
\end{align}

To determine the probability $p_{\text{\#1}}(w_t,w_s)$, we need to integrate the joint probability distribution $p(\tilde{x},x_1,...,x_{v-2})$ over the entire subspace of $\mathbb{R}^{v-1}$ where the condition (\ref{eq_condition_nearest}) is fulfilled. To be able to carry out this integration, we need to make one more assumption, namely that the different \textsl{random variables $\tilde{x},x_1,...,x_{v-2}$ are independent}\footnote{This assumption is backed by the observations we describe in Section \ref{sec_cosine_similarity_gaussian}.}. This allows us to write the joint probability distribution as:
\begin{align}
p(\tilde{x},x_1,...,x_{v-2})=p(\tilde{x})\cdot p(x_1)\cdot ... \cdot p(x_{v-2})
\end{align}

Now, for a given value of $\tilde{x}$, all $x_j$ can assume any value smaller than $\tilde{x}$, for condition (\ref{eq_condition_nearest}) to hold, hence:\footnote{In practice, we can substitute the integration limits $-\infty$ and $+\infty$ for $0$ and $1$ respectively, since all probability distributions we have seen in our experiments are decreasing sufficiently fast.}
\begin{align}
p_{\text{\#1}}(w_t,w_s) = \int_{-\infty}^{\infty} \left[\prod_{j=1}^{v-2}\int_{-\infty}^{\tilde{x}} f(x_j,\mu_j,\sigma_j)\, \text{d}x_j \right]f(\tilde{x},\tilde{\mu},\tilde{\sigma})\,\text{d}\tilde{x}
\label{eq_probability_top1}
\end{align}
Where the $f(x,\mu,\sigma)$ denote the \textsl{probability density function} of the Normal distribution with mean $\mu$ and variance $\sigma^2$:
\begin{align}
f(x,\mu,\sigma) = \frac{1}{\sqrt{2\pi}\sigma} \exp \left[-\frac{1}{2}\left(\frac{x-\mu}{\sigma}\right)^2\right]
\end{align}
Therefore:
\begin{align}
\begin{split}
\int_{-\infty}^{\tilde{x}}& f(x_j,\mu_j,\sigma_j)\, \text{d}x_j = 
 \int_{-\infty}^{\tilde{x}} \frac{1}{\sqrt{2\pi}\sigma} \exp \left[-\frac{1}{2}\left(\frac{x-\mu}{\sigma}\right)^2\right]\, \text{d}x_j \\
&=\frac{1}{2}\cdot \text{erf}\left(\frac{x_j-\mu_j}{\sqrt{2}\sigma_j}\right) \bigg \vert_{-\infty}^{\tilde{x}} = \frac{1}{2}\left[ \text{erf}\left(\frac{\tilde{x}-\mu_j}{\sqrt{2}\sigma_j}\right)+1\right]
\label{eq_gauss_integral}
\end{split}
\end{align}
Inserting this back into Equation (\ref{eq_probability_top1}) leaves us with:
\begin{align}
\begin{split}
&p_{\text{\#1}}(w_t,w_s) =\\
&\int_{-\infty}^{\infty} \left\{\prod_{j=1}^{v-2} \frac{1}{2}\left[ \text{erf}\left(\frac{\tilde{x}-\mu_j}{\sqrt{2}\sigma_j}\right)+1\right] \right\} \frac{1}{\sqrt{2\pi}\tilde{\sigma}} \exp \left[-\frac{1}{2}\left(\frac{\tilde{x}-\tilde{\mu}}{\tilde{\sigma}}\right)^2\right]\,\text{d}\tilde{x}
\label{eq_probability_top1_final} 
\end{split}
\end{align}

Although it is generally possible to derive closed-form expressions for these types of integrals, we found it more convenient to use numerical integration. While the evaluation of this integral looks rather resource-intensive at first glance, we show in Appendix \ref{app_reducing_complexity_prediction} that the relevant terms in Equation (\ref{eq_probability_top1_final}) will assume trivial values for most pairs of words, which renders the calculations considerably simpler.

Finally, predicting $p_{\text{@}n}$ and hence also $j_{\text{@}n}$ for a target word $w_t$ from the probabilities $p_{\text{\#n}}(w_t,w_s)$ for all relevant query words $w_s$ is rather straightforward.\footnote{As outlined in Section \ref{sec_previous_approaches}, one can derive $j_{\text{@}n}$ from $p_{\text{@}n}$.} The probability of any query word $w_s$ to make the top-$n$-list of the target word in two subsequent runs, is given by the square of $p_{\text{\#n}}(w_t,w_s)$. The expected overlap is therefore:
\begin{align}
p_{\text{@}n}(w_t) =\sum_{w_s\in \mathcal{V}\setminus\{w_t\}}p_{\text{\#n}}(w_t,w_s)^2
\label{eq_expected_overlap}
\end{align}

The agreement between this theoretical prediction and the measurements of $p_{\text{@}n}$ for all languages and techniques included in our experiments (Pearson's $\rho > 0.95$) is outlined in Appendix \ref{app_agreement_prediction_instability}.

The derivation shows that the metrics $p_{\text{@}n}$ and $j_{\text{@}n}$ depend on two qualitatively different sets of parameters of the distribution of the word embeddings: The mean values $\mu$, and the respective variances $\sigma^2$. The nature of the Normal distribution implies, that the mean values $\mu$ are unrelated to the expected difference between the embeddings of a word $w$ over multiple runs. Hence, the requirement of \textsl{independence} that is examined here, demands that $p_{\text{@}n}$ and $j_{\text{@}n}$ are independent of these values. 

However, we observe that this is not the case. In order to understand the influence of the mean values $\mu$ on the measurements of $p_{\text{@}n}$ and $j_{\text{@}n}$, we introduce a new quantity: The \textsl{structure factor $\rho_{\text{@}n}(w_t)$ of a target word} $w_t$ at the threshold $n$. If we take any set of embedding spaces $\{\mathbf{V}_1^{(A)}, \mathbf{V}_2^{(A)},... , \mathbf{V}_{128}^{(A)}\}$ from our experiments, which are obtained by applying the technique $\mathcal{T}^{(A)}$ to \textsl{shuffled} versions of the corpus $\mathcal{C}^{(A)}$, we can estimate the parameters $\mu_{ij}^{(A)}$ and $\sigma_{ij}^{(A)}$ of the distribution of the cosine similarity of any pair of words $w_i,w_j\in\mathcal{V}_A$ and use these estimates to get a very accurate prediction of $p_{\text{@}n}^{(A)}(w_t)$ and $j_{\text{@}n}^{(A)}(w_t)$ for every target word $w_t$ in the vocabulary, as illustrated in Figure \ref{fig_prediction_observation_overlap}. Now, we assume to have an imaginary set of embeddings, called $B$, with the same vocabulary and identical means $\mu_{ij}^{(B)}=\mu_{ij}^{(A)}$, however with $\sigma_{ij}^{(B)}=:\gamma=\text{const.}$ for all word pairs $w_i,w_j\in\mathcal{V}_B$.\footnote{The specific value of $\gamma$ that we use is the mean of $\sigma_{ij}^{(A)}$ over all word pairs.} We call this prediction, i.e. the expected overlap for a specific target word, if the embeddings are in principle identical to $A$, but the variance of the distribution $\Psi$ for every word pair is constant, the \textsl{structure factor $\rho_{\text{@}n}^{(A)}(w_t)$ of the target word}.

\begin{figure}[htbp]
\center
\input{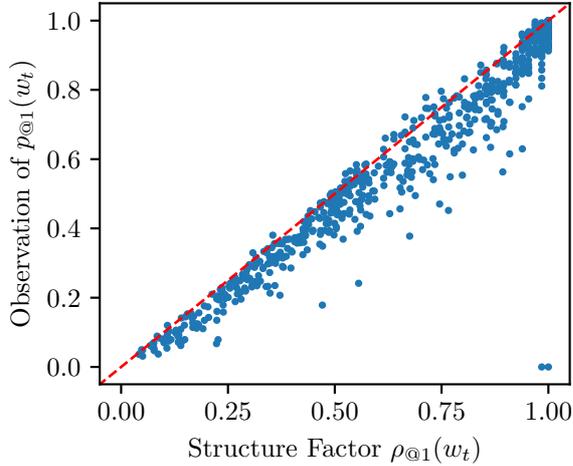}
\vspace{-0.7cm}
\caption{Plot of the structure factor $\rho_{\text{@}1}(w_t)$ over the measurements of $p_{\text{@}1}(w_t)=j_{\text{@}1}(w_t)$, for 1000 randomly sampled target words $w_t$ obtained from 128 runs of \ftt\ on a Portuguese Wikipedia extract.}
\label{fig_structural_instability}
\end{figure}

The \textsl{structure factor} of a word $w_t$ is unrelated to the expected distance of the embedding of $w_t$ between different embedding spaces: It depends solely on the mean values $\mu$ and the constant $\gamma$. Figure \ref{fig_structural_instability} shows the measured overlap $p_{\text{@}1}$ against the structure factor $\rho_{\text{@}1}$ for 1000 randomly sampled target words over 128 runs of \ftt\ on the Portuguese Wikipedia: Quite surprising, the two quantities, are nearly identical for all target words. As Table \ref{tab_structural_instability} shows, the same is true for all languages and embedding techniques we have tested, i.e.:
\begin{align}
p_{@n}(w_t) \approx \rho_{\text{@}n}(w_t)
\end{align}

Hence, the requirement of \textsl{independence} is not fulfilled: On the contrary, the metrics $p_{@n}(w_t)$ and $j_{@n}(w_t)$ are \textsl{virtually identical to a quantity which is unrelated to the expected distance} of the embedding of $w_t$ over multiple runs. Thus, we conclude that the metrics are practically independent of the distance itself. 

\begin{table}[htbp]
\center
\input{tables/tab_structural_instability.tex}
\caption{Pearson correlation coefficient between the structure factor $\rho_{\text{@}1}(w_t)$ and the measurements of $p_{\text{@}1}(w_t)=j_{\text{@}1}(w_t)$, for 1000 randomly sampled target words $w_t$ obtained from 128 runs of \wtv\ (skip-gram), \glv\ and \ftt\ (skip-gram) on Wikipedia corpora in seven different languages.}
\label{tab_structural_instability}
\end{table}

\subsection{Global Distance Metrics}
\label{sec_global_distance_metrics}

We have shown in the previous section that nearest-neighbor based approaches, which were used in most of the previous work on the stability of word embeddings, have several flaws (for a summary, please refer to the conclusion at the end of this section). This observation was our initial motivation to look for other methods to quantify stability and finally led us to a metric that quantifies the distance between two embedding spaces and is sensitive to changes in the global structure: The Pairwise Inner Product (PIP) loss, introduced by \citet{yin2018}.

\subsubsection{The Pairwise Inner Product (PIP) Loss}

The definition of the \textsl{PIP loss} as a metric to quantify the distance between two embedding spaces is based on the rotation-invariance of embedding spaces, which is outlined in Section \ref{sec_random_orientation_of_embedding_spaces}: Two embedding spaces $\mathbf{V}_i,\mathbf{V}_j\in \mathbb{R}^{v\times d}$ are equivalent in regards to all practical purposes if one can be obtained from the other by applying an orthogonal transformation $\mathbf{A}\in\mathbb{R}^{d\times d}: \mathbf{A}\mathbf{A}^\top = \mathbf{I}$. If $\mathbf{V}_i$ and $\mathbf{V}_j$ are rotated versions of one another, we can write:
\begin{align}
\exists \mathbf{A} \text{ orthogonal, with } \mathbf{V}_i \mathbf{A} = \mathbf{V}_j
\end{align}
If this is given, one can see:
\begin{align}
\label{eq_pip_loss_zero_identical}
\mathbf{V}_j\mathbf{V}_j^\top = \mathbf{V}_i \mathbf{A}(\mathbf{V}_i \mathbf{A})^\top=\mathbf{V}_i \mathbf{A}\mathbf{A}^\top \mathbf{V}_i^\top=\mathbf{V}_i\mathbf{V}_i^\top
\end{align}

The $(k,l)$-th entry of the matrix $\mathbf{V}_i\mathbf{V}_i^\top$, which \citet{yin2018} also call the PIP matrix, corresponds to the inner product between the embeddings of the words with index $k$ and $l$ respectively. If the word embeddings are normalized, i.e. $\vec{v}\vec{v}^\top=1$ for all embeddings $\vec{v}$, these entries are identical to the \textsl{cosine similarities} of the word pairs. 

Equation (\ref{eq_pip_loss_zero_identical}) shows, that if two embedding spaces are equivalent, their PIP matrices are equal. Hence, the PIP loss $\mathbf{D}_{\text{PIP}}$ between the embedding spaces $\mathbf{V}_i$ and $\mathbf{V}_j$ is defined as the norm of the difference between their PIP matrices:
\begin{align}
\begin{split}
&\mathbf{D}_{\text{PIP}}: \mathbb{R}^{v\times d} \times \mathbb{R}^{v\times d}\to\mathbb{R},\quad (\mathbf{V}_i,\mathbf{V}_j)\mapsto \mathbf{D}_{\text{PIP}}(\mathbf{V}_i,\mathbf{V}_j) \\
&\mathbf{D}_{\text{PIP}}(\mathbf{V}_i,\mathbf{V}_j) = ||\mathbf{V}_i\mathbf{V}_i^\top-\mathbf{V}_j\mathbf{V}_j^\top||=\sqrt{\sum_{w_k,w_l\in\mathcal{V}}\left(\vec{v_{ik}}\vec{v}_{il}^\top -\vec{v_{jk}}\vec{v}_{jl}^\top \right)^2}
\end{split}
\end{align} 

Where $\vec{v}_{ik}$ is the embedding of the word $w_k$ in the space $\mathbf{V}_i$, i.e. the $k$-th row of the matrix $\mathbf{V}_i$. The sum in the equation above consists of $|\mathcal{V}|^2$ terms, i.e. it scales with the size of the vocabulary of the embeddings. To compare measurements between embeddings with different-sized vocabularies, we introduce the \textsl{reduced PIP loss}:
\begin{align}
\label{eq_rPIP}
\mathbf{D}_{\text{rPIP}}(\mathbf{V}_i,\mathbf{V}_j) =\frac{1}{2\cdot |\mathcal{V}|} \mathbf{D}_{\text{PIP}}(\mathbf{V}_i,\mathbf{V}_j)
\end{align}
which measures the squared mean of the expression $\frac{1}{2}\left(\vec{v_{ik}}\vec{v}_{il}^\top -\vec{v_{jk}}\vec{v}_{jl}^\top \right)$ over all word pairs.\footnote{As shown in the sections below, the additional factor $2$ is necessary to ensure the desired codomain $[0,1]$.}

\subsubsection{The Word-Wise PIP Loss}
\label{sec_def_wwr_pip_loss}

As mentioned several times in the sections above, we are interested in a metric $\mathbf{d}$ that captures the distance of the embedding of one word $w_k$ between two embedding spaces $\mathbf{V}_i$ and $\mathbf{V}_j$. While the PIP loss does not match this format, we can derive a metric of the desired shape from the (reduced) PIP loss, which we call the \textsl{word-wise reduced PIP loss} $\mathbf{d}_{\text{PIP}}$:
\begin{align}
\label{eq_wwrPIP}
\begin{split}
&\mathbf{d}_{\text{PIP}}: \mathcal{V}\times \mathbb{R}^{v\times d} \times \mathbb{R}^{v\times d}\to \mathbb{R},\quad (w_k,\mathbf{V}_i,\mathbf{V}_j)\mapsto \mathbf{d}_{\text{PIP}}(w_k,\mathbf{V}_i,\mathbf{V}_j) \\
&\mathbf{d}_{\text{PIP}}(w_k,\mathbf{V}_i,\mathbf{V}_j) = \frac{1}{2\cdot\sqrt{|\mathcal{V}|}} \sqrt{\sum_{w_l\in\mathcal{V}}\left(\vec{v_{ik}}\vec{v}_{il}^\top -\vec{v_{jk}}\vec{v}_{jl}^\top \right)^2}
\end{split}
\end{align}
Here, in contrast to the (reduced) PIP loss, the word $w_k$ is fixed; we obtain the mean of the expression $\frac{1}{2}\left(\vec{v_{ik}}\vec{v}_{il}^\top -\vec{v_{jk}}\vec{v}_{jl}^\top \right)$ for the given $w_k$ with all other words $w_l\in\mathcal{V}$ of the vocabulary.\footnote{Technically there is no difference in excluding the target word $w_k$ from $\mathcal{V}$ or keeping it. As long as the embeddings are normalized, $\vec{v}_{ik}\vec{v}_{ik}^\top - \vec{v}_{jk}\vec{v}_{jk}^\top = 1-1=0$.}

Intuitively, the word-wise reduced PIP loss of a word $w_k$ between the embedding spaces $\mathbf{V}_i$ and $\mathbf{V}_j$ measures the mean squared difference in cosine similarity\footnote{Under the condition that the embeddings are normalized.} of the word $w_k$ with all words in the vocabulary between the two spaces $\mathbf{V}_i$ and $\mathbf{V}_j$. As shown in theory (Section \ref{sec_random_orientation_of_embedding_spaces}) and experiment (Section \ref{sec_cosine_similarity_gaussian}), the cosine similarity of any two words is a rotation-invariance property, hence expected to be stable over subsequent runs. Measuring the difference of this quantity between two embedding spaces thus makes a prospective candidate for the distance metric we are looking for. We want to note that while the use of this derivate of the PIP loss by \citet{yin2018} to measure stability is -- to our knowledge -- a novel approach, \citet{egger2016} already used a similar metric to compare embeddings derived from different corpora.

\subsubsection{Reducing the Computational Complexity}

Before we begin to assess if the metric fulfils the requirements introduced above, let us introduce an approach to reduce the resource utilization for the computation of the metrics significantly: Both, the reduced PIP loss as well as the word-wise reduced PIP loss, measure the squared mean of the expression $\frac{1}{2}\left(\vec{v_{ik}}\vec{v}_{il}^\top -\vec{v_{jk}}\vec{v}_{jl}^\top \right)$ over all possible pairs (with $w_k$ fixed for the word-wise reduced PIP loss) of words from the vocabulary $\mathcal{V}$. Naturally, we expect this mean value to be independent of the vocabulary size. Hence, we can obtain a \textsl{proxy} of the (word-wise) reduced PIP loss by calculating the mean of this expression over a randomly sampled subset $\mathcal{V'}\subset\mathcal{V}$. The time complexity of calculating the reduced PIP loss is $\mathcal{O}(|\mathcal{V}|^2)$, that of the word-wise reduced PIP loss $\mathcal{O}(|\mathcal{V}|)$. Thus, sampling a random $\mathcal{V'}$ with $|\mathcal{V'}|\ll |\mathcal{V}|$ yields a substantial reduction in complexity for both metrics.

One could argue that this introduces a free variable to this distance metrics, comparable to the scope $n$ of the nearest neighbor approaches: the size of the subset $\mathcal{V'}$. However, we show in the section below that the metric is consistent over this variable, if $\mathcal{V'}$ has a sufficiently large size ($\geq 10^3$ words).

\subsubsection*{(I) Formal Criteria}

The definition of the \textsl{word-wise reduced PIP loss} in Section \ref{sec_def_wwr_pip_loss} matches the format of the distance metric we are looking for. However, we still need to verify that the codomain of $\mathbf{d}_{\text{PIP}}$ coincides with the interval $[0,1]$. From now on, we will assume that all word vectors are normalized, i.e. $\vec{v}_k\vec{v}_k^\top=1 \quad \forall w_k \in\mathcal{V}$. If this is given, any two word vectors $\vec{v}_k$ and $\vec{v}_l$ fulfil:
\begin{align}
\vec{v}_k\vec{v}_l^\top=\frac{\vec{v}_k\vec{v}_l^\top}{1\cdot 1}=\frac{\vec{v}_k\vec{v}_l^\top}{\left(\vec{v}_k\vec{v}_k^\top\right)\cdot\left(\vec{v}_l\vec{v}_l^\top\right)}=\cos\left(\angle (\vec{v}_k,\vec{v}_l)\right) 
\end{align}
We know that the codomain of the cosine function is $[-1,1]$, hence for any set of normalized word embeddings $\vec{v}_k,\vec{v}_l,\vec{v}_m,\vec{v}_n$:
\begin{align}
\vec{v}_k\vec{v}_{l}^\top -\vec{v}_{m}\vec{v}_{n}^\top \in [-2,2] \implies \left(\frac{\vec{v}_k\vec{v}_{l}^\top -\vec{v}_{m}\vec{v}_{n}^\top}{2}\right)^2 \in [0,1]
\end{align}
And since the mean of any set of real numbers will not fall short of the smallest or exceed the largest one:
\begin{align}
\mathbf{d}_{\text{PIP}}: \mathcal{V}\times \mathbb{R}^{v\times d} \times \mathbb{R}^{v\times d}\to [0,1],\quad (w_k,\mathbf{V}_i,\mathbf{V}_j)\mapsto \mathbf{d}_{\text{PIP}}(w_k,\mathbf{V}_i,\mathbf{V}_j)
\end{align}

\subsubsection*{(II) Consistency}

The word-wise reduced PIP loss as defined in Equation (\ref{eq_wwrPIP}), does not have any free parameters that could cause inconsistencies; hence this requirement is fulfilled.

However, when computing the metrics in practice, we generally do not calculate the mean over the whole vocabulary $\mathcal{V}$, but instead over a randomly sampled subset of words $\mathcal{V'}\subset \mathcal{V}$ of fixed size $|\mathcal{V'}|$. In order to examine the consistency of $\mathbf{d}_{\text{PIP}}$ over this free variable, given a sufficiently large size ($|\mathcal{V'}|\geq 10^3$ words), we ran the following experiments: 

\begin{enumerate}
\itemsep0em 
\item For each language and embedding technique described in Section \ref{sec_exp_setup}, we sample 1000 target words randomly from the joint vocabulary of all runs in the respective language.
\item Now we randomly pick a pair of runs\footnote{To ensure the statistical significance of the results, we picked 10 random pairs and ultimately calculate the mean Spearman correlation values over these pairs.} over independently shuffled corpora and calculate the word-wise reduced PIP loss $\mathbf{d}_{\text{PIP}}$ for every target word at different values of $|\mathcal{V'}|\in\{10^3,10^4,10^5,|\mathcal{V}|\}$.
\item Finally, we calculate the Spearman correlation of $\mathbf{d}_{\text{PIP}}$ for the 1000 target words between different values of $|\mathcal{V'}|$.
\end{enumerate} 

\begin{table}[htbp]
\begin{center}
\input{tables/tab_pip_consistent.tex}
\caption{Spearman correlation of the metrics $\mathbf{d}_{\text{PIP}}$ for 1000 target words for different values of $|\mathcal{V'}|\in\{10^3,10^4,10^5,|\mathcal{V}|\}$ for \wtv\ (top), \glv\ (middle) and \ftt\ (bottom), obtained as outlined in Section \ref{sec_global_distance_metrics}. For each of the techniques, we show the average of the correlation over all languages mentioned in Section \ref{sec_exp_setup}. One might ask why the values on the main diagonal are different from 1: This is the case, since we repeated the random sampling of the proxy words twice and calculated the correlation between these two samples. This helps to understand not only how stable the measures are over different numbers of proxy words, but also over subsequent runs with a fixed number of proxy words.}
\label{tab_pip_consistent}
\end{center}
\end{table}

The results -- please refer to Table \ref{tab_pip_consistent} -- show that the metric is highly consistent over the different sizes of $|\mathcal{V'}|$, with all correlation values larger than $0.99$ for $|\mathcal{V'}|\geq 10^4$.\footnote{Thus, we choose $|\mathcal{V'}|\geq 2\cdot 10^4$ in practice.} Hence, our approach to increase the computational efficiency by calculating the word-wise reduced PIP loss over a randomly sampled subset $\mathcal{V'}$ does not interfere with the meaningfulness of the metric.

\subsubsection*{(III) Independence}

Finally, we want to understand if the word-wise reduced PIP loss of a word $w$ between two embedding spaces $\mathbf{V}_i$ and $\mathbf{V}_j$, is influenced by any quantities which are unrelated to the distance. Hence, similarly to Section \ref{sec_previous_approaches}, we attempt to predict $\mathbf{d}_{\text{PIP}}$ simply based on the assumption that the cosine similarity of any two words $w_1$ and $w_2$ follows a Normal distribution:
\begin{align}
\cos(w_1,w_2)_{i} \sim \mathcal{N}\left(\mu_{12},\sigma_{12}^2 \right)
\end{align}
Under this assumption, the expression $\vec{v_{ik}}\vec{v}_{il}^\top -\vec{v_{jk}}\vec{v}_{jl}^\top$ is just the difference of two Normally distributed random variables, which is again a Normal distribution \cite{lemons2002introduction}:
\begin{align}
\vec{v_{ik}}\vec{v}_{il}^\top\sim\mathcal{N}\left(\mu_{kl},\sigma_{kl}^2 \right), \vec{v_{jk}}\vec{v}_{jl}^\top\sim\mathcal{N}\left(\mu_{kl},\sigma_{kl}^2 \right) \implies \vec{v_{ik}}\vec{v}_{il}^\top -\vec{v_{jk}}\vec{v}_{jl}^\top \sim\mathcal{N}\left(0, 2\sigma_{kl}^2 \right)
\label{eq_ww_pip_loss_gauss}
\end{align}
Hence, we can write the expectation of $\mathbf{d}_{\text{PIP}}$ for a word $w_k$ over randomly sampled embedding spaces $\mathbf{V}_i,\mathbf{V}_j$ as:\footnote{In the last step we use the definition of the variance $\sigma^2$ for a randomly distributed variable $x\sim\mathcal{N}(\mu,\sigma^2)$, that yields $\langle x^2 \rangle = \mu^2+\sigma^2$.}
\begin{align}
\begin{split}
\langle \mathbf{d}_{\text{PIP}} (w_k,\mathbf{V}_i,\mathbf{V}_j) \rangle_{ij} =&\bigg \langle \sqrt{\frac{1}{4\cdot |\mathcal{V}|}\sum_{w_l\in\mathcal{V}}\left(\vec{v_{ik}}\vec{v}_{il}^\top -\vec{v_{jk}}\vec{v}_{jl}^\top \right)^2} \bigg \rangle_{ij} \\
=&\sqrt{\frac{1}{4\cdot |\mathcal{V}|}\sum_{w_l\in\mathcal{V}}		\bigg \langle \left(\vec{v_{ik}}\vec{v}_{il}^\top -\vec{v_{jk}}\vec{v}_{jl}^\top \right)^2\bigg \rangle_{ij}} \\
=&\sqrt{\frac{1}{4\cdot |\mathcal{V}|}\sum_{w_l\in\mathcal{V}}		2\sigma_{kl}^2} \\
=& \sqrt{\frac{1}{2\cdot |\mathcal{V}|}\sum_{w_l\in\mathcal{V}}	\sigma_{kl}^2}
\end{split}
\label{eq_ww_pip_loss_expectation}
\end{align}
This means, the expectation of the word-wise reduced PIP loss of a word $w_k$ is a multiple of \textsl{the squared mean of the standard deviation of the cosine similarity between the target word and all other words of the vocabulary}. The variance $\sigma_{kl}$ is a measure of the expected difference of the cosine similarity of the word pair $w_k$,$w_l$ over two independent embedding spaces $\mathbf{V}_i,\mathbf{V}_j$ sampled from the same probability distribution. Hence, the expectation of the word-wise reduced PIP loss is independent of any quantities that do not measure the distance between the embedding spaces: The third and last \textsl{requirement of independence is fulfilled.}

\subsection*{Conclusion}

In the last two sections, we introduced and compared two methods to measure the distance of a word $w$ between two embedding spaces $\mathbf{V}_i$ and $\mathbf{V}_j$: 
\begin{description}
\item[Nearest-neighbor based approaches] Namely, measuring the percentage overlap (or Jaccard metric) of the $n$ nearest neighbors (by cosine distance) of the word $w$ in the two spaces $\mathbf{V}_i$ and $\mathbf{V}_j$.
\item[Word-wise reduced PIP loss] Which is defined as the squared mean of the difference in the cosine similarity of the word $w$ with all other words of the vocabulary between the two spaces $\mathbf{V}_i$ and $\mathbf{V}_j$.
\end{description}
We find that nearest-neighbor based approaches were used in all previous work that attempts to quantify the stability of word embeddings \citep{hellrich-hahn-2016-assessment,hellrich-hahn-2016-bad,hellrich-hahn-2017-fool,hellrich-etal-2019-influence,antoniak2018,Chugh2018StabilityOW,wendlandt2018,pierrejean-tanguy-2018-predicting}. The word-wise reduced PIP loss is a novel measure, based on the PIP loss, introduced by \citet{yin2018}.\footnote{As mentioned above, \citet{egger2016} already used a similar metric to measure the difference between embeddings derived from different corpora.}

The comparison of these two methods with respect to the requirements for a distance metric that were introduced in Section \ref{sec_requirements} is summarized in Table \ref{tab_comparison_approaches}. Whereas both types of metrics fulfil the necessary formal criteria, the nearest-neighbor based approaches are inconsistent over the chosen threshold value $n$ and strongly -- if not exclusively -- depend on the structure factor of the embedding, which is unrelated to the distance between multiple runs. Some of these limitations were already touched upon by \citet{pierrejean-tanguy-2018-predicting} but they argued the convenience and simplicity of the measure justifies its use. 

However, we argue that the word-wise reduced PIP loss, which does not exhibit these problems and has lower computational complexity than the nearest-neighbor based metrics, is better suited for this task and -- unless there are limitations to this metric that we overlooked -- should be the method of choice in the future. In our studies on the stability of word embeddings in the following section, we will hence predominantly use the word-wise reduced PIP loss.

\begin{table}[htbp]
\input{tables/tab_comparison_approaches.tex}
\caption{Comparison of the two types of approaches for a metric $\mathbf{d}(w,\mathbf{V}_i,\mathbf{V}_j)$ to measure the distance between the embeddings of a word $w$ in the two embedding spaces $\mathbf{V}_i$ and $\mathbf{V}_j$: nearest-neighbor based approaches and the word-wise reduced PIP loss.}
\label{tab_comparison_approaches}
\end{table}

\section{Understanding the Instability}
\label{sec_understanding_instability}

Now that we have a tool at hand to measure the distance between embedding spaces and individual embeddings within them -- the reduced PIP loss and word-wise reduced PIP loss, we can finally tackle the task we set out to do at the beginning of this chapter: To quantify and understand the instability of word embeddings. 

We begin by calculating the reduced PIP loss for 120 pairs of embedding spaces, composed of 16 subsequent runs with each type of document sampling (fixed, shuffled, and bootstrapped) for every language and embedding technique outlined in Section \ref{sec_exp_setup}. The computations are based on a random sample of $|\mathcal{V}'|=2\times10^4$ target words for every language. The results are summarized in Table \ref{tab_overview_measured_pip}.

\begin{table}[htbp]
\begin{center}
\input{tables/tab_overview_measured_pip.tex}
\caption{Reduced PIP loss $\mathbf{D}_\text{rPIP}$ for different types of document sampling (fixed, shuffled, bootstrapped) and every language and embedding technique outlined in Section \ref{sec_exp_setup}. The calculation of mean $\mu$ and standard deviation $\sigma$ is based on 120 pairs of embeddings for each setting, composed of 16 independent runs, each based on a random sample of $2\times 10^4$ target words. The values for \glv\ trained on the English corpus must be treated with caution: As mentioned in Section \ref{sec_exp_setup}, we had to restrict the iterations for these runs to 25. Therefore, the results are not directly comparable to the other languages.}
\label{tab_overview_measured_pip}
\end{center}
\end{table}

\subsubsection*{Observation 1: The variance of the measurements is small}
The relative variance of the reduced PIP loss between different pairs of embedding spaces is rather small (between around $10^{-2}$ and $10^{-3}$). The explanation for this can be found in our calculations in \ref{sec_global_distance_metrics}: As Equation \ref{eq_ww_pip_loss_gauss} shows, the expectation of the word-wise reduced PIP loss is the squared sum over $|\mathcal{V}|$ samples of Gaussian probability distributions $\mathcal{N}\left(0,\sigma_{kl}^2\right)$, with zero mean and variances $\sigma_{kl}^2$. We consider the special case of constant variances ($\sigma_{kl}=\tilde{\sigma}$ $\forall i$) for a second:
\begin{align}
\mathbf{d}_{\text{PIP}} (w_k,\mathbf{V}_i,\mathbf{V}_j)=& \sqrt{\frac{1}{2\cdot |\mathcal{V}|}\sum_{w_l\in\mathcal{V}} \sigma_{kl}^2}= \frac{\tilde{\sigma}}{\sqrt{2\cdot |\mathcal{V}|}}\sqrt{\sum_{i=1}^{|\mathcal{V}|} \tilde{x}_i^2} \quad \text{with}\quad x_i \sim \mathcal{N}(0,1)
\end{align}
The term on the right, i.e. the squared sum over $|\mathcal{V}|$ random variables following a normal distribution $\mathcal{N}(0,1)$ corresponds to a Chi distribution with $|\mathcal{V}|$ degrees of freedom. The mean $\mu_k$ and variance $\sigma_k^2$ of the Chi distribution with $k$ degrees of freedom are given by \cite{Walck1996HandbookOS}:
\begin{align}
&\mu_k = \sqrt{2}\frac{\Gamma((k+1)/2)}{\Gamma(k/2)} &\sigma^2_k = k-\mu_k^2
\end{align}
Where $\Gamma$ is the gamma function. We can obtain an estimation of these parameters for large $k$ through an expansion of the gamma function around $+\infty$:
\begin{align}
&\mu_k = \sqrt{k} + \mathcal{O}\left(\frac{1}{\sqrt{k}}\right) &\sigma^2_k = k - (k - \mathcal{O}(1)) = \mathcal{O}(1)
\end{align}
Hence, the relative width of this distribution for large $k$ scales with:
\begin{align}
\frac{\sigma_k}{\mu_k}\propto \frac{1}{\sqrt{k}}
\end{align}
For the word-wise reduced PIP loss, this means:
\begin{align}
\frac{\sigma (\mathbf{d}_{\text{PIP}})}{\mu (\mathbf{d}_{\text{PIP}})} \propto \frac{1}{\sqrt{|\mathcal{V}|}}
\end{align}
Since we typically deal with large vocabularies $|\mathcal{V}|>10^5$, the relative width of this distribution converges to zero. This effect is even stronger for the reduced PIP loss, which equals the mean of the word-wise reduced PIP loss over all words of the vocabulary. This means in practice, that to get an accurate estimate of the overall instability of an embedding space, obtained by applying the embedding technique $\mathcal{T}$ to a corpus $\mathcal{C}$, it is sufficient to perform only two independent runs, and measure the reduced PIP loss between the two resulting spaces. The resources required to conduct 128 independent runs, as done in this work, should thus not prevent anyone from obtaining a practical understanding of the instability of any combination $(\mathcal{T},\mathcal{C})$. \textsl{Two subsequent runs on an independently shuffled corpus suffice in most cases.}

\subsubsection*{Observation 2: We can identify patterns of stability for the different techniques and sampling types that are consistent for all languages}

Another, not quite surprising observation, is that the \glv\ embeddings show no statistically significant difference in the distribution of the reduced PIP loss between \textsl{fixed} and \textsl{shuffled} document sampling. This technique contains one step, where the word co-occurrence matrix of the input text is randomly shuffled, hence even in the \textsl{fixed} setting, the data is shuffled implicitly. Therefore, we do not expect any difference in the distance of the embedding spaces between the \textsl{fixed} and the \textsl{shuffled} sampling.

Comparing the stability of our three embedding techniques for the different sampling types shows: For each of the three sampling types, we find a \textsl{distinct order, that is consistent over all seven languages}, as summarized in Table \ref{tab_instability_order_sampling}.

\begin{table}[htbp]
\begin{center}
\input{tables/tab_instability_order_sampling.tex}
\caption{For any type of document sampling, we find that the order of stability of the different embedding techniques, as measured by the reduced PIP loss, is consistent over all languages.}
\label{tab_instability_order_sampling}
\end{center}
\end{table}

For the \textsl{fixed} setting, \wtv\ is the most stable method, \glv\ for the \textsl{shuffled} setting, and finally when training the embeddings on \textsl{bootstrapped corpora}, \ftt\ is the most stable one. This leads us to the following interpretation: While \ftt\ has the largest \textsl{intrinsic instability} of all methods, as seen for the \textsl{fixed} and \textsl{shuffled} settings, it seems to be better at abstracting and capturing the semantic relationships of words within a language from a limited set of documents that is sampled from this language. And this abstraction is one of the main objectives when training word embeddings!

This capability is also demonstrated in the word analogy tasks in the different languages (see Table \ref{tab_overview_analogy_scores}): \ftt\ outperforms \wtv\ and \glv\ in every language except for Finnish. And the results on the Finnish word analogy task set need to be treated with caution, as it consists of only around $10^3$ tasks, around 20-times fewer than in any other language.

\begin{table}[htbp]
\begin{center}
\input{tables/tab_overview_analogy_scores.tex}
\caption{Analogy scores for every language and embedding technique outlined in Section \ref{sec_exp_setup}. The calculation of mean $\mu$ and standard deviation $\sigma$ is based on 128 independent runs.}
\label{tab_overview_analogy_scores}
\end{center}
\end{table}

Further experiments are necessary to confirm these findings, and especially examine the correlation of the PIP loss over bootstrapped corpora with the performance of the embeddings on various downstream tasks. Our preliminary results allow for the following statement: If we want to compare the \textsl{quality of embeddings} produces by different techniques on one corpus, especially for languages where no analogy task set is available, the \textsl{reduced PIP loss} between multiple sets of embeddings that were trained on bootstrapped corpora \textsl{could be an indicator}.

\subsubsection*{Observation 3: We can differentiate between two main causes of the instability}

We already used the term \textsl{intrinsic instability} in the section above to describe the PIP loss of an embedding technique over \textsl{fixed} or \textsl{shuffled} corpora. Even when the order of the documents within a corpus is randomly shuffled, the semantics of the different words of the vocabulary do not change. Therefore, one would expect that for an imaginary embedding technique, which has no method-induced instability at all, the PIP loss between embedding spaces trained on \textsl{fixed} and \textsl{shuffled} corpora is zero.

However, even for this perfectly stable technique, one would expect differences between embedding spaces trained on \textsl{bootstrapped} corpora -- as these inhibit actual differences in the semantics of the words (for example one meaning of a homonym might be dropped through bootstrapping, while another one is amplified). Now, let us assume to have an embedding technique of great practical value, which is able to abstract the semantics of a language by training on a given corpus sampled from this language. Then one would expect these differences to be comparably small.

This assessment leads us to introduce the following distinction between two types of instability for an embedding technique $\mathcal{T}$ trained on a corpus $\mathcal{C}$:

\begin{description}

\item[Intrinsic Instability] The mean of the reduced PIP loss of a sample of embedding spaces obtained by applying the technique $\mathcal{T}$ on independently \textsl{shuffled} versions of the corpus $\mathcal{C}$:
\begin{align}
\mathcal{I}_{\text{int}}(\mathcal{T},\mathcal{C}) = \langle \mathbf{D}_\text{rPIP}(\mathbf{V}_i,\mathbf{V}_j) \rangle \quad \text{with} \quad \mathbf{V}_i,\mathbf{V}_j\sim \Omega_{\textsl{shuf.}}(\mathcal{T},\mathcal{C})
\end{align}
This measure describes the instability of the technique $\mathcal{T}$ trained on the corpus $\mathcal{C}$.

\item[Extrinsic Instability] The quadratic difference between the mean of the reduced PIP loss over \textsl{bootstrapped} samples and the intrinsic instability:
\begin{align}
\mathcal{I}_{\text{ext}}(\mathcal{T},\mathcal{C})=\sqrt{\langle \mathbf{D}_\text{rPIP}(\mathbf{V}_k,\mathbf{V}_l)\rangle - \mathcal{I}_{\text{int}}(\mathcal{T},\mathcal{C})}\quad \text{with} \quad \mathbf{V}_k,\mathbf{V}_l\sim \Omega_{\textsl{boot.}}(\mathcal{T},\mathcal{C})
\end{align}
This measure describes the instability of the technique $\mathcal{T}$ towards variations in the corpus $\mathcal{C}$.

\end{description}

This distinction may be even more insightful on the level of individual words -- see Section \ref{sec_stability_words}.

\subsubsection*{Observation 4: We observe a weak correlation between extrinsic instability, corpus and vocabulary size}

Table \ref{tab_overview_measured_pip} contains the results of the evaluation of the intrinsic instability for all techniques and languages. The values of the extrinsic instability can be found in Table \ref{tab_extrinsic_instabilities} below.

\begin{table}[htbp]
\begin{center}
\input{tables/tab_extrinsic_instabilities.tex}
\caption{Extrinsic instability for every language and embedding technique outlined in Section \ref{sec_exp_setup}. The calculation of mean $\mu$ and standard deviation $\sigma$ is based on 120 pairs of embeddings for each setting, composed of 16 independent runs, each based on a random sample of $2\times 10^4$ target words. The value for \glv\ trained on the English corpus must be treated with caution: As mentioned in Section \ref{sec_exp_setup}, we had to restrict the iterations for these runs to 25. Therefore, the results are not directly comparable to the other languages.}
\label{tab_extrinsic_instabilities}
\end{center}
\end{table}

To determine if there is a correlation between the extrinsic instability of a language with the word count and vocabulary size of the respective corpus (outlined in Table \ref{tab_embedding_languages}), we calculated Spearman's $\rho$ of $\mathcal{I}_{\text{ext}}(\mathcal{T},\mathcal{C})$ with the quotient $|\mathcal{V}|/|\mathcal{C}|$ of vocabulary and corpus size for the three different embedding techniques:\footnote{For \glv\, the values for English were excluded from the calculation, since these runs are based on different model parameters, as outlined in Section \ref{sec_exp_setup}.}
\begin{align}
\rho_{\text{\wtv}} = 0.715,\quad \rho_{\text{\glv}} = 0.143,\quad \rho_{\text{\ftt}} = 0.536
\label{eq_results_rho_techniques}
\end{align}
The extrinsic instability $\mathcal{I}_{\text{ext}}(\mathcal{T},\mathcal{C})$ seems to decrease with corpus size and increase with vocabulary size for \wtv, and \ftt. For \glv, we cannot confirm a correlation based on the data we have. 

We intuitively expect a correlation like this; hence the values in Equation (\ref{eq_results_rho_techniques}) are smaller than one might have thought. The correlation seems evident for \wtv\ (with a $p$ value of $0.071$), but less so for \ftt\ ($p=0.22$) and looks entirely random for \glv \ ($p=0.78$).\footnote{The $p$ value measures the probability to observe the present correlation if the two datasets are in fact independent of each other -- in other words, a small $p$ value is a good indicator for true correlation.} Since these results are based on a rather small sample size of 7 languages/corpora, additional experiments are necessary to confirm or refute these findings and understand the phenomenon in detail.

\subsection{Instability of Individual Words}
\label{sec_stability_words}

\begin{figure}[htbp]
\center
\vspace{-1cm}
\input{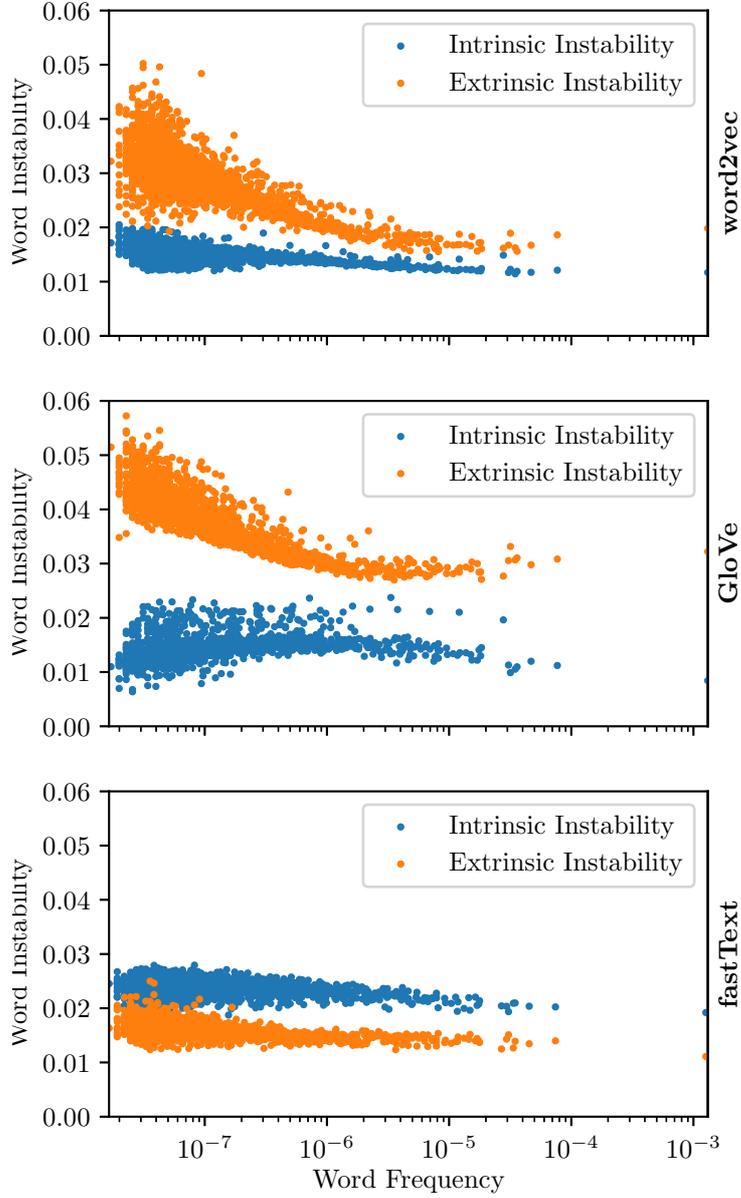}
\vspace{-1cm}
\caption{Intrinsic instability $\mathcal{J}_{\text{int}}(\mathcal{T},\mathcal{C},w)$ and extrinsic instability $\mathcal{J}_{\text{ext}}(\mathcal{T},\mathcal{C},w)$ for \wtv,\glv\ and \ftt\ on the Hindi Wikipedia corpus for 2,000 randomly sampled words as a function of word frequency. The calculation is based on 120 pairs of embeddings for each setting, composed of 16 independent runs, and the word-wise reduced PIP loss is calculated over $2\times 10^4$ randomly sampled target words.}
\label{fig_ww_instability_hindi}
\end{figure}

\begin{figure}[htbp]
\center
\vspace{-1cm}
\input{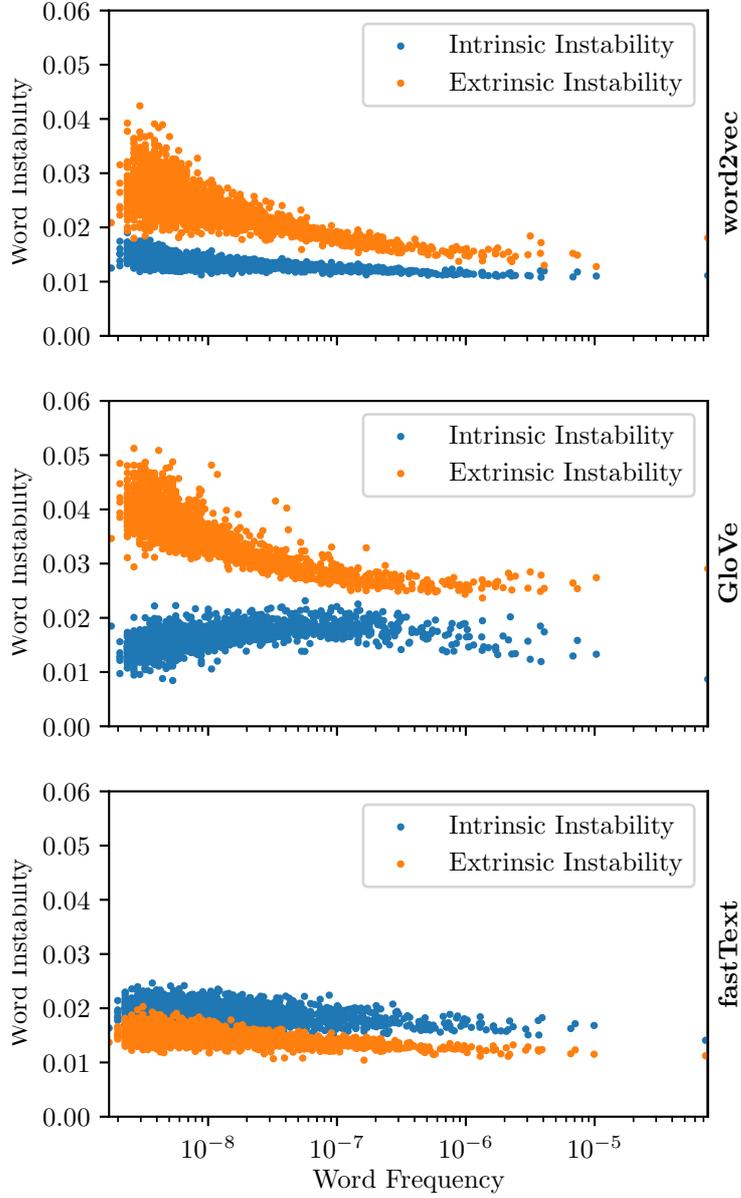}
\vspace{-1cm}
\caption{Intrinsic instability $\mathcal{J}_{\text{int}}(\mathcal{T},\mathcal{C},w)$ and extrinsic instability $\mathcal{J}_{\text{ext}}(\mathcal{T},\mathcal{C},w)$ for \wtv, \glv, and \ftt\ on the Polish Wikipedia corpus for 2,000 randomly sampled words as a function of word frequency. The calculation is based on 120 pairs of embeddings for each setting, composed of 16 independent runs, and the word-wise reduced PIP loss is calculated over $2\times 10^4$ randomly sampled target words.}
\label{fig_ww_instability_polish}
\end{figure}

What we found to be even more insightful than the analysis of the instability of embedding spaces, is to examine the instability of the embeddings of individual words over multiple runs. First, we extend our definitions of the intrinsic and extrinsic instability of embedding spaces to individual words, using the word-wise reduced PIP loss $\mathbf{d}_{\text{PIP}}$:
\begin{align}
\begin{split}
\mathcal{J}_{\text{int}}(\mathcal{T},\mathcal{C},w) &= \langle \mathbf{d}_\text{PIP}(\mathbf{V}_i,\mathbf{V}_j,w) \rangle \\
\mathcal{J}_{\text{ext}}(\mathcal{T},\mathcal{C},w) &= \sqrt{\langle \mathbf{d}_\text{PIP}(\mathbf{V}_k,\mathbf{V}_l,w)\rangle - \mathcal{J}_{\text{int}}(\mathcal{T},\mathcal{C},w)}
\end{split}
\end{align}
where $\mathbf{V}_i,\mathbf{V}_j\sim \Omega_{\textsl{shuf.}}(\mathcal{T},\mathcal{C})$ and $\mathbf{V}_k,\mathbf{V}_l\sim \Omega_{\textsl{boot.}}(\mathcal{T},\mathcal{C})$. Based on these definitions, we calculated the intrinsic and extrinsic instability for 2,000 randomly sampled words over 120 pairs of embedding spaces, for every language and embedding technique. Both quantities are plotted over the word frequency in Figure \ref{fig_ww_instability_hindi} for Hindi and \ref{fig_ww_instability_polish} for Polish. The experiments yield several observations, that are outlined below.

\subsubsection*{Observation 1: The distribution of intrinsic/extrinsic instability over word frequency over is different for each technique, but similar for all languages}

Figures \ref{fig_ww_instability_hindi} and \ref{fig_ww_instability_polish} demonstrate this. Both show distinctive patterns for the intrinsic and extrinsic instability of the three different techniques, but as a whole, the two figures look much alike. All languages we examined in this work (see Section \ref{sec_exp_setup}) show the same pattern, hence we conclude that the shape of the curves depends primarily on the technique and is only slightly affected by the corpus. Therefore, we argue, most of the observations below can be generalized to any corpus.

\subsubsection*{Observation 2: Intrinsic instability is constant over word frequency for all techniques}

This observation might come as a surprise, as e.g. \citet{wendlandt2018} claimed that word stability increases with frequency, which is a somewhat intuitive expectation. However, \citet{hellrich-hahn-2016-bad} made a similar observation for skip-gram embeddings trained on a German corpus.

The shape of the curves in Figures \ref{fig_ww_instability_hindi} and \ref{fig_ww_instability_polish} indicates that the mean of the intrinsic instability for \wtv, \glv\ and \ftt, is constant over the whole frequency interval. A statistical analysis -- i.e. dividing the data into 20 batches by frequency and calculating the Spearman correlation between the mean values of the intrinsic stability and the mean frequency of the respective batch -- supports this: For \glv\ and \ftt, we find no statistically significant correlation ($p<0.05$) in any language. For \wtv\ on the other hand, we find a significant correlation, however, the change of the intrinsic instability over the whole frequency interval is still relatively small ($<20\%$). The plots furthermore suggest that the variance of the instability increases for small frequencies, but this effect is visually enhanced in our plots by the abundance of low-frequency words in the vocabulary compared to high-frequency words. 

Altogether we conclude that the intrinsic instability of words hardly depends on their frequency, at least for the techniques and languages examined in this work.

\subsubsection*{Observation 3: The extrinsic instability of word-based techniques decreases with word frequency}

Whereas the intrinsic instability of low-frequency words is not significantly higher than for high-frequency words, the extrinsic instability of word-based embedding techniques (\wtv, \glv) offers quite a different picture: The extrinsic instability decreases with word frequency. 

Since any corpus -- independent of its size -- is only a snapshot of the respective language, this observation leads us to the somewhat expectable conclusion, that the quality of the embedding of a word based on \wtv\ and \glv, i.e. how well it resembles the meaning of the word in the language as a whole and not only in the given corpus, increases with the word frequency. 

\subsubsection*{Observation 4: The extrinsic instability of \ftt\ (sub-word based) is constant over word frequency}

For the sub-word based embedding technique \ftt\ not only the intrinsic but also the extrinsic instability is independent of the word frequency. We suspect the following reason for this: Since the technique is implicitly learning sub-word embeddings and the embeddings of vocabulary words are derived from these, the number of training samples that are used to construct the embeddings of a word does not directly depend on the frequency of the word itself.

Combining the conclusion from above with this observation yields, that the quality of the \ftt\ embedding of a word is not expected to decrease for rare words -- hence we can expect especially the embeddings of low-frequency words to be superior to word-based approaches like \wtv\ and \glv. This might explain why \ftt\ seems to be able to better abstract from the corpus it was trained on to the underlying language, as shown in Section \ref{sec_understanding_instability}.

%% file: tables/tab_ftt_baseline_differences.tex
\begin{tabular}{c|cc|c}
\textbf{Language} & \textbf{Lowest Score} & \textbf{Highest Score} & \textbf{Rel. Difference} \\
\hline
\textsc{Hi} 		& $16.07$ 	& $18.24$ 	& $13.5\,\%$ 	\\
\textsc{Fi} 		& $38.63$ 	& $47.40$ 	& $22.7\,\%$ 	\\
\textsc{Zh} 		& $52.55$ 	& $59.50$ 	& $13.2\,\%$ 	\\
\textsc{Cs} 		& $61.60$ 	& $64.36$ 	& $4.5\,\%$	 	\\
\textsc{Pl} 		& $55.44$ 	& $60.20$ 	& $8.6\,\%$	 	\\
\textsc{Pt} 		& $55.42$ 	& $57.67$ 	& $4.1\,\%$ 	\\
\textsc{En}			& $73.74$	& $74.83$	& $1.5\,\%$ 	\\
\end{tabular}

%% file: tikz/fig_random_orientation_of_embeddings.tex
\begin{tikzpicture}
\begin{axis}
[empty, name = plot1]
\coordinate (center_1) 	at (axis cs:+0.00,+0.00) {};
\coordinate (dog_1) 	at (axis cs:+1.00,+1.00) {};
\coordinate (cat_1) 	at (axis cs:+1.05,+0.75) {};
\coordinate (car_1) 	at (axis cs:-1.10,-0.50) {};
\coordinate (label_1) 	at (axis cs:+0.00,-1.25) {};
\end{axis}		
\draw [blue,  dotted, -latex] (center_1) -- (dog_1)  node [anchor=south] 	{\small{\texttt{dog}}};	
\draw [red,   dotted, -latex] (center_1) -- (cat_1)  node [anchor=west] 	{\small{\texttt{cat}}};	
\draw [green, dotted, -latex] (center_1) -- (car_1)  node [anchor=east] 	{\small{\texttt{car}}};	
\node at					   (label_1) [black]{Embedding space $\mathbf{V}_i$};

\begin{axis}
[empty, name = plot2, at=(plot1.right of south east), anchor=left of south west]
\coordinate (center_2) 	at (axis cs:+0.00,+0.00) {};
\coordinate (dog_2) 	at (axis cs:-1.00,+1.00) {};
\coordinate (cat_2) 	at (axis cs:-1.08,+0.60) {};
\coordinate (car_2) 	at (axis cs:+1.10,-0.50) {};
\coordinate (label_2) 	at (axis cs:+0.00,-1.25) {};
\end{axis}		
\draw [blue,  dashed, -latex] (center_2) -- (dog_2)  node [anchor=south] 	{\small{\texttt{dog}}};	
\draw [red,   dashed, -latex] (center_2) -- (cat_2)  node [anchor=east] 	{\small{\texttt{cat}}};	
\draw [green, dashed, -latex] (center_2) -- (car_2)  node [anchor=west] 	{\small{\texttt{car}}};	
\node at					   (label_2) [black]{Embedding space $\mathbf{V}_j$};				
\end{tikzpicture}

%% file: plots/gauss_cat_dog.pgf
\begingroup%
\makeatletter%
\begin{pgfpicture}%
\pgfpathrectangle{\pgfpointorigin}{\pgfqpoint{3.500000in}{2.800000in}}%
\pgfusepath{use as bounding box, clip}%
\begin{pgfscope}%
\pgfsetbuttcap%
\pgfsetmiterjoin%
\definecolor{currentfill}{rgb}{1.000000,1.000000,1.000000}%
\pgfsetfillcolor{currentfill}%
\pgfsetlinewidth{0.000000pt}%
\definecolor{currentstroke}{rgb}{1.000000,1.000000,1.000000}%
\pgfsetstrokecolor{currentstroke}%
\pgfsetdash{}{0pt}%
\pgfpathmoveto{\pgfqpoint{0.000000in}{0.000000in}}%
\pgfpathlineto{\pgfqpoint{3.500000in}{0.000000in}}%
\pgfpathlineto{\pgfqpoint{3.500000in}{2.800000in}}%
\pgfpathlineto{\pgfqpoint{0.000000in}{2.800000in}}%
\pgfpathclose%
\pgfusepath{fill}%
\end{pgfscope}%
\begin{pgfscope}%
\pgfsetbuttcap%
\pgfsetmiterjoin%
\definecolor{currentfill}{rgb}{1.000000,1.000000,1.000000}%
\pgfsetfillcolor{currentfill}%
\pgfsetlinewidth{0.000000pt}%
\definecolor{currentstroke}{rgb}{0.000000,0.000000,0.000000}%
\pgfsetstrokecolor{currentstroke}%
\pgfsetstrokeopacity{0.000000}%
\pgfsetdash{}{0pt}%
\pgfpathmoveto{\pgfqpoint{0.619028in}{0.582778in}}%
\pgfpathlineto{\pgfqpoint{3.350000in}{0.582778in}}%
\pgfpathlineto{\pgfqpoint{3.350000in}{2.650000in}}%
\pgfpathlineto{\pgfqpoint{0.619028in}{2.650000in}}%
\pgfpathclose%
\pgfusepath{fill}%
\end{pgfscope}%
\begin{pgfscope}%
\pgfpathrectangle{\pgfqpoint{0.619028in}{0.582778in}}{\pgfqpoint{2.730972in}{2.067222in}}%
\pgfusepath{clip}%
\pgfsetbuttcap%
\pgfsetmiterjoin%
\definecolor{currentfill}{rgb}{0.121569,0.466667,0.705882}%
\pgfsetfillcolor{currentfill}%
\pgfsetlinewidth{1.003750pt}%
\definecolor{currentstroke}{rgb}{0.000000,0.000000,0.000000}%
\pgfsetstrokecolor{currentstroke}%
\pgfsetdash{}{0pt}%
\pgfpathmoveto{\pgfqpoint{0.743163in}{0.582778in}}%
\pgfpathlineto{\pgfqpoint{0.908676in}{0.582778in}}%
\pgfpathlineto{\pgfqpoint{0.908676in}{0.646287in}}%
\pgfpathlineto{\pgfqpoint{0.743163in}{0.646287in}}%
\pgfpathclose%
\pgfusepath{stroke,fill}%
\end{pgfscope}%
\begin{pgfscope}%
\pgfpathrectangle{\pgfqpoint{0.619028in}{0.582778in}}{\pgfqpoint{2.730972in}{2.067222in}}%
\pgfusepath{clip}%
\pgfsetbuttcap%
\pgfsetmiterjoin%
\definecolor{currentfill}{rgb}{0.121569,0.466667,0.705882}%
\pgfsetfillcolor{currentfill}%
\pgfsetlinewidth{1.003750pt}%
\definecolor{currentstroke}{rgb}{0.000000,0.000000,0.000000}%
\pgfsetstrokecolor{currentstroke}%
\pgfsetdash{}{0pt}%
\pgfpathmoveto{\pgfqpoint{0.908676in}{0.582778in}}%
\pgfpathlineto{\pgfqpoint{1.074190in}{0.582778in}}%
\pgfpathlineto{\pgfqpoint{1.074190in}{0.582778in}}%
\pgfpathlineto{\pgfqpoint{0.908676in}{0.582778in}}%
\pgfpathclose%
\pgfusepath{stroke,fill}%
\end{pgfscope}%
\begin{pgfscope}%
\pgfpathrectangle{\pgfqpoint{0.619028in}{0.582778in}}{\pgfqpoint{2.730972in}{2.067222in}}%
\pgfusepath{clip}%
\pgfsetbuttcap%
\pgfsetmiterjoin%
\definecolor{currentfill}{rgb}{0.121569,0.466667,0.705882}%
\pgfsetfillcolor{currentfill}%
\pgfsetlinewidth{1.003750pt}%
\definecolor{currentstroke}{rgb}{0.000000,0.000000,0.000000}%
\pgfsetstrokecolor{currentstroke}%
\pgfsetdash{}{0pt}%
\pgfpathmoveto{\pgfqpoint{1.074190in}{0.582778in}}%
\pgfpathlineto{\pgfqpoint{1.239703in}{0.582778in}}%
\pgfpathlineto{\pgfqpoint{1.239703in}{0.773305in}}%
\pgfpathlineto{\pgfqpoint{1.074190in}{0.773305in}}%
\pgfpathclose%
\pgfusepath{stroke,fill}%
\end{pgfscope}%
\begin{pgfscope}%
\pgfpathrectangle{\pgfqpoint{0.619028in}{0.582778in}}{\pgfqpoint{2.730972in}{2.067222in}}%
\pgfusepath{clip}%
\pgfsetbuttcap%
\pgfsetmiterjoin%
\definecolor{currentfill}{rgb}{0.121569,0.466667,0.705882}%
\pgfsetfillcolor{currentfill}%
\pgfsetlinewidth{1.003750pt}%
\definecolor{currentstroke}{rgb}{0.000000,0.000000,0.000000}%
\pgfsetstrokecolor{currentstroke}%
\pgfsetdash{}{0pt}%
\pgfpathmoveto{\pgfqpoint{1.239703in}{0.582778in}}%
\pgfpathlineto{\pgfqpoint{1.405217in}{0.582778in}}%
\pgfpathlineto{\pgfqpoint{1.405217in}{0.773305in}}%
\pgfpathlineto{\pgfqpoint{1.239703in}{0.773305in}}%
\pgfpathclose%
\pgfusepath{stroke,fill}%
\end{pgfscope}%
\begin{pgfscope}%
\pgfpathrectangle{\pgfqpoint{0.619028in}{0.582778in}}{\pgfqpoint{2.730972in}{2.067222in}}%
\pgfusepath{clip}%
\pgfsetbuttcap%
\pgfsetmiterjoin%
\definecolor{currentfill}{rgb}{0.121569,0.466667,0.705882}%
\pgfsetfillcolor{currentfill}%
\pgfsetlinewidth{1.003750pt}%
\definecolor{currentstroke}{rgb}{0.000000,0.000000,0.000000}%
\pgfsetstrokecolor{currentstroke}%
\pgfsetdash{}{0pt}%
\pgfpathmoveto{\pgfqpoint{1.405217in}{0.582778in}}%
\pgfpathlineto{\pgfqpoint{1.570730in}{0.582778in}}%
\pgfpathlineto{\pgfqpoint{1.570730in}{0.900323in}}%
\pgfpathlineto{\pgfqpoint{1.405217in}{0.900323in}}%
\pgfpathclose%
\pgfusepath{stroke,fill}%
\end{pgfscope}%
\begin{pgfscope}%
\pgfpathrectangle{\pgfqpoint{0.619028in}{0.582778in}}{\pgfqpoint{2.730972in}{2.067222in}}%
\pgfusepath{clip}%
\pgfsetbuttcap%
\pgfsetmiterjoin%
\definecolor{currentfill}{rgb}{0.121569,0.466667,0.705882}%
\pgfsetfillcolor{currentfill}%
\pgfsetlinewidth{1.003750pt}%
\definecolor{currentstroke}{rgb}{0.000000,0.000000,0.000000}%
\pgfsetstrokecolor{currentstroke}%
\pgfsetdash{}{0pt}%
\pgfpathmoveto{\pgfqpoint{1.570730in}{0.582778in}}%
\pgfpathlineto{\pgfqpoint{1.736244in}{0.582778in}}%
\pgfpathlineto{\pgfqpoint{1.736244in}{1.217869in}}%
\pgfpathlineto{\pgfqpoint{1.570730in}{1.217869in}}%
\pgfpathclose%
\pgfusepath{stroke,fill}%
\end{pgfscope}%
\begin{pgfscope}%
\pgfpathrectangle{\pgfqpoint{0.619028in}{0.582778in}}{\pgfqpoint{2.730972in}{2.067222in}}%
\pgfusepath{clip}%
\pgfsetbuttcap%
\pgfsetmiterjoin%
\definecolor{currentfill}{rgb}{0.121569,0.466667,0.705882}%
\pgfsetfillcolor{currentfill}%
\pgfsetlinewidth{1.003750pt}%
\definecolor{currentstroke}{rgb}{0.000000,0.000000,0.000000}%
\pgfsetstrokecolor{currentstroke}%
\pgfsetdash{}{0pt}%
\pgfpathmoveto{\pgfqpoint{1.736244in}{0.582778in}}%
\pgfpathlineto{\pgfqpoint{1.901757in}{0.582778in}}%
\pgfpathlineto{\pgfqpoint{1.901757in}{1.471906in}}%
\pgfpathlineto{\pgfqpoint{1.736244in}{1.471906in}}%
\pgfpathclose%
\pgfusepath{stroke,fill}%
\end{pgfscope}%
\begin{pgfscope}%
\pgfpathrectangle{\pgfqpoint{0.619028in}{0.582778in}}{\pgfqpoint{2.730972in}{2.067222in}}%
\pgfusepath{clip}%
\pgfsetbuttcap%
\pgfsetmiterjoin%
\definecolor{currentfill}{rgb}{0.121569,0.466667,0.705882}%
\pgfsetfillcolor{currentfill}%
\pgfsetlinewidth{1.003750pt}%
\definecolor{currentstroke}{rgb}{0.000000,0.000000,0.000000}%
\pgfsetstrokecolor{currentstroke}%
\pgfsetdash{}{0pt}%
\pgfpathmoveto{\pgfqpoint{1.901757in}{0.582778in}}%
\pgfpathlineto{\pgfqpoint{2.067271in}{0.582778in}}%
\pgfpathlineto{\pgfqpoint{2.067271in}{1.789451in}}%
\pgfpathlineto{\pgfqpoint{1.901757in}{1.789451in}}%
\pgfpathclose%
\pgfusepath{stroke,fill}%
\end{pgfscope}%
\begin{pgfscope}%
\pgfpathrectangle{\pgfqpoint{0.619028in}{0.582778in}}{\pgfqpoint{2.730972in}{2.067222in}}%
\pgfusepath{clip}%
\pgfsetbuttcap%
\pgfsetmiterjoin%
\definecolor{currentfill}{rgb}{0.121569,0.466667,0.705882}%
\pgfsetfillcolor{currentfill}%
\pgfsetlinewidth{1.003750pt}%
\definecolor{currentstroke}{rgb}{0.000000,0.000000,0.000000}%
\pgfsetstrokecolor{currentstroke}%
\pgfsetdash{}{0pt}%
\pgfpathmoveto{\pgfqpoint{2.067271in}{0.582778in}}%
\pgfpathlineto{\pgfqpoint{2.232784in}{0.582778in}}%
\pgfpathlineto{\pgfqpoint{2.232784in}{2.551561in}}%
\pgfpathlineto{\pgfqpoint{2.067271in}{2.551561in}}%
\pgfpathclose%
\pgfusepath{stroke,fill}%
\end{pgfscope}%
\begin{pgfscope}%
\pgfpathrectangle{\pgfqpoint{0.619028in}{0.582778in}}{\pgfqpoint{2.730972in}{2.067222in}}%
\pgfusepath{clip}%
\pgfsetbuttcap%
\pgfsetmiterjoin%
\definecolor{currentfill}{rgb}{0.121569,0.466667,0.705882}%
\pgfsetfillcolor{currentfill}%
\pgfsetlinewidth{1.003750pt}%
\definecolor{currentstroke}{rgb}{0.000000,0.000000,0.000000}%
\pgfsetstrokecolor{currentstroke}%
\pgfsetdash{}{0pt}%
\pgfpathmoveto{\pgfqpoint{2.232784in}{0.582778in}}%
\pgfpathlineto{\pgfqpoint{2.398298in}{0.582778in}}%
\pgfpathlineto{\pgfqpoint{2.398298in}{1.725942in}}%
\pgfpathlineto{\pgfqpoint{2.232784in}{1.725942in}}%
\pgfpathclose%
\pgfusepath{stroke,fill}%
\end{pgfscope}%
\begin{pgfscope}%
\pgfpathrectangle{\pgfqpoint{0.619028in}{0.582778in}}{\pgfqpoint{2.730972in}{2.067222in}}%
\pgfusepath{clip}%
\pgfsetbuttcap%
\pgfsetmiterjoin%
\definecolor{currentfill}{rgb}{0.121569,0.466667,0.705882}%
\pgfsetfillcolor{currentfill}%
\pgfsetlinewidth{1.003750pt}%
\definecolor{currentstroke}{rgb}{0.000000,0.000000,0.000000}%
\pgfsetstrokecolor{currentstroke}%
\pgfsetdash{}{0pt}%
\pgfpathmoveto{\pgfqpoint{2.398298in}{0.582778in}}%
\pgfpathlineto{\pgfqpoint{2.563811in}{0.582778in}}%
\pgfpathlineto{\pgfqpoint{2.563811in}{1.598924in}}%
\pgfpathlineto{\pgfqpoint{2.398298in}{1.598924in}}%
\pgfpathclose%
\pgfusepath{stroke,fill}%
\end{pgfscope}%
\begin{pgfscope}%
\pgfpathrectangle{\pgfqpoint{0.619028in}{0.582778in}}{\pgfqpoint{2.730972in}{2.067222in}}%
\pgfusepath{clip}%
\pgfsetbuttcap%
\pgfsetmiterjoin%
\definecolor{currentfill}{rgb}{0.121569,0.466667,0.705882}%
\pgfsetfillcolor{currentfill}%
\pgfsetlinewidth{1.003750pt}%
\definecolor{currentstroke}{rgb}{0.000000,0.000000,0.000000}%
\pgfsetstrokecolor{currentstroke}%
\pgfsetdash{}{0pt}%
\pgfpathmoveto{\pgfqpoint{2.563811in}{0.582778in}}%
\pgfpathlineto{\pgfqpoint{2.729324in}{0.582778in}}%
\pgfpathlineto{\pgfqpoint{2.729324in}{0.900323in}}%
\pgfpathlineto{\pgfqpoint{2.563811in}{0.900323in}}%
\pgfpathclose%
\pgfusepath{stroke,fill}%
\end{pgfscope}%
\begin{pgfscope}%
\pgfpathrectangle{\pgfqpoint{0.619028in}{0.582778in}}{\pgfqpoint{2.730972in}{2.067222in}}%
\pgfusepath{clip}%
\pgfsetbuttcap%
\pgfsetmiterjoin%
\definecolor{currentfill}{rgb}{0.121569,0.466667,0.705882}%
\pgfsetfillcolor{currentfill}%
\pgfsetlinewidth{1.003750pt}%
\definecolor{currentstroke}{rgb}{0.000000,0.000000,0.000000}%
\pgfsetstrokecolor{currentstroke}%
\pgfsetdash{}{0pt}%
\pgfpathmoveto{\pgfqpoint{2.729324in}{0.582778in}}%
\pgfpathlineto{\pgfqpoint{2.894838in}{0.582778in}}%
\pgfpathlineto{\pgfqpoint{2.894838in}{0.709796in}}%
\pgfpathlineto{\pgfqpoint{2.729324in}{0.709796in}}%
\pgfpathclose%
\pgfusepath{stroke,fill}%
\end{pgfscope}%
\begin{pgfscope}%
\pgfpathrectangle{\pgfqpoint{0.619028in}{0.582778in}}{\pgfqpoint{2.730972in}{2.067222in}}%
\pgfusepath{clip}%
\pgfsetbuttcap%
\pgfsetmiterjoin%
\definecolor{currentfill}{rgb}{0.121569,0.466667,0.705882}%
\pgfsetfillcolor{currentfill}%
\pgfsetlinewidth{1.003750pt}%
\definecolor{currentstroke}{rgb}{0.000000,0.000000,0.000000}%
\pgfsetstrokecolor{currentstroke}%
\pgfsetdash{}{0pt}%
\pgfpathmoveto{\pgfqpoint{2.894838in}{0.582778in}}%
\pgfpathlineto{\pgfqpoint{3.060351in}{0.582778in}}%
\pgfpathlineto{\pgfqpoint{3.060351in}{0.582778in}}%
\pgfpathlineto{\pgfqpoint{2.894838in}{0.582778in}}%
\pgfpathclose%
\pgfusepath{stroke,fill}%
\end{pgfscope}%
\begin{pgfscope}%
\pgfpathrectangle{\pgfqpoint{0.619028in}{0.582778in}}{\pgfqpoint{2.730972in}{2.067222in}}%
\pgfusepath{clip}%
\pgfsetbuttcap%
\pgfsetmiterjoin%
\definecolor{currentfill}{rgb}{0.121569,0.466667,0.705882}%
\pgfsetfillcolor{currentfill}%
\pgfsetlinewidth{1.003750pt}%
\definecolor{currentstroke}{rgb}{0.000000,0.000000,0.000000}%
\pgfsetstrokecolor{currentstroke}%
\pgfsetdash{}{0pt}%
\pgfpathmoveto{\pgfqpoint{3.060351in}{0.582778in}}%
\pgfpathlineto{\pgfqpoint{3.225865in}{0.582778in}}%
\pgfpathlineto{\pgfqpoint{3.225865in}{0.646287in}}%
\pgfpathlineto{\pgfqpoint{3.060351in}{0.646287in}}%
\pgfpathclose%
\pgfusepath{stroke,fill}%
\end{pgfscope}%
\begin{pgfscope}%
\pgfsetbuttcap%
\pgfsetroundjoin%
\definecolor{currentfill}{rgb}{0.000000,0.000000,0.000000}%
\pgfsetfillcolor{currentfill}%
\pgfsetlinewidth{0.803000pt}%
\definecolor{currentstroke}{rgb}{0.000000,0.000000,0.000000}%
\pgfsetstrokecolor{currentstroke}%
\pgfsetdash{}{0pt}%
\pgfsys@defobject{currentmarker}{\pgfqpoint{0.000000in}{-0.048611in}}{\pgfqpoint{0.000000in}{0.000000in}}{%
\pgfpathmoveto{\pgfqpoint{0.000000in}{0.000000in}}%
\pgfpathlineto{\pgfqpoint{0.000000in}{-0.048611in}}%
\pgfusepath{stroke,fill}%
}%
\begin{pgfscope}%
\pgfsys@transformshift{1.294970in}{0.582778in}%
\pgfsys@useobject{currentmarker}{}%
\end{pgfscope}%
\end{pgfscope}%
\begin{pgfscope}%
\definecolor{textcolor}{rgb}{0.000000,0.000000,0.000000}%
\pgfsetstrokecolor{textcolor}%
\pgfsetfillcolor{textcolor}%
\pgftext[x=1.294970in,y=0.485556in,,top]{\color{textcolor}\rmfamily\fontsize{10.000000}{12.000000}\selectfont 0.66}%
\end{pgfscope}%
\begin{pgfscope}%
\pgfsetbuttcap%
\pgfsetroundjoin%
\definecolor{currentfill}{rgb}{0.000000,0.000000,0.000000}%
\pgfsetfillcolor{currentfill}%
\pgfsetlinewidth{0.803000pt}%
\definecolor{currentstroke}{rgb}{0.000000,0.000000,0.000000}%
\pgfsetstrokecolor{currentstroke}%
\pgfsetdash{}{0pt}%
\pgfsys@defobject{currentmarker}{\pgfqpoint{0.000000in}{-0.048611in}}{\pgfqpoint{0.000000in}{0.000000in}}{%
\pgfpathmoveto{\pgfqpoint{0.000000in}{0.000000in}}%
\pgfpathlineto{\pgfqpoint{0.000000in}{-0.048611in}}%
\pgfusepath{stroke,fill}%
}%
\begin{pgfscope}%
\pgfsys@transformshift{2.061868in}{0.582778in}%
\pgfsys@useobject{currentmarker}{}%
\end{pgfscope}%
\end{pgfscope}%
\begin{pgfscope}%
\definecolor{textcolor}{rgb}{0.000000,0.000000,0.000000}%
\pgfsetstrokecolor{textcolor}%
\pgfsetfillcolor{textcolor}%
\pgftext[x=2.061868in,y=0.485556in,,top]{\color{textcolor}\rmfamily\fontsize{10.000000}{12.000000}\selectfont 0.68}%
\end{pgfscope}%
\begin{pgfscope}%
\pgfsetbuttcap%
\pgfsetroundjoin%
\definecolor{currentfill}{rgb}{0.000000,0.000000,0.000000}%
\pgfsetfillcolor{currentfill}%
\pgfsetlinewidth{0.803000pt}%
\definecolor{currentstroke}{rgb}{0.000000,0.000000,0.000000}%
\pgfsetstrokecolor{currentstroke}%
\pgfsetdash{}{0pt}%
\pgfsys@defobject{currentmarker}{\pgfqpoint{0.000000in}{-0.048611in}}{\pgfqpoint{0.000000in}{0.000000in}}{%
\pgfpathmoveto{\pgfqpoint{0.000000in}{0.000000in}}%
\pgfpathlineto{\pgfqpoint{0.000000in}{-0.048611in}}%
\pgfusepath{stroke,fill}%
}%
\begin{pgfscope}%
\pgfsys@transformshift{2.828767in}{0.582778in}%
\pgfsys@useobject{currentmarker}{}%
\end{pgfscope}%
\end{pgfscope}%
\begin{pgfscope}%
\definecolor{textcolor}{rgb}{0.000000,0.000000,0.000000}%
\pgfsetstrokecolor{textcolor}%
\pgfsetfillcolor{textcolor}%
\pgftext[x=2.828767in,y=0.485556in,,top]{\color{textcolor}\rmfamily\fontsize{10.000000}{12.000000}\selectfont 0.70}%
\end{pgfscope}%
\begin{pgfscope}%
\definecolor{textcolor}{rgb}{0.000000,0.000000,0.000000}%
\pgfsetstrokecolor{textcolor}%
\pgfsetfillcolor{textcolor}%
\pgftext[x=1.984514in,y=0.307345in,,top]{\color{textcolor}\rmfamily\fontsize{10.000000}{12.000000}\selectfont Cosine Similarity}%
\end{pgfscope}%
\begin{pgfscope}%
\pgfsetbuttcap%
\pgfsetroundjoin%
\definecolor{currentfill}{rgb}{0.000000,0.000000,0.000000}%
\pgfsetfillcolor{currentfill}%
\pgfsetlinewidth{0.803000pt}%
\definecolor{currentstroke}{rgb}{0.000000,0.000000,0.000000}%
\pgfsetstrokecolor{currentstroke}%
\pgfsetdash{}{0pt}%
\pgfsys@defobject{currentmarker}{\pgfqpoint{-0.048611in}{0.000000in}}{\pgfqpoint{0.000000in}{0.000000in}}{%
\pgfpathmoveto{\pgfqpoint{0.000000in}{0.000000in}}%
\pgfpathlineto{\pgfqpoint{-0.048611in}{0.000000in}}%
\pgfusepath{stroke,fill}%
}%
\begin{pgfscope}%
\pgfsys@transformshift{0.619028in}{0.582778in}%
\pgfsys@useobject{currentmarker}{}%
\end{pgfscope}%
\end{pgfscope}%
\begin{pgfscope}%
\definecolor{textcolor}{rgb}{0.000000,0.000000,0.000000}%
\pgfsetstrokecolor{textcolor}%
\pgfsetfillcolor{textcolor}%
\pgftext[x=0.452378in,y=0.534950in,left,base]{\color{textcolor}\rmfamily\fontsize{10.000000}{12.000000}\selectfont 0}%
\end{pgfscope}%
\begin{pgfscope}%
\pgfsetbuttcap%
\pgfsetroundjoin%
\definecolor{currentfill}{rgb}{0.000000,0.000000,0.000000}%
\pgfsetfillcolor{currentfill}%
\pgfsetlinewidth{0.803000pt}%
\definecolor{currentstroke}{rgb}{0.000000,0.000000,0.000000}%
\pgfsetstrokecolor{currentstroke}%
\pgfsetdash{}{0pt}%
\pgfsys@defobject{currentmarker}{\pgfqpoint{-0.048611in}{0.000000in}}{\pgfqpoint{0.000000in}{0.000000in}}{%
\pgfpathmoveto{\pgfqpoint{0.000000in}{0.000000in}}%
\pgfpathlineto{\pgfqpoint{-0.048611in}{0.000000in}}%
\pgfusepath{stroke,fill}%
}%
\begin{pgfscope}%
\pgfsys@transformshift{0.619028in}{0.900323in}%
\pgfsys@useobject{currentmarker}{}%
\end{pgfscope}%
\end{pgfscope}%
\begin{pgfscope}%
\definecolor{textcolor}{rgb}{0.000000,0.000000,0.000000}%
\pgfsetstrokecolor{textcolor}%
\pgfsetfillcolor{textcolor}%
\pgftext[x=0.452378in,y=0.852496in,left,base]{\color{textcolor}\rmfamily\fontsize{10.000000}{12.000000}\selectfont 5}%
\end{pgfscope}%
\begin{pgfscope}%
\pgfsetbuttcap%
\pgfsetroundjoin%
\definecolor{currentfill}{rgb}{0.000000,0.000000,0.000000}%
\pgfsetfillcolor{currentfill}%
\pgfsetlinewidth{0.803000pt}%
\definecolor{currentstroke}{rgb}{0.000000,0.000000,0.000000}%
\pgfsetstrokecolor{currentstroke}%
\pgfsetdash{}{0pt}%
\pgfsys@defobject{currentmarker}{\pgfqpoint{-0.048611in}{0.000000in}}{\pgfqpoint{0.000000in}{0.000000in}}{%
\pgfpathmoveto{\pgfqpoint{0.000000in}{0.000000in}}%
\pgfpathlineto{\pgfqpoint{-0.048611in}{0.000000in}}%
\pgfusepath{stroke,fill}%
}%
\begin{pgfscope}%
\pgfsys@transformshift{0.619028in}{1.217869in}%
\pgfsys@useobject{currentmarker}{}%
\end{pgfscope}%
\end{pgfscope}%
\begin{pgfscope}%
\definecolor{textcolor}{rgb}{0.000000,0.000000,0.000000}%
\pgfsetstrokecolor{textcolor}%
\pgfsetfillcolor{textcolor}%
\pgftext[x=0.382951in,y=1.170041in,left,base]{\color{textcolor}\rmfamily\fontsize{10.000000}{12.000000}\selectfont 10}%
\end{pgfscope}%
\begin{pgfscope}%
\pgfsetbuttcap%
\pgfsetroundjoin%
\definecolor{currentfill}{rgb}{0.000000,0.000000,0.000000}%
\pgfsetfillcolor{currentfill}%
\pgfsetlinewidth{0.803000pt}%
\definecolor{currentstroke}{rgb}{0.000000,0.000000,0.000000}%
\pgfsetstrokecolor{currentstroke}%
\pgfsetdash{}{0pt}%
\pgfsys@defobject{currentmarker}{\pgfqpoint{-0.048611in}{0.000000in}}{\pgfqpoint{0.000000in}{0.000000in}}{%
\pgfpathmoveto{\pgfqpoint{0.000000in}{0.000000in}}%
\pgfpathlineto{\pgfqpoint{-0.048611in}{0.000000in}}%
\pgfusepath{stroke,fill}%
}%
\begin{pgfscope}%
\pgfsys@transformshift{0.619028in}{1.535415in}%
\pgfsys@useobject{currentmarker}{}%
\end{pgfscope}%
\end{pgfscope}%
\begin{pgfscope}%
\definecolor{textcolor}{rgb}{0.000000,0.000000,0.000000}%
\pgfsetstrokecolor{textcolor}%
\pgfsetfillcolor{textcolor}%
\pgftext[x=0.382951in,y=1.487587in,left,base]{\color{textcolor}\rmfamily\fontsize{10.000000}{12.000000}\selectfont 15}%
\end{pgfscope}%
\begin{pgfscope}%
\pgfsetbuttcap%
\pgfsetroundjoin%
\definecolor{currentfill}{rgb}{0.000000,0.000000,0.000000}%
\pgfsetfillcolor{currentfill}%
\pgfsetlinewidth{0.803000pt}%
\definecolor{currentstroke}{rgb}{0.000000,0.000000,0.000000}%
\pgfsetstrokecolor{currentstroke}%
\pgfsetdash{}{0pt}%
\pgfsys@defobject{currentmarker}{\pgfqpoint{-0.048611in}{0.000000in}}{\pgfqpoint{0.000000in}{0.000000in}}{%
\pgfpathmoveto{\pgfqpoint{0.000000in}{0.000000in}}%
\pgfpathlineto{\pgfqpoint{-0.048611in}{0.000000in}}%
\pgfusepath{stroke,fill}%
}%
\begin{pgfscope}%
\pgfsys@transformshift{0.619028in}{1.852960in}%
\pgfsys@useobject{currentmarker}{}%
\end{pgfscope}%
\end{pgfscope}%
\begin{pgfscope}%
\definecolor{textcolor}{rgb}{0.000000,0.000000,0.000000}%
\pgfsetstrokecolor{textcolor}%
\pgfsetfillcolor{textcolor}%
\pgftext[x=0.382951in,y=1.805133in,left,base]{\color{textcolor}\rmfamily\fontsize{10.000000}{12.000000}\selectfont 20}%
\end{pgfscope}%
\begin{pgfscope}%
\pgfsetbuttcap%
\pgfsetroundjoin%
\definecolor{currentfill}{rgb}{0.000000,0.000000,0.000000}%
\pgfsetfillcolor{currentfill}%
\pgfsetlinewidth{0.803000pt}%
\definecolor{currentstroke}{rgb}{0.000000,0.000000,0.000000}%
\pgfsetstrokecolor{currentstroke}%
\pgfsetdash{}{0pt}%
\pgfsys@defobject{currentmarker}{\pgfqpoint{-0.048611in}{0.000000in}}{\pgfqpoint{0.000000in}{0.000000in}}{%
\pgfpathmoveto{\pgfqpoint{0.000000in}{0.000000in}}%
\pgfpathlineto{\pgfqpoint{-0.048611in}{0.000000in}}%
\pgfusepath{stroke,fill}%
}%
\begin{pgfscope}%
\pgfsys@transformshift{0.619028in}{2.170506in}%
\pgfsys@useobject{currentmarker}{}%
\end{pgfscope}%
\end{pgfscope}%
\begin{pgfscope}%
\definecolor{textcolor}{rgb}{0.000000,0.000000,0.000000}%
\pgfsetstrokecolor{textcolor}%
\pgfsetfillcolor{textcolor}%
\pgftext[x=0.382951in,y=2.122678in,left,base]{\color{textcolor}\rmfamily\fontsize{10.000000}{12.000000}\selectfont 25}%
\end{pgfscope}%
\begin{pgfscope}%
\pgfsetbuttcap%
\pgfsetroundjoin%
\definecolor{currentfill}{rgb}{0.000000,0.000000,0.000000}%
\pgfsetfillcolor{currentfill}%
\pgfsetlinewidth{0.803000pt}%
\definecolor{currentstroke}{rgb}{0.000000,0.000000,0.000000}%
\pgfsetstrokecolor{currentstroke}%
\pgfsetdash{}{0pt}%
\pgfsys@defobject{currentmarker}{\pgfqpoint{-0.048611in}{0.000000in}}{\pgfqpoint{0.000000in}{0.000000in}}{%
\pgfpathmoveto{\pgfqpoint{0.000000in}{0.000000in}}%
\pgfpathlineto{\pgfqpoint{-0.048611in}{0.000000in}}%
\pgfusepath{stroke,fill}%
}%
\begin{pgfscope}%
\pgfsys@transformshift{0.619028in}{2.488052in}%
\pgfsys@useobject{currentmarker}{}%
\end{pgfscope}%
\end{pgfscope}%
\begin{pgfscope}%
\definecolor{textcolor}{rgb}{0.000000,0.000000,0.000000}%
\pgfsetstrokecolor{textcolor}%
\pgfsetfillcolor{textcolor}%
\pgftext[x=0.382951in,y=2.440224in,left,base]{\color{textcolor}\rmfamily\fontsize{10.000000}{12.000000}\selectfont 30}%
\end{pgfscope}%
\begin{pgfscope}%
\definecolor{textcolor}{rgb}{0.000000,0.000000,0.000000}%
\pgfsetstrokecolor{textcolor}%
\pgfsetfillcolor{textcolor}%
\pgftext[x=0.327395in,y=1.616389in,,bottom,rotate=90.000000]{\color{textcolor}\rmfamily\fontsize{10.000000}{12.000000}\selectfont Count}%
\end{pgfscope}%
\begin{pgfscope}%
\pgfsetrectcap%
\pgfsetmiterjoin%
\pgfsetlinewidth{0.803000pt}%
\definecolor{currentstroke}{rgb}{0.000000,0.000000,0.000000}%
\pgfsetstrokecolor{currentstroke}%
\pgfsetdash{}{0pt}%
\pgfpathmoveto{\pgfqpoint{0.619028in}{0.582778in}}%
\pgfpathlineto{\pgfqpoint{0.619028in}{2.650000in}}%
\pgfusepath{stroke}%
\end{pgfscope}%
\begin{pgfscope}%
\pgfsetrectcap%
\pgfsetmiterjoin%
\pgfsetlinewidth{0.803000pt}%
\definecolor{currentstroke}{rgb}{0.000000,0.000000,0.000000}%
\pgfsetstrokecolor{currentstroke}%
\pgfsetdash{}{0pt}%
\pgfpathmoveto{\pgfqpoint{3.350000in}{0.582778in}}%
\pgfpathlineto{\pgfqpoint{3.350000in}{2.650000in}}%
\pgfusepath{stroke}%
\end{pgfscope}%
\begin{pgfscope}%
\pgfsetrectcap%
\pgfsetmiterjoin%
\pgfsetlinewidth{0.803000pt}%
\definecolor{currentstroke}{rgb}{0.000000,0.000000,0.000000}%
\pgfsetstrokecolor{currentstroke}%
\pgfsetdash{}{0pt}%
\pgfpathmoveto{\pgfqpoint{0.619028in}{0.582778in}}%
\pgfpathlineto{\pgfqpoint{3.350000in}{0.582778in}}%
\pgfusepath{stroke}%
\end{pgfscope}%
\begin{pgfscope}%
\pgfsetrectcap%
\pgfsetmiterjoin%
\pgfsetlinewidth{0.803000pt}%
\definecolor{currentstroke}{rgb}{0.000000,0.000000,0.000000}%
\pgfsetstrokecolor{currentstroke}%
\pgfsetdash{}{0pt}%
\pgfpathmoveto{\pgfqpoint{0.619028in}{2.650000in}}%
\pgfpathlineto{\pgfqpoint{3.350000in}{2.650000in}}%
\pgfusepath{stroke}%
\end{pgfscope}%
\end{pgfpicture}%
\makeatother%
\endgroup%

%% file: tables/tab_momentum_nn.tex
\begin{tabular}{lc|cccc|cccc}
\multicolumn{10}{c}{target word: \textbf{\texttt{momentum}}}\\[10pt]
\multirow{2}{*}{\textbf{word}} & \quad & \quad &
\multicolumn{2}{c}{\textbf{run \# 1}} & \quad & \quad &
\multicolumn{2}{c}{\textbf{run \# 2}} & \quad \\
&&& \textsl{rank} & \textsl{cos} &&&  \textsl{rank} & \textsl{cos} & \\
\hline 
&&&&&&&\\[-15pt]
\texttt{inertia}	 	&&&	1 	&   0.639	&&&	1 				&	0.639 & \\
\texttt{kinetic} 		&&&	2 	&	0.630	&&&	3 				&	0.613 & \\
\texttt{momenta} 		&&&	3	&	0.626	&&&	2 				&	0.615 & \\
\texttt{energy}		 	&&&	4	&	0.593	&&&	6				&	0.590 & \\
\texttt{centripetal} 	&&&	5	&	0.586	&&&	5 				&	0.592 & \\
\texttt{mass-energy}	&&&	6	&	0.581	&&&	9 				&	0.575 & \\
\texttt{vorticity}		&&&	7	&	0.578	&&&	4				&	0.593 & \\
\texttt{gravitational}	&&&	8	&	0.577	&&&	7				&	0.587 & \\
\texttt{angular}	 	&&&	9	&	0.576	&&&	\textbf{11}		&	0.570 & \\
\texttt{relativistic}	&&&	10	&	0.572	&&&	\textbf{16}	 	&	0.564 & \\
\hdashline[0.3pt/2.2pt] 
\texttt{eigenstate}		&&&	11	&	0.571	&&&	\textbf{18}		&	0.563 & \\
\texttt{spin}			&&&	12	&	0.569	&&&	\textbf{29}		&	0.546 & \\
\texttt{accelerating} 	&&&	13	&	0.568	&&&	\textbf{17}		&	0.564 & \\
\texttt{eigenstates}	&&&	14	&	0.566	&&&	14 				&	0.565 & \\
\texttt{velocity}		&&&	15	&	0.564	&&&	8			 	&	0.582 & \\
\hdashline[0.3pt/2.2pt] 
\end{tabular}

%% file: tikz/fig_previous_work_values_n.tex
\begin{tikzpicture}

\def\vertspacing{0.70}
\node [text width = 5cm] at (1,6*\vertspacing){\citet{antoniak2018}};
\node [text width = 5cm] at (1,5*\vertspacing){\citet{hellrich-hahn-2016-assessment}};
\node [text width = 5cm] at (1,4*\vertspacing){\citet{hellrich-hahn-2016-bad}};
\node [text width = 5cm] at (1,3*\vertspacing){\citet{hellrich-hahn-2017-fool}};
\node [text width = 5cm] at (1,2*\vertspacing){\citet{hellrich-etal-2019-influence}};
\node [text width = 5cm] at (1,1*\vertspacing){\citet{pierrejean-tanguy-2018-predicting}};
\node [text width = 5cm] at (1,0*\vertspacing){\citet{wendlandt2018}};

\draw (3.25,-0.5*\vertspacing) -- 	(3.25,6.5*\vertspacing);
\draw (10.0,-0.5*\vertspacing) -- 	(10.0,6.5*\vertspacing);

\draw [-] (3.25,-0.5*\vertspacing) 		-- 	(6.97,-0.5*\vertspacing);
\draw [-] (7.04,-0.3*\vertspacing) 		--	(6.90,-0.7*\vertspacing);
\draw [-] (7.10,-0.3*\vertspacing) 		-- 	(6.96,-0.7*\vertspacing);
\draw [-] (7.03,-0.5*\vertspacing) 		-- 	(8.47,-0.5*\vertspacing);
\draw [-] (8.54,-0.3*\vertspacing) 		--	(8.40,-0.7*\vertspacing);
\draw [-] (8.60,-0.3*\vertspacing) 		-- 	(8.46,-0.7*\vertspacing);

\foreach \x/\xtext in {1/1, 5/5, 10/10}
    \draw[shift={(\x / 5 + 4,-0.5*\vertspacing)}] (0pt,0pt) -- (0pt,-4pt) node[below] {$\xtext$};
\draw [-] (8.53,-0.5*\vertspacing) 	-- 	(10,-0.5*\vertspacing);   
\foreach \x/\xtext in {0/25}
    \draw[shift={(7.75,-0.5*\vertspacing)}] (0pt,0pt) -- (0pt,-4pt) node[below] {$\xtext$};
\foreach \x/\xtext in {0/50}
    \draw[shift={(9.25,-0.5*\vertspacing)}] (0pt,0pt) -- (0pt,-4pt) node[below] {$\xtext$};

\draw [-] 	(3.25, { ( -0.5 + 7 ) * \vertspacing }) 	-- 	(6.97, { ( -0.5 + 7 ) * \vertspacing });
\draw [-] 	(7.04, { ( -0.3 + 7 ) * \vertspacing }) 	--	(6.90, { ( -0.7 + 7 ) * \vertspacing });
\draw [-] 	(7.10, { ( -0.3 + 7 ) * \vertspacing }) 	-- 	(6.96, { ( -0.7 + 7 ) * \vertspacing });
\draw [-] 	(7.03, { ( -0.5 + 7 ) * \vertspacing }) 	-- 	(8.47, { ( -0.5 + 7 ) * \vertspacing });
\draw [-] 	(8.54, { ( -0.3 + 7 ) * \vertspacing }) 	--	(8.40, { ( -0.7 + 7 ) * \vertspacing });
\draw [-] 	(8.60, { ( -0.3 + 7 ) * \vertspacing }) 	-- 	(8.46, { ( -0.7 + 7 ) * \vertspacing });
\draw [-] 	(8.53, { ( -0.5 + 7 ) * \vertspacing }) 	-- 	(10,   { ( -0.5 + 7 ) * \vertspacing });   

\foreach \x/\xtext in {1/1, 5/5, 10/10}
    \draw[shift={(\x / 5 + 4, 	6.5 * \vertspacing)}] 	(0pt,0pt) -- (0pt,+4pt);
\foreach \x/\xtext in {0/25}
    \draw[shift={(7.75, 		6.5 * \vertspacing)}] 	(0pt,0pt) -- (0pt,+4pt);
\foreach \x/\xtext in {0/50}
	\draw[shift={(9.25, 		6.5 * \vertspacing)}] 	(0pt,0pt) -- (0pt,+4pt);

\tikzset{
	data/.style = {
		draw=black,
		fill=blue,
		fill opacity=0.2
	}
}

\def\rsize{0.20}
\foreach \y/\x in {6/2, 6/10, 3/10, 0/10}
	\draw [data] (4 - \rsize + \x/5, \y*\vertspacing - \rsize) rectangle (4 + \rsize + \x/5,\y*\vertspacing + \rsize) 
	node[opacity = 1, text = black,pos=.5] {\small \x};
\foreach \y/\x in {1/25}
	\draw [data] (7.75 - \rsize, \y*\vertspacing - \rsize) rectangle (7.75 + \rsize ,\y*\vertspacing + \rsize) 
	node[opacity = 1, text = black,pos=.5] {\small \x};	
\foreach \y/\xstart/\xend in {2/1/5, 5/1/10}
	\draw [data] (4 - \rsize + \xstart/5, \y*\vertspacing - \rsize) rectangle (4 + \rsize + \xend/5,\y*\vertspacing + \rsize) 
	node[opacity = 1, text = black,pos=.5] {\small \xstart\ - \xend};
\foreach \y/\xstart/\xend in {4/1/50}
	\draw [data] (4 - \rsize + \xstart/5, \y*\vertspacing - \rsize) rectangle (9.25 + \rsize,\y*\vertspacing + \rsize) 
	node[opacity = 1, text = black, pos=.5] {\small \xstart\ - \xend};
	
\end{tikzpicture}

%% file: tables/tab_dependance_of_pj_on_n.tex
\setlength\tabcolsep{3.0pt}
\resizebox{\columnwidth}{!}{
\begin{tabular}{p{0.70cm}|x{0.70cm}x{0.70cm}x{0.70cm}x{0.70cm}x{0.70cm}}
			 		& $p_{\text{@}2}$ 	& $p_{\text{@}5}$ 	& $p_{\text{@}10}$ 	& $p_{\text{@}25}$	& $p_{\text{@}50}$ \\
\hline
$p_{\text{@}2}$		&	1	&	0.98	&	0.82	&	0.71	&	0.65	\\
$p_{\text{@}5}$		&		&	1		&	0.95	&	0.86	&	0.79	\\
$p_{\text{@}10}$	&		&			&	1		&	0.95	&	0.89	\\
$p_{\text{@}20}$	&		&			&			&	1		&	0.97	\\
$p_{\text{@}50}$	&		&			&			&			&	1		\\
\end{tabular}
\quad\quad
\begin{tabular}{p{0.70cm}|x{0.70cm}x{0.70cm}x{0.70cm}x{0.70cm}x{0.70cm}}
			 	& $j_{\text{@}2}$ 	& $j_{\text{@}5}$ 	& $j_{\text{@}10}$ 	& $j_{\text{@}25}$	& $j_{\text{@}50}$ \\
\hline
$j_{\text{@}2}$		&	1	&	0.99	&	0.82	&	0.72	&	0.66	\\
$j_{\text{@}5}$		&		&	1		&	0.95	&	0.86	&	0.79	\\
$j_{\text{@}10}$	&		&			&	1		&	0.95	&	0.89	\\
$j_{\text{@}20}$	&		&			&			&	1		&	0.97	\\
$j_{\text{@}50}$	&		&			&			&			&	1		\\
\end{tabular}
}
\\
\vspace{0.1cm}
Embedding Technique: \wtv\ (skip-gram) \\
\vspace{0.3cm}
\resizebox{\columnwidth}{!}{
\begin{tabular}{p{0.70cm}|x{0.70cm}x{0.70cm}x{0.70cm}x{0.70cm}x{0.70cm}}
			 	& $p_{\text{@}2}$ 	& $p_{\text{@}5}$ 	& $p_{\text{@}10}$ 	& $p_{\text{@}25}$	& $p_{\text{@}50}$ \\
\hline
$p_{\text{@}2}$		&	1	&	0.67	&	0.56	&	0.47	&	0.42	\\
$p_{\text{@}5}$		&		&	1		&	0.83	&	0.69	&	0.63	\\
$p_{\text{@}10}$	&		&			&	1		&	0.86	&	0.77	\\
$p_{\text{@}20}$	&		&			&			&	1		&	0.93	\\
$p_{\text{@}50}$	&		&			&			&			&	1		\\
\end{tabular}
\quad\quad
\begin{tabular}{p{0.70cm}|x{0.70cm}x{0.70cm}x{0.70cm}x{0.70cm}x{0.70cm}}
			 	& $j_{\text{@}2}$ 	& $j_{\text{@}5}$ 	& $j_{\text{@}10}$ 	& $j_{\text{@}25}$	& $j_{\text{@}50}$ \\
\hline
$j_{\text{@}2}$		&	1	&	0.68	&	0.56	&	0.47	&	0.43	\\
$j_{\text{@}5}$		&		&	1		&	0.83	&	0.69	&	0.63	\\
$j_{\text{@}10}$	&		&			&	1		&	0.86	&	0.78	\\
$j_{\text{@}20}$	&		&			&			&	1		&	0.93	\\
$j_{\text{@}50}$	&		&			&			&			&	1		\\
\end{tabular}
}
\\
\vspace{0.1cm}
Embedding Technique: \glv \\
\vspace{0.3cm}
\resizebox{\columnwidth}{!}{
\begin{tabular}{p{0.70cm}|x{0.70cm}x{0.70cm}x{0.70cm}x{0.70cm}x{0.70cm}}
				& $p_{\text{@}2}$ 	& $p_{\text{@}5}$ 	& $p_{\text{@}10}$ 	& $p_{\text{@}25}$	& $p_{\text{@}50}$ \\
\hline
$p_{\text{@}2}$		&	1	&	0.56	&	0.42	&	0.33	&	0.29	\\
$p_{\text{@}5}$		&		&	1		&	0.78	&	0.61	&	0.53	\\
$p_{\text{@}10}$	&		&			&	1		&	0.83	&	0.72	\\
$p_{\text{@}20}$	&		&			&			&	1		&	0.94	\\
$p_{\text{@}50}$	&		&			&			&			&	1		\\
\end{tabular}
\quad\quad
\begin{tabular}{p{0.70cm}|x{0.70cm}x{0.70cm}x{0.70cm}x{0.70cm}x{0.70cm}}
			 	& $j_{\text{@}2}$ 	& $j_{\text{@}5}$ 	& $j_{\text{@}10}$ 	& $j_{\text{@}25}$	& $j_{\text{@}50}$ \\
\hline
$j_{\text{@}2}$		&	1	&	0.56	&	0.43	&	0.33	&	0.29	\\
$j_{\text{@}5}$		&		&	1		&	0.78	&	0.61	&	0.53	\\
$j_{\text{@}10}$	&		&			&	1		&	0.83	&	0.72	\\
$j_{\text{@}20}$	&		&			&			&	1		&	0.95	\\
$j_{\text{@}50}$	&		&			&			&			&	1		\\
\end{tabular}
}
\\
\vspace{0.1cm}
Embedding Technique: \ftt\ (skip-gram) \\
\vspace{0.3cm}

%% file: tables/tab_gaussian_parameters.tex
\begin{tabular}{c|l|cc|cc}
\multicolumn{6}{c}{target word $w_t=$\ \textbf{\texttt{momentum}}}\\[10pt]
\textbf{rank} & \multicolumn{1}{c|}{\textbf{query word}} & \textbf{mean} $\mu$ & \textbf{std.} $\sigma$ & $\bm{p_{\text{\#1}}(w_t,w_s)}$ & $\bm{p_{\text{\#}2}(w_t,w_s)}$ \\
\hline  
    1 & \texttt{inertia}				& 0.650 & 0.010 & $0.867$ 				& $0.991$ 				\\
    2 & \texttt{momenta}		     	& 0.633 & 0.011 & $0.124$ 				& $0.801$ 				\\
    3 & \texttt{kinetic}				& 0.621 & 0.009 & $9.24\cdot 10^{-3}$ 	& $0.204$ 				\\
    4 & \texttt{centripetal}			& 0.587 & 0.015 & $1.14\cdot 10^{-4}$	& $3.39\cdot 10^{-3}$	\\
    5 & \texttt{vorticity}				& 0.584 & 0.011 & $<10^{-6}$			& $1.56\cdot 10^{-4}$	\\
   10 & \texttt{massless}				& 0.567 & 0.011 & $<10^{-10}$			& $<10^{-6}$			\\
   50 & \texttt{spherically}			& 0.527 & 0.011 & $<10^{-17}$			& $<10^{-12}$			\\
  100 & \texttt{inelastic}				& 0.489 & 0.009 & $<10^{-64}$			& $<10^{-64}$			\\
  500 & \texttt{joule}					& 0.386 & 0.012 & $<10^{-64}$			& $<10^{-64}$			\\
 1000 & \texttt{power}					& 0.383 & 0.009 & $<10^{-64}$			& $<10^{-64}$			\\
\end{tabular}

%% file: tables/tab_structural_instability.tex
\begin{tabular}{c|x{2cm}x{2cm}x{2cm}}
\textbf{Language}				& \wtv			 		& \glv 				& \ftt 				\\
\hline
\textsc{Hi} 					& 0.969					& 0.968				& 0.984				\\
\textsc{Fi} 					& 0.976					& 0.995				& 0.974				\\
\textsc{Zh} 					& 0.977					& 0.990				& 0.985				\\
\textsc{Cs} 					& 0.979					& 0.991				& 0.979				\\
\textsc{Pl} 					& 0.976					& 0.989				& 0.968				\\
\textsc{Pt} 					& 0.981					& 0.975				& 0.976				\\
\textsc{En} 					& 0.975					& 0.973				& 0.985				\\
\end{tabular}

%% file: tables/tab_pip_consistent.tex
\setlength\tabcolsep{3.0pt}
\begin{tabular}{c|x{1.00cm}x{1.00cm}x{1.00cm}x{1.00cm}x{1.00cm}}
			 		& $10^3$ 	& $10^4$		& $10^5$ 	& $|\mathcal{V}|$	\\
\hline
$10^3$  			&	0.958	&	0.976	&	0.978	&	0.978	\\
$10^4$  			&			&	0.995	&	0.997	&	0.998	\\
$10^5$  			&			&			&	0.999	&	1.000	\\
$|\mathcal{V}|$		&			&			&			&	1		\\
\end{tabular} \\
\vspace{0.3cm}
Embedding Technique: \wtv\ (skip-gram) \\
\vspace{0.5cm}
\begin{tabular}{c|x{1.00cm}x{1.00cm}x{1.00cm}x{1.00cm}x{1.00cm}}
			 		& $10^3$ 	& $10^4$ 	& $10^5$ 	& $|\mathcal{V}|$	\\
\hline
$10^3$  			&	0.961	&	0.979	&	0.981	&	0.981	\\
$10^4$  			&			&	0.996	&	0.998	&	0.998	\\
$10^5$  			&			&			&	1.000	&	1.000	\\
$|\mathcal{V}|$		&			&			&			&	1		\\
\end{tabular} \\
\vspace{0.3cm}
Embedding Technique: \glv \\
\vspace{0.5cm}
\begin{tabular}{c|x{1.00cm}x{1.00cm}x{1.00cm}x{1.00cm}x{1.00cm}}
			 		& $10^3$ 	& $10^4$		& $10^5$ 	& $|\mathcal{V}|$	\\
\hline
$10^3$  			&	0.949	&	0.971	&	0.974	&	0.974	\\
$10^4$  			&			&	0.994	&	0.997	&	0.997	\\
$10^5$  			&			&			&	0.999	&	1.000	\\
$|\mathcal{V}|$		&			&			&			&	1		\\
\end{tabular} \\
\vspace{0.3cm}
Embedding Technique: \ftt\ (skip-gram)
\vspace{0.5cm}

%% file: tables/tab_comparison_approaches.tex
\begin{tabular}{l|x{4cm}x{4cm}}
						& \textbf{n.n. based metrics} 			& \textbf{w.w.r. PIP loss}				 	\\
\hline 
Formal criteria 		& Yes 									& Yes 										\\
Consistency				& No									& Yes										\\
Independence 			& No 									& Yes 										\\
Complexity	& $\mathcal{O}(|\mathcal{V}|)\sim 10^6-10^8$		& $\mathcal{O}(|\mathcal{V'}|)\sim 10^4$	\\
\end{tabular}

%% file: tables/tab_overview_measured_pip.tex
\hspace{-0.2cm}
\begin{tabular}{x{0.5cm}x{1.6cm}|cc|cc|cc}
\multirow{3}{*}{}	& \multirow{3}{*}{} &\multicolumn{2}{c|}{\wtv}	& \multicolumn{2}{c|}{\glv} & \multicolumn{2}{c}{\ftt}  \\
&&\multicolumn{2}{c|}{$\mathbf{D}_\text{rPIP}$}		&\multicolumn{2}{c|}{$\mathbf{D}_\text{rPIP}$}		&\multicolumn{2}{c}{$\mathbf{D}_\text{rPIP}$}	\\
&& $\mu \times 10^2$	& $\sigma \times 10^4$	& $\mu \times 10^2$	& $\sigma \times 10^4$	& $\mu \times 10^2$	& $\sigma \times 10^4$	 \\
\hline \multirow{3}{*}{\textsc{Hi}}	
			& \textsl{fixed} 		&	0.771	&	2.8	&	1.278	&	1.0	&	2.170	&	4.2	\\
			& \textsl{shuffle} 		&	1.805	&	5.3	&	1.275	&	0.9	&	2.367	&	0.2	\\
			& \textsl{bootstrap}	&	3.417	&	0.9	&	4.272	&	0.4	&	2.879	&	0.3	\\
\hline \multirow{3}{*}{\textsc{Fi}}	
			& \textsl{fixed} 		&	0.611	&	0.9	&	1.564	&	1.0	&	1.743	&	1.5	\\
			& \textsl{shuffle} 		&	1.665	&	2.5	&	1.558	&	1.2	&	1.963	&	0.2	\\
			& \textsl{bootstrap}	&	3.258	&	1.2	&	4.108	&	0.4	&	2.483	&	0.7	\\
\hline \multirow{3}{*}{\textsc{Zh}}	
			& \textsl{fixed} 		&	0.653	&	0.7	&	1.560	&	1.3	&	2.117	&	0.5	\\
			& \textsl{shuffle} 		&	1.634	&	2.1	&	1.543	&	0.9	&	2.428	&	0.6	\\
			& \textsl{bootstrap}	&	3.124	&	1.0	&	4.084	&	0.5	&	2.951	&	0.3	\\
\hline \multirow{3}{*}{\textsc{Cs}}	
			& \textsl{fixed} 		&	0.661	&	2.4	&	1.416	&	1.3	&	1.819	&	1.7	\\
			& \textsl{shuffle} 		&	1.543	&	2.5	&	1.417	&	1.4	&	2.044	&	0.2	\\
			& \textsl{bootstrap}	&	2.987	&	0.8	&	3.938	&	0.4	&	2.544	&	0.2	\\
\hline \multirow{3}{*}{\textsc{Pl}}
			& \textsl{fixed} 		&	0.644	&	1.2	&	1.469	&	1.9	&	1.704	&	0.7	\\
			& \textsl{shuffle} 		&	1.507	&	2.4	&	1.465	&	1.4	&	1.943	&	0.2	\\
			& \textsl{bootstrap}	&	2.853	&	1.3	&	3.947	&	0.7	&	2.426	&	0.3	\\
\hline \multirow{3}{*}{\textsc{Pt}}	
			& \textsl{fixed} 		&	0.702	&	0.9	&	1.334	&	1.1	&	1.828	&	0.9	\\
			& \textsl{shuffle} 		&	1.609	&	2.9	&	1.352	&	1.7	&	2.059	&	0.2	\\
			& \textsl{bootstrap}	&	3.063	&	1.1	&	4.065	&	0.7	&	2.575	&	0.2	\\
\hline \multirow{3}{*}{\textsc{En}}	
			& \textsl{fixed} 		&	0.725 	&	0.3	&	\textsl{1.201}	&  \textsl{4.5}		&	1.664	&	2.0	\\
			& \textsl{shuffle}		&	1.543	&	2.3	&	\textsl{1.208}	&  \textsl{4.0}		&	1.891	&	0.2	\\
			& \textsl{bootstrap}	&	2.883	&	0.8	&	\textsl{4.672}	&  \textsl{1.8} 	&	2.388	&	0.6	\\
\end{tabular}

%% file: tables/tab_instability_order_sampling.tex
\begin{tabular}{c|ccc}
\multicolumn{1}{c|}{\textbf{Sampling}}		& \multicolumn{3}{c}{\textbf{Instability}}	\\
\multicolumn{1}{c|}{\textbf{Method}}			& \textsl{Small}	&$\longrightarrow$	& \textsl{Large}		\\
\hline
fixed			& \wtv		& GloVe			& fastText	\\
shuffled		& \glv			& word2vec		& fastText	\\
bootstrapped	& \ftt		& word2vec		& GloVe 	\\
\end{tabular}

%% file: tables/tab_overview_analogy_scores.tex
\hspace{-0.2cm}
\begin{tabular}{c|cc|cc|cc}
\multirow{3}{*}{\textbf{Language}}		&\multicolumn{2}{c|}{\wtv}	& \multicolumn{2}{c|}{\glv} & \multicolumn{2}{c}{\ftt}  \\
&\multicolumn{2}{c|}{Analogy Score}		&\multicolumn{2}{c|}{Analogy Score}		&\multicolumn{2}{c}{Analogy Score}	\\
& $\mu$	& $\sigma$	& $\mu$	& $\sigma$	& $\mu$	& $\sigma$	 \\
\hline
\textsc{Hi}	& $	14.45	$ & $	0.38	$ & $	8.19		$ & $	0.24	$ & $	17.06	$ & $	0.46	$ \\
\textsc{Fi}	& $	45.69	$ & $	1.19	$ & $	26.16	$ & $	1.30	$ & $	42.82	$ & $	1.51	$ \\
\textsc{Zh}	& $	50.81	$ & $	1.10	$ & $	36.18	$ & $	1.62	$ & $	57.01	$ & $	1.21	$ \\
\textsc{Cs}	& $	48.54	$ & $	0.57	$ & $	41.80	$ & $	0.50	$ & $	62.89	$ & $	0.55	$ \\
\textsc{Pl}	& $	45.21	$ & $	0.53	$ & $	16.50	$ & $	0.38	$ & $	58.16	$ & $	0.78	$ \\
\textsc{Pt}	& $	50.48	$ & $	0.32	$ & $	33.26	$ & $	0.38	$ & $	56.52	$ & $	0.42	$ \\
\textsc{En}	& $	71.89	$ & $	0.20	$ & $	68.37	$ & $	0.28	$ & $	74.21	$ & $	0.21	$ \\
\end{tabular}

%% file: tables/tab_extrinsic_instabilities.tex
\hspace{-0.2cm}
\begin{tabular}{c|cc|cc|cc}
\multirow{3}{*}{\textbf{Language}}	 &\multicolumn{2}{c|}{\wtv}	& \multicolumn{2}{c|}{\glv} & \multicolumn{2}{c}{\ftt}  \\
&\multicolumn{2}{c|}{$\mathcal{I}_{\text{ext}}(\mathcal{T},\mathcal{C})$}
&\multicolumn{2}{c|}{$\mathcal{I}_{\text{ext}}(\mathcal{T},\mathcal{C})$}
&\multicolumn{2}{c}{$\mathcal{I}_{\text{ext}}(\mathcal{T},\mathcal{C})$}	\\
& $\mu	\times 10^2$ & $\sigma \times 10^4$ & $\mu	\times 10^2$ & $\sigma \times 10^4$ & $\mu	\times 10^2$ & $\sigma \times 10^4$  \\
\hline
\textsc{Hi}	&	2.901	&	3.4	&	4.077	&	0.5	&	1.639	&	0.6	\\
\textsc{Fi}	&	2.801	&	2.0	&	3.801	&	0.6	&	1.521	&	1.1	\\
\textsc{Zh}	&	2.663	&	1.8	&	3.781	&	0.6	&	1.677	&	1.0	\\
\textsc{Cs}	&	2.557	&	1.8	&	3.674	&	0.7	&	1.515	&	0.5	\\
\textsc{Pl}	&	2.423	&	2.1	&	3.665	&	0.9	&	1.453	&	0.5	\\
\textsc{Pt}	&	2.606	&	2.2	&	3.833	&	1.0	&	1.546	&	0.5	\\
\textsc{En}	&	2.435	&	1.8	&	\textsl{4.513}	&	\textsl{2.4}	&	1.458	&	1.0	\\
\end{tabular}

%% file: chapters/3_reducing.tex
\chapter{Minimizing the Instability}

Now that we developed a mathematical model to describe the randomness within embedding spaces (Section \ref{sec_cosine_similarity_gaussian}), decided on a method to quantify its extent (Section \ref{sec_measuring_distances}) and analysed different embedding techniques and languages, we turn our focus towards the potential actions one can take to minimize the effects of randomness on the embedding spaces.

\section{Model and Parameter Choices}
\label{sec_model_param_choices}

As outlined in the previous Chapter, the stability of a set of word embeddings trained on a corpus $\mathcal{C}$ depends heavily on the choice of embedding technique. However, for each of the three types of document sampling we examined, a different technique was found to be the most stable one. Hence, we cannot make a generic proposal on which technique to use in order to minimize the instability of the embedding spaces; the specific scenario and objectives need to be considered.

Apart from the choice of technique itself, each implementation -- at least the ones utilized for this work -- offers a selection of configurable parameters, which can be further optimized to minimize the instability of the method: \citet{hellrich-etal-2019-influence} found a strong influence of down-sampling strategies on the instability of embeddings for the \textbf{SVD}$_{\text{PPMI}}$ technique introduced by \citet{levy-etal-2015-improving}. \citet{yin2018} write, that the PIP loss -- i.e. our definition of instability -- depends on the number of dimensions of the embedding spaces, and exhibits a distinct minimum, depending on the corpus, for both \wtv\ and \glv.

We are interested in the correlation between training time and stability, i.e. in answering the question: Will the distribution of embeddings eventually converge, given sufficient training time? Due to limited resources, we had to restrict the experiments in this section to \wtv\ and \ftt\, trained on Hindi, Finnish, Chinese, Czech, Polish and Portuguese Wikipedia Corpora (i.e. \glv, as well as English are dropped). We can influence the training time for \wtv\ and \ftt\ mainly through two parameters: The number of training epochs and the number of negative samples. Table \ref{tab_ftt_training_time_pip} and Table \ref {tab_wtv_training_time_pip} show the influence of these parameters on the reduced PIP loss measured between two subsequent runs on independently shuffled corpora. An increase in any of the two parameters over the default values yields a decrease in the PIP loss, hence more stable embeddings for both, \wtv\ and \ftt. However, especially for \wtv, the PIP loss reaches a plateau at a certain point and does not further decrease -- and can even increase -- with longer training time. 

\begin{table}[htbp]
\begin{center}
\input{tables/tab_wtv_training_time_pip.tex}
\caption{Reduced PIP loss and percentage change due to increasing the training time -- i.e. number of epochs and negative samples -- of \wtv\ models in six different languages over the default. The depicted values correspond to the mean reduced PIP loss measured between four independent runs on shuffled corpora for each set of parameters in any language.}
\label{tab_wtv_training_time_pip}
\end{center}
\end{table} 
 
\begin{table}[htbp]
\begin{center}
\input{tables/tab_ftt_training_time_pip.tex}
\caption{Reduced PIP loss and percentage change due to increasing the training time -- i.e. number of epochs and negative samples -- of \ftt\ models in six different languages over the default. The depicted values correspond to the mean reduced PIP loss measured between four independent runs on shuffled corpora for each set of parameters in any language.}
\label{tab_ftt_training_time_pip}
\end{center}
\end{table}

Altogether, we observe a positive effect when increasing the training time for \wtv\ and \ftt\ on the stability of the embeddings. This is mirrored in the mean scores of the respective models on word analogy tasks -- outlined in Tables \ref{tab_wtv_training_time_scores} and \ref{tab_ftt_training_time_scores}.
 
\begin{table}[htbp]
\begin{center}
\input{tables/tab_wtv_training_time_scores.tex}
\caption{Scores on word analogy tasks and change due to increasing the training time -- i.e. number of epochs and negative samples -- of \wtv\ models in six different languages over the default. The depicted values correspond to the mean score of four independent runs on shuffled corpora for each set of parameters in any language.}
\label{tab_wtv_training_time_scores}
\end{center}
\end{table}

\begin{table}[htbp]
\begin{center}
\input{tables/tab_ftt_training_time_scores.tex}
\caption{Scores on word analogy tasks and change due to increasing the training time -- i.e. number of epochs and negative samples -- of \ftt\ models in six different languages over the default. The depicted values correspond to the mean score of four independent runs on shuffled corpora for each set of parameters in any language.}\label{tab_ftt_training_time_scores}
\end{center}
\end{table}

\section{Sample Average over Multiple Embedding Spaces}
\label{sec_average_multiple_embedding_spaces}
We have shown in Section \ref{sec_cosine_similarity_gaussian}, that embedding spaces derived from subsequent runs of a technique $\mathcal{T}$ over shuffled versions of a corpus $\mathcal{C}$ follow a particular probability distribution: The cosine similarity of any pair of words $w_k$ and $w_l$ is normally distributed, with a characteristic mean $\mu_{kl}$ and variance $\sigma_{kl}$ for every pair of words.

Furthermore, the experiments presented in Sections \ref{sec_understanding_instability} and \ref{sec_model_param_choices} demonstrated that a higher quality of embedding spaces (as measured by their performance on word analogy tasks) seems to correlate with a lower mean variance $\sigma_{kl}$ , and hence lower (reduced) PIP loss.

These two findings prompted us to try a novel approach to increase the quality of word embeddings: If we could compute a meaningful average of the embeddings over multiple runs on fixed, shuffled or bootstrapped corpora, we would expect the mean variance $\sigma_{kl}$ to decrease, and thus the quality of the embeddings to increase. This idea is supported by recent work on machine learning for image processing \cite{cirecsan2012}; \citet{izmailov2019} found that ``averaging weights lead to wider optima and better generalization'' for various neural network architectures. This could be particularly valuable for the task we will tackle in the following chapter -- detecting semantic change -- since the variance of the embeddings in the individual corpora leads to errors when measuring the difference of the embeddings between distinct corpora.

\subsection{A Meaningful Average of Two Embedding Spaces}
\label{sec_averaging_two_spaces}
We begin by looking for a meaningful average over two embedding spaces $\mathbf{V}_i$ and $\mathbf{V}_j$, trained by applying the same technique $\mathcal{T}$ on two independently shuffled versions of a corpus $\mathcal{C}$. 

The first problem we encounter is the random orientation of embedding spaces (see Figure \ref{fig_random_orientation_of_embeddings}), hence naively averaging over the embeddings does not yield meaningful results. However, since the random orientation is the result of the rotation-invariance of embedding spaces, we can make use of this characteristic and ``align'' the embeddings with another, before averaging. This alignment of embedding spaces was first proposed almost simultaneously by \citet{kulkarni2015} and \citet{zhang2015} to compare the embeddings of words trained on different corpora. We follow the approach of \citet{hamilton2016}, i.e. solving the orthogonal Procrustes problem:
\begin{align}
\mathbf{A}_{ij}=\underset{\mathbf{A}\mathbf{A}^\top=\mathbf{I}}{\text{arg min}} ||\mathbf{V}_i\mathbf{A}-\mathbf{V}_j||
\label{eq_procrustes}
\end{align}
where $\mathbf{A}_{ij}\in\mathbb{R}^{d\times d}$ is an orthogonal matrix that corresponds to rotating $\mathbf{V}_i$ to minimize the Frobenius norm between $\mathbf{V}_i\mathbf{A}$ and $\mathbf{V}_j$. The solution can be obtained efficiently using the SVD-based approach from \citet{schoenemann1966}.

Using this transformation, we define the \textsl{aligned average} $\mathbf{M}_{ij}$ of the two embedding spaces $\mathbf{V}_i$ and $\mathbf{V}_j$ as:
\begin{align}
\mathbf{M}_{ij}=\frac{1}{2}\left(\mathbf{V}_i\mathbf{A}_{ij}+\mathbf{V}_j\right)
\end{align} 

The distribution of the cosine similarity of the embeddings of two words $w_k$ and $w_l$ in the embedding space $\mathbf{M}_{ij}$, i.e. their dot product -- assuming normalized vectors -- reads:
\begin{align}
\vec{m}_{ij}(w_k)\cdot \vec{m}_{ij}^\top(w_l)&=&\frac{1}{2}\left(\vec{v}_i(w_k)\mathbf{A}_{ij}+\vec{v}_j(w_k)\right)\cdot \frac{1}{2}\left(\vec{v}_i(w_l)\mathbf{A}_{ij}+\vec{v}_j(w_l)\right)^\top 
\end{align}
For the sake of readability, we continue with the notation $\mathbf{A}:=\mathbf{A}_{ij}$, $\vec{v}_{k,l}:=\vec{v}_i(w_{k,l})$ and $\vec{u}_{k,l}:=\vec{v}_j(w_{k,l})$. Then:
\begin{align}
\label{eq_dot_product_average}
\begin{split}
\vec{m}_{ij}(w_k)\cdot \vec{m}_{ij}^\top(w_l)
&= \frac{1}{4} \left(\vec{v}_{k} \mathbf{A} \mathbf{A}^{\top} \vec{v}_{l}^{\top} + \vec{u}_{k} \mathbf{A}^{\top} \vec{v}_{l}^{\top} + \vec{v}_{k} \mathbf{A} \vec{u}_{l}^{\top} + \vec{u}_{k}\vec{u}_{l}^{\top} \right) \\ 
&= \frac{1}{4} \left[ \vec{v}_{k}\vec{v}_{l}^{\top} + \vec{u}_{k}\vec{u}_{l}^{\top} + (\vec{u}_{k}+\vec{v}_{k} \mathbf{A} - \vec{v}_{k} \mathbf{A}) \mathbf{A}^{\top} \vec{v}_{l}^{\top} \right. \\
& \quad \left.+ (\vec{v}_{k} \mathbf{A} +\vec{u}_{k}-\vec{u}_{k}) \vec{u}_{l}^{\top} \right] \\
&= \frac{1}{2} \left( \vec{v}_{k}\vec{v}_{l}^{\top} + \vec{u}_{k}\vec{u}_{l}^{\top} \right) + \frac{1}{4} \left[ (\vec{u}_{k}-\vec{v}_{k} \mathbf{A} ) \mathbf{A} ^{\top}\vec{v}_{l}^{\top} + (\vec{v}_{k} \mathbf{A} -\vec{u}_{k})\vec{u}_{l}^{\top} \right] \\
&= \frac{1}{2} \left( \vec{v}_{k}\vec{v}_{l}^{\top} + \vec{u}_{k}\vec{u}_{l}^{\top} \right) - \frac{1}{4} (\vec{v}_{k} \mathbf{A} -\vec{u}_{k})\left( \mathbf{A} ^{\top}\vec{v}_{l}^{\top}-\vec{u}_{l}^{\top}\right) \\ 
&= \frac{1}{2} \left( \vec{v}_{k}\vec{v}_{l}^{\top} + \vec{u}_{k}\vec{u}_{l}^{\top} \right) -\frac{1}{4} (\vec{v}_{k} \mathbf{A} -\vec{u}_{k})\left(\vec{v}_{l} \mathbf{A} -\vec{u}_{l}\right)^{\top}
\end{split}
\end{align}
The result is quite intuitive: The first term is the mean of the cosine similarity of the words $w_k$ and $w_l$ in the two embedding spaces $\mathbf{V}_i$ and $\mathbf{V}_j$ respectively; the second term corresponds to the variance introduced by the numerical alignment of the two embedding spaces. 

We know that both $\vec{v}_{k}\vec{v}_{l}^{\top}$ and $\vec{u}_{k}\vec{u}_{l}^{\top}$ are sampled from the same Gaussian distribution $\mathcal{N}\left(\mu_{kl},\sigma_{kl}^2\right)$, hence the first term is a Gaussian with mean $\tilde{\mu}$ and variance $\tilde{\sigma}^2$:
\begin{align}
\tilde{\mu}=\frac{1}{2}(\mu_{kl}+\mu_{kl})=\mu_{kl} && \tilde{\sigma}^2=\left(\frac{1}{2}\sqrt{\sigma_{kl}^2+\sigma_{kl}^2}\right)^2 = \frac{1}{2}\sigma_{kl}^2
\end{align}
Furthermore, we find in our experiments, that the second term is again a Normal distribution, with zero mean and a variance smaller than $\frac{1}{2}\sigma_{kl}^2$, for all languages and techniques examined in this work. 

Altogether, this means that the distribution of the dot product $\vec{m}_{ij}(w_k)\cdot \vec{m}_{ij}^\top(w_l)$ of the embeddings of the words $w_k$ and $w_l$ in the aligned average space $\mathbf{M}_{ij}$ has the same mean, but a smaller variance, than the distributions in the original spaces $\mathbf{V}_i$ and $\mathbf{V}_j$. 

This is an indicator that averaging over embedding spaces might be beneficial for their quality: It seems like the cosine similarities are converging towards the mean of the underlying distribution.

\subsubsection{Normalization and the Bias-Variance Trade-off}
\label{sec_bias_variance_tradeoff}
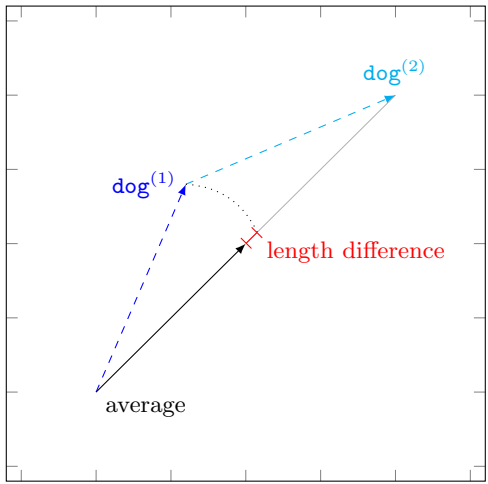
\begin{figure}[htbp]
\center
\input{tikz/fig_normalization_bias_averaging.tex}
\caption{Illustration of the normalization problem that occurs when we average over aligned embedding spaces. If the two vectors $\vec{v}_{\texttt{dog}^{(1)}}$ and $\vec{v}_{\texttt{dog}^{(2)}}$ are normalized and not parallel, the length of their average $\frac{1}{2}\left(\vec{v}_{\texttt{dog}^{(1)}}+\vec{v}_{\texttt{dog}^{(2)}}\right)$ is smaller than $1$.}
\label{fig_normalization_bias_averaging}
\end{figure}

We made one -- as we realized during our experiments -- rather naive assumption in the section above, namely, that the aligned average of two normalized vectors is still normalized. However, as illustrated in Figure \ref{fig_normalization_bias_averaging}, the aligned average is not only not normalized, but the length distribution is systematically lopsided. Since the two normalized vectors $\vec{v}_j(w_k)$ and $\vec{v}_i(w_k)\mathbf{A}_{ij}$ are generally not parallel, the triangle inequality yields that the length of $\vec{m}_{ij}(w_k)$ can only be smaller (or equal, in case of two parallel vectors) to $1$.

To calculate the cosine similarity of the embeddings of any two words $w_k$ and $w_l$ in the aligned average space $\mathbf{M}_{ij}$, the result of Equation (\ref{eq_dot_product_average}) must be divided by the length of the vectors $\vec{m}_{ij}(w_k)$ and $\vec{m}_{ij}(w_l)$, which are generally smaller than $1$, hence the distribution of the cosine similarity is displaced towards larger values than in the original spaces $\mathbf{V}_i$ and $\mathbf{V}_j$.

\begin{figure}[htbp]
\center
\input{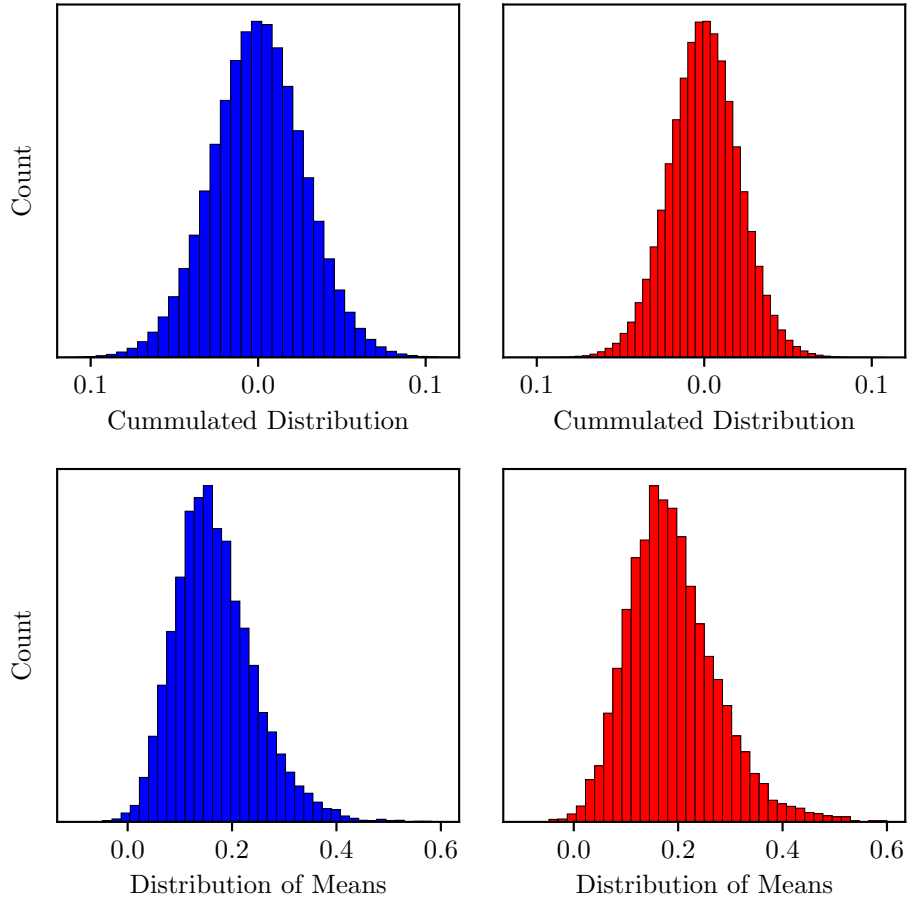}
\vspace{-1cm}
\caption{Distribution of cosine similarities for 5,000 randomly sampled word pairs over 128 runs of \ftt\ embeddings trained on independently shuffled versions of the Finnish Wikipedia. The column on the left shows the distribution for the 128 individual models, the column on the right shows the distribution for the 64 embedding spaces we computed as an aligned average over two models each. The histograms in the top row depict the cumulated centred distribution over all word pairs. The variance of the averaged models is significantly smaller than that of the original ones. The bottom row shows the distribution of the mean values of the cosine similarity of the different word pairs; with the expected bias towards higher values. Table \ref{tab_momentum_nn_avg} illustrates these distributions for a specific example.}
\label{fig_average_gauss_accumulated}
\end{figure}

Hence, we find that averaging leads to a \textsl{bias-variance trade-off}: The variance of the cosine similarity distribution in the aligned average spaces is smaller than in the original ones, however, the means are systematically biased towards larger values. This is illustrated in Figure \ref{fig_average_gauss_accumulated}. For the embeddings and techniques examined in this work, we found the differences of $\mu_{kl}$ and $\sigma_{kl}$ between the aligned average space $\mathbf{M}_{ij}$ and the two original spaces $\mathbf{V}_i$ and $\mathbf{V}_j$ (averaged over 10,000 randomly sampled word pairs) to fall within the following limits:
\begin{align}
\frac{1}{1.39} < \frac{\langle \sigma_{kl} \rangle ^{\mathbf{M}}}{\langle \sigma_{kl} \rangle ^{\mathbf{V}}} < \frac{1}{1.28} 
&& 
1.01 < \frac{\langle \mu_{kl} \rangle ^{\mathbf{M}}}{\langle \mu_{kl} \rangle ^{\mathbf{V}}} < 1.22
\end{align}
Before we analyse the effect of this trade-off on the quality of the embeddings, we want to find a way to average over samples that consist of more than two arbitrarily oriented embedding spaces.

\subsection{Increasing the Sample Size}
\label{sec_tree_approach}
\begin{figure}[htbp]
\vspace{0.5cm}
\center
\input{tikz/fig_schema_averaging_unused.tex}
\caption{Illustration of a naive and discarded approach to calculate the aligned average over more than two embedding spaces.}
\label{fig_schema_averaging_unused}
\end{figure}
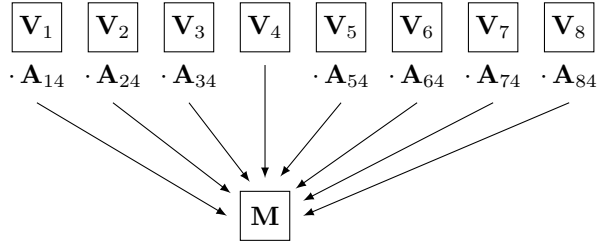

A naive method -- and our own initial approach -- to compute the aligned average over a set $\mathcal{S}=\{\mathbf{V}_i\text{ for }i=1,...,r\}$ of $r>2$ embedding spaces is illustrated in Figure \ref{fig_schema_averaging_unused}: We randomly select one embedding space $\mathbf{V}_j\in \mathcal{S}$, calculate the closest orthogonal transformation $\mathbf{A}_{ij}$ -- using Equation (\ref{eq_procrustes}) -- from any other space $\mathbf{V}_i \in \mathcal{S} \setminus \{\mathbf{V}_j\}$ to $\mathbf{V}_j$, and define the aligned average as:
\begin{align}
\mathbf{M}=\frac{1}{r}\left(\mathbf{V}_j + \sum_{j\neq i=1}^r \mathbf{V}_i\mathbf{A}_{ij} \right)
\end{align}
However, both our experiments and a theoretical analysis yield that this approach creates more random variations than it can reduce. The result depends heavily on the initial, random choice of $\mathbf{V}_j$ and the sum of the deviations from transforming the embeddings of the $r-1$ remaining embedding to $\mathbf{V}_j$ can cause substantial variations in the resulting space $\mathbf{M}$.

\begin{figure}[htbp]
\center
\input{tikz/fig_schema_averaging_used.tex}
\caption{Illustration of the tree-based approach used to calculate the aligned average over more than two embedding spaces.}
\label{fig_schema_averaging_used}
\end{figure}
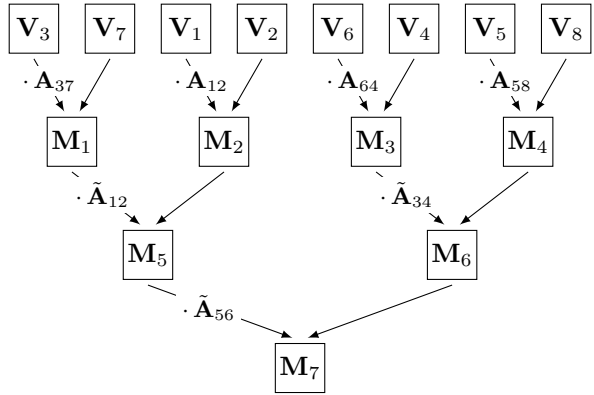

We have shown in Section \ref{sec_averaging_two_spaces} that the aligned average $\mathbf{M} _{ij}$ of two embedding spaces $\mathbf{V}_{i}$ and $\mathbf{V}_{j}$ exhibits smaller variances $\sigma_{kl}$ of the cosine similarities of arbitrary word pairs $w_k$ and $w_l$ than the original spaces -- but also some bias. Since this was found to be true for any set of initial embedding spaces, we can extend the finding to an approach for averaging over more than two spaces: By averaging only two spaces at a time -- in a binary-tree fashion (see Figure \ref{fig_schema_averaging_used}).

\subsection{Influence of Averaging on Stability and Quality}
\label{sec_influence_aa_stability}
We have applied the approach outlined in the previous sections to \textsl{compute the aligned average over 128 independent runs} on fixed, shuffled and bootstrapped corpora for all techniques and languages\footnote{Except for English, which was omitted due to limited computational resources.} outlined in Section \ref{sec_exp_setup}.

Using the tree-based approach outlined in Section \ref{sec_tree_approach}, and storing all intermediate spaces computed in the process, means that for every language, technique, and sampling method we have 128 initial spaces, 64 2-fold average spaces, 32 4-fold average spaces, etc., and finally, one embedding space that constitutes the average over all 128 initial spaces. 

\begin{figure}[htbp]
\center
\vspace{-2cm}
\input{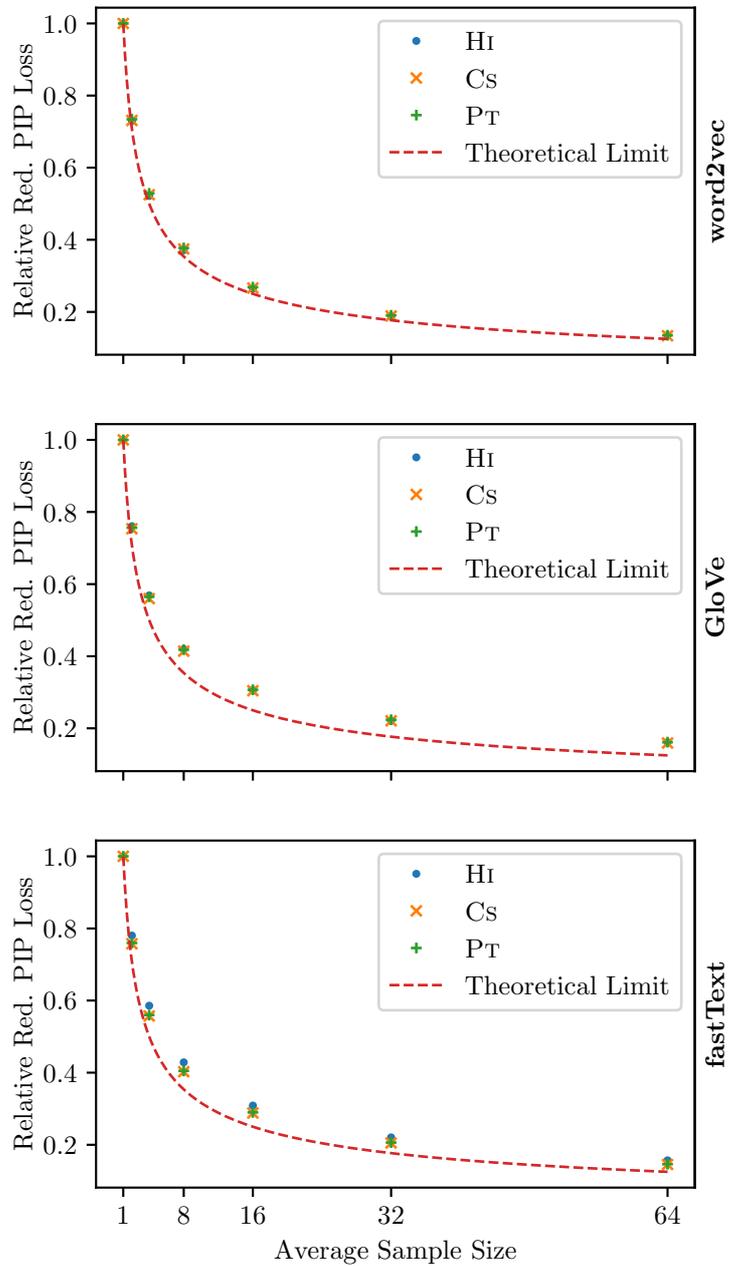}
\vspace{-1cm}
\caption{Mean reduced PIP loss observed between pairs of averaged embeddings of different sample sizes, as a fraction of the mean reduced PIP loss of the individual models. For every technique, the decrease in reduced PIP loss is close to the theoretical lower limit, i.e. what we would expect, if the deviations caused by the orthogonal transformation of one embedding space into another vanish.}
\label{fig_pip_loss_average}
\end{figure}

Figure \ref{fig_pip_loss_average} illustrates the reduced PIP loss measured between pairs of spaces at different levels of the averaging process. We observe a strong decline of the reduced PIP loss, close to the theoretical limit, i.e. the course one would expect if the deviations caused by the orthogonal transformation of one embedding space into another -- the second term in Equation \ref{eq_dot_product_average} -- would vanish. Hence, averaging over aligned embedding spaces trained in multiple runs seems to reduce the random noise inherent to the training process of any embedding technique.

Table \ref{tab_momentum_nn_avg} contains an anecdotal illustration of the increased stability of the averaged embeddings: It comprises the 15 nearest neighbors of the target word \texttt{momentum} for two embedding spaces, each the aligned average over 32 subsequently trained models on independently shuffled versions of the English Wikipedia. Table \ref{tab_momentum_nn} holds the same comparison for two of the initial models. For the average-based models, one can observe less variations in the ranking, smaller differences between the cosine similarity values and the bias towards higher similarities outlined in Section \ref{sec_bias_variance_tradeoff}. 

\begin{table}[htbp]
\center
\input{tables/tab_momentum_nn_avg.tex}
\captionsetup{singlelinecheck=off}
\caption[.]{Most similar words to the target word \texttt{momentum} for two embedding spaces, each the aligned average over 32 subsequently trained models on independently shuffled versions of the English Wikipedia, with a reduced vocabulary size of 200,000 words. Although we argue against the use of nearest-neighbor based metrics to quantify the instability of word embeddings, they can be used to illustrate the increase in stability over the models depicted in Table \ref{tab_momentum_nn}:
\begin{align}
\begin{split}
p_{\text{@}10} = 0.9, \quad j_{\text{@}10} \approx 0.818, \quad p_{\text{@}15} \approx 0.933, \quad j_{\text{@}15} \approx 0.875
\end{split}
\end{align}
}
\label{tab_momentum_nn_avg}
\end{table}

\begin{figure}[htbp]
\center
\vspace{-1cm}
\input{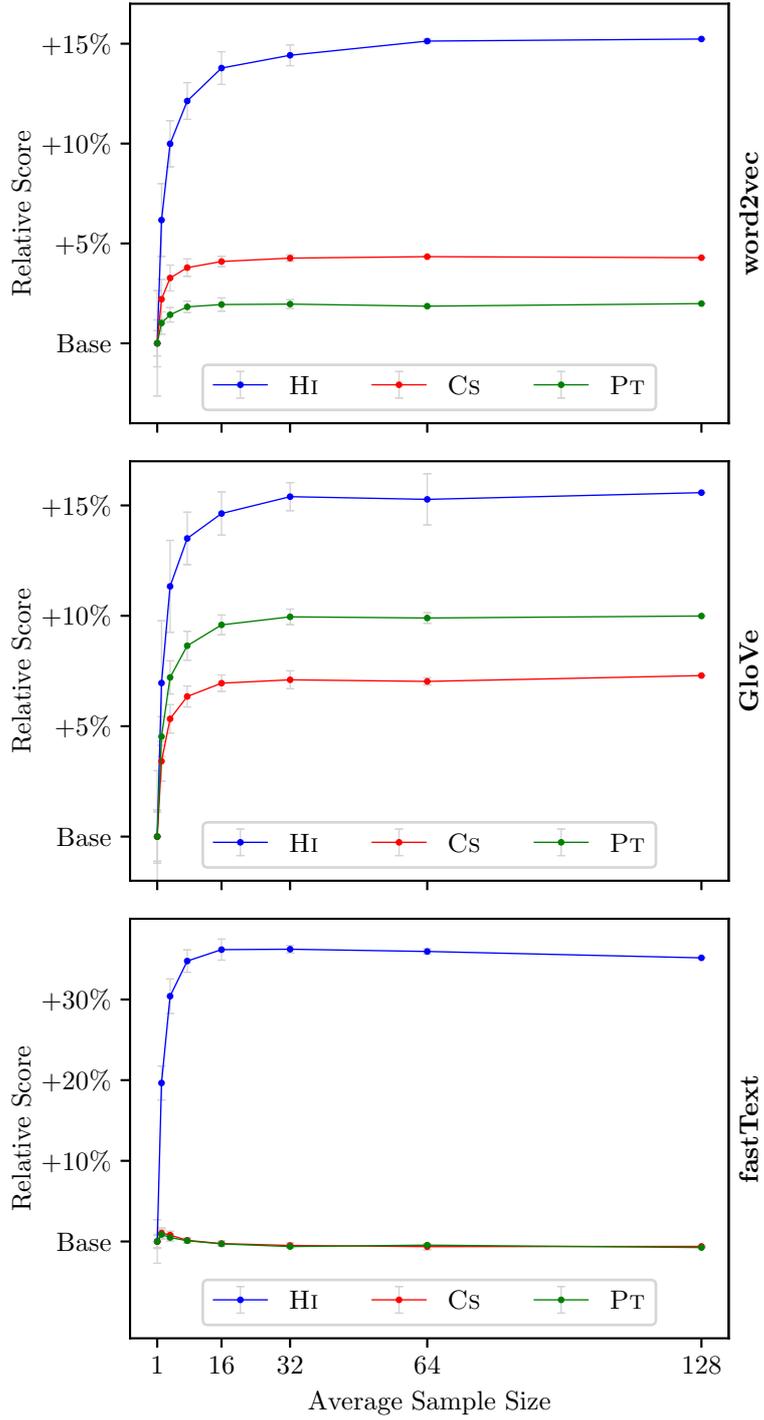}
\vspace{-0.5cm}
\caption{Mean score on the word analogy tasks in Hindi, Czech and Portuguese for \wtv, \glv, and \ftt\ models, as a function of the averaging sample size. The $y$-axis measures the relative difference of the score, compared to the individually trained models.}
\label{fig_score_average}
\end{figure}

However, \textsl{more stable embeddings are not necessarily better} suited for the use in downstream tasks. As outlined in the beginning of Chapter \ref{chp_measuring}, the inherent random processes of any embedding technique could be replaced by deterministic alternatives, which would only create a false sense of reliability. Hence, we need to determine if the embeddings compiled by averaging have a higher quality, as measured by their score on word analogy tasks. Figure \ref{fig_score_average} shows the improvements in the score for \wtv, \glv, and \ftt\ models for increasing sample size. The results for all languages that were examined in this work are compiled in Table \ref{tab_improvements_score_average}.

For \glv, one can see an increase in score compared to the individual runs, for all languages, up to a sample size of 32. Increasing the size further does not yield significant improvements, the score plateaus. For \wtv\, we observe a similar behaviour -- although the relative increase in score is smaller than for \glv\ -- for all languages expect Chinese. Finally, for \ftt, the results are mixed: We see a strong increase of the score for Hindi and Finnish, with a significant decrease for Chinese, on the other hand.

One possible reason for this observation is that \ftt\ embeddings -- in contrast to \wtv\ and \glv\ -- are sub-word based. The approach we have chosen is based on naively averaging over the embeddings of vocabulary words, however. Hence, averaging over the underlying sub-word embeddings might be the more promising approach for \ftt. 

\begin{table}[htbp]
\begin{center}
\input{tables/tab_improvements_score_average.tex}
\hspace{-0.2cm}
\caption{Difference in the mean score on the word analogy task between the 128 models trained in individual runs and the 16 models computed by averaging over 8 samples each. }
\label{tab_improvements_score_average}
\end{center}
\end{table}

\subsection*{Conclusion}
In our experiments on six different languages, we found that increasing the training time of \wtv\ and \ftt, i.e. the number of epochs and negative samples, generally reduces the instability (as measured by the PIP loss), and improves the quality (as measured by the score on word analogy tasks) of the trained word embeddings. However, this general observation does not strictly hold in all scenarios -- increasing the training time can also have the opposite effect. We do not argue to have a satisfactory explanation for these mixed observations.

Furthermore, we introduced a novel method to compute a meaningful average over several embedding spaces. This method proved to be efficient in reducing the instability of the embeddings and -- for most, but not all combinations of technique and language -- also increasing the quality of the embeddings. 

%% file: tables/tab_wtv_training_time_pip.tex
\setlength\tabcolsep{3.0pt}
\resizebox{\columnwidth}{!}{
\begin{tabular}{x{0.5cm}x{0.6cm}|x{1.8cm}x{1.1cm}x{1.1cm}}	
\multicolumn{2}{c|}{\multirow{2}{*}{\textsc{Hi}}}	&
\multicolumn{3}{c}{\textbf{Negative Samples}} \\
&		&	5	&	10	&	20	\\
\hline
\multirow{4}{*}{\rotatebox[origin=c]{90}{\textbf{Epochs}}} 
&	5	& $ \mathbf{	1.79	 } \times 10^{-2}$ & $ - $ 	11.6	& $ - $ 	18.9	\\
&	10	& $ - $ 	14.0	& $ - $ 	18.6	& $ - $ 	19.7	\\
&	20	& $ - $ 	17.0	& $ - $ 	17.6	& $ - $ 	15.6	\\
&	40	& $ - $ 	15.0	& $ - $ 	12.3	& $ - $ 	8.9	\\
\end{tabular}
\quad\quad
\begin{tabular}{x{0.5cm}x{0.6cm}|x{1.8cm}x{1.1cm}x{1.1cm}}	
\multicolumn{2}{c|}{\multirow{2}{*}{\textsc{Fi}}}	&
\multicolumn{3}{c}{\textbf{Negative Samples}} \\
&		&	5	&	10	&	20	\\
\hline
\multirow{4}{*}{\rotatebox[origin=c]{90}{\textbf{Epochs}}} 
&	5	& $ \mathbf{	1.65	 } \times 10^{-2}$ & $ - $ 	12.4	& $ - $ 	18.7	\\
&	10	& $ - $ 	9.4	& $ - $ 	17.9	& $ - $ 	20.2	\\
&	20	& $ - $ 	15.0	& $ - $ 	14.4	& $ - $ 	14.5	\\
&	40	& $ - $ 	11.6	& $ - $ 	9.4	& $ - $ 	5.6	\\
\end{tabular}
}
\\
\vspace{0.5cm}
\resizebox{\columnwidth}{!}{
\begin{tabular}{x{0.5cm}x{0.6cm}|x{1.8cm}x{1.1cm}x{1.1cm}}	
\multicolumn{2}{c|}{\multirow{2}{*}{\textsc{Zh}}}	&
\multicolumn{3}{c}{\textbf{Negative Samples}} \\
&		&	5	&	10	&	20	\\
\hline
\multirow{4}{*}{\rotatebox[origin=c]{90}{\textbf{Epochs}}} 
&	5	& $ \mathbf{	1.52	 } \times 10^{-2}$ & $ - $ 	12.6	& $ - $ 	18.1	\\
&	10	& $ - $ 	14.0	& $ - $ 	19.7	& $ - $ 	21.5	\\
&	20	& $ - $ 	18.0	& $ - $ 	19.4	& $ - $ 	17.7	\\
&	40	& $ - $ 	17.5	& $ - $ 	15.9	& $ - $ 	11.6	\\
\end{tabular}
\quad\quad
\begin{tabular}{x{0.5cm}x{0.6cm}|x{1.8cm}x{1.1cm}x{1.1cm}}	
\multicolumn{2}{c|}{\multirow{2}{*}{\textsc{Cs}}}	&
\multicolumn{3}{c}{\textbf{Negative Samples}} \\
&		&	5	&	10	&	20	\\
\hline
\multirow{4}{*}{\rotatebox[origin=c]{90}{\textbf{Epochs}}} 
&	5	& $ \mathbf{	1.63	 } \times 10^{-2}$ & $ - $ 	12.9	& $ - $ 	16.9	\\
&	10	& $ - $ 	16.9	& $ - $ 	24.9	& $ - $ 	26.2	\\
&	20	& $ - $ 	23.0	& $ - $ 	25.3	& $ - $ 	24.0	\\
&	40	& $ - $ 	24.1	& $ - $ 	21.7	& $ - $ 	17.8	\\
\end{tabular}
}
\\
\vspace{0.5cm}
\resizebox{\columnwidth}{!}{
\begin{tabular}{x{0.5cm}x{0.6cm}|x{1.8cm}x{1.1cm}x{1.1cm}}	
\multicolumn{2}{c|}{\multirow{2}{*}{\textsc{Pl}}}	&
\multicolumn{3}{c}{\textbf{Negative Samples}} \\
&		&	5	&	10	&	20	\\
\hline
\multirow{4}{*}{\rotatebox[origin=c]{90}{\textbf{Epochs}}} 
&	5	& $ \mathbf{	1.51	 } \times 10^{-2}$ & $ - $ 	12.0	& $ - $ 	17.0	\\
&	10	& $ - $ 	13.0	& $ - $ 	18.8	& $ - $ 	19.9	\\
&	20	& $ - $ 	17.3	& $ - $ 	18.2	& $ - $ 	16.6	\\
&	40	& $ - $ 	17.8	& $ - $ 	15.9	& $ - $ 	9.0	\\
\end{tabular}
\quad\quad
\begin{tabular}{x{0.5cm}x{0.6cm}|x{1.9cm}x{1.1cm}x{1.1cm}}	
\multicolumn{2}{c|}{\multirow{2}{*}{\textsc{Pt}}}	&
\multicolumn{3}{c}{\textbf{Negative Samples}} \\
&		&	5	&	10	&	20	\\
\hline
\multirow{4}{*}{\rotatebox[origin=c]{90}{\textbf{Epochs}}} 
&	5	& $ \mathbf{	1.60	 } \times 10^{-2}$ & $ - $ 	11.0	& $ - $ 	15.0	\\
&	10	& $ - $ 	11.9	& $ - $ 	17.4	& $ - $ 	19.1	\\
&	20	& $ - $ 	17.0	& $ - $ 	18.6	& $ - $ 	17.6	\\
&	40	& $ - $ 	18.9	& $ - $ 	17.8	& $ - $ 	14.3	\\
\end{tabular}
}

%% file: tables/tab_ftt_training_time_pip.tex
\setlength\tabcolsep{3.0pt}
\resizebox{\columnwidth}{!}{
\begin{tabular}{x{0.5cm}x{0.6cm}|x{1.8cm}x{1.1cm}x{1.1cm}}	
\multicolumn{2}{c|}{\multirow{2}{*}{\textsc{Hi}}}	&
\multicolumn{3}{c}{\textbf{Negative Samples}} \\
&		&	5								&	10			&	20			\\
\hline
\multirow{4}{*}{\rotatebox[origin=c]{90}{\textbf{Epochs}}} 
&	5	& $	\mathbf{2.37} \times 10^{-2}$ 	& $ - $	0.8 	& $ - $	1.7		\\
&	10	& $ - $ 3.2							& $ - $ 3.7	 	& $ - $ 4.2		\\
&	20	& $ - $	7.9							& $ - $ 8.4		& $ - $ 8.1		\\
&	40	& $ - $	13.9						& $ - $ 14.3	& $ - $ 14.2	\\
\end{tabular}
\quad\quad
\begin{tabular}{x{0.5cm}x{0.6cm}|x{1.8cm}x{1.1cm}x{1.1cm}}	
\multicolumn{2}{c|}{\multirow{2}{*}{\textsc{Fi}}}	&
\multicolumn{3}{c}{\textbf{Negative Samples}} \\
&		&	5								&	10			&	20			\\
\hline
\multirow{4}{*}{\rotatebox[origin=c]{90}{\textbf{Epochs}}} 
&	5	& $ \mathbf{1.96} \times 10^{-2}$ 	& $ - $ 1.6		& $ - $ 3.4		\\
&	10	& $ - $ 2.1							& $ - $ 3.5		& $ - $ 6.2		\\
&	20	& $ - $ 6.5							& $ - $ 8.9		& $ - $ 10.4	\\
&	40	& $ - $ 13.1						& $ - $ 15.4	& $ - $ 16.9	\\
\end{tabular}
}
\\
\vspace{0.5cm}
\resizebox{\columnwidth}{!}{
\begin{tabular}{x{0.5cm}x{0.6cm}|x{1.8cm}x{1.1cm}x{1.1cm}}	
\multicolumn{2}{c|}{\multirow{2}{*}{\textsc{Zh}}}	&
\multicolumn{3}{c}{\textbf{Negative Samples}} \\
&		&	5								&	10			&	20			\\
\hline
\multirow{4}{*}{\rotatebox[origin=c]{90}{\textbf{Epochs}}} 
&	5	& $ \mathbf{2.05} \times 10^{-2}$ 	& $ - $ 2.2		& $ - $ 3.9		\\
&	10	& $ - $ 4.9							& $ - $ 6.2		& $ - $ 8.5		\\
&	20	& $ - $ 11.2						& $ - $ 13.5	& $ - $ 15.3	\\
&	40	& $ - $ 19.6						& $ - $ 21.5	& $ - $ 22.1	\\
\end{tabular}
\quad\quad
\begin{tabular}{x{0.5cm}x{0.6cm}|x{1.8cm}x{1.1cm}x{1.1cm}}	
\multicolumn{2}{c|}{\multirow{2}{*}{\textsc{Cs}}}	&
\multicolumn{3}{c}{\textbf{Negative Samples}} \\
&		&	5								&	10			&	20			\\
\hline
\multirow{4}{*}{\rotatebox[origin=c]{90}{\textbf{Epochs}}} 
&	5	& $ \mathbf{2.43} \times 10^{-2}$ 	& $ - $	2.1		& $ - $ 3.8		\\
&	10	& $ - $ 5.0							& $ - $	6.6		& $ - $ 8.0		\\
&	20	& $ - $ 10.7						& $ - $	12.6	& $ - $ 13.8	\\
&	40	& $ - $ 18.4						& $ - $	20.1	& $ - $ 21.1	\\
\end{tabular}
}
\\
\vspace{0.5cm}
\resizebox{\columnwidth}{!}{
\begin{tabular}{x{0.5cm}x{0.6cm}|x{1.8cm}x{1.1cm}x{1.1cm}}	
\multicolumn{2}{c|}{\multirow{2}{*}{\textsc{Pl}}}	&
\multicolumn{3}{c}{\textbf{Negative Samples}} \\
&		&	5								&	10			&	20			\\
\hline
\multirow{4}{*}{\rotatebox[origin=c]{90}{\textbf{Epochs}}} 
&	5	& $ \mathbf{1.94} \times 10^{-2}$	& $ - $ 2.4		& $ - $ 4.5		\\
&	10	& $ - $ 5.5							& $ - $ 8.1		& $ - $ 10.3	\\
&	20	& $ - $ 12.4						& $ - $ 15.0	& $ - $ 17.0	\\
&	40	& $ - $ 21.0						& $ - $ 23.3	& $ - $ 18.8	\\
\end{tabular}
\quad\quad
\begin{tabular}{x{0.5cm}x{0.6cm}|x{1.9cm}x{1.1cm}x{1.1cm}}	
\multicolumn{2}{c|}{\multirow{2}{*}{\textsc{Pt}}}	&
\multicolumn{3}{c}{\textbf{Negative Samples}} \\
&		&	5								&	10			&	20			\\
\hline
\multirow{4}{*}{\rotatebox[origin=c]{90}{\textbf{Epochs}}} 
&	5	& $ \mathbf{2.06} \times 10^{-2}$ 	& $ - $ 2.5		& $ - $	4.7		\\
&	10	& $ - $ 5.8							& $ - $ 8.4		& $ - $	10.1	\\
&	20	& $ - $ 12.7						& $ - $ 15.1	& $ - $	17.0	\\
&	40	& $ - $ 21.0						& $ - $ 23.1	& $ - $	18.2	\\
\end{tabular}
}

%% file: tables/tab_wtv_training_time_scores.tex
\setlength\tabcolsep{3.0pt}
\resizebox{\columnwidth}{!}{
\begin{tabular}{x{0.5cm}x{0.6cm}|x{1.1cm}x{1.1cm}x{1.1cm}}	
\multicolumn{2}{c|}{\multirow{2}{*}{\textsc{Hi}}}	&
\multicolumn{3}{c}{\textbf{Negative Samples}} \\
&		&	5	&	10	&	20	\\
\hline
\multirow{4}{*}{\rotatebox[origin=c]{90}{\textbf{Epochs}}} 
&	5	& \textbf{14.5}	& +	1.2	& +	2.0	\\
&	10	& +	4.5	& +	5.7	& +	5.8	\\
&	20	& +	7.4	& +	8.3	& +	9.3	\\
&	40	& +	9.6	& +	10.8	& +	11.7	\\
\end{tabular}
\quad\quad
\begin{tabular}{x{0.5cm}x{0.6cm}|x{1.1cm}x{1.1cm}x{1.1cm}}	
\multicolumn{2}{c|}{\multirow{2}{*}{\textsc{Fi}}}	&
\multicolumn{3}{c}{\textbf{Negative Samples}} \\
&		&	5	&	10	&	20	\\
\hline
\multirow{4}{*}{\rotatebox[origin=c]{90}{\textbf{Epochs}}} 
&	5	& \textbf{45.7} & +	1.4	& +	2.7	\\
&	10	& +	3.6	& +	5.6	& +	7.1	\\
&	20	& +	5.6	& +	6.9	& +	8.7	\\
&	40	& +	7.3	& +	7.8	& +	9.5	\\

\end{tabular}
}
\\
\vspace{0.5cm}
\resizebox{\columnwidth}{!}{
\begin{tabular}{x{0.5cm}x{0.6cm}|x{1.1cm}x{1.1cm}x{1.1cm}}	
\multicolumn{2}{c|}{\multirow{2}{*}{\textsc{Zh}}}	&
\multicolumn{3}{c}{\textbf{Negative Samples}} \\
&		&	5	&	10	&	20	\\
\hline
\multirow{4}{*}{\rotatebox[origin=c]{90}{\textbf{Epochs}}} 
&	5	& \textbf{48.5}	& +	1.9	& +	2.7	\\
&	10	& +	1.7	& +	3.5	& +	4.6	\\
&	20	& +	1.4	& +	2.7	& +	3.9	\\
&	40	& +	1.1	& +	2.0	& +	3.0	\\
\end{tabular}
\quad\quad
\begin{tabular}{x{0.5cm}x{0.6cm}|x{1.1cm}x{1.1cm}x{1.1cm}}	
\multicolumn{2}{c|}{\multirow{2}{*}{\textsc{Cs}}}	&
\multicolumn{3}{c}{\textbf{Negative Samples}} \\
&		&	5	&	10	&	20	\\
\hline
\multirow{4}{*}{\rotatebox[origin=c]{90}{\textbf{Epochs}}} 
&	5	& \textbf{50.8} & +	5.0	& +	6.4	\\
&	10	& +	7.1	& +	10.4	& +	11.6	\\
&	20	& +	8.9	& +	12.3	& +	12.8	\\
&	40	& +	8.9	& +	12.4	& +	14.7	\\
\end{tabular}
}
\\
\vspace{0.5cm}
\resizebox{\columnwidth}{!}{
\begin{tabular}{x{0.5cm}x{0.6cm}|x{1.1cm}x{1.1cm}x{1.1cm}}	
\multicolumn{2}{c|}{\multirow{2}{*}{\textsc{Pl}}}	&
\multicolumn{3}{c}{\textbf{Negative Samples}} \\
&		&	5	&	10	&	20	\\
\hline
\multirow{4}{*}{\rotatebox[origin=c]{90}{\textbf{Epochs}}} 
&	5	&\textbf{45.2}	& +	1.7	& +	4.2	\\
&	10	& +	2.6	& +	5.0	& +	7.0	\\
&	20	& +	4.0	& +	5.7	& +	7.9	\\
&	40	& +	4.8	& +	5.8	& +	8.3	\\
\end{tabular}
\quad\quad
\begin{tabular}{x{0.5cm}x{0.6cm}|x{1.1cm}x{1.1cm}x{1.1cm}}	
\multicolumn{2}{c|}{\multirow{2}{*}{\textsc{Pt}}}	&
\multicolumn{3}{c}{\textbf{Negative Samples}} \\
&		&	5	&	10	&	20	\\
\hline
\multirow{4}{*}{\rotatebox[origin=c]{90}{\textbf{Epochs}}} 
&	5	& \textbf{50.5}	& +	1.5	& +	2.8	\\
&	10	& +	2.1	& +	3.4	& +	4.6	\\
&	20	& +	2.6	& +	3.7	& +	4.7	\\
&	40	& +	2.7	& +	3.8	& +	4.6	\\
\end{tabular}
}

%% file: tables/tab_ftt_training_time_scores.tex
\setlength\tabcolsep{3.0pt}
\resizebox{\columnwidth}{!}{
\begin{tabular}{x{0.5cm}x{0.6cm}|x{1.1cm}x{1.1cm}x{1.1cm}}	
\multicolumn{2}{c|}{\multirow{2}{*}{\textsc{Hi}}}	&
\multicolumn{3}{c}{\textbf{Negative Samples}} \\
&		&	5	&	10	&	20	\\
\hline
\multirow{4}{*}{\rotatebox[origin=c]{90}{\textbf{Epochs}}} 
&	5	& \textbf{17.1}	& +	2.8	& +	4.9	\\
&	10	& +	4.4	& +	6.7	& +	8.9	\\
&	20	& +	7.0	& +	9.5	& +	12.0	\\
&	40	& +	9.6	& +	12.0	& +	13.6	\\
\end{tabular}
\quad\quad
\begin{tabular}{x{0.5cm}x{0.6cm}|x{1.1cm}x{1.1cm}x{1.1cm}}	
\multicolumn{2}{c|}{\multirow{2}{*}{\textsc{Fi}}}	&
\multicolumn{3}{c}{\textbf{Negative Samples}} \\
&		&	5	&	10	&	20	\\
\hline
\multirow{4}{*}{\rotatebox[origin=c]{90}{\textbf{Epochs}}} 
&	5	& \textbf{42.8}	& +	4.2	& +	4.9	\\
&	10	& +	4.5	& +	5.8	& +	7.8	\\
&	20	& +	6.4	& +	9.2	& +	10.0	\\
&	40	& +	10.0	& +	12.2	& +	12.7	\\
\end{tabular}
}
\\
\vspace{0.5cm}
\resizebox{\columnwidth}{!}{
\begin{tabular}{x{0.5cm}x{0.6cm}|x{1.1cm}x{1.1cm}x{1.1cm}}	
\multicolumn{2}{c|}{\multirow{2}{*}{\textsc{Zh}}}	&
\multicolumn{3}{c}{\textbf{Negative Samples}} \\
&		&	5	&	10	&	20	\\
\hline
\multirow{4}{*}{\rotatebox[origin=c]{90}{\textbf{Epochs}}} 
&	5	& \textbf{57.0}	& +	2.8	& +	5.2	\\
&	10	& +	2.8	& +	5.6	& +	6.7	\\
&	20	& +	5.0	& +	7.3	& +	8.8	\\
&	40	& +	5.3	& +	7.8	& +	8.9	\\
\end{tabular}
\quad\quad
\begin{tabular}{x{0.5cm}x{0.6cm}|x{1.1cm}x{1.1cm}x{1.1cm}}	
\multicolumn{2}{c|}{\multirow{2}{*}{\textsc{Cs}}}	&
\multicolumn{3}{c}{\textbf{Negative Samples}} \\
&		&	5	&	10	&	20	\\
\hline
\multirow{4}{*}{\rotatebox[origin=c]{90}{\textbf{Epochs}}} 
&	5	& \textbf{62.9}	& +	0.4	& +	0.2	\\
&	10	& $-$ 1.8	& $-$ 1.3	& $-$ 1.5	\\
&	20	& $-$ 3.9	& $-$ 3.9	& $-$ 4.0	\\
&	40	& $-$ 6.9	& $-$ 6.6	& $-$ 6.1	\\
\end{tabular}
}
\\
\vspace{0.5cm}
\resizebox{\columnwidth}{!}{
\begin{tabular}{x{0.5cm}x{0.6cm}|x{1.1cm}x{1.1cm}x{1.1cm}}	
\multicolumn{2}{c|}{\multirow{2}{*}{\textsc{Pl}}}	&
\multicolumn{3}{c}{\textbf{Negative Samples}} \\
&		&	5	&	10	&	20	\\
\hline
\multirow{4}{*}{\rotatebox[origin=c]{90}{\textbf{Epochs}}} 
&	5	& \textbf{58.2}	& +	2.2	& +	3.4	\\
&	10	& +	2.7	& +	4.0	& +	4.3	\\
&	20	& +	2.0	& +	3.1	& +	4.3	\\
&	40	& +	1.3	& +	2.9	& +	3.5	\\
\end{tabular}
\quad\quad
\begin{tabular}{x{0.5cm}x{0.6cm}|x{1.1cm}x{1.1cm}x{1.1cm}}	
\multicolumn{2}{c|}{\multirow{2}{*}{\textsc{Pt}}}	&
\multicolumn{3}{c}{\textbf{Negative Samples}} \\
&		&	5	&	10	&	20	\\
\hline
\multirow{4}{*}{\rotatebox[origin=c]{90}{\textbf{Epochs}}} 
&	5	& \textbf{56.5}	& +	1.4	& +	2.9	\\
&	10	& +	2.0	& +	3.0	& +	3.9	\\
&	20	& +	2.4	& +	3.4	& +	4.3	\\
&	40	& +	2.4	& +	3.5	& +	4.1	\\
\end{tabular}
}

%% file: tikz/fig_normalization_bias_averaging.tex
\begin{tikzpicture}
\begin{axis}
[empty, width = 0.65 * \textwidth, height = 0.65 * \textwidth, name = plot1]
\coordinate (str_dog_1) 	at (axis cs:-1.00,-1.00) {};
\coordinate (end_dog_1) 	at (axis cs:-0.40,+0.40) {};
\coordinate (end_dog_2) 	at (axis cs:+1.00,+1.00) {};
\coordinate (way_point) 	at (axis cs:+0.00,+0.00) {};
\coordinate (vrt_point) 	at (axis cs:+0.077,+0.077) {};

\end{axis}		
\draw [blue,  dashed, -latex] 		(str_dog_1) -- (end_dog_1)	node [anchor=east] 	{\small{\texttt{dog$^{(1)}$}}};	
\draw [cyan,   dashed, -latex] 	(end_dog_1) -- (end_dog_2)	node [anchor=south] 	{\small{\texttt{dog$^{(2)}$}}};	
\draw [black, -latex] 				(str_dog_1) node [anchor=north west] 	{\small{average}} -- (way_point)	;	
\draw [black, opacity = 0.3, -] 	(way_point) -- (end_dog_2)	;	

\draw [black, dotted, -] (end_dog_1) to [bend left] (vrt_point) ;
\draw [red, |-|] (way_point) to (vrt_point) node [anchor=north west] 	{\small{length difference}} ;
	
\end{tikzpicture}

%% file: tikz/fig_schema_averaging_unused.tex
\begin{tikzpicture}
\draw (-4 - 0.325, 0.5 - 0.325) rectangle (-4 + 0.325, 0.5 + 0.325) node[pos=.5] {$\mathbf{V}_1$};	
\draw (-3 - 0.325, 0.5 - 0.325) rectangle (-3 + 0.325, 0.5 + 0.325) node[pos=.5] {$\mathbf{V}_2$};
\draw (-2 - 0.325, 0.5 - 0.325) rectangle (-2 + 0.325, 0.5 + 0.325) node[pos=.5] {$\mathbf{V}_3$};
\draw (-1 - 0.325, 0.5 - 0.325) rectangle (-1 + 0.325, 0.5 + 0.325) node[pos=.5] {$\mathbf{V}_4$};
\draw (-0 - 0.325, 0.5 - 0.325) rectangle (-0 + 0.325, 0.5 + 0.325) node[pos=.5] {$\mathbf{V}_5$};
\draw (+1 - 0.325, 0.5 - 0.325) rectangle (+1 + 0.325, 0.5 + 0.325) node[pos=.5] {$\mathbf{V}_6$};
\draw (+2 - 0.325, 0.5 - 0.325) rectangle (+2 + 0.325, 0.5 + 0.325) node[pos=.5] {$\mathbf{V}_7$};
\draw (+3 - 0.325, 0.5 - 0.325) rectangle (+3 + 0.325, 0.5 + 0.325) node[pos=.5] {$\mathbf{V}_8$};

\draw (-1 - 0.325, -2 - 0.325)  rectangle (-1 + 0.325, -2  + 0.325) node[pos=.5] {$\mathbf{M}$};

\draw [-latex] (-1,+0) -- (-1,-1.5);

\node at (-4,-0.15) {$\cdot  \, \mathbf{A}_{14}$};
\node at (-3,-0.15) {$\cdot  \, \mathbf{A}_{24}$};
\node at (-2,-0.15) {$\cdot  \, \mathbf{A}_{34}$};
\node at (-0,-0.15) {$\cdot  \, \mathbf{A}_{54}$};
\node at (+1,-0.15) {$\cdot  \, \mathbf{A}_{64}$};
\node at (+2,-0.15) {$\cdot  \, \mathbf{A}_{74}$};
\node at (+3,-0.15) {$\cdot  \, \mathbf{A}_{84}$};

\draw [-latex] (-4,-0.5) -- (-1.5,-2);		
\draw [-latex] (-3,-0.5) -- (-1.45,-1.7);		
\draw [-latex] (-2,-0.5) -- (-1.2,-1.55);		
\draw [-latex] (+0,-0.5) -- (-0.8,-1.48);		
\draw [-latex] (+1,-0.5) -- (-0.6,-1.64);		
\draw [-latex] (+2,-0.5) -- (-0.5,-1.8);	
\draw [-latex] (+3,-0.5) -- (-0.5,-2.0);					
\end{tikzpicture}

%% file: tikz/fig_schema_averaging_used.tex
\begin{tikzpicture}
\draw (-4 - 0.325, 0.5 - 0.325) rectangle (-4 + 0.325, 0.5 + 0.325) node[pos=.5] {$\mathbf{V}_3$};
\draw (-3 - 0.325, 0.5 - 0.325) rectangle (-3 + 0.325, 0.5 + 0.325) node[pos=.5] {$\mathbf{V}_7$};
\draw (-2 - 0.325, 0.5 - 0.325) rectangle (-2 + 0.325, 0.5 + 0.325) node[pos=.5] {$\mathbf{V}_1$};
\draw (-1 - 0.325, 0.5 - 0.325) rectangle (-1 + 0.325, 0.5 + 0.325) node[pos=.5] {$\mathbf{V}_2$};
\draw (-0 - 0.325, 0.5 - 0.325) rectangle (-0 + 0.325, 0.5 + 0.325) node[pos=.5] {$\mathbf{V}_6$};
\draw (+1 - 0.325, 0.5 - 0.325) rectangle (+1 + 0.325, 0.5 + 0.325) node[pos=.5] {$\mathbf{V}_4$};
\draw (+2 - 0.325, 0.5 - 0.325) rectangle (+2 + 0.325, 0.5 + 0.325) node[pos=.5] {$\mathbf{V}_5$};
\draw (+3 - 0.325, 0.5 - 0.325) rectangle (+3 + 0.325, 0.5 + 0.325) node[pos=.5] {$\mathbf{V}_8$};

\draw [-latex] (-3,0.1) -- (-3.4,-0.6);
\draw [-latex] (-1,0.1) -- (-1.4,-0.6);	
\draw [-latex] (+1,0.1) -- (+0.6,-0.6);	
\draw [-latex] (+3,0.1) -- (+2.6,-0.6);

\draw [-latex] (-4,0.1) -- (-3.6,-0.6);
\draw [-latex] (-2,0.1) -- (-1.6,-0.6);	
\draw [-latex] (+0,0.1) -- (+0.4,-0.6);	
\draw [-latex] (+2,0.1) -- (+2.4,-0.6);			

\node at (-3.8, -0.2)	[fill=white, inner sep=2pt] {\small{$\cdot  \, \mathbf{A}_{37}$}};
\node at (-1.8, -0.2)	[fill=white, inner sep=2pt] {\small{$\cdot  \, \mathbf{A}_{12}$}};
\node at (+0.2, -0.2)	[fill=white, inner sep=2pt] {\small{$\cdot  \, \mathbf{A}_{64}$}};
\node at (+2.2, -0.2)	[fill=white, inner sep=2pt] {\small{$\cdot  \, \mathbf{A}_{58}$}};			

\draw (-3.5 - 0.325, -1.0 - 0.325) rectangle (-3.5 + 0.325, -1.0 + 0.325) node[pos=.5] {$\mathbf{M}_{1}$};
\draw (-1.5 - 0.325, -1.0 - 0.325) rectangle (-1.5 + 0.325, -1.0 + 0.325) node[pos=.5] {$\mathbf{M}_{2}$};
\draw (+0.5 - 0.325, -1.0 - 0.325) rectangle (+0.5 + 0.325, -1.0 + 0.325) node[pos=.5] {$\mathbf{M}_{3}$};
\draw (+2.5 - 0.325, -1.0 - 0.325) rectangle (+2.5 + 0.325, -1.0 + 0.325) node[pos=.5] {$\mathbf{M}_{4}$};	

\draw [-latex] (-3.5,-1.4) -- (-2.6,-2.1);
\draw [-latex] (-1.5,-1.4) -- (-2.4,-2.1);	
\draw [-latex] (+0.5,-1.4) -- (+1.4,-2.1);	
\draw [-latex] (+2.5,-1.4) -- (+1.6,-2.1);			

\node at (-3.1,-1.7)	[fill=white, inner sep=2pt] {\small{$\cdot  \, \tilde{\mathbf{A}}_{12}$}};
\node at (+0.9,-1.7)	[fill=white, inner sep=2pt] {\small{$\cdot  \, \tilde{\mathbf{A}}_{34}$}};

\draw (-2.5 - 0.325, -2.5 - 0.325) rectangle ( -2.5 + 0.325, -2.5 + 0.325) node[pos=.5] {$\mathbf{M}_{5}$};
\draw (+1.5 - 0.325, -2.5 - 0.325) rectangle ( +1.5 + 0.325, -2.5 + 0.325) node[pos=.5] {$\mathbf{M}_{6}$};

\draw [-latex] (-2.5,-2.9) -- (-0.6,-3.6);	
\draw [-latex] (+1.5,-2.9) -- (-0.4,-3.6);			

\node at (-1.7,-3.2)[fill=white, inner sep=2pt] {\small{$\cdot  \, \tilde{\mathbf{A}}_{56}$}};

\draw (-0.5 - 0.325, -4.0 - 0.325) rectangle (-0.5 + 0.325, -4.0 + 0.325) node[pos=.5] {$\mathbf{M}_{7}$};
\end{tikzpicture}

%% file: tables/tab_momentum_nn_avg.tex
\begin{tabular}{lc|cccc|cccc}
\multicolumn{10}{c}{target word: \textbf{\texttt{momentum}}}\\[10pt]
\multirow{2}{*}{\textbf{word}} & \quad & \quad &
\multicolumn{2}{c}{\textbf{run \# 1}} & \quad & \quad &
\multicolumn{2}{c}{\textbf{run \# 2}} & \quad \\
&&& rank & cos &&&  rank & cos & \\
\hline 
&&&&&&&\\[-15pt]
\texttt{inertia}	 	&&&	1 	& 	0.662	&&&	1 				&	0.668 & \\
\texttt{momenta} 		&&&	2 	&	0.651	&&&	2 				&	0.648 & \\
\texttt{kinetic} 		&&&	3	&	0.632	&&&	3 				&	0.633 & \\
\texttt{centripetal} 	&&&	4	&	0.608	&&&	4				&	0.608 & \\
\texttt{vorticity} 		&&&	5	&	0.603	&&&	5 				&	0.606 & \\
\texttt{gravitational}	&&&	6	&	0.596	&&&	6 				&	0.599 & \\
\texttt{energy}			&&&	7	&	0.594	&&&	7				&	0.598 & \\
\texttt{mass-energy}	&&&	8	&	0.594	&&&	8				&	0.594 & \\
\texttt{accelerating} 	&&&	9	&	0.590	&&&	9				&	0.591 & \\
\texttt{flux}			&&&	10	&	0.589	&&&	\textbf{11}	 	&	0.588 & \\
\hdashline[0.5pt/5pt] 
\texttt{angular}		&&&	11	&	0.587	&&&	\textbf{10}		&	0.589 & \\
\texttt{massless}		&&&	12	&	0.586	&&&	12				&	0.585 & \\
\texttt{velocity} 		&&&	13	&	0.584	&&&	\textsl{14}		&	0.584 & \\
\texttt{eigenstate}		&&&	14	&	0.583	&&&	\textsl{13} 	&	0.585 & \\
\texttt{decelerating}	&&&	15	&	0.583	&&&	\textbf{20}	 	&	0.574 & \\
\hdashline[0.5pt/5pt] 
\end{tabular}

%% file: tables/tab_improvements_score_average.tex
\begin{tabular}{c|ccc}
\textbf{Language}	 	&\wtv		& \glv 		& \ftt \\
\hline
\textsc{Hi}	& +	1.8		& +	1.1		& +	5.9		\\
\textsc{Fi}	& +	1.0		& +	3.1		& +	8.1		\\
\textsc{Zh}	& $-$ 0.2	& +	6.4		& $-$ 3.9	\\
\textsc{Cs}	& +	1.8		& +	2.7		& +	0.1		\\
\textsc{Pt}	& +	0.9		& +	2.9		& +	0.1		\\
\textsc{Pl}	& +	0.9		& +	2.6		& +	1.9		\\
\end{tabular}

%% file: chapters/4_semantic.tex
\chapter{Semantic Change}

The meaning of words in a language can change severely over time, reflecting complex developments in the respective society. ``Examples include both changes to the core meaning of words (like the word \textsl{gay} shifting from meaning \textsl{carefree} to \textsl{homosexual} during the 20th century) and subtle shifts of cultural associations (like \textsl{Iraq} or \textsl{Syria} being associated with the concept of \textsl{war} after armed conflicts had started in these countries)'' -- \citet{kutuzov2018}.

Understanding these changes has long been a topic of interest in linguistic research: In one of the earliest works on this topic, \citet{breal1899} documented and categorized semantic shifts. Over the past decade this field has changed dramatically through the use of prediction-based word embedding techniques \cite{tang_2018, kutuzov2018, tahmasebi2018}. \citet{kim-etal-2014-temporal} were the first to use prediction-based (\wtv) embeddings to trace diachronic shifts, and after \citet{zhang2015} and \citet{kulkarni2015} proposed a method to align word embeddings trained on different corpora (see Section \ref{sec_averaging_two_spaces}), many followed this approach \cite{hamilton2016,hamilton2016b,dubossarsky2017}. 

The approach can be summarized as follows: First, one obtains the embedding spaces $\mathbf{V}_{t_1}$ and $\mathbf{V}_{t_2}$ by applying the same embedding technique $\mathcal{T}$ on two corpora $\mathcal{C}_{t_1}$ and $\mathcal{C}_{t_2}$ from different epochs $t_1$ and $t_2$:
\begin{align}
\mathbf{V}_{t_1} \sim \Omega(\mathcal{T}, \mathcal{C}_{t_1}) && \mathbf{V}_{t_2} \sim \Omega(\mathcal{T}, \mathcal{C}_{t_2})
\end{align}
Next, the two embedding spaces $\mathbf{V}_{t_1}$ and $\mathbf{V}_{t_2}$ are aligned by solving the orthogonal Procrustes problem, that yields the orthogonal transformation $\mathbf{A}_{t_1t_2}$ corresponding to the closest match between $\mathbf{V}_{t_1}\mathbf{A}_{t_1t_2}$ and $\mathbf{V}_{t_2}$ -- see Equation (\ref{eq_procrustes}). Finally, the semantic change $\Delta_{t_1t_2}$ of a word $w$ between the epochs $t_1$ and $t_2$ is defined as the cosine distance\footnote{The term cosine distance refers to $1$ minus the cosine similarity.} of the two word vectors $\vec{v}_{t_1}(w)\cdot\mathbf{A}_{t_1t_2}$ and $\vec{v}_{t_2}(w)$, i.e.:
\begin{align}
\Delta_{t_1t_2}(w) = \text{cos-dist}\,\left[\vec{v}_{t_1}(w)\cdot\mathbf{A}_{t_1t_2},\vec{v}_{t_2}(w)\right]
\label{eq_semantic_change}
\end{align}

Applying this approach to 20 decades of documents in American English, \citet{hamilton2016} proposed multiple \textsl{Statistical Laws of Semantic Change}, e.g. that more frequently used words change slower than less frequently used ones, but \citet{dubossarsky2017} contested these findings and argued that they are products of the inherent instability of the embedding techniques. 

In this chapter, we employ diachronic word embeddings, utilizing the understanding of the stability of word embeddings outlined in Chapter 2, along with the methods introduced in Chapter 3 to minimize the instability in order to differentiate between model artifacts and actual semantic shifts.

\section{Semantic Change Detection}
\label{sec_semeval}

The large methodological changes in research on semantic change that are outlined in the section above, prompted \citet{schlechtweg2020semeval} to call for a competition on semantic change detection: Task 1 of the 14th International Workshop on Semantic Evaluation, taking place in Barcelona, Spain in September 2020 comprises the unsupervised detection of lexical semantic change.

The organizers provide corpora from two distinct epochs in each of the following four languages: English, German, Latin and Swedish. The corpora are described in more detail in Section \ref{sec_SEMEVAL_corpus}. The competition consists of two tasks: a classification task and a ranking task, both on the same two corpora in each language, and the same set of 30 to 50 target words per language. The goal of the classification task is to decide which of the target words have lost or gained senses between the two epochs, whereas the goal of the ranking task is to sort the target words according to their degree of semantic change.

The submitted solutions for each task are evaluated against annotations by experts (native speakers for English, German and Swedish, and scholars of Latin for Latin). The annotations were produced using the framework developed by \citet{schlechtweg-etal-2018-diachronic}, i.e. by ranking the relatedness of pairs of the usage of one word $w$ in two different contexts (an example provided by the authors is illustrated in Table \ref{tab_donnerwetter}). \citet{schlechtweg-etal-2018-diachronic} asked five annotators to evaluate the use relatedness of more than 1000 use-pairs of different words and found an inter-annotator agreement (measured by Spearman's $\rho$) of between $0.57$ and $0.68$. This demonstrates that there are limits to the performance of any model on these tasks, since even human experts do not fully agree on a ``single version of the truth''.

\begin{table}[htbp]
\center
\newcolumntype{P}[1]{>{\hspace{0pt}}p{#1}}
\begin{tabular}{P{5.3cm}P{0.2cm}P{5.3cm}}
\underline{Target Context 1:} && \underline{Target Context 2:} \\
\textsl{Ein \textbf{Donnerwetter} in Paris ist mit so vielen Verdrieslichkeiten verkn\"upft, daß ichs hier anf\"uhren muß.} &&
\textsl{Der andre observirte sch\"arfer mit dem Ausruf: \glqq\textbf{Donnerwetter}, sollte ich mich irren!\grqq}
\end{tabular}
\caption{A pair of use-pair of the German word \textsl{Donnerwetter}, with a small relatedness, as evaluated by humans \cite{schlechtweg-etal-2018-diachronic}.}
\label{tab_donnerwetter}
\end{table}

\begin{table}[htbp]
\input{tables/tab_semeval_words_disp.tex}
\caption{The five words with the largest semantic change between the two epochs $t_1$ and $t_2$ in the four different languages, out of the set of target words of the respective language. The results in the rightmost column correspond to an aligned average over 32 runs of \ftt\ on bootstrapped corpora.}
\label{tab_semeval_words_disp}
\end{table}

\begin{table}[htbp]
\center
\input{tables/tab_semeval_scores.tex}
\caption{Results of different \wtv, \glv\ and \ftt\ models on the two tasks of the semantic change detection outlined in Section \ref{sec_semeval}. The accuracy of the binary classification of the target words is reported for Task 1, and Spearman's $\rho$ between the ranking produced by the model and the human annotation for Task 2 (averaged over the four languages in both cases). For each of the three embedding techniques, we report the mean and standard deviation of six different models: The individual runs (IR) over shuffled and bootstrapped corpora, the aligned average (AA) over the 32 runs (see Section \ref{sec_average_multiple_embedding_spaces}) and the ensemble average (EA) proposed by \citet{antoniak2018}. The standard deviations for the AA and EA models, are calculated on size-16 models (we can only produce one size-32 model) and hence an upper limit to the true standard deviation. The overall best-performing model is the aligned average over a set of 32 \ftt\ embeddings trained on bootstrapped corpora.}
\label{tab_semeval_scores}
\end{table}

We employed the following approach for the two tasks: We trained \wtv, \glv\ and \ftt\ embeddings on 32 shuffled, as well as 32 bootstrapped versions of the corpora from both epochs in all four languages. Out of these embedding spaces, we produced multiple models that were evaluated on the two tasks:

\begin{enumerate}
\itemsep0em
\item \textsl{Individual runs} (IR) trained on shuffled, as well as bootstrapped corpora.
\item The \textsl{aligned average} (AA) of the 32 embeddings trained on shuffled and bootstrapped\footnote{Training embedding spaces on bootstrapped versions of a corpus might lead to partially disjoint vocabularies. In this case, we compute the average for all words that occur in both vocabularies and keep the original embeddings for all words that occur only in one of the two vocabularies.} corpora, as introduced in Section \ref{sec_average_multiple_embedding_spaces}.
\item The \textsl{ensemble average} (EA) of the 32 embeddings trained on shuffled and bootstrapped corpora, as proposed by \citet{antoniak2018}: Here, we independently apply each of the 32 models to the tasks at hand and finally average over the 32 sets of results (for the classification in Task 1, the majority vote is used).
\end{enumerate}

For each of the models and languages, we calculate the semantic change for any word that appears in the corpus of both epochs $t_1$ and $t_2$, as outlined in Equation (\ref{eq_semantic_change}). The ranking of the semantic change of the target words (Task 2) is computed directly from this measure. For the binary classification (Task 1), we compute a model dependent threshold $\tau$ and assume that every target word that exceeds this threshold has gained or lost a sense between the two epochs.\footnote{In preliminary experiments we found that the threshold $\tau = \mu + \sigma/2$ performs well, where $\mu$ and $\sigma$ are mean and standard deviation of the semantic change of all words appearing in both corpora.}

The performance of the different models on the two tasks is outlined in Table \ref{tab_semeval_scores}; Table \ref{tab_semeval_words_disp} illustrates the results of our best-performing model on the ranking task. 

The standard deviations of the scores of the individual runs (see Table \ref{tab_semeval_scores}) emphasize once more a recommendation already made in Chapter 2: When reporting any score on an NLP task that is based on word embeddings, the mean and standard deviation over -- at least five -- subsequent runs should be provided. For \wtv, the ensemble average over shuffled corpora yields the highest overall score, but most of the scores fall within the $3\sigma$-confidence interval\footnote{The confidence intervals are based on the standard deviation of the mean $\sigma_{\mu}$; in Table \ref{tab_semeval_scores} only the standard deviation of the distribution $\sigma$ is reported. $\sigma_{\mu}$ is given by $\sigma / \sqrt{n}$, with $n=32$ for the models of size 1 and $n=2$ for the models of size 32.} of each other, hence the significance of this result is not entirely clear. For \glv\ and \ftt, the aligned average over bootstrapped corpora produces the highest scores, for \ftt\ the differences between this and any other model exceed $3\sigma$ and can therefore be considered significant. Comparing the different models over all three embedding techniques, we consider the \textsl{aligned average over embeddings trained on bootstrapped corpora} to be the most promising approach to detect semantic change -- it also yields the highest overall score (with \ftt\ embeddings). 

The scores of this model on the two tasks in each of the four languages is presented in Table \ref{tab_semeval_best_scores}: The results for German and Swedish are significantly better, than for English and Latin. This is not only true for the best-performing model, but for virtually any model we tried. A likely explanation for this can be found in Table \ref{tab_semeval_corpus_numbers}: The size of the training corpora for German and Swedish is more than 10 times larger than for English and Latin.

\begin{table}[htbp]
\begin{center}
\input{tables/tab_semeval_best_scores.tex}
\caption{Scores of our best-performing model (aligned average over 32 \ftt\ embeddings trained on bootstrapped corpora) on the two tasks in the four different languages.}
\label{tab_semeval_best_scores}
\end{center}
\end{table}

The models our team submitted during the official evaluation phase -- when the annotated results were not yet published, i.e. verifying the models was not possible -- had severe problems, hence we could not produce any meaningful contributions within the official competition. In the post-evaluation phase, i.e. after the annotations were published, our best submission ranks -- as of June 11th, 2020 -- 7th on Task 1 and 6th on Task 2 out of 34 participating teams (each team can submit an arbitrary number of models).\footnote{The leaderboard is publicly visible at \url{https://competitions.codalab.org/competitions/20948\#results} -- our submission was made under the team name \#hitsters.} Given the comparably small size of the test sets, it is fair to say our models are generally competitive, but they do not quite reach the state-of-the-art. We have to wait until September to find out how the better-performing models are built.

\section{Laws of Semantic Change}
\label{sec_laws_semantic_change}
Based on a study of \textbf{PPMI}, \textbf{SVD} and \wtv\ embeddings trained on different historical corpora, \citet{hamilton2016} proposed the \textsl{law of conformity}: Rarely used words exhibit -- on average -- higher rates of semantic change than more frequently used words. However, \citet{dubossarsky2017} contested the validity of these findings, by showing that the same correlation is observed in a control condition, i.e. on a corpus that is randomly split into multiple batches, which are then treated like different epochs of a genuine historical corpus.

In this section, we apply the instability-reducing technique of averaging over aligned embedding spaces, introduced in Section \ref{sec_average_multiple_embedding_spaces}, to the genuine historical COHA corpus, as well as a randomly composed control corpus. Thus, we try to differentiate between true semantic change as found on a genuine historical corpus and the intrinsic variability also observed on a randomly composed corpus, to put the \textsl{law of conformity} proposed by \citet{hamilton2016} to the test.\footnote{\citet{hamilton2016} and \citet{dubossarsky2017} furthermore examined the influence of polysemy and prototypicality on the rates of semantic change. However, \citet{dubossarsky2017} showed that neither of these word properties significantly improves the explained variance by the fixed effects, hence we place our focus on the effect of frequency.}

We trained \wtv, \glv\ and \ftt\ embeddings on 32 shuffled and bootstrapped versions of the 20 decades of the historical COHA corpus, that was also used by \citet{hamilton2016} and \citet{dubossarsky2017}. For the control condition, we repeated this procedure on a randomized historical corpus, which was compiled by accumulating the texts of the 20 decades of the COHA corpus and randomly splitting it into 20 batches. Out of the 32 embedding spaces for each decade/batch and type of document sampling, we computed the aligned average of size 2,4,8,16 and 32.

\begin{figure}[htbp]
\center
\vspace{-0.5cm}
\input{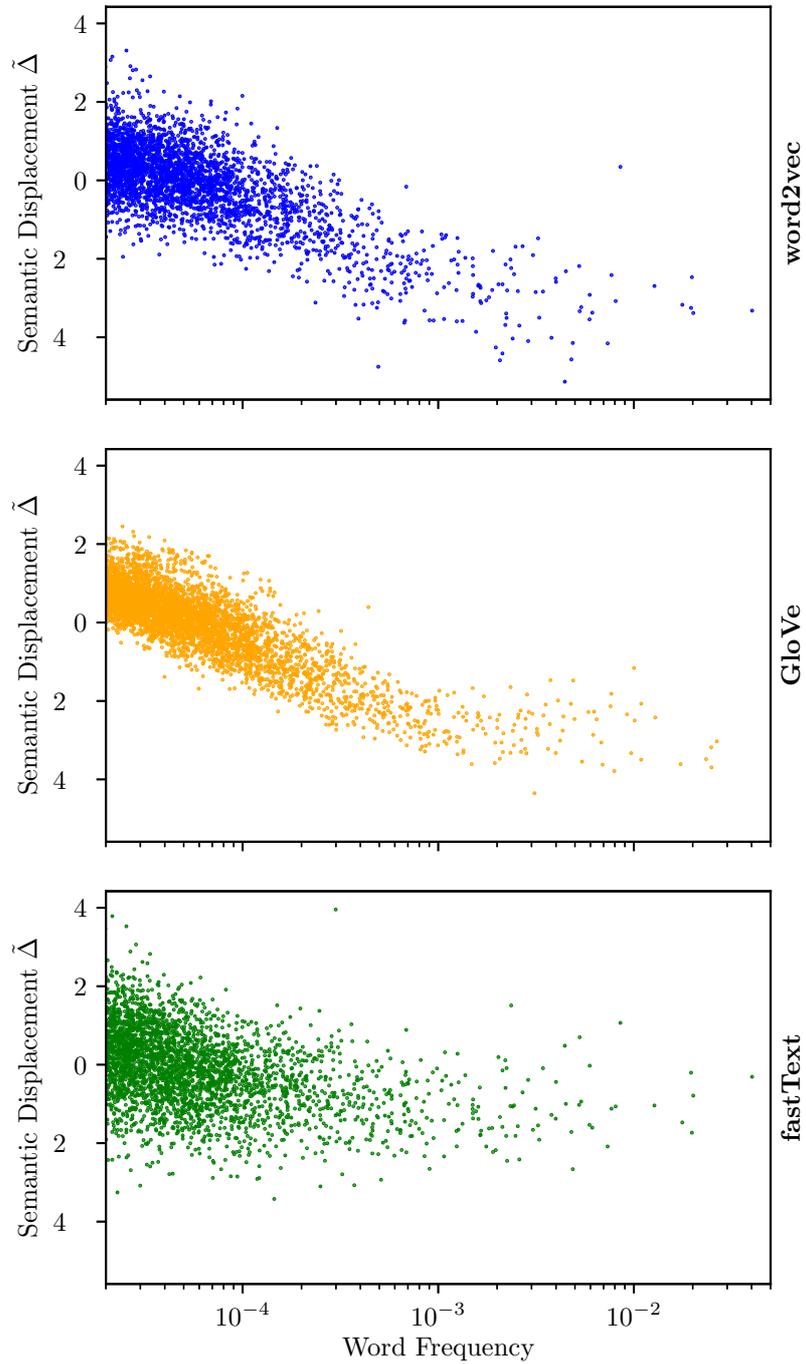}
\vspace{-0.5cm}
\caption{Semantic Displacement -- measured as the normalized $\log$ of the cosine distance of \wtv, \glv\ and \ftt\ embeddings of individual words from the 1990s to the 2000s -- over Word Frequency. The models correspond to individual runs of the embedding techniques on the respective decades of the COHA corpus.}
\label{fig_semantic_change_scatter}
\end{figure}

Then, the semantic change $\Delta(w,t)$ of every word $w$ between the epoch $t$ and $t+1$ -- if the word appears at least 500 times both epochs -- is computed based on Equation (\ref{eq_semantic_change}). Following the previous work, the rate of semantic change, as well as word frequency is log-transformed and standardized: The variables are then denoted as $\tilde{\Delta}$ and $\tilde{f}$ respectively. Figure \ref{fig_semantic_change_scatter} shows the semantic displacement of words between the 1990s and the 2000s as a function of their frequency. 

The influence of word frequency on semantic change is -- again following the previous work -- treated with a linear mixed effects model:
\begin{align}
\tilde{\Delta}(w,t) = \beta_0 + \beta_f \tilde{f}(w,t) + z(w) + \varepsilon(w,t)
\end{align}
where $\tilde{\Delta}(w,t)$ is the (log-transformed and standardized) rate of semantic change of the word $w$ between the temporal epochs $t$ and $t+1$, $\beta_0$ is the fixed intercept, $\beta_f$ is the fixed effect of word frequency, $z(w)\sim\mathcal{N}(0,\sigma)$ is a random, time-independent, intercept for word $w$ and $\varepsilon(w,t)$ is an error term associated with the individual measurement.

\begin{figure}[htbp]
\center
\vspace{-3.5cm}
\input{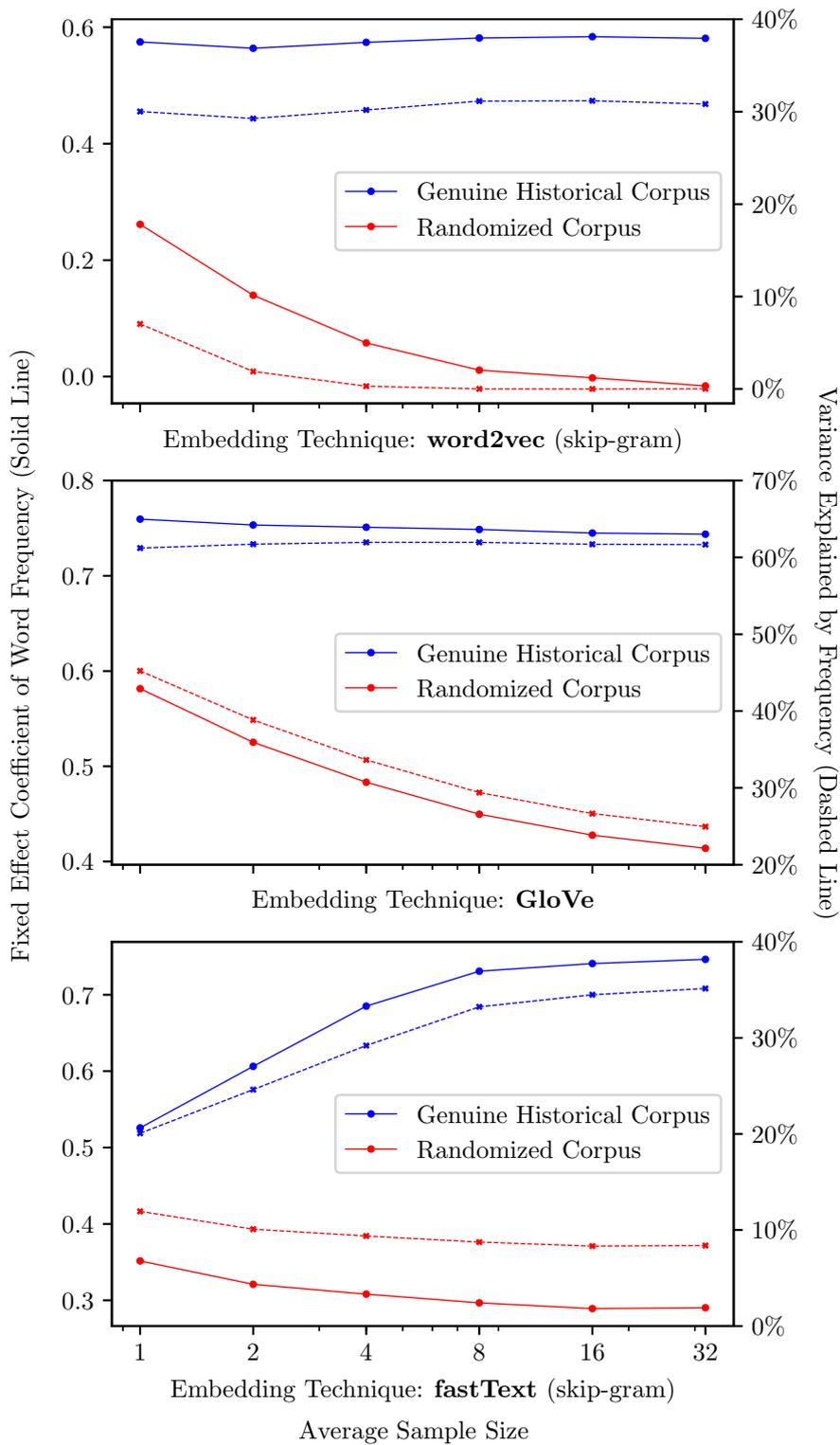}
\vspace{-0.5cm}
\caption{Fixed effect coefficient $\beta_f$ and variance explained by frequency of the mixed linear effects model as a function of the sample size of the aligned average of \wtv, \glv\ and \ftt\ embeddings trained on shuffled corpora.}
\label{fig_semantic_change_law}
\end{figure}

The fixed-effect predictor coefficient for frequency $\beta_f$, as well as the fraction of variance explained\footnote{Variance explained is the generalized $R^2$ for linear mixed effect models as defined by \citet{nakagawa2013}.} for the different embedding techniques, is plotted against the sample size of the aligned average in Figure \ref{fig_semantic_change_law}: The difference in the variance explained by frequency is significantly increasing through averaging for all three embedding techniques. 

The numerical results are summarized in Table \ref{tab_semantic_change_law}. \citet{dubossarsky2017} found only an $8\,\%$ difference in the variance explained by frequency between the genuine historical corpus and the control condition, and concluded, that the effect of frequency on the rate of semantic change ``may be real, but to a far lesser extent than had be claimed''. We argue, that averaging over aligned samples, hence reducing the intrinsic instability of word embedding models -- as shown in Section \ref{sec_influence_aa_stability} -- yields clearly distinct results for the genuine historical corpus and the random control condition. The difference in explained variance by frequency for the 32-fold average models are $31\,\%$, $37\,\%$ and $27\,\%$ for \wtv, \glv\ and \ftt, respectively. The fixed-effect coefficient $\beta_f$, which \citet{hamilton2016} placed in the interval $[-1.26,-0.27]$ is restricted to:
\begin{align}
\beta_f = [-0.75,-0.58]
\end{align}
in our experiments.
\begin{table}[htbp]
\center
\input{tables/tab_semantic_change_law.tex}
\caption{Fixed effect coefficient $\beta_f$ and variance explained by frequency of the mixed linear effects model for individual runs of \wtv, \glv\ and \ftt, as well as the aligned average over 32 embeddings spaces trained on shuffled corpora.}
\label{tab_semantic_change_law}
\end{table}

\section*{Conclusion}
Our experiments on Task 1 of the SemEval 2020 workshop showed, that word embeddings, trained on diachronic corpora, are a valid tool for detecting semantic change -- as judged by human experts -- and are competitive with the state-of-the-art. We found that the performance of this approach can be significantly improved by using the aligned average of multiple embedding spaces trained independently on bootstrapped versions of the corpus.

Furthermore, the minimization of the intrinsic instability of the embedding techniques through this approach enables to make the distinction between true semantic change and artifacts produces by the inherent instability of the embedding techniques. The values in Table \ref{tab_semantic_change_law} show differences of around $30\%$ of variance explained by frequency between the genuine historical corpus and the control condition -- for \wtv, \glv\ and \ftt\ embeddings. This supports the \textsl{law of conformity}: Word frequency correlates negatively with the rate of semantic change of a word.

%% file: tables/tab_semeval_words_disp.tex
\setlength{\dashlinedash}{0.3pt}
\setlength{\dashlinegap}{2.2pt}
\setlength{\arrayrulewidth}{0.3pt}

\begin{tabular}{l|x{0.7cm}p{2.7cm}|x{0.7cm}p{2.7cm}}
\multirow{2}{*}{\textbf{Corpus}} & \multicolumn{2}{c|}{\textbf{Human Annotations}} & \multicolumn{2}{c}{\textbf{Word Embeddings}} \\
& \multicolumn{1}{c}{\textsl{Rank}} &  \multicolumn{1}{l|}{\textsl{Word}} 
& \multicolumn{1}{c}{\textsl{Rank}} &  \multicolumn{1}{l}{\textsl{Word}} \\ 
\hline
								& 1 & \texttt{plane} 		& 1 &  \texttt{plane} 		\\
English							& 2 & \texttt{tip} 			& 2 &  \texttt{prop} 		\\
$t_1=1810-1860$					& 3 & \texttt{prop} 			& 3 &  \texttt{graft} 		\\
$t_2=1960-2010$					& 4 & \texttt{graft} 		& 4 &  \texttt{record} 		\\
								& 5 & \texttt{record} 		& 5 &  \texttt{player} 		\\
\hline
								& 1 & \texttt{abgebr\"{u}ht}	& 1 & \texttt{Engpaß} 		\\
German							& 2 & \texttt{Ohrwurm} 		& 2 & \texttt{Ohrwurm} 		\\
$t_1=1800-1899$					& 3 & \texttt{Engpaß} 		& 3 & \texttt{artikulieren} 	\\
$t_2=1946-1990$					& 4 & \texttt{abbauen} 		& 4 & \texttt{Sensation} 	\\
								& 5 & \texttt{ausspannen} 	& 5 & \texttt{abbauen} 		\\
\hline
								& 1 & \texttt{pontifex} 		& 1 &  \texttt{sanctus} 		\\
Latin							& 2 & \texttt{imperator} 	& 2 &  \texttt{titulus} 		\\
$t_1=200\,\text{BC}-0$			& 3 & \texttt{beatus} 		& 3 &  \texttt{adsumo} 		\\
$t_2=0-2000$						& 4 & \texttt{sacramentum} 	& 4 &  \texttt{sacramentum} 	\\
								& 5 & \texttt{titulus} 		& 5 &  \texttt{beatus} 		\\
\hline
								& 1 & \texttt{medium} 		& 1 &  \texttt{konduktör} 		\\
Swedish							& 2 & \texttt{krita} 		& 2 &  \texttt{antyda} 		\\
$t_1=1790-1830$					& 3 & \texttt{motiv} 		& 3 &  \texttt{medium} 		\\
$t_2=1895-1903$					& 4 & \texttt{ledning} 		& 4 &  \texttt{central} 		\\
								& 5 & \texttt{granskare} 	& 5 &  \texttt{aktiv} 		\\
\end{tabular}

%% file: tables/tab_semeval_scores.tex
\setlength{\dashlinedash}{0.3pt}
\setlength{\dashlinegap}{2.2pt}
\setlength{\arrayrulewidth}{0.3pt}

\begin{tabular}{cccc|cc|cc}
\multirow{2}{*}{\textbf{Model}} & \multirow{2}{*}{\textbf{Type}} & \multirow{2}{*}{\textbf{Size}} & \multirow{2}{*}{\textbf{Sampling}} &
\multicolumn{2}{c|}{	Task 1: \textbf{Binary}} & 
\multicolumn{2}{c}{		Task 2: \textbf{Ranking}} \\
&&&& \textsl{Mean} & \textsl{Std.} & \textsl{Mean} & \textsl{Std.} \\
\hline
\multirow{6}{*}{\rotatebox[origin=c]{90}{\wtv}} 
	& \multirow{2}{*}{IR}	& \multirow{2}{*}{1}	& shuffle		& $0.644$ 			& $0.018$	& $0.487$			& $0.018$ 	\\
	& 						& 						& bootstrap		& $0.635$ 			& $0.020$	& $0.467$			& $0.018$ 	\\
	\cdashline{2-8}
	& \multirow{2}{*}{AA}	& \multirow{2}{*}{32}	& shuffle 		& $0.615$			& $<0.004$	& $0.483$			& $<0.002$ 	\\
	& 		 				&						& bootstrap		& $0.651$		 	& $<0.007$	& $0.465$			& $<0.008$ 	\\
	\cdashline{2-8}
	& \multirow{2}{*}{EA}	& \multirow{2}{*}{32}	& \textbf{shuffle} & $\mathbf{0.657}$ & $\mathbf{<0.003}$ & $\mathbf{0.491}$ &  $\mathbf{<0.008}$ \\
	& 						& 						& bootstrap		& $0.642$		 	& $<0.006$	& $0.485$			& $<0.005$	\\
\hline
\multirow{6}{*}{\rotatebox[origin=c]{90}{\glv}}
	& \multirow{2}{*}{IR}	& \multirow{2}{*}{1}	& shuffle		& $0.585$ 			& $0.009$	& $0.258$			& $0.025$ 	\\
	& 						& 						& bootstrap		& $0.589$ 			& $0.016$	& $0.223$			& $0.046$ 	\\
	\cdashline{2-8}
	& \multirow{2}{*}{AA}	& \multirow{2}{*}{32}	& shuffle 		& $0.580$			& $<0.003$	& $0.241$			& $<0.006$ 	\\
	& 						& 						& \textbf{bootstrap} & $\mathbf{0.600}$ 	& $\mathbf{<0.010}$	& $\mathbf{0.311}$	& $\mathbf{<0.019}$ 	\\
	\cdashline{2-8}
	& \multirow{2}{*}{EA}	& \multirow{2}{*}{32}	& shuffle 		& $0.584$			& $<0.001$	& $0.267$			& $<0.005$	\\
	& 						& 						& bootstrap		& $0.587$		 	& $<0.001$	& $0.275$		 	& $<0.009$	\\
\hline
\multirow{6}{*}{\rotatebox[origin=c]{90}{\ftt}} 
	& \multirow{2}{*}{IR}	& \multirow{2}{*}{1}	& shuffle		& $0.653$ 			& $0.017$	& $0.431$			& $0.025$ 	\\
	& 						& 						& bootstrap		& $0.638$ 			& $0.023$	& $0.408$			& $0.034$ 	\\
	\cdashline{2-8}
	& \multirow{2}{*}{AA}	& \multirow{2}{*}{32}	& shuffle		& $0.650$			& $<0.001$	& $0.471$			& $<0.007$ 	\\
	& 						& 						& \textbf{bootstrap} & $\mathbf{0.674}$ 	& $\mathbf{<0.002}$	& $\mathbf{0.483}$	& $\mathbf{<0.005}$ 	\\
	\cdashline{2-8}
	& \multirow{2}{*}{EA}	& \multirow{2}{*}{32}	& shuffle		& $0.644$			& $<0.011$	& $0.458$			& $<0.007$	\\
	& 						& 						& bootstrap		& $0.637$		 	& $<0.011$	& $0.448$		 	& $<0.007$	\\
\end{tabular}

%% file: tables/tab_semeval_best_scores.tex
\begin{tabular}{l|cc|c}
\textbf{Language} 	& Task 1: \textbf{Binary} 	& 	Task 2: \textbf{Ranking} &	\textbf{Average}	\\
\hline
English 			& $0.703$					& $0.356$						& $0.530$				\\
German				& $0.750$					& $0.679$						& $0.714$				\\
Latin				& $0.500$					& $0.300$						& $0.400$				\\ 
Swedish				& $0.742$					& $0.597$						& $0.670$				\\	
\end{tabular}

%% file: tables/tab_semantic_change_law.tex
\setlength{\dashlinedash}{0.3pt}
\setlength{\dashlinegap}{2.2pt}
\setlength{\arrayrulewidth}{0.3pt}

\begin{tabular}{cc|cc|cc}
\multirow{2}{*}{\textbf{Model}} & \multirow{2}{*}{\textbf{Size}} & \multicolumn{2}{c|}{\textbf{Coefficient} $\beta_f$} & \multicolumn{2}{c}{\textbf{Expl. Variance}} \\
& & \textsl{Genuine} & \textsl{Random} & \textsl{Genuine} & \textsl{Random} \\
\hline
\multirow{2}{*}{\wtv}	& 1	 		&	$-0.57$ & $- 0.26$	& $30\,\%$	& $7\,\%$ 	\\
						& 32		&	$-0.58$ & $+ 0.02$	& $31\,\%$ 	& $0\,\%$	\\
\hdashline
\multirow{2}{*}{\glv}	& 1	 		&	$-0.76$ & $- 0.58$	& $61\,\%$ 	& $45\,\%$ 	\\
						& 32		&	$-0.74$ & $- 0.41$	& $62\,\%$ 	& $25\,\%$	\\
\hdashline
\multirow{2}{*}{\ftt}	& 1	 		&	$-0.53$ & $- 0.35$	& $20\,\%$	& $11\,\%$ 	\\
						& 32		&	$-0.75$ & $- 0.29$	& $35\,\%$ 	& $8\,\%$	\\
\end{tabular}

%% file: chapters/5_discussion.tex
\chapter{Discussion}
In this last chapter, we briefly discuss what we deem the most important results presented in this work and place them in the greater context of NLP research. 

We could reproduce the findings of \citet{hellrich-hahn-2016-assessment,hellrich-hahn-2016-bad,hellrich-hahn-2017-fool,hellrich-etal-2019-influence,antoniak2018,Chugh2018StabilityOW,wendlandt2018,pierrejean-tanguy-2018-predicting}, that the training processes of prediction-, as well as count-based non-contextualized word embeddings exhibit significant instability. The extent of this instability, that we found even for comparably large corpora might still come as a surprise to the reader. For example, out of 128 \ftt\ models \cite{bojanowski2016}, trained on independently shuffled versions of the Polish Wikipedia (470 million words), the relative difference in the score on the Polish word analogy dataset published by \citet{grave-etal-2018-learning}, between the best and worst performing models were close to $10\%$ (see Table \ref{tab_ftt_baseline_differences}). This supports the case, that every time a score on a word analogy dataset -- or any task that depends on word embeddings for that matter –- is published, \textsl{the results of at least five independent runs should be reported}. The current practice in research is to provide a single score, without information on its variance; our data indicates that this is insufficient.

The large number of experiments we conducted -- in total, over 10,000 embedding models were trained -- allow us to conclude that the distribution of the cosine similarity of any word pair over multiple runs of the same technique on independently shuffled corpora can be \textsl{closely approximated by a normal distribution} (see Section \ref{sec_cosine_similarity_gaussian}). We found this to be true for \wtv, \glv\ and \ftt\ embeddings, trained on any of the corpora outlined in Section \ref{sec_wiki_coprora}. This might not come as a surprise, but the observation is helpful in understanding the observed variability. In particular, it made us question the validity of nearest-neighbor based metrics, used in most of the previous work, to measure the instability of the embedding of a word $w$ over multiple runs. The \textsl{word-wise PIP loss}, based on the PIP loss of \citet{yin2018}, was proposed as an alternative.

Furthermore, the Gaussian nature of the instability suggests that averaging might help to reduce it. We used the methods developed by \citet{kulkarni2015} and \citet{zhang2015} and later modified by \citet{hamilton2016} to ``align'' two sets of embeddings and proposed a novel method to compute the aligned average over multiple embedding spaces. We found that this approach comes with a bias-variance trade-off, i.e. while the embeddings clearly converge through averaging (see Figure \ref{fig_pip_loss_average}), hence we can in fact minimize the instability, this does not always result in a higher quality of the embeddings, as measured by their score on word analogy tasks (see Table \ref{tab_improvements_score_average}). Therefore, less instability is not necessarily better.

Applying this method to the task of detecting and quantifying semantic change produced significantly better results than the individual runs for \ftt\ and \glv\ embeddings. For \wtv\ embeddings, the ensemble average over bootstrapped corpora proposed by \citet{antoniak2018} lead to the best results. Their approach captivates through its simplicity, but the computational expense is a significant drawback: One needs to store several embedding models and apply each of them individually to the respective downstream task. 

Finally, training several embedding models on 20 decades of historical American English, showed that the proposed averaging procedure significantly reduces the observed, artificial, effect of frequency in a control condition introduced by \citet{dubossarsky2017}, whereas the effect of frequency found in the genuine historical corpus was not diminished for \wtv\ and \glv, and even increased for \ftt\ (see Figure \ref{fig_semantic_change_law}). We argue that this supports the \textsl{law of conformity}, proposed by \citet{hamilton2016} at least for historical American English between 1800 and 2000.

Nevertheless, several questions concerning the instability of NLP models in general, remain unanswered: Specifically, the influence of the instability on more complex downstream tasks, and furthermore, the extent of the instability of more recently developed attention-based language models, like \textbf{ELMo} and \textbf{BERT}. 

%% file: chapters/appendix.tex
\nocite{*}
\bibliography{references}
\bibliographystyle{acl_natbib}
\appendix


\chapter{Supporting Content}

\section{Minimum Sample Size to Evaluate the Consistency of \texorpdfstring{$p_{\text{@}n}$}{p@n}}
\label{app_average_size_suffices}
To prove that the sample size of 16 subsequent runs is sufficient to draw conclusions on the underlying distribution and the values in Table \ref{tab_dependance_of_pj_on_n} are not a result of the inherent variations of the embeddings, we repeat the experiment described in Section \ref{sec_nn_consistency} with another set of 16 independent runs. We denote the evaluation of the two metrics $p_{\text{@}n}$ and $j_{\text{@}n}$ on the same target words, for this set of runs, as $\underline{p_{\text{@}n}}$ and $\underline{j_{\text{@}n}}$. The Spearman correlations between $\underline{p_{\text{@}n}}$ and $p_{\text{@}n}$, as well as $\underline{j_{\text{@}n}}$ and $j_{\text{@}n}$, i.e. the consistency of the results over the two independent experiments are shown in Table \ref{tab_average_size_suffices}. All correlation values are higher than $0.92$, which shows that the chosen sample size of 16 runs is indeed sufficient to conclude the underlying distribution.

\begin{table}[htbp]
\begin{center}
\input{tables/tab_average_size_suffices.tex}
\caption{Spearman correlation of the metrics $p_{\text{@}n}$ and $j_{\text{@}n}$ for 1000 target words for different values of $n\in \{ 2,5,10,25,50 \}$ for \wtv\ (top), \glv\ (middle) and \ftt\ (bottom) between two sets of 16 independent runs, obtained as outlined in Section \ref{sec_nn_consistency}. For each of the techniques, we show the average of the correlation for all languages mentioned in Section \ref{sec_exp_setup}.}
\label{tab_average_size_suffices}
\end{center}
\end{table}

\section{Reducing the Complexity of the Prediction of \texorpdfstring{$p_{\text{@}n}$}{p@n}}
\label{app_reducing_complexity_prediction}
As described in Section \ref{sec_previous_approaches}, we rely on numerical integration when predicting $p_{\text{@}1}$ with Equation \ref{eq_probability_top1_final}. Whereas this looks very resource-intensive at first glance, we can show that the relevant terms in the equation assume trivial values for most pairs of words, which renders the calculations considerably simpler.

A specific example from Table \ref{tab_gaussian_parameters} helps to illustrate this. We look at two words: Firstly, the -- on average -- nearest neighbor $w_{\text{\#}1}=\texttt{inertia}$ of the target word $w_t = \texttt{momentum}$ and secondly, the word with the 100th largest mean cosine similarity, $w_{\text{\#}100}=\texttt{inelastic}$. Now we want to estimate $p_{\text{\#1}}(w_t,w_{\text{\#}100})$, i.e. the probability that a word which usually ranks around position $100$, ends up on rank $1$ for one run. We can get an upper bound for this probability, by evaluating a weaker condition -- the probability that $\cos(w_t,w_{\#100})$ is larger than $\cos(w_t,w_{\#1})$. 

This probability, for a target word $w_t$ and the query words $w_s$ and $w_{s'}$, again assuming both cosine similarities are independent and normally distributed, is given by (please refer to Appendix \ref{app_proof_gaussian} for the derivation):
\begin{align}
p\left[\cos(w_t,w_s) > \cos(w_t,w_{s'})\right] = \frac{1}{2}\left[1+\text{erf}\left(\frac{\mu_{ts}-\mu_{ts'}}{\sqrt{2\left(\sigma_{ts}^2+\sigma_{ts'}^2\right)}}\right)\right]
\label{eq_criterion_relevance}
\end{align}

With the values from Table \ref{tab_gaussian_parameters} we find:\footnote{The vocabulary size of the models illustrated in Table \ref{tab_gaussian_parameters} was restricted to 200,000 to simplify the calculations for this demonstrative example. However, this leads to a higher stability than we generally observe in practice.}
\begin{align}
p\left[\cos(w_t,w_{\#100}) > \cos(w_t,w_{\#1})\right] = \frac{1}{2}\left[1+\text{erf}\left(A\right)\right]&\approx 2.74\times 10^{-33}\\
\text{with} \quad A = \frac{0.489-0.650}{\sqrt{2\left(0.009^2+0.010^2\right)}}&\approx - 8.46
\end{align}
And this is an upper bound on the probability $p_{\text{\#1}}(w_t,w_{\text{\#}100})$, thus:
\begin{align}
p_{\text{\#1}}(w_t,w_{\text{\#}100})\leq 2.74\times 10^{-33}
\end{align}

This shows that the probability (\ref{eq_probability_top1_final}) will practically be zero for most query words $w_s$. Hence, for any practical calculations of the probabilities in Equation (\ref{eq_probability_top1_final}), we can disregard most words of the vocabulary, and reduce the scope to the nearest neighbors of the target word.\footnote{In practice, we evaluate (\ref{eq_criterion_relevance}) for all words in the vocabulary for a given target word $w_t$ and disregard any query word $w_s$ in the subsequent calculation if $p\left[\cos(w_t,w_s) > \cos(w_t,w_{\text{\#}1})\right]$ is smaller than a certain threshold -- usually $1.0\times 10^{-5}$. }

\section{Probability of One Normally Distributed Random Variable to be Larger Than Another}
\label{app_proof_gaussian}

Two random variables $x$ and $y$ both follow a normal distribution:
\begin{align}
x\sim\mathcal{N}(\mu_x,\sigma_x) \quad \quad \quad y\sim\mathcal{N}(\mu_y,\sigma_y)
\end{align}

Now we want to calculate $p(x>y)$, i.e. the probability, that if we randomly sample the value $x_i$ from $\mathcal{N}(\mu_x,\sigma_x)$ and $y_i$ from $\mathcal{N}(\mu_y,\sigma_y)$, the condition $x_i>y_i$ is fulfilled. 

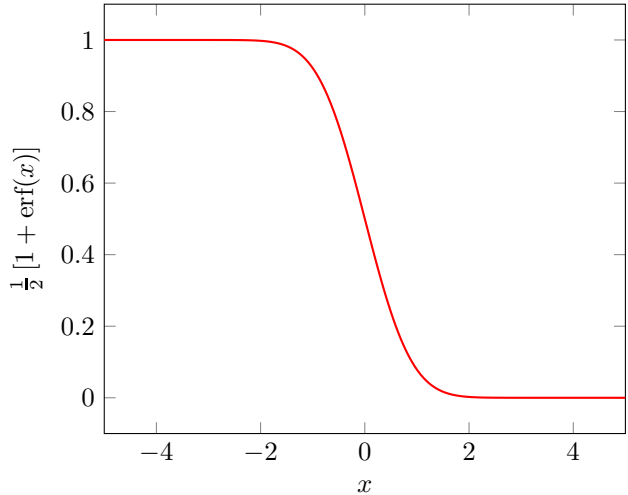
\begin{figure}
\center
\input{tikz/fig_error_function.tex}
\caption{Illustration of the expression in Equation \ref{eq_proof_gaussian}, and its strong convergence based on the Gaussian error function.}
\label{fig_error_function}
\end{figure}

First, let us introduce the variable $z:=x-y$. We know that the \textsl{difference of two normally distributed random variables is again a normal distribution} \cite{lemons2002introduction}, i.e. $z\sim\mathcal{N}(\mu_z,\sigma_z)$ with the parameters:
\begin{align}
\mu_z = \mu_x-\mu_y,\quad \sigma_z = \sqrt{\sigma_x^2+\sigma_y^2}
\end{align}
As $x>y\iff z>0$, we can determine $p(x>y)$ by integrating the distribution $\mathcal{N}(\mu_z,\sigma_z)$ from $0$ to $+\infty$:
\begin{align}
p(x>y) = p(z>0)=\int_0^{+\infty} \frac{1}{\sqrt{2\pi}\sigma_z}\exp\left[- \frac{1}{2}\left(\frac{z-\mu_z}{\sigma_z}\right)^2\right]\ \text{d}z
\end{align} 
This non-trivial integral of the normal distribution yields the Gaussian error function \cite{andrews1998special}:
\begin{align}
p(x>y) = \frac{1}{2}\left[1+\text{erf}\left(\frac{\mu_z}{\sqrt{2\sigma_z}}\right)\right] = \frac{1}{2}\left[1+\text{erf}\left(\frac{\mu_x-\mu_y}{\sqrt{2\left(\sigma_x^2+\sigma_y^2\right)}}\right)\right]
\label{eq_proof_gaussian}
\end{align}
The course of the function and its strong convergence is illustrated in Figure \ref{fig_error_function}.

\section{Comparison of the Prediction and Observation of \texorpdfstring{$p_{\text{@}n}$}{p@n}}
\label{app_agreement_prediction_instability}

\begin{figure}[htbp]
\center
\vspace{-0.6cm}
\input{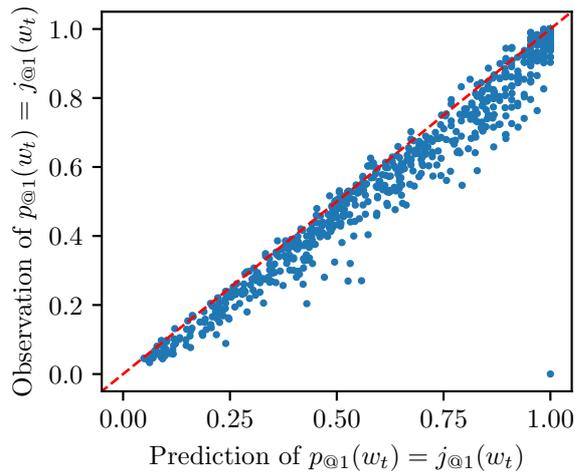}
\vspace{-0.7cm}
\caption{Plot of the predicted against the observed values of $p_{\text{@}1}=j_{\text{@}1}$ for 1000 randomly sampled target words, obtained from 128 runs of \ftt\ on a Portuguese Wikipedia extract. Aside from the expected random fluctuations, prediction and observation coincide approximately, which suggests that our theory and the underlying assumptions on the distribution of the embedding spaces are -- in good approximation -- valid. However, since most of the data points (blue) fall below the bisection of the coordinate axes (red), the prediction seems to have a slight systematic error in overestimating $p_{\text{@}1}=j_{\text{@}1}$. Since we only use the nearest neighbors of the target word in our prediction (to increase the computational performance) we would expect to overestimate the stability, but we cannot say with certainty that this is the only reason for the observed difference between prediction and measurement.}
\label{fig_prediction_observation_overlap}
\end{figure}

In order to test the assumptions and the subsequent derivation of $p_{\text{@}n}$ in Section \ref{sec_previous_approaches}, we compare the measurements of $p_{\text{@}1}=j_{\text{@}1}$ with the expectation based on Equation (\ref{eq_expected_overlap}) for various sets of embeddings. For every technique and language outlined in Section \ref{sec_exp_setup}, we randomly sample 1000 sampled target words, measure the overlap $p_{\text{@}1}$ over 128 runs on independently shuffled corpora and predict the same property based on the estimation of the Gaussian parameters. Figure \ref{fig_prediction_observation_overlap} shows the agreement between prediction and observation for 128 runs of \ftt\ on the Portuguese Wikipedia. In Table \ref{tab_pearson_overlap}, one can see the Pearson correlation coefficient $\rho$ \cite{freedman2007statistics} between prediction and measurement for the 1000 target words in every language and technique. In all our experiments, predictions and observations agree ($\rho > 0.95$), which indicates that the derivations above, as well as the underlying assumptions, are valid.

\begin{table}[htbp]
\center
\input{tables/tab_pearson_overlap.tex}
\caption{Pearson correlation coefficient between prediction and measurements of $p_{\text{@}1}(w_t)=j_{\text{@}1}(w_t)$, for 1000 randomly sampled target words $w_t$ obtained from 128 runs of \wtv\ (skip-gram), \glv\ and \ftt\ (skip-gram) on Wikipedia corpora in seven different languages.}
\label{tab_pearson_overlap}
\end{table}

\begin{figure}[htbp]
\center
\vspace{-0.6cm}
\input{"./plots/prediction_observation_overlap_n=2_fi_fasttext.pgf"}
\vspace{-0.7cm}
\caption{Plot of the predicted against the observed values of $p_{\text{@}2}$ for 200 randomly sampled target words $w_t$, obtained from 128 runs of \ftt\ on a Finnish Wikipedia extract.}
\label{fig_prediction_observation_overlap_n=2}
\end{figure}
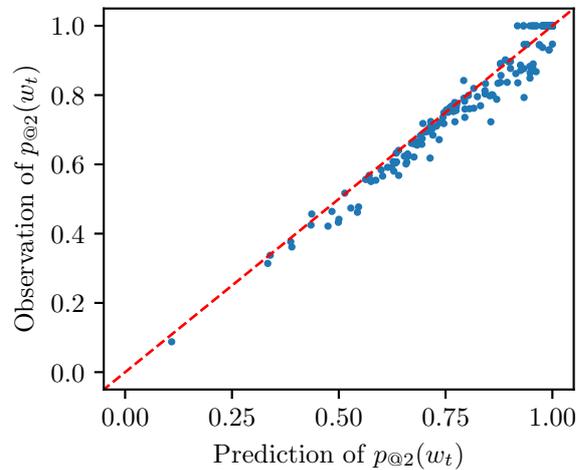

As mentioned before, predicting the overlap for larger $n$ is becoming increasingly complex: The derivation of the prediction of $p_{\text{@}n}$ and $j_{\text{@}n}$ for $n=2$ is outlined in Appendix \ref{app_prediction_n=2}. Figure \ref{fig_prediction_observation_overlap_n=2} shows the prediction and observation of $p_{\text{@}n}$ for $n=2$ for 200 target words from \ftt\ embeddings obtained from the Finnish Wikipedia. Similar to $n=1$, the two quantities match rather well. Altogether, we are not aware of a reason to expect any different outcome for $n>1$ and will therefore, for sake of simplicity, continue to focus on $n=1$.

\section{Prediction of \texorpdfstring{$p_{\text{@}n}$ and $j_{\text{@}n}$ for $n>1$}{p@n and j@n for n>1}}
\label{app_prediction_n=2}

In this section, we give an outlook on how the metrics $p_{\text{@}n}$ and $j_{\text{@}n}$ can be predicted from the parameters $\mu_{ij},\sigma_{ij}$ of the Normal distributions of the cosine similarities $\cos(w_i,w_j)\sim\mathcal{N}(\mu_{ij},\sigma_{ij})\quad \forall w_i,w_j\in\mathcal{V}$, \textsl{for the case $n>1$}. We specifically show how the case $n=2$ can be derived from $n=1$ and hence provide the instruments to handle any value of $n$ in an iterative manner.

Let $p_{\text{\#}2}(w_t,w_s)$ denote the probability that $w_s$ is one of the two nearest neighbors of $w_t$, for one randomly sampled embedding space $\mathbf{V}_k\sim \Omega(\mathcal{T},\mathcal{C})$. Naturally, this is the sum of the probabilities $p_{\text{\#}1}(w_t,w_s)$, i.e. of $w_s$ being the nearest neighbor and $p_{\text{\#}\underline{2}}(w_t,w_s)$ -- the chance that $w_s$ is exactly the second closest word to $w_t$ by cosine distance:
\begin{align}
p_{\text{\#}2}(w_t,w_s)= p_{\text{\#}1}(w_t,w_s) +p_{\text{\#}\underline{2}}(w_t,w_s)
\end{align}

Equation (\ref{eq_probability_top1_final}) already yields the first term, hence we only need to derive $p_{\text{\#}\underline{2}}(w_t,w_s)$: If $w_s$ is the second nearest neighbor of the target word, there is exactly one word, which we call $w_n$, that is closer to the target word. In principle, this could be any word of the vocabulary and to obtain $p_{\text{\#}\underline{2}}(w_t,w_s)$ we need to derive the sum of the probability of all possible constellations, i.e. for all $w_{n}\in\mathcal{V}\setminus \{w_t,w_s\}$. For any word $w_{n}$, we are therefore interested in the probability of the case:
\begin{align}
\cos(w_t,w_n) > \cos(w_t,w_s) > \cos(w_t,w_{s'}) \quad \forall w_{s'}\in \mathcal{V}\setminus\{w_t,w_s,w_n\}
\end{align}
For the sake of readability, we fix an arbitrary pair $w_t$, $w_s$ and introduce the following notation:
\begin{align}
\begin{split}
\cos(w_t,w_n)&=:\bar{x}_n\sim\mathcal{N}\left(\bar{\mu}_n,\bar{\sigma}_n^2\right) \\
\cos(w_t,w_s)&=:\tilde{x}\sim\mathcal{N}\left(\tilde{\mu},\tilde{\sigma}^2\right) \\
\cos(w_t,w_{s'})&=:x_j \sim\mathcal{N}\left(\mu_j,\sigma_j^2\right) \quad \text{with} \quad j \in \{1,2,...,v-3\}
\end{split}
\label{eq_condition_top2}
\end{align}

To determine the probability $p_{\text{\#}1,2}(w_t,w_s,w_n)$ that a triple of words $w_t,w_s,w_n$ fulfils condition (\ref{eq_condition_top2}), we need to integrate the joint probability distribution $p(\bar{x}_n,\tilde{x},x_1,...,x_{v-2})$ over the respective subspace of $\mathbb{R}^{v-1}$. Here we need the assumption of independence for the random variables $\bar{x}_n,\tilde{x},x_1,...,x_{v-2}$ once more.\footnote{See Section \ref{sec_cosine_similarity_gaussian} for evidence on the validity of this assumption.} Then:
\begin{align}
p(\bar{x}_n,\tilde{x},x_1,...,x_{v-2})=p(\bar{x}_n)\cdot p(\tilde{x})\cdot p(x_1)\cdot ... \cdot p(x_{v-2})
\end{align}
Now, for a given value of $\bar{x}_n$, $\tilde{x}$ can assume any value smaller than $\bar{x}_n$, while the $x_j$ need to be smaller than $\tilde{x}$. This means:
\begin{align}
\begin{split}
&p_{\text{\#}1,2}(w_t,w_s,w_n) = \\
 &\int_{-\infty}^{\infty} \left\{\int_{-\infty}^{\bar{x}_n} \left[\prod_{j=1}^{v-3}\int_{-\infty}^{\tilde{x}} f(x_j,\mu_j,\sigma_j)\, \text{d}x_j   \right]f(\tilde{x},\tilde{\mu},\tilde{\sigma})\,\text{d}\tilde{x} \right\}f(\bar{x}_n,\bar{\mu}_n,\bar{\sigma}_n)\,\text{d}\bar{x}_n
\end{split}
\end{align}
Where the $f(x,\mu,\sigma)$ denote the \textsl{probability density function} of the Normal distribution with mean $\mu$ and variance $\sigma^2$. Using Equation (\ref{eq_gauss_integral}) yields:
\begin{align}
\begin{split}
&p_{\text{\#}1,2}(w_t,w_s,w_n) = \\
&\int_{-\infty}^{\infty} \left( \int_{-\infty}^{\bar{x}_n} \left\{ \prod_{j=1}^{v-3}\frac{1}{2}\left[ \text{erf}\left(\frac{\tilde{x}-\mu_j}{\sqrt{2}\sigma_j}\right)+1 \right] \right\}f(\tilde{x},\tilde{\mu},\tilde{\sigma})\,\text{d}\tilde{x} \right) f(\bar{x}_n,\bar{\mu}_n,\bar{\sigma}_n)\,\text{d}\bar{x}_n \end{split}
\end{align}
To compute this, we use numerical integration. Finally:
\begin{align}
p_{\text{\#}2}(w_t,w_s)=p_{\text{\#}1}(w_t,w_s) + \sum_{w_n\in\mathcal{V}\setminus \{w_t,w_s\}}&p_{\text{\#}1,2}(w_t,w_s,w_n)
\label{eq_top2}
\end{align}
And this allows us to predict the metric $p_{\text{@}2}$ for the target word $w_t$ as:
\begin{align}
p_{\text{@}2}(w_t)=\frac{1}{2}\left(\sum_{w_s\in\mathcal{V}\setminus \{w_t\}} \left[p_{\text{\#2}}(w_t,w_s)\right]^2\right)
\end{align} 


\selectlanguage{ngerman}
\newpage

Erkl\"{a}rung:\par
\vspace{3\baselineskip}
Ich versichere, dass ich diese Arbeit selbstst\"{a}ndig verfasst habe und keine
anderen als die angegebenen Quellen und Hilfsmittel benutzt habe.\par
\vspace{5\baselineskip}
Heidelberg, den \today\hspace{3cm}\dotfill

%% file: tables/tab_average_size_suffices.tex
\setlength\tabcolsep{3.0pt}
\resizebox{\columnwidth}{!}{
\begin{tabular}{p{0.70cm}|x{0.70cm}x{0.70cm}x{0.70cm}x{0.70cm}x{0.70cm}}
			 	& $\underline{p_{\text{@}2}}$ 	& $\underline{p_{\text{@}5}}$ 	& $\underline{p_{\text{@}10}}$ 	& $\underline{p_{\text{@}25}}$	& $\underline{p_{\text{@}50}}$ \\
\hline
$p_{\text{@}2}$ 	& 0.93				& 	 				& 					& 					& 					\\ 
$p_{\text{@}5}$ 	& 					& 0.96 				& 	 				& 					& 					\\
$p_{\text{@}10}$	&					& 	 				& 0.98 				& 					& 					\\
$p_{\text{@}25}$	&  					& 	 				& 	 				& 0.99				& 					\\
$p_{\text{@}50}$	&  					& 	 				& 	 				& 					& 0.99				\\
\end{tabular}
\quad\quad
\begin{tabular}{p{0.70cm}|x{0.70cm}x{0.70cm}x{0.70cm}x{0.70cm}x{0.70cm}}
			 	& $\underline{j_{\text{@}2}}$ 	& $\underline{j_{\text{@}5}}$ 	& $\underline{j_{\text{@}10}}$ 	& $\underline{j_{\text{@}25}}$	& $\underline{j_{\text{@}50}}$ \\
\hline
$j_{\text{@}2}$ 	& 0.93				& 	 				& 					&					& 					\\ 
$j_{\text{@}5}$ 	& 					& 0.96 				& 	 				& 					& 					\\
$j_{\text{@}10}$	&					& 	 				& 0.98 				&					&					\\
$j_{\text{@}25}$	&  					& 	 				& 	 				& 0.99				& 					\\
$j_{\text{@}50}$	&  					& 	 				& 	 				& 					& 0.99				\\
\end{tabular}
}
\\
\vspace{0.1cm}
Embedding Technique: \wtv\ (skip-gram) \\
\vspace{0.3cm}
\resizebox{\columnwidth}{!}{
\begin{tabular}{p{0.70cm}|x{0.70cm}x{0.70cm}x{0.70cm}x{0.70cm}x{0.70cm}}
			 	& $\underline{p_{\text{@}2}}$ 	& $\underline{p_{\text{@}5}}$ 	& $\underline{p_{\text{@}10}}$ 	& $\underline{p_{\text{@}25}}$	& $\underline{p_{\text{@}50}}$ \\
\hline
$p_{\text{@}2}$ 	& 0.92				& 	 				& 					& 					& 					\\ 
$p_{\text{@}5}$ 	& 					& 0.95 				& 	 				& 					& 					\\
$p_{\text{@}10}$	&					& 	 				& 0.97				& 					& 					\\
$p_{\text{@}25}$	&  					& 	 				& 	 				& 0.99				& 					\\
$p_{\text{@}50}$	&  					& 	 				& 	 				& 					& 0.99				\\
\end{tabular}
\quad\quad
\begin{tabular}{p{0.70cm}|x{0.70cm}x{0.70cm}x{0.70cm}x{0.70cm}x{0.70cm}}
			 	& $\underline{j_{\text{@}2}}$ 	& $\underline{j_{\text{@}5}}$ 	& $\underline{j_{\text{@}10}}$ 	& $\underline{j_{\text{@}25}}$	& $\underline{j_{\text{@}50}}$ \\
\hline
$j_{\text{@}2}$ 	& 0.92				& 	 				& 					&					& 					\\ 
$j_{\text{@}5}$ 	& 					& 0.95 				& 	 				& 					& 					\\
$j_{\text{@}10}$	&					& 	 				& 0.97 				&					&					\\
$j_{\text{@}25}$	&  					& 	 				& 	 				& 0.99				& 					\\
$j_{\text{@}50}$	&  					& 	 				& 	 				& 					& 0.99				\\
\end{tabular}
}
\\
\vspace{0.1cm}
Embedding Technique: \glv \\
\vspace{0.3cm}
\resizebox{\columnwidth}{!}{
\begin{tabular}{p{0.70cm}|x{0.70cm}x{0.70cm}x{0.70cm}x{0.70cm}x{0.70cm}}
			 	& $\underline{p_{\text{@}2}}$ 	& $\underline{p_{\text{@}5}}$ 	& $\underline{p_{\text{@}10}}$ 	& $\underline{p_{\text{@}25}}$	& $\underline{p_{\text{@}50}}$ \\
\hline
$p_{\text{@}2}$ 	& 0.94				& 	 				& 					& 					& 					\\ 
$p_{\text{@}5}$ 	& 					& 0.97 				& 	 				& 					& 					\\
$p_{\text{@}10}$	&					& 	 				& 0.99 				& 					& 					\\
$p_{\text{@}25}$	&  					& 	 				& 	 				& 0.99				& 					\\
$p_{\text{@}50}$	&  					& 	 				& 	 				& 					& 0.99				\\
\end{tabular}
\quad\quad
\begin{tabular}{p{0.70cm}|x{0.70cm}x{0.70cm}x{0.70cm}x{0.70cm}x{0.70cm}}
			 	& $\underline{j_{\text{@}2}}$ 	& $\underline{j_{\text{@}5}}$ 	& $\underline{j_{\text{@}10}}$ 	& $\underline{j_{\text{@}25}}$	& $\underline{j_{\text{@}50}}$ \\
\hline
$j_{\text{@}2}$ 	& 0.94				& 	 				& 					&					& 					\\ 
$j_{\text{@}5}$ 	& 					& 0.97 				& 	 				& 					& 					\\
$j_{\text{@}10}$	&					& 	 				& 0.99 				&					&					\\
$j_{\text{@}25}$	&  					& 	 				& 	 				& 0.99				& 					\\
$j_{\text{@}50}$	&  					& 	 				& 	 				& 					& 0.99				\\
\end{tabular}
}
\\
\vspace{0.1cm}
Embedding Technique: \ftt\ (skip-gram) \\
\vspace{0.3cm}

%% file: tikz/fig_error_function.tex
\begin{tikzpicture}
\begin{axis}
[xmin = -5, xmax = 5, ymin = -0.1, ymax = 1.1, xlabel = {$x$}, ylabel = {$\frac{1}{2}\left[1+\text{erf}(x)\right]$}]
\addplot[domain = -5:5, samples = 400, color = red, style = thick] (\x,{0.5 * ( 1 - erf(\x))});
\end{axis}
\end{tikzpicture}

%% file: tables/tab_pearson_overlap.tex
\begin{tabular}{c|x{2cm}x{2cm}x{2cm}}
\textbf{Language}				& \wtv			 		& \glv 				& \ftt 				\\
\hline
\textsc{Hi} 					& 0.969					& 0.968				& 0.984				\\
\textsc{Fi} 					& 0.976					& 0.995				& 0.974				\\
\textsc{Zh} 					& 0.980					& 0.988				& 0.989				\\
\textsc{Cs} 					& 0.978					& 0.992				& 0.983				\\
\textsc{Pl} 					& 0.980					& 0.991				& 0.970				\\
\textsc{Pt} 					& 0.983					& 0.978				& 0.984				\\
\textsc{En} 					& 0.975					& 0.973				& 0.985				\\
\end{tabular}

%% file: plots/prediction_observation_overlap_n=2_fi_fasttext.pgf
\begingroup%
\makeatletter%
\begin{pgfpicture}%
\pgfpathrectangle{\pgfpointorigin}{\pgfqpoint{3.500000in}{2.800000in}}%
\pgfusepath{use as bounding box, clip}%
\begin{pgfscope}%
\pgfsetbuttcap%
\pgfsetmiterjoin%
\definecolor{currentfill}{rgb}{1.000000,1.000000,1.000000}%
\pgfsetfillcolor{currentfill}%
\pgfsetlinewidth{0.000000pt}%
\definecolor{currentstroke}{rgb}{1.000000,1.000000,1.000000}%
\pgfsetstrokecolor{currentstroke}%
\pgfsetdash{}{0pt}%
\pgfpathmoveto{\pgfqpoint{0.000000in}{0.000000in}}%
\pgfpathlineto{\pgfqpoint{3.500000in}{0.000000in}}%
\pgfpathlineto{\pgfqpoint{3.500000in}{2.800000in}}%
\pgfpathlineto{\pgfqpoint{0.000000in}{2.800000in}}%
\pgfpathclose%
\pgfusepath{fill}%
\end{pgfscope}%
\begin{pgfscope}%
\pgfsetbuttcap%
\pgfsetmiterjoin%
\definecolor{currentfill}{rgb}{1.000000,1.000000,1.000000}%
\pgfsetfillcolor{currentfill}%
\pgfsetlinewidth{0.000000pt}%
\definecolor{currentstroke}{rgb}{0.000000,0.000000,0.000000}%
\pgfsetstrokecolor{currentstroke}%
\pgfsetstrokeopacity{0.000000}%
\pgfsetdash{}{0pt}%
\pgfpathmoveto{\pgfqpoint{0.916403in}{0.633903in}}%
\pgfpathlineto{\pgfqpoint{3.350000in}{0.633903in}}%
\pgfpathlineto{\pgfqpoint{3.350000in}{2.623000in}}%
\pgfpathlineto{\pgfqpoint{0.916403in}{2.623000in}}%
\pgfpathclose%
\pgfusepath{fill}%
\end{pgfscope}%
\begin{pgfscope}%
\pgfpathrectangle{\pgfqpoint{0.916403in}{0.633903in}}{\pgfqpoint{2.433597in}{1.989097in}}%
\pgfusepath{clip}%
\pgfsetbuttcap%
\pgfsetroundjoin%
\definecolor{currentfill}{rgb}{0.121569,0.466667,0.705882}%
\pgfsetfillcolor{currentfill}%
\pgfsetlinewidth{1.003750pt}%
\definecolor{currentstroke}{rgb}{0.121569,0.466667,0.705882}%
\pgfsetstrokecolor{currentstroke}%
\pgfsetdash{}{0pt}%
\pgfsys@defobject{currentmarker}{\pgfqpoint{-0.012028in}{-0.012028in}}{\pgfqpoint{0.012028in}{0.012028in}}{%
\pgfpathmoveto{\pgfqpoint{0.000000in}{-0.012028in}}%
\pgfpathcurveto{\pgfqpoint{0.003190in}{-0.012028in}}{\pgfqpoint{0.006250in}{-0.010761in}}{\pgfqpoint{0.008505in}{-0.008505in}}%
\pgfpathcurveto{\pgfqpoint{0.010761in}{-0.006250in}}{\pgfqpoint{0.012028in}{-0.003190in}}{\pgfqpoint{0.012028in}{0.000000in}}%
\pgfpathcurveto{\pgfqpoint{0.012028in}{0.003190in}}{\pgfqpoint{0.010761in}{0.006250in}}{\pgfqpoint{0.008505in}{0.008505in}}%
\pgfpathcurveto{\pgfqpoint{0.006250in}{0.010761in}}{\pgfqpoint{0.003190in}{0.012028in}}{\pgfqpoint{0.000000in}{0.012028in}}%
\pgfpathcurveto{\pgfqpoint{-0.003190in}{0.012028in}}{\pgfqpoint{-0.006250in}{0.010761in}}{\pgfqpoint{-0.008505in}{0.008505in}}%
\pgfpathcurveto{\pgfqpoint{-0.010761in}{0.006250in}}{\pgfqpoint{-0.012028in}{0.003190in}}{\pgfqpoint{-0.012028in}{0.000000in}}%
\pgfpathcurveto{\pgfqpoint{-0.012028in}{-0.003190in}}{\pgfqpoint{-0.010761in}{-0.006250in}}{\pgfqpoint{-0.008505in}{-0.008505in}}%
\pgfpathcurveto{\pgfqpoint{-0.006250in}{-0.010761in}}{\pgfqpoint{-0.003190in}{-0.012028in}}{\pgfqpoint{0.000000in}{-0.012028in}}%
\pgfpathclose%
\pgfusepath{stroke,fill}%
}%
\begin{pgfscope}%
\pgfsys@transformshift{2.427303in}{1.816664in}%
\pgfsys@useobject{currentmarker}{}%
\end{pgfscope}%
\begin{pgfscope}%
\pgfsys@transformshift{2.741955in}{2.109105in}%
\pgfsys@useobject{currentmarker}{}%
\end{pgfscope}%
\begin{pgfscope}%
\pgfsys@transformshift{2.302776in}{1.724393in}%
\pgfsys@useobject{currentmarker}{}%
\end{pgfscope}%
\begin{pgfscope}%
\pgfsys@transformshift{2.784688in}{2.138722in}%
\pgfsys@useobject{currentmarker}{}%
\end{pgfscope}%
\begin{pgfscope}%
\pgfsys@transformshift{3.239382in}{2.532586in}%
\pgfsys@useobject{currentmarker}{}%
\end{pgfscope}%
\begin{pgfscope}%
\pgfsys@transformshift{2.677173in}{2.022193in}%
\pgfsys@useobject{currentmarker}{}%
\end{pgfscope}%
\begin{pgfscope}%
\pgfsys@transformshift{3.239382in}{2.532586in}%
\pgfsys@useobject{currentmarker}{}%
\end{pgfscope}%
\begin{pgfscope}%
\pgfsys@transformshift{2.617291in}{1.995693in}%
\pgfsys@useobject{currentmarker}{}%
\end{pgfscope}%
\begin{pgfscope}%
\pgfsys@transformshift{3.171879in}{2.431916in}%
\pgfsys@useobject{currentmarker}{}%
\end{pgfscope}%
\begin{pgfscope}%
\pgfsys@transformshift{2.915067in}{2.167354in}%
\pgfsys@useobject{currentmarker}{}%
\end{pgfscope}%
\begin{pgfscope}%
\pgfsys@transformshift{2.832186in}{2.205695in}%
\pgfsys@useobject{currentmarker}{}%
\end{pgfscope}%
\begin{pgfscope}%
\pgfsys@transformshift{2.272427in}{1.730308in}%
\pgfsys@useobject{currentmarker}{}%
\end{pgfscope}%
\begin{pgfscope}%
\pgfsys@transformshift{2.726031in}{2.084940in}%
\pgfsys@useobject{currentmarker}{}%
\end{pgfscope}%
\begin{pgfscope}%
\pgfsys@transformshift{2.360480in}{1.747947in}%
\pgfsys@useobject{currentmarker}{}%
\end{pgfscope}%
\begin{pgfscope}%
\pgfsys@transformshift{2.686836in}{2.084339in}%
\pgfsys@useobject{currentmarker}{}%
\end{pgfscope}%
\begin{pgfscope}%
\pgfsys@transformshift{3.239382in}{2.532586in}%
\pgfsys@useobject{currentmarker}{}%
\end{pgfscope}%
\begin{pgfscope}%
\pgfsys@transformshift{2.097817in}{1.563604in}%
\pgfsys@useobject{currentmarker}{}%
\end{pgfscope}%
\begin{pgfscope}%
\pgfsys@transformshift{3.222098in}{2.532586in}%
\pgfsys@useobject{currentmarker}{}%
\end{pgfscope}%
\begin{pgfscope}%
\pgfsys@transformshift{3.239382in}{2.532586in}%
\pgfsys@useobject{currentmarker}{}%
\end{pgfscope}%
\begin{pgfscope}%
\pgfsys@transformshift{1.993160in}{1.550121in}%
\pgfsys@useobject{currentmarker}{}%
\end{pgfscope}%
\begin{pgfscope}%
\pgfsys@transformshift{3.060145in}{2.235006in}%
\pgfsys@useobject{currentmarker}{}%
\end{pgfscope}%
\begin{pgfscope}%
\pgfsys@transformshift{3.154867in}{2.293073in}%
\pgfsys@useobject{currentmarker}{}%
\end{pgfscope}%
\begin{pgfscope}%
\pgfsys@transformshift{2.485143in}{1.820578in}%
\pgfsys@useobject{currentmarker}{}%
\end{pgfscope}%
\begin{pgfscope}%
\pgfsys@transformshift{3.092535in}{2.435494in}%
\pgfsys@useobject{currentmarker}{}%
\end{pgfscope}%
\begin{pgfscope}%
\pgfsys@transformshift{2.972908in}{2.273399in}%
\pgfsys@useobject{currentmarker}{}%
\end{pgfscope}%
\begin{pgfscope}%
\pgfsys@transformshift{2.785913in}{2.054595in}%
\pgfsys@useobject{currentmarker}{}%
\end{pgfscope}%
\begin{pgfscope}%
\pgfsys@transformshift{1.884420in}{1.405391in}%
\pgfsys@useobject{currentmarker}{}%
\end{pgfscope}%
\begin{pgfscope}%
\pgfsys@transformshift{2.969914in}{2.237620in}%
\pgfsys@useobject{currentmarker}{}%
\end{pgfscope}%
\begin{pgfscope}%
\pgfsys@transformshift{3.239382in}{2.532586in}%
\pgfsys@useobject{currentmarker}{}%
\end{pgfscope}%
\begin{pgfscope}%
\pgfsys@transformshift{3.092535in}{2.158272in}%
\pgfsys@useobject{currentmarker}{}%
\end{pgfscope}%
\begin{pgfscope}%
\pgfsys@transformshift{2.716777in}{2.113218in}%
\pgfsys@useobject{currentmarker}{}%
\end{pgfscope}%
\begin{pgfscope}%
\pgfsys@transformshift{2.890843in}{2.176964in}%
\pgfsys@useobject{currentmarker}{}%
\end{pgfscope}%
\begin{pgfscope}%
\pgfsys@transformshift{2.740321in}{2.090633in}%
\pgfsys@useobject{currentmarker}{}%
\end{pgfscope}%
\begin{pgfscope}%
\pgfsys@transformshift{3.022174in}{2.346313in}%
\pgfsys@useobject{currentmarker}{}%
\end{pgfscope}%
\begin{pgfscope}%
\pgfsys@transformshift{2.804286in}{2.168797in}%
\pgfsys@useobject{currentmarker}{}%
\end{pgfscope}%
\begin{pgfscope}%
\pgfsys@transformshift{2.654854in}{2.027216in}%
\pgfsys@useobject{currentmarker}{}%
\end{pgfscope}%
\begin{pgfscope}%
\pgfsys@transformshift{3.188346in}{2.532586in}%
\pgfsys@useobject{currentmarker}{}%
\end{pgfscope}%
\begin{pgfscope}%
\pgfsys@transformshift{1.778266in}{1.334562in}%
\pgfsys@useobject{currentmarker}{}%
\end{pgfscope}%
\begin{pgfscope}%
\pgfsys@transformshift{3.222098in}{2.532586in}%
\pgfsys@useobject{currentmarker}{}%
\end{pgfscope}%
\begin{pgfscope}%
\pgfsys@transformshift{2.708611in}{2.105723in}%
\pgfsys@useobject{currentmarker}{}%
\end{pgfscope}%
\begin{pgfscope}%
\pgfsys@transformshift{2.291752in}{1.729023in}%
\pgfsys@useobject{currentmarker}{}%
\end{pgfscope}%
\begin{pgfscope}%
\pgfsys@transformshift{2.472759in}{1.851882in}%
\pgfsys@useobject{currentmarker}{}%
\end{pgfscope}%
\begin{pgfscope}%
\pgfsys@transformshift{3.022174in}{2.309945in}%
\pgfsys@useobject{currentmarker}{}%
\end{pgfscope}%
\begin{pgfscope}%
\pgfsys@transformshift{3.239382in}{2.532586in}%
\pgfsys@useobject{currentmarker}{}%
\end{pgfscope}%
\begin{pgfscope}%
\pgfsys@transformshift{2.077402in}{1.486586in}%
\pgfsys@useobject{currentmarker}{}%
\end{pgfscope}%
\begin{pgfscope}%
\pgfsys@transformshift{3.239382in}{2.532586in}%
\pgfsys@useobject{currentmarker}{}%
\end{pgfscope}%
\begin{pgfscope}%
\pgfsys@transformshift{3.124109in}{2.286410in}%
\pgfsys@useobject{currentmarker}{}%
\end{pgfscope}%
\begin{pgfscope}%
\pgfsys@transformshift{3.239382in}{2.532586in}%
\pgfsys@useobject{currentmarker}{}%
\end{pgfscope}%
\begin{pgfscope}%
\pgfsys@transformshift{2.131024in}{1.506625in}%
\pgfsys@useobject{currentmarker}{}%
\end{pgfscope}%
\begin{pgfscope}%
\pgfsys@transformshift{3.154595in}{2.532586in}%
\pgfsys@useobject{currentmarker}{}%
\end{pgfscope}%
\begin{pgfscope}%
\pgfsys@transformshift{3.005162in}{2.237512in}%
\pgfsys@useobject{currentmarker}{}%
\end{pgfscope}%
\begin{pgfscope}%
\pgfsys@transformshift{2.686836in}{2.081091in}%
\pgfsys@useobject{currentmarker}{}%
\end{pgfscope}%
\begin{pgfscope}%
\pgfsys@transformshift{3.049938in}{2.283149in}%
\pgfsys@useobject{currentmarker}{}%
\end{pgfscope}%
\begin{pgfscope}%
\pgfsys@transformshift{3.171879in}{2.434665in}%
\pgfsys@useobject{currentmarker}{}%
\end{pgfscope}%
\begin{pgfscope}%
\pgfsys@transformshift{3.222098in}{2.532586in}%
\pgfsys@useobject{currentmarker}{}%
\end{pgfscope}%
\begin{pgfscope}%
\pgfsys@transformshift{2.744949in}{2.127148in}%
\pgfsys@useobject{currentmarker}{}%
\end{pgfscope}%
\begin{pgfscope}%
\pgfsys@transformshift{2.851103in}{2.162888in}%
\pgfsys@useobject{currentmarker}{}%
\end{pgfscope}%
\begin{pgfscope}%
\pgfsys@transformshift{3.188346in}{2.532586in}%
\pgfsys@useobject{currentmarker}{}%
\end{pgfscope}%
\begin{pgfscope}%
\pgfsys@transformshift{2.810819in}{2.099725in}%
\pgfsys@useobject{currentmarker}{}%
\end{pgfscope}%
\begin{pgfscope}%
\pgfsys@transformshift{3.239382in}{2.532586in}%
\pgfsys@useobject{currentmarker}{}%
\end{pgfscope}%
\begin{pgfscope}%
\pgfsys@transformshift{2.352859in}{1.780296in}%
\pgfsys@useobject{currentmarker}{}%
\end{pgfscope}%
\begin{pgfscope}%
\pgfsys@transformshift{3.239382in}{2.532586in}%
\pgfsys@useobject{currentmarker}{}%
\end{pgfscope}%
\begin{pgfscope}%
\pgfsys@transformshift{3.205086in}{2.532586in}%
\pgfsys@useobject{currentmarker}{}%
\end{pgfscope}%
\begin{pgfscope}%
\pgfsys@transformshift{3.188346in}{2.532585in}%
\pgfsys@useobject{currentmarker}{}%
\end{pgfscope}%
\begin{pgfscope}%
\pgfsys@transformshift{3.239382in}{2.532586in}%
\pgfsys@useobject{currentmarker}{}%
\end{pgfscope}%
\begin{pgfscope}%
\pgfsys@transformshift{3.107778in}{2.435357in}%
\pgfsys@useobject{currentmarker}{}%
\end{pgfscope}%
\begin{pgfscope}%
\pgfsys@transformshift{3.239382in}{2.532586in}%
\pgfsys@useobject{currentmarker}{}%
\end{pgfscope}%
\begin{pgfscope}%
\pgfsys@transformshift{2.786594in}{2.097914in}%
\pgfsys@useobject{currentmarker}{}%
\end{pgfscope}%
\begin{pgfscope}%
\pgfsys@transformshift{3.239382in}{2.532586in}%
\pgfsys@useobject{currentmarker}{}%
\end{pgfscope}%
\begin{pgfscope}%
\pgfsys@transformshift{2.677445in}{2.078378in}%
\pgfsys@useobject{currentmarker}{}%
\end{pgfscope}%
\begin{pgfscope}%
\pgfsys@transformshift{3.239382in}{2.532587in}%
\pgfsys@useobject{currentmarker}{}%
\end{pgfscope}%
\begin{pgfscope}%
\pgfsys@transformshift{2.581498in}{1.978656in}%
\pgfsys@useobject{currentmarker}{}%
\end{pgfscope}%
\begin{pgfscope}%
\pgfsys@transformshift{2.291072in}{1.751739in}%
\pgfsys@useobject{currentmarker}{}%
\end{pgfscope}%
\begin{pgfscope}%
\pgfsys@transformshift{2.300054in}{1.719699in}%
\pgfsys@useobject{currentmarker}{}%
\end{pgfscope}%
\begin{pgfscope}%
\pgfsys@transformshift{2.957529in}{2.232637in}%
\pgfsys@useobject{currentmarker}{}%
\end{pgfscope}%
\begin{pgfscope}%
\pgfsys@transformshift{2.432610in}{1.822154in}%
\pgfsys@useobject{currentmarker}{}%
\end{pgfscope}%
\begin{pgfscope}%
\pgfsys@transformshift{3.138263in}{2.532586in}%
\pgfsys@useobject{currentmarker}{}%
\end{pgfscope}%
\begin{pgfscope}%
\pgfsys@transformshift{2.691327in}{2.094122in}%
\pgfsys@useobject{currentmarker}{}%
\end{pgfscope}%
\begin{pgfscope}%
\pgfsys@transformshift{2.606404in}{2.007118in}%
\pgfsys@useobject{currentmarker}{}%
\end{pgfscope}%
\begin{pgfscope}%
\pgfsys@transformshift{2.734605in}{2.031742in}%
\pgfsys@useobject{currentmarker}{}%
\end{pgfscope}%
\begin{pgfscope}%
\pgfsys@transformshift{2.568842in}{2.022670in}%
\pgfsys@useobject{currentmarker}{}%
\end{pgfscope}%
\begin{pgfscope}%
\pgfsys@transformshift{2.931399in}{2.169502in}%
\pgfsys@useobject{currentmarker}{}%
\end{pgfscope}%
\begin{pgfscope}%
\pgfsys@transformshift{2.563126in}{1.914177in}%
\pgfsys@useobject{currentmarker}{}%
\end{pgfscope}%
\begin{pgfscope}%
\pgfsys@transformshift{2.595380in}{1.989050in}%
\pgfsys@useobject{currentmarker}{}%
\end{pgfscope}%
\begin{pgfscope}%
\pgfsys@transformshift{2.926227in}{2.172937in}%
\pgfsys@useobject{currentmarker}{}%
\end{pgfscope}%
\begin{pgfscope}%
\pgfsys@transformshift{3.239382in}{2.532586in}%
\pgfsys@useobject{currentmarker}{}%
\end{pgfscope}%
\begin{pgfscope}%
\pgfsys@transformshift{3.089677in}{2.532586in}%
\pgfsys@useobject{currentmarker}{}%
\end{pgfscope}%
\begin{pgfscope}%
\pgfsys@transformshift{2.971411in}{2.330712in}%
\pgfsys@useobject{currentmarker}{}%
\end{pgfscope}%
\begin{pgfscope}%
\pgfsys@transformshift{2.522297in}{1.917112in}%
\pgfsys@useobject{currentmarker}{}%
\end{pgfscope}%
\begin{pgfscope}%
\pgfsys@transformshift{3.222098in}{2.532586in}%
\pgfsys@useobject{currentmarker}{}%
\end{pgfscope}%
\begin{pgfscope}%
\pgfsys@transformshift{3.239382in}{2.532586in}%
\pgfsys@useobject{currentmarker}{}%
\end{pgfscope}%
\begin{pgfscope}%
\pgfsys@transformshift{3.239382in}{2.532586in}%
\pgfsys@useobject{currentmarker}{}%
\end{pgfscope}%
\begin{pgfscope}%
\pgfsys@transformshift{3.239382in}{2.532586in}%
\pgfsys@useobject{currentmarker}{}%
\end{pgfscope}%
\begin{pgfscope}%
\pgfsys@transformshift{3.090358in}{2.532586in}%
\pgfsys@useobject{currentmarker}{}%
\end{pgfscope}%
\begin{pgfscope}%
\pgfsys@transformshift{2.229693in}{1.559087in}%
\pgfsys@useobject{currentmarker}{}%
\end{pgfscope}%
\begin{pgfscope}%
\pgfsys@transformshift{2.779925in}{2.247023in}%
\pgfsys@useobject{currentmarker}{}%
\end{pgfscope}%
\begin{pgfscope}%
\pgfsys@transformshift{1.989349in}{1.492399in}%
\pgfsys@useobject{currentmarker}{}%
\end{pgfscope}%
\begin{pgfscope}%
\pgfsys@transformshift{2.783872in}{2.154247in}%
\pgfsys@useobject{currentmarker}{}%
\end{pgfscope}%
\begin{pgfscope}%
\pgfsys@transformshift{3.239382in}{2.532586in}%
\pgfsys@useobject{currentmarker}{}%
\end{pgfscope}%
\begin{pgfscope}%
\pgfsys@transformshift{3.239382in}{2.532586in}%
\pgfsys@useobject{currentmarker}{}%
\end{pgfscope}%
\begin{pgfscope}%
\pgfsys@transformshift{3.239382in}{2.532586in}%
\pgfsys@useobject{currentmarker}{}%
\end{pgfscope}%
\begin{pgfscope}%
\pgfsys@transformshift{1.765473in}{1.291997in}%
\pgfsys@useobject{currentmarker}{}%
\end{pgfscope}%
\begin{pgfscope}%
\pgfsys@transformshift{3.222098in}{2.532586in}%
\pgfsys@useobject{currentmarker}{}%
\end{pgfscope}%
\begin{pgfscope}%
\pgfsys@transformshift{3.239382in}{2.532586in}%
\pgfsys@useobject{currentmarker}{}%
\end{pgfscope}%
\begin{pgfscope}%
\pgfsys@transformshift{2.534546in}{1.928972in}%
\pgfsys@useobject{currentmarker}{}%
\end{pgfscope}%
\begin{pgfscope}%
\pgfsys@transformshift{3.239382in}{2.532586in}%
\pgfsys@useobject{currentmarker}{}%
\end{pgfscope}%
\begin{pgfscope}%
\pgfsys@transformshift{2.733244in}{2.132296in}%
\pgfsys@useobject{currentmarker}{}%
\end{pgfscope}%
\begin{pgfscope}%
\pgfsys@transformshift{1.267909in}{0.882986in}%
\pgfsys@useobject{currentmarker}{}%
\end{pgfscope}%
\begin{pgfscope}%
\pgfsys@transformshift{2.414510in}{1.823934in}%
\pgfsys@useobject{currentmarker}{}%
\end{pgfscope}%
\begin{pgfscope}%
\pgfsys@transformshift{3.139760in}{2.333649in}%
\pgfsys@useobject{currentmarker}{}%
\end{pgfscope}%
\begin{pgfscope}%
\pgfsys@transformshift{3.222098in}{2.532586in}%
\pgfsys@useobject{currentmarker}{}%
\end{pgfscope}%
\begin{pgfscope}%
\pgfsys@transformshift{2.509776in}{1.919437in}%
\pgfsys@useobject{currentmarker}{}%
\end{pgfscope}%
\begin{pgfscope}%
\pgfsys@transformshift{3.188074in}{2.532586in}%
\pgfsys@useobject{currentmarker}{}%
\end{pgfscope}%
\begin{pgfscope}%
\pgfsys@transformshift{2.970322in}{2.336358in}%
\pgfsys@useobject{currentmarker}{}%
\end{pgfscope}%
\begin{pgfscope}%
\pgfsys@transformshift{3.239382in}{2.532586in}%
\pgfsys@useobject{currentmarker}{}%
\end{pgfscope}%
\begin{pgfscope}%
\pgfsys@transformshift{3.239382in}{2.532586in}%
\pgfsys@useobject{currentmarker}{}%
\end{pgfscope}%
\begin{pgfscope}%
\pgfsys@transformshift{2.695546in}{2.083514in}%
\pgfsys@useobject{currentmarker}{}%
\end{pgfscope}%
\begin{pgfscope}%
\pgfsys@transformshift{3.188346in}{2.419576in}%
\pgfsys@useobject{currentmarker}{}%
\end{pgfscope}%
\begin{pgfscope}%
\pgfsys@transformshift{3.239382in}{2.532586in}%
\pgfsys@useobject{currentmarker}{}%
\end{pgfscope}%
\begin{pgfscope}%
\pgfsys@transformshift{2.556049in}{1.959041in}%
\pgfsys@useobject{currentmarker}{}%
\end{pgfscope}%
\begin{pgfscope}%
\pgfsys@transformshift{2.538356in}{1.909771in}%
\pgfsys@useobject{currentmarker}{}%
\end{pgfscope}%
\begin{pgfscope}%
\pgfsys@transformshift{3.122476in}{2.320791in}%
\pgfsys@useobject{currentmarker}{}%
\end{pgfscope}%
\begin{pgfscope}%
\pgfsys@transformshift{3.091719in}{2.308968in}%
\pgfsys@useobject{currentmarker}{}%
\end{pgfscope}%
\begin{pgfscope}%
\pgfsys@transformshift{3.075251in}{2.299318in}%
\pgfsys@useobject{currentmarker}{}%
\end{pgfscope}%
\begin{pgfscope}%
\pgfsys@transformshift{2.522705in}{1.847588in}%
\pgfsys@useobject{currentmarker}{}%
\end{pgfscope}%
\begin{pgfscope}%
\pgfsys@transformshift{3.188074in}{2.532586in}%
\pgfsys@useobject{currentmarker}{}%
\end{pgfscope}%
\begin{pgfscope}%
\pgfsys@transformshift{2.832322in}{2.102226in}%
\pgfsys@useobject{currentmarker}{}%
\end{pgfscope}%
\begin{pgfscope}%
\pgfsys@transformshift{2.133882in}{1.522486in}%
\pgfsys@useobject{currentmarker}{}%
\end{pgfscope}%
\begin{pgfscope}%
\pgfsys@transformshift{3.239382in}{2.532586in}%
\pgfsys@useobject{currentmarker}{}%
\end{pgfscope}%
\begin{pgfscope}%
\pgfsys@transformshift{3.239382in}{2.532586in}%
\pgfsys@useobject{currentmarker}{}%
\end{pgfscope}%
\begin{pgfscope}%
\pgfsys@transformshift{2.731067in}{2.100512in}%
\pgfsys@useobject{currentmarker}{}%
\end{pgfscope}%
\begin{pgfscope}%
\pgfsys@transformshift{2.920511in}{2.031495in}%
\pgfsys@useobject{currentmarker}{}%
\end{pgfscope}%
\begin{pgfscope}%
\pgfsys@transformshift{2.594564in}{1.981369in}%
\pgfsys@useobject{currentmarker}{}%
\end{pgfscope}%
\begin{pgfscope}%
\pgfsys@transformshift{3.239382in}{2.532586in}%
\pgfsys@useobject{currentmarker}{}%
\end{pgfscope}%
\begin{pgfscope}%
\pgfsys@transformshift{3.123293in}{2.259402in}%
\pgfsys@useobject{currentmarker}{}%
\end{pgfscope}%
\begin{pgfscope}%
\pgfsys@transformshift{3.138263in}{2.532586in}%
\pgfsys@useobject{currentmarker}{}%
\end{pgfscope}%
\begin{pgfscope}%
\pgfsys@transformshift{2.195805in}{1.581029in}%
\pgfsys@useobject{currentmarker}{}%
\end{pgfscope}%
\begin{pgfscope}%
\pgfsys@transformshift{3.188346in}{2.532586in}%
\pgfsys@useobject{currentmarker}{}%
\end{pgfscope}%
\begin{pgfscope}%
\pgfsys@transformshift{3.058512in}{2.532586in}%
\pgfsys@useobject{currentmarker}{}%
\end{pgfscope}%
\begin{pgfscope}%
\pgfsys@transformshift{3.239382in}{2.532586in}%
\pgfsys@useobject{currentmarker}{}%
\end{pgfscope}%
\begin{pgfscope}%
\pgfsys@transformshift{2.620558in}{1.962175in}%
\pgfsys@useobject{currentmarker}{}%
\end{pgfscope}%
\begin{pgfscope}%
\pgfsys@transformshift{3.205086in}{2.532586in}%
\pgfsys@useobject{currentmarker}{}%
\end{pgfscope}%
\begin{pgfscope}%
\pgfsys@transformshift{3.205086in}{2.532586in}%
\pgfsys@useobject{currentmarker}{}%
\end{pgfscope}%
\begin{pgfscope}%
\pgfsys@transformshift{2.235817in}{1.587149in}%
\pgfsys@useobject{currentmarker}{}%
\end{pgfscope}%
\begin{pgfscope}%
\pgfsys@transformshift{2.781150in}{2.154089in}%
\pgfsys@useobject{currentmarker}{}%
\end{pgfscope}%
\begin{pgfscope}%
\pgfsys@transformshift{3.222098in}{2.407314in}%
\pgfsys@useobject{currentmarker}{}%
\end{pgfscope}%
\begin{pgfscope}%
\pgfsys@transformshift{2.890026in}{2.193995in}%
\pgfsys@useobject{currentmarker}{}%
\end{pgfscope}%
\begin{pgfscope}%
\pgfsys@transformshift{2.672274in}{2.054280in}%
\pgfsys@useobject{currentmarker}{}%
\end{pgfscope}%
\begin{pgfscope}%
\pgfsys@transformshift{2.653084in}{1.938305in}%
\pgfsys@useobject{currentmarker}{}%
\end{pgfscope}%
\begin{pgfscope}%
\pgfsys@transformshift{3.014417in}{2.343955in}%
\pgfsys@useobject{currentmarker}{}%
\end{pgfscope}%
\begin{pgfscope}%
\pgfsys@transformshift{1.890136in}{1.377971in}%
\pgfsys@useobject{currentmarker}{}%
\end{pgfscope}%
\begin{pgfscope}%
\pgfsys@transformshift{3.239382in}{2.532586in}%
\pgfsys@useobject{currentmarker}{}%
\end{pgfscope}%
\begin{pgfscope}%
\pgfsys@transformshift{2.430841in}{1.867839in}%
\pgfsys@useobject{currentmarker}{}%
\end{pgfscope}%
\begin{pgfscope}%
\pgfsys@transformshift{3.222098in}{2.406178in}%
\pgfsys@useobject{currentmarker}{}%
\end{pgfscope}%
\begin{pgfscope}%
\pgfsys@transformshift{3.239382in}{2.437024in}%
\pgfsys@useobject{currentmarker}{}%
\end{pgfscope}%
\begin{pgfscope}%
\pgfsys@transformshift{2.606132in}{1.841944in}%
\pgfsys@useobject{currentmarker}{}%
\end{pgfscope}%
\begin{pgfscope}%
\pgfsys@transformshift{2.695546in}{2.094231in}%
\pgfsys@useobject{currentmarker}{}%
\end{pgfscope}%
\begin{pgfscope}%
\pgfsys@transformshift{2.489770in}{1.863480in}%
\pgfsys@useobject{currentmarker}{}%
\end{pgfscope}%
\begin{pgfscope}%
\pgfsys@transformshift{3.222098in}{2.532586in}%
\pgfsys@useobject{currentmarker}{}%
\end{pgfscope}%
\begin{pgfscope}%
\pgfsys@transformshift{3.239382in}{2.532586in}%
\pgfsys@useobject{currentmarker}{}%
\end{pgfscope}%
\begin{pgfscope}%
\pgfsys@transformshift{2.164503in}{1.658953in}%
\pgfsys@useobject{currentmarker}{}%
\end{pgfscope}%
\begin{pgfscope}%
\pgfsys@transformshift{3.205086in}{2.532586in}%
\pgfsys@useobject{currentmarker}{}%
\end{pgfscope}%
\begin{pgfscope}%
\pgfsys@transformshift{3.239382in}{2.532586in}%
\pgfsys@useobject{currentmarker}{}%
\end{pgfscope}%
\begin{pgfscope}%
\pgfsys@transformshift{2.540534in}{1.926838in}%
\pgfsys@useobject{currentmarker}{}%
\end{pgfscope}%
\begin{pgfscope}%
\pgfsys@transformshift{2.325095in}{1.726322in}%
\pgfsys@useobject{currentmarker}{}%
\end{pgfscope}%
\begin{pgfscope}%
\pgfsys@transformshift{2.867162in}{2.115179in}%
\pgfsys@useobject{currentmarker}{}%
\end{pgfscope}%
\begin{pgfscope}%
\pgfsys@transformshift{3.239382in}{2.532586in}%
\pgfsys@useobject{currentmarker}{}%
\end{pgfscope}%
\begin{pgfscope}%
\pgfsys@transformshift{3.222098in}{2.532586in}%
\pgfsys@useobject{currentmarker}{}%
\end{pgfscope}%
\begin{pgfscope}%
\pgfsys@transformshift{3.239382in}{2.532586in}%
\pgfsys@useobject{currentmarker}{}%
\end{pgfscope}%
\begin{pgfscope}%
\pgfsys@transformshift{2.565031in}{1.943898in}%
\pgfsys@useobject{currentmarker}{}%
\end{pgfscope}%
\begin{pgfscope}%
\pgfsys@transformshift{3.239382in}{2.532586in}%
\pgfsys@useobject{currentmarker}{}%
\end{pgfscope}%
\begin{pgfscope}%
\pgfsys@transformshift{3.239382in}{2.532586in}%
\pgfsys@useobject{currentmarker}{}%
\end{pgfscope}%
\begin{pgfscope}%
\pgfsys@transformshift{2.529782in}{1.916310in}%
\pgfsys@useobject{currentmarker}{}%
\end{pgfscope}%
\begin{pgfscope}%
\pgfsys@transformshift{3.239382in}{2.532586in}%
\pgfsys@useobject{currentmarker}{}%
\end{pgfscope}%
\begin{pgfscope}%
\pgfsys@transformshift{2.407025in}{1.793470in}%
\pgfsys@useobject{currentmarker}{}%
\end{pgfscope}%
\begin{pgfscope}%
\pgfsys@transformshift{2.609398in}{2.033048in}%
\pgfsys@useobject{currentmarker}{}%
\end{pgfscope}%
\begin{pgfscope}%
\pgfsys@transformshift{3.239382in}{2.532586in}%
\pgfsys@useobject{currentmarker}{}%
\end{pgfscope}%
\begin{pgfscope}%
\pgfsys@transformshift{2.995500in}{2.354901in}%
\pgfsys@useobject{currentmarker}{}%
\end{pgfscope}%
\begin{pgfscope}%
\pgfsys@transformshift{2.481469in}{1.811501in}%
\pgfsys@useobject{currentmarker}{}%
\end{pgfscope}%
\begin{pgfscope}%
\pgfsys@transformshift{3.222098in}{2.532586in}%
\pgfsys@useobject{currentmarker}{}%
\end{pgfscope}%
\begin{pgfscope}%
\pgfsys@transformshift{2.443634in}{1.751604in}%
\pgfsys@useobject{currentmarker}{}%
\end{pgfscope}%
\begin{pgfscope}%
\pgfsys@transformshift{3.137447in}{2.314889in}%
\pgfsys@useobject{currentmarker}{}%
\end{pgfscope}%
\begin{pgfscope}%
\pgfsys@transformshift{3.239382in}{2.532586in}%
\pgfsys@useobject{currentmarker}{}%
\end{pgfscope}%
\begin{pgfscope}%
\pgfsys@transformshift{3.222098in}{2.532586in}%
\pgfsys@useobject{currentmarker}{}%
\end{pgfscope}%
\begin{pgfscope}%
\pgfsys@transformshift{2.699493in}{2.081063in}%
\pgfsys@useobject{currentmarker}{}%
\end{pgfscope}%
\begin{pgfscope}%
\pgfsys@transformshift{3.188346in}{2.532586in}%
\pgfsys@useobject{currentmarker}{}%
\end{pgfscope}%
\begin{pgfscope}%
\pgfsys@transformshift{3.239382in}{2.532586in}%
\pgfsys@useobject{currentmarker}{}%
\end{pgfscope}%
\begin{pgfscope}%
\pgfsys@transformshift{2.494670in}{1.853872in}%
\pgfsys@useobject{currentmarker}{}%
\end{pgfscope}%
\begin{pgfscope}%
\pgfsys@transformshift{3.239382in}{2.532586in}%
\pgfsys@useobject{currentmarker}{}%
\end{pgfscope}%
\begin{pgfscope}%
\pgfsys@transformshift{3.123293in}{2.532586in}%
\pgfsys@useobject{currentmarker}{}%
\end{pgfscope}%
\begin{pgfscope}%
\pgfsys@transformshift{3.239382in}{2.532586in}%
\pgfsys@useobject{currentmarker}{}%
\end{pgfscope}%
\begin{pgfscope}%
\pgfsys@transformshift{2.639203in}{2.009107in}%
\pgfsys@useobject{currentmarker}{}%
\end{pgfscope}%
\begin{pgfscope}%
\pgfsys@transformshift{2.892748in}{2.230519in}%
\pgfsys@useobject{currentmarker}{}%
\end{pgfscope}%
\begin{pgfscope}%
\pgfsys@transformshift{3.154731in}{2.532586in}%
\pgfsys@useobject{currentmarker}{}%
\end{pgfscope}%
\begin{pgfscope}%
\pgfsys@transformshift{2.443362in}{1.882274in}%
\pgfsys@useobject{currentmarker}{}%
\end{pgfscope}%
\begin{pgfscope}%
\pgfsys@transformshift{2.950997in}{2.149000in}%
\pgfsys@useobject{currentmarker}{}%
\end{pgfscope}%
\begin{pgfscope}%
\pgfsys@transformshift{3.074026in}{2.327303in}%
\pgfsys@useobject{currentmarker}{}%
\end{pgfscope}%
\begin{pgfscope}%
\pgfsys@transformshift{3.205086in}{2.532586in}%
\pgfsys@useobject{currentmarker}{}%
\end{pgfscope}%
\begin{pgfscope}%
\pgfsys@transformshift{2.385794in}{1.793765in}%
\pgfsys@useobject{currentmarker}{}%
\end{pgfscope}%
\begin{pgfscope}%
\pgfsys@transformshift{2.418184in}{1.773981in}%
\pgfsys@useobject{currentmarker}{}%
\end{pgfscope}%
\end{pgfscope}%
\begin{pgfscope}%
\pgfsetbuttcap%
\pgfsetroundjoin%
\definecolor{currentfill}{rgb}{0.000000,0.000000,0.000000}%
\pgfsetfillcolor{currentfill}%
\pgfsetlinewidth{0.803000pt}%
\definecolor{currentstroke}{rgb}{0.000000,0.000000,0.000000}%
\pgfsetstrokecolor{currentstroke}%
\pgfsetdash{}{0pt}%
\pgfsys@defobject{currentmarker}{\pgfqpoint{0.000000in}{-0.048611in}}{\pgfqpoint{0.000000in}{0.000000in}}{%
\pgfpathmoveto{\pgfqpoint{0.000000in}{0.000000in}}%
\pgfpathlineto{\pgfqpoint{0.000000in}{-0.048611in}}%
\pgfusepath{stroke,fill}%
}%
\begin{pgfscope}%
\pgfsys@transformshift{1.027021in}{0.633903in}%
\pgfsys@useobject{currentmarker}{}%
\end{pgfscope}%
\end{pgfscope}%
\begin{pgfscope}%
\definecolor{textcolor}{rgb}{0.000000,0.000000,0.000000}%
\pgfsetstrokecolor{textcolor}%
\pgfsetfillcolor{textcolor}%
\pgftext[x=1.027021in,y=0.536681in,,top]{\color{textcolor}\rmfamily\fontsize{10.000000}{12.000000}\selectfont 0.00}%
\end{pgfscope}%
\begin{pgfscope}%
\pgfsetbuttcap%
\pgfsetroundjoin%
\definecolor{currentfill}{rgb}{0.000000,0.000000,0.000000}%
\pgfsetfillcolor{currentfill}%
\pgfsetlinewidth{0.803000pt}%
\definecolor{currentstroke}{rgb}{0.000000,0.000000,0.000000}%
\pgfsetstrokecolor{currentstroke}%
\pgfsetdash{}{0pt}%
\pgfsys@defobject{currentmarker}{\pgfqpoint{0.000000in}{-0.048611in}}{\pgfqpoint{0.000000in}{0.000000in}}{%
\pgfpathmoveto{\pgfqpoint{0.000000in}{0.000000in}}%
\pgfpathlineto{\pgfqpoint{0.000000in}{-0.048611in}}%
\pgfusepath{stroke,fill}%
}%
\begin{pgfscope}%
\pgfsys@transformshift{1.580111in}{0.633903in}%
\pgfsys@useobject{currentmarker}{}%
\end{pgfscope}%
\end{pgfscope}%
\begin{pgfscope}%
\definecolor{textcolor}{rgb}{0.000000,0.000000,0.000000}%
\pgfsetstrokecolor{textcolor}%
\pgfsetfillcolor{textcolor}%
\pgftext[x=1.580111in,y=0.536681in,,top]{\color{textcolor}\rmfamily\fontsize{10.000000}{12.000000}\selectfont 0.25}%
\end{pgfscope}%
\begin{pgfscope}%
\pgfsetbuttcap%
\pgfsetroundjoin%
\definecolor{currentfill}{rgb}{0.000000,0.000000,0.000000}%
\pgfsetfillcolor{currentfill}%
\pgfsetlinewidth{0.803000pt}%
\definecolor{currentstroke}{rgb}{0.000000,0.000000,0.000000}%
\pgfsetstrokecolor{currentstroke}%
\pgfsetdash{}{0pt}%
\pgfsys@defobject{currentmarker}{\pgfqpoint{0.000000in}{-0.048611in}}{\pgfqpoint{0.000000in}{0.000000in}}{%
\pgfpathmoveto{\pgfqpoint{0.000000in}{0.000000in}}%
\pgfpathlineto{\pgfqpoint{0.000000in}{-0.048611in}}%
\pgfusepath{stroke,fill}%
}%
\begin{pgfscope}%
\pgfsys@transformshift{2.133201in}{0.633903in}%
\pgfsys@useobject{currentmarker}{}%
\end{pgfscope}%
\end{pgfscope}%
\begin{pgfscope}%
\definecolor{textcolor}{rgb}{0.000000,0.000000,0.000000}%
\pgfsetstrokecolor{textcolor}%
\pgfsetfillcolor{textcolor}%
\pgftext[x=2.133201in,y=0.536681in,,top]{\color{textcolor}\rmfamily\fontsize{10.000000}{12.000000}\selectfont 0.50}%
\end{pgfscope}%
\begin{pgfscope}%
\pgfsetbuttcap%
\pgfsetroundjoin%
\definecolor{currentfill}{rgb}{0.000000,0.000000,0.000000}%
\pgfsetfillcolor{currentfill}%
\pgfsetlinewidth{0.803000pt}%
\definecolor{currentstroke}{rgb}{0.000000,0.000000,0.000000}%
\pgfsetstrokecolor{currentstroke}%
\pgfsetdash{}{0pt}%
\pgfsys@defobject{currentmarker}{\pgfqpoint{0.000000in}{-0.048611in}}{\pgfqpoint{0.000000in}{0.000000in}}{%
\pgfpathmoveto{\pgfqpoint{0.000000in}{0.000000in}}%
\pgfpathlineto{\pgfqpoint{0.000000in}{-0.048611in}}%
\pgfusepath{stroke,fill}%
}%
\begin{pgfscope}%
\pgfsys@transformshift{2.686292in}{0.633903in}%
\pgfsys@useobject{currentmarker}{}%
\end{pgfscope}%
\end{pgfscope}%
\begin{pgfscope}%
\definecolor{textcolor}{rgb}{0.000000,0.000000,0.000000}%
\pgfsetstrokecolor{textcolor}%
\pgfsetfillcolor{textcolor}%
\pgftext[x=2.686292in,y=0.536681in,,top]{\color{textcolor}\rmfamily\fontsize{10.000000}{12.000000}\selectfont 0.75}%
\end{pgfscope}%
\begin{pgfscope}%
\pgfsetbuttcap%
\pgfsetroundjoin%
\definecolor{currentfill}{rgb}{0.000000,0.000000,0.000000}%
\pgfsetfillcolor{currentfill}%
\pgfsetlinewidth{0.803000pt}%
\definecolor{currentstroke}{rgb}{0.000000,0.000000,0.000000}%
\pgfsetstrokecolor{currentstroke}%
\pgfsetdash{}{0pt}%
\pgfsys@defobject{currentmarker}{\pgfqpoint{0.000000in}{-0.048611in}}{\pgfqpoint{0.000000in}{0.000000in}}{%
\pgfpathmoveto{\pgfqpoint{0.000000in}{0.000000in}}%
\pgfpathlineto{\pgfqpoint{0.000000in}{-0.048611in}}%
\pgfusepath{stroke,fill}%
}%
\begin{pgfscope}%
\pgfsys@transformshift{3.239382in}{0.633903in}%
\pgfsys@useobject{currentmarker}{}%
\end{pgfscope}%
\end{pgfscope}%
\begin{pgfscope}%
\definecolor{textcolor}{rgb}{0.000000,0.000000,0.000000}%
\pgfsetstrokecolor{textcolor}%
\pgfsetfillcolor{textcolor}%
\pgftext[x=3.239382in,y=0.536681in,,top]{\color{textcolor}\rmfamily\fontsize{10.000000}{12.000000}\selectfont 1.00}%
\end{pgfscope}%
\begin{pgfscope}%
\definecolor{textcolor}{rgb}{0.000000,0.000000,0.000000}%
\pgfsetstrokecolor{textcolor}%
\pgfsetfillcolor{textcolor}%
\pgftext[x=2.133201in,y=0.358470in,,top]{\color{textcolor}\rmfamily\fontsize{10.000000}{12.000000}\selectfont Prediction of \(\displaystyle p_{@2}(w_t)\)}%
\end{pgfscope}%
\begin{pgfscope}%
\pgfsetbuttcap%
\pgfsetroundjoin%
\definecolor{currentfill}{rgb}{0.000000,0.000000,0.000000}%
\pgfsetfillcolor{currentfill}%
\pgfsetlinewidth{0.803000pt}%
\definecolor{currentstroke}{rgb}{0.000000,0.000000,0.000000}%
\pgfsetstrokecolor{currentstroke}%
\pgfsetdash{}{0pt}%
\pgfsys@defobject{currentmarker}{\pgfqpoint{-0.048611in}{0.000000in}}{\pgfqpoint{0.000000in}{0.000000in}}{%
\pgfpathmoveto{\pgfqpoint{0.000000in}{0.000000in}}%
\pgfpathlineto{\pgfqpoint{-0.048611in}{0.000000in}}%
\pgfusepath{stroke,fill}%
}%
\begin{pgfscope}%
\pgfsys@transformshift{0.916403in}{0.724316in}%
\pgfsys@useobject{currentmarker}{}%
\end{pgfscope}%
\end{pgfscope}%
\begin{pgfscope}%
\definecolor{textcolor}{rgb}{0.000000,0.000000,0.000000}%
\pgfsetstrokecolor{textcolor}%
\pgfsetfillcolor{textcolor}%
\pgftext[x=0.641755in,y=0.676489in,left,base]{\color{textcolor}\rmfamily\fontsize{10.000000}{12.000000}\selectfont 0.0}%
\end{pgfscope}%
\begin{pgfscope}%
\pgfsetbuttcap%
\pgfsetroundjoin%
\definecolor{currentfill}{rgb}{0.000000,0.000000,0.000000}%
\pgfsetfillcolor{currentfill}%
\pgfsetlinewidth{0.803000pt}%
\definecolor{currentstroke}{rgb}{0.000000,0.000000,0.000000}%
\pgfsetstrokecolor{currentstroke}%
\pgfsetdash{}{0pt}%
\pgfsys@defobject{currentmarker}{\pgfqpoint{-0.048611in}{0.000000in}}{\pgfqpoint{0.000000in}{0.000000in}}{%
\pgfpathmoveto{\pgfqpoint{0.000000in}{0.000000in}}%
\pgfpathlineto{\pgfqpoint{-0.048611in}{0.000000in}}%
\pgfusepath{stroke,fill}%
}%
\begin{pgfscope}%
\pgfsys@transformshift{0.916403in}{1.085970in}%
\pgfsys@useobject{currentmarker}{}%
\end{pgfscope}%
\end{pgfscope}%
\begin{pgfscope}%
\definecolor{textcolor}{rgb}{0.000000,0.000000,0.000000}%
\pgfsetstrokecolor{textcolor}%
\pgfsetfillcolor{textcolor}%
\pgftext[x=0.641755in,y=1.038143in,left,base]{\color{textcolor}\rmfamily\fontsize{10.000000}{12.000000}\selectfont 0.2}%
\end{pgfscope}%
\begin{pgfscope}%
\pgfsetbuttcap%
\pgfsetroundjoin%
\definecolor{currentfill}{rgb}{0.000000,0.000000,0.000000}%
\pgfsetfillcolor{currentfill}%
\pgfsetlinewidth{0.803000pt}%
\definecolor{currentstroke}{rgb}{0.000000,0.000000,0.000000}%
\pgfsetstrokecolor{currentstroke}%
\pgfsetdash{}{0pt}%
\pgfsys@defobject{currentmarker}{\pgfqpoint{-0.048611in}{0.000000in}}{\pgfqpoint{0.000000in}{0.000000in}}{%
\pgfpathmoveto{\pgfqpoint{0.000000in}{0.000000in}}%
\pgfpathlineto{\pgfqpoint{-0.048611in}{0.000000in}}%
\pgfusepath{stroke,fill}%
}%
\begin{pgfscope}%
\pgfsys@transformshift{0.916403in}{1.447624in}%
\pgfsys@useobject{currentmarker}{}%
\end{pgfscope}%
\end{pgfscope}%
\begin{pgfscope}%
\definecolor{textcolor}{rgb}{0.000000,0.000000,0.000000}%
\pgfsetstrokecolor{textcolor}%
\pgfsetfillcolor{textcolor}%
\pgftext[x=0.641755in,y=1.399797in,left,base]{\color{textcolor}\rmfamily\fontsize{10.000000}{12.000000}\selectfont 0.4}%
\end{pgfscope}%
\begin{pgfscope}%
\pgfsetbuttcap%
\pgfsetroundjoin%
\definecolor{currentfill}{rgb}{0.000000,0.000000,0.000000}%
\pgfsetfillcolor{currentfill}%
\pgfsetlinewidth{0.803000pt}%
\definecolor{currentstroke}{rgb}{0.000000,0.000000,0.000000}%
\pgfsetstrokecolor{currentstroke}%
\pgfsetdash{}{0pt}%
\pgfsys@defobject{currentmarker}{\pgfqpoint{-0.048611in}{0.000000in}}{\pgfqpoint{0.000000in}{0.000000in}}{%
\pgfpathmoveto{\pgfqpoint{0.000000in}{0.000000in}}%
\pgfpathlineto{\pgfqpoint{-0.048611in}{0.000000in}}%
\pgfusepath{stroke,fill}%
}%
\begin{pgfscope}%
\pgfsys@transformshift{0.916403in}{1.809278in}%
\pgfsys@useobject{currentmarker}{}%
\end{pgfscope}%
\end{pgfscope}%
\begin{pgfscope}%
\definecolor{textcolor}{rgb}{0.000000,0.000000,0.000000}%
\pgfsetstrokecolor{textcolor}%
\pgfsetfillcolor{textcolor}%
\pgftext[x=0.641755in,y=1.761451in,left,base]{\color{textcolor}\rmfamily\fontsize{10.000000}{12.000000}\selectfont 0.6}%
\end{pgfscope}%
\begin{pgfscope}%
\pgfsetbuttcap%
\pgfsetroundjoin%
\definecolor{currentfill}{rgb}{0.000000,0.000000,0.000000}%
\pgfsetfillcolor{currentfill}%
\pgfsetlinewidth{0.803000pt}%
\definecolor{currentstroke}{rgb}{0.000000,0.000000,0.000000}%
\pgfsetstrokecolor{currentstroke}%
\pgfsetdash{}{0pt}%
\pgfsys@defobject{currentmarker}{\pgfqpoint{-0.048611in}{0.000000in}}{\pgfqpoint{0.000000in}{0.000000in}}{%
\pgfpathmoveto{\pgfqpoint{0.000000in}{0.000000in}}%
\pgfpathlineto{\pgfqpoint{-0.048611in}{0.000000in}}%
\pgfusepath{stroke,fill}%
}%
\begin{pgfscope}%
\pgfsys@transformshift{0.916403in}{2.170932in}%
\pgfsys@useobject{currentmarker}{}%
\end{pgfscope}%
\end{pgfscope}%
\begin{pgfscope}%
\definecolor{textcolor}{rgb}{0.000000,0.000000,0.000000}%
\pgfsetstrokecolor{textcolor}%
\pgfsetfillcolor{textcolor}%
\pgftext[x=0.641755in,y=2.123105in,left,base]{\color{textcolor}\rmfamily\fontsize{10.000000}{12.000000}\selectfont 0.8}%
\end{pgfscope}%
\begin{pgfscope}%
\pgfsetbuttcap%
\pgfsetroundjoin%
\definecolor{currentfill}{rgb}{0.000000,0.000000,0.000000}%
\pgfsetfillcolor{currentfill}%
\pgfsetlinewidth{0.803000pt}%
\definecolor{currentstroke}{rgb}{0.000000,0.000000,0.000000}%
\pgfsetstrokecolor{currentstroke}%
\pgfsetdash{}{0pt}%
\pgfsys@defobject{currentmarker}{\pgfqpoint{-0.048611in}{0.000000in}}{\pgfqpoint{0.000000in}{0.000000in}}{%
\pgfpathmoveto{\pgfqpoint{0.000000in}{0.000000in}}%
\pgfpathlineto{\pgfqpoint{-0.048611in}{0.000000in}}%
\pgfusepath{stroke,fill}%
}%
\begin{pgfscope}%
\pgfsys@transformshift{0.916403in}{2.532586in}%
\pgfsys@useobject{currentmarker}{}%
\end{pgfscope}%
\end{pgfscope}%
\begin{pgfscope}%
\definecolor{textcolor}{rgb}{0.000000,0.000000,0.000000}%
\pgfsetstrokecolor{textcolor}%
\pgfsetfillcolor{textcolor}%
\pgftext[x=0.641755in,y=2.484759in,left,base]{\color{textcolor}\rmfamily\fontsize{10.000000}{12.000000}\selectfont 1.0}%
\end{pgfscope}%
\begin{pgfscope}%
\definecolor{textcolor}{rgb}{0.000000,0.000000,0.000000}%
\pgfsetstrokecolor{textcolor}%
\pgfsetfillcolor{textcolor}%
\pgftext[x=0.586199in,y=1.628451in,,bottom,rotate=90.000000]{\color{textcolor}\rmfamily\fontsize{10.000000}{12.000000}\selectfont Observation of \(\displaystyle p_{@2}(w_t)\)}%
\end{pgfscope}%
\begin{pgfscope}%
\pgfpathrectangle{\pgfqpoint{0.916403in}{0.633903in}}{\pgfqpoint{2.433597in}{1.989097in}}%
\pgfusepath{clip}%
\pgfsetbuttcap%
\pgfsetroundjoin%
\pgfsetlinewidth{1.003750pt}%
\definecolor{currentstroke}{rgb}{1.000000,0.000000,0.000000}%
\pgfsetstrokecolor{currentstroke}%
\pgfsetdash{{3.700000pt}{1.600000pt}}{0.000000pt}%
\pgfpathmoveto{\pgfqpoint{0.906403in}{0.625729in}}%
\pgfpathlineto{\pgfqpoint{0.928694in}{0.643949in}}%
\pgfpathlineto{\pgfqpoint{0.959980in}{0.669520in}}%
\pgfpathlineto{\pgfqpoint{0.991266in}{0.695092in}}%
\pgfpathlineto{\pgfqpoint{1.022551in}{0.720663in}}%
\pgfpathlineto{\pgfqpoint{1.053837in}{0.746235in}}%
\pgfpathlineto{\pgfqpoint{1.085123in}{0.771806in}}%
\pgfpathlineto{\pgfqpoint{1.116409in}{0.797378in}}%
\pgfpathlineto{\pgfqpoint{1.147695in}{0.822949in}}%
\pgfpathlineto{\pgfqpoint{1.178981in}{0.848521in}}%
\pgfpathlineto{\pgfqpoint{1.210267in}{0.874092in}}%
\pgfpathlineto{\pgfqpoint{1.241553in}{0.899664in}}%
\pgfpathlineto{\pgfqpoint{1.272839in}{0.925235in}}%
\pgfpathlineto{\pgfqpoint{1.304125in}{0.950807in}}%
\pgfpathlineto{\pgfqpoint{1.335411in}{0.976378in}}%
\pgfpathlineto{\pgfqpoint{1.366696in}{1.001950in}}%
\pgfpathlineto{\pgfqpoint{1.397982in}{1.027521in}}%
\pgfpathlineto{\pgfqpoint{1.429268in}{1.053093in}}%
\pgfpathlineto{\pgfqpoint{1.460554in}{1.078664in}}%
\pgfpathlineto{\pgfqpoint{1.491840in}{1.104236in}}%
\pgfpathlineto{\pgfqpoint{1.523126in}{1.129807in}}%
\pgfpathlineto{\pgfqpoint{1.554412in}{1.155379in}}%
\pgfpathlineto{\pgfqpoint{1.585698in}{1.180950in}}%
\pgfpathlineto{\pgfqpoint{1.616984in}{1.206522in}}%
\pgfpathlineto{\pgfqpoint{1.648270in}{1.232093in}}%
\pgfpathlineto{\pgfqpoint{1.679556in}{1.257665in}}%
\pgfpathlineto{\pgfqpoint{1.710842in}{1.283236in}}%
\pgfpathlineto{\pgfqpoint{1.742127in}{1.308808in}}%
\pgfpathlineto{\pgfqpoint{1.773413in}{1.334379in}}%
\pgfpathlineto{\pgfqpoint{1.804699in}{1.359951in}}%
\pgfpathlineto{\pgfqpoint{1.835985in}{1.385522in}}%
\pgfpathlineto{\pgfqpoint{1.867271in}{1.411094in}}%
\pgfpathlineto{\pgfqpoint{1.898557in}{1.436665in}}%
\pgfpathlineto{\pgfqpoint{1.929843in}{1.462237in}}%
\pgfpathlineto{\pgfqpoint{1.961129in}{1.487808in}}%
\pgfpathlineto{\pgfqpoint{1.992415in}{1.513380in}}%
\pgfpathlineto{\pgfqpoint{2.023701in}{1.538951in}}%
\pgfpathlineto{\pgfqpoint{2.054987in}{1.564523in}}%
\pgfpathlineto{\pgfqpoint{2.086273in}{1.590094in}}%
\pgfpathlineto{\pgfqpoint{2.117558in}{1.615666in}}%
\pgfpathlineto{\pgfqpoint{2.148844in}{1.641237in}}%
\pgfpathlineto{\pgfqpoint{2.180130in}{1.666809in}}%
\pgfpathlineto{\pgfqpoint{2.211416in}{1.692380in}}%
\pgfpathlineto{\pgfqpoint{2.242702in}{1.717952in}}%
\pgfpathlineto{\pgfqpoint{2.273988in}{1.743523in}}%
\pgfpathlineto{\pgfqpoint{2.305274in}{1.769095in}}%
\pgfpathlineto{\pgfqpoint{2.336560in}{1.794666in}}%
\pgfpathlineto{\pgfqpoint{2.367846in}{1.820238in}}%
\pgfpathlineto{\pgfqpoint{2.399132in}{1.845809in}}%
\pgfpathlineto{\pgfqpoint{2.430418in}{1.871381in}}%
\pgfpathlineto{\pgfqpoint{2.461703in}{1.896952in}}%
\pgfpathlineto{\pgfqpoint{2.492989in}{1.922524in}}%
\pgfpathlineto{\pgfqpoint{2.524275in}{1.948095in}}%
\pgfpathlineto{\pgfqpoint{2.555561in}{1.973667in}}%
\pgfpathlineto{\pgfqpoint{2.586847in}{1.999238in}}%
\pgfpathlineto{\pgfqpoint{2.618133in}{2.024810in}}%
\pgfpathlineto{\pgfqpoint{2.649419in}{2.050381in}}%
\pgfpathlineto{\pgfqpoint{2.680705in}{2.075953in}}%
\pgfpathlineto{\pgfqpoint{2.711991in}{2.101524in}}%
\pgfpathlineto{\pgfqpoint{2.743277in}{2.127096in}}%
\pgfpathlineto{\pgfqpoint{2.774563in}{2.152667in}}%
\pgfpathlineto{\pgfqpoint{2.805849in}{2.178239in}}%
\pgfpathlineto{\pgfqpoint{2.837134in}{2.203810in}}%
\pgfpathlineto{\pgfqpoint{2.868420in}{2.229382in}}%
\pgfpathlineto{\pgfqpoint{2.899706in}{2.254953in}}%
\pgfpathlineto{\pgfqpoint{2.930992in}{2.280525in}}%
\pgfpathlineto{\pgfqpoint{2.962278in}{2.306096in}}%
\pgfpathlineto{\pgfqpoint{2.993564in}{2.331668in}}%
\pgfpathlineto{\pgfqpoint{3.024850in}{2.357239in}}%
\pgfpathlineto{\pgfqpoint{3.056136in}{2.382811in}}%
\pgfpathlineto{\pgfqpoint{3.087422in}{2.408382in}}%
\pgfpathlineto{\pgfqpoint{3.118708in}{2.433954in}}%
\pgfpathlineto{\pgfqpoint{3.149994in}{2.459525in}}%
\pgfpathlineto{\pgfqpoint{3.181280in}{2.485097in}}%
\pgfpathlineto{\pgfqpoint{3.212565in}{2.510668in}}%
\pgfpathlineto{\pgfqpoint{3.243851in}{2.536240in}}%
\pgfpathlineto{\pgfqpoint{3.275137in}{2.561811in}}%
\pgfpathlineto{\pgfqpoint{3.306423in}{2.587383in}}%
\pgfpathlineto{\pgfqpoint{3.337709in}{2.612954in}}%
\pgfpathlineto{\pgfqpoint{3.360000in}{2.631173in}}%
\pgfusepath{stroke}%
\end{pgfscope}%
\begin{pgfscope}%
\pgfsetrectcap%
\pgfsetmiterjoin%
\pgfsetlinewidth{0.803000pt}%
\definecolor{currentstroke}{rgb}{0.000000,0.000000,0.000000}%
\pgfsetstrokecolor{currentstroke}%
\pgfsetdash{}{0pt}%
\pgfpathmoveto{\pgfqpoint{0.916403in}{0.633903in}}%
\pgfpathlineto{\pgfqpoint{0.916403in}{2.623000in}}%
\pgfusepath{stroke}%
\end{pgfscope}%
\begin{pgfscope}%
\pgfsetrectcap%
\pgfsetmiterjoin%
\pgfsetlinewidth{0.803000pt}%
\definecolor{currentstroke}{rgb}{0.000000,0.000000,0.000000}%
\pgfsetstrokecolor{currentstroke}%
\pgfsetdash{}{0pt}%
\pgfpathmoveto{\pgfqpoint{3.350000in}{0.633903in}}%
\pgfpathlineto{\pgfqpoint{3.350000in}{2.623000in}}%
\pgfusepath{stroke}%
\end{pgfscope}%
\begin{pgfscope}%
\pgfsetrectcap%
\pgfsetmiterjoin%
\pgfsetlinewidth{0.803000pt}%
\definecolor{currentstroke}{rgb}{0.000000,0.000000,0.000000}%
\pgfsetstrokecolor{currentstroke}%
\pgfsetdash{}{0pt}%
\pgfpathmoveto{\pgfqpoint{0.916403in}{0.633903in}}%
\pgfpathlineto{\pgfqpoint{3.350000in}{0.633903in}}%
\pgfusepath{stroke}%
\end{pgfscope}%
\begin{pgfscope}%
\pgfsetrectcap%
\pgfsetmiterjoin%
\pgfsetlinewidth{0.803000pt}%
\definecolor{currentstroke}{rgb}{0.000000,0.000000,0.000000}%
\pgfsetstrokecolor{currentstroke}%
\pgfsetdash{}{0pt}%
\pgfpathmoveto{\pgfqpoint{0.916403in}{2.623000in}}%
\pgfpathlineto{\pgfqpoint{3.350000in}{2.623000in}}%
\pgfusepath{stroke}%
\end{pgfscope}%
\end{pgfpicture}%
\makeatother%
\endgroup%